\documentclass[sigconf]{aamas} 

\usepackage{balance} %

\usepackage[utf8]{inputenc} %
\usepackage[T1]{fontenc}    %
\usepackage{booktabs} %
\usepackage{hyperref}
\usepackage{url}            %
\usepackage{amsfonts}       %
\usepackage{amsmath}       %
\usepackage{nicefrac}       %
\usepackage{microtype}      %
\usepackage{xcolor}         %
\usepackage{xspace}
\usepackage{enumerate}
\usepackage{tabularx}       %
\usepackage{array,multirow,graphicx}
\usepackage{algorithm,algorithmic}
\usepackage{float}
\usepackage{dsfont}
\usepackage{amsthm}
\usepackage{mathtools}
\usepackage{breqn}
\usepackage{cleveref} %
\crefname{figure}{figure}{figures}
\crefname{equation}{equation}{equations}
\crefname{table}{table}{tables}
\crefname{section}{section}{sections}
\crefname{theorem}{theorem}{theorems}
\crefname{corollary}{corollary}{corollaries}
\usepackage{pgf}
\usepackage{subcaption}
\usepackage{tikz}
\usetikzlibrary{arrows,automata}
\usepackage[
    separate-uncertainty=true,
    detect-weight,
    table-number-alignment = center,
	table-text-alignment = center,
	table-format=1.2(3),
    table-sign-mantissa,
]{siunitx}
\usepackage{etoolbox}
\usepackage[symbol]{footmisc}

\newcommand{\pibeh}{$\pi_{\beta}$\xspace}
\newcommand{\piexp}{$\pi_e$\xspace}

\input{widebar}

\setcopyright{ifaamas}
\acmConference[AAMAS '22]{Proc.\@ of the 21st International Conference
on Autonomous Agents and Multiagent Systems (AAMAS 2022)}{May 9--13, 2022}
{Online}{P.~Faliszewski, V.~Mascardi, C.~Pelachaud,
M.E.~Taylor (eds.)}
\copyrightyear{2022}
\acmYear{2022}
\acmDOI{}
\acmPrice{}
\acmISBN{}

\acmSubmissionID{347}

\title[Decoupled Reinforcement Learning to Stabilise Intrinsically-Motivated Exploration]{Decoupled Reinforcement Learning to Stabilise Intrinsically-Motivated Exploration}

\author{Lukas Sch\"afer}
\affiliation{
  \institution{University of Edinburgh}
  \city{Edinburgh}
  \country{United Kingdom}}
\email{l.schaefer@ed.ac.uk}

\author{Filippos Christianos}
\affiliation{
  \institution{University of Edinburgh}
  \city{Edinburgh}
  \country{United Kingdom}}
\email{f.christianos@ed.ac.uk}

\author{Josiah P. Hanna}
\affiliation{
  \institution{University of Wisconsin--Madison}
  \city{Madison}
  \country{United States of America}}
\email{jphanna@cs.wisc.edu}

\author{Stefano V. Albrecht}
\affiliation{
  \institution{University of Edinburgh}
  \city{Edinburgh}
  \country{United Kingdom}}
\email{s.albrecht@ed.ac.uk}

\begin{abstract}
Intrinsic rewards can improve exploration in reinforcement learning, but the exploration process may suffer from instability caused by non-stationary reward shaping and strong dependency on hyperparameters. In this work, we introduce Decoupled RL (DeRL) as a general framework which trains separate policies for intrinsically-motivated exploration and exploitation. Such decoupling allows DeRL to leverage the benefits of intrinsic rewards for exploration while demonstrating improved robustness and sample efficiency. We evaluate DeRL algorithms in two sparse-reward environments with multiple types of intrinsic rewards. Our results show that DeRL is more robust to varying scale and rate of decay of intrinsic rewards and converges to the same evaluation returns than intrinsically-motivated baselines in fewer interactions. Lastly, we discuss the challenge of distribution shift and show that divergence constraint regularisers can successfully minimise instability caused by divergence of exploration and exploitation policies.
\end{abstract}


\newcommand{\BibTeX}{\rm B\kern-.05em{\sc i\kern-.025em b}\kern-.08em\TeX}

\begin{document}

\pagestyle{fancy}
\fancyhead{}

\maketitle

\section{Introduction}

Exploration is one of the essential challenges in reinforcement learning (RL). RL algorithms often use simple randomised methods, e.g. applying $\epsilon$-greedy policies \citep{watkins1989learning} %
or adding random noise to continuous actions \citep{lillicrap2015continuous}.
Such exploration techniques may be inefficient on tasks where rewards are sparse. One category of exploration techniques which has been found to be particularly effective in sparse-reward environments are intrinsic rewards~\citep{flet-berliac2021adversarially,raileanu2020ride,burda2018exploration,pathak2017curiosity,oudeyer2007intrinsic,chentanez2005intrinsically}. %
These additional rewards $r^i$ are computed by the agent and added to the extrinsic reward $r^e$ provided by the environment for a combined reward signal $r = r^e + \lambda r^i$ with some weighting factor $\lambda$. Intrinsic rewards incentivise the exploration of novel or underexplored parts of the environment commonly using self-supervised predictions in the environment~\citep{raileanu2020ride,burda2018exploration,pathak2017curiosity,schmidhuber1991curious} or (pseudo-) counts of states~\citep{ostrovski2017count,tang2017exploration,strehl2008analysis}.

Unfortunately, optimising for this combined feedback introduces three challenges. (1) \textbf{Intrinsic rewards lead to non-stationary rewards} as they are designed to diminish with more completed exploration. Such non-stationary reward shaping violates the Markov assumption and can cause the learning progress to be inconsistent. (2) \textbf{Intrinsically-motivated exploration is sensitive to the scale $\lambda$}. If intrinsic rewards are too large, they might heavily distort training and introduce non-stationary noise to the optimisation. On the other hand, we show that small intrinsic rewards have no sufficient impact and do not incentivise exploration as intended (\Cref{fig:sensitivity_intcoefs}). (3) \textbf{Intrinsically-motivated exploration is sensitive to the rate of decay} which intrinsic rewards rely on throughout training. Similar to their scale, we show that slowly decaying intrinsic rewards disrupt training whereas quickly vanishing intrinsic rewards have insufficient impact on exploration (\Cref{fig:sensitivity_decay}).

These challenges lead to a significant dependency of intrinsic rewards on hyperparameters. Additionally, determining these hyperparameters for scale and rate of decay is task-dependent due to the scale of extrinsic rewards and required exploration in the respective task. Current approaches usually address the difficulties caused by sensitivity to hyperparameters using a large hyperparameter search to find effective parameterisation of a method. However, such a search can be considered an exploration by itself, and introduces bias in reported results focusing only on runs with best-identified hyperparameters and disregarding the considerable computational cost involved~\citep{papoudakis2021benchmarking}. We argue that this bias is particularly harmful in approaches focusing on exploration as best-identified hyperparameters may steer exploration towards the solution of an environment, effectively shifting the achieved exploration from the proposed method to the hyperparameter search. All these properties make the practical application of such methods difficult~\citep{taiga2019benchmarking} and motivate the need for more robust approaches.%

Motivated by these challenges and success in off-policy RL~\citep{zhong2021robust,fujimoto2018addressing,haarnoja2018soft,silver2014deterministic,degris2012off}, we propose to separate the RL training into two separate policies. We train an \textit{exploration policy} \pibeh with the combined signal of extrinsic and intrinsic rewards. Simultaneously, we train an \textit{exploitation policy} \piexp using only extrinsic rewards on the data collected by the exploration policy. We refer to this approach as \textbf{Decoupled RL} (DeRL)\footnote{We provide an implementation of DeRL and the Hallway environment at \url{https://github.com/uoe-agents/derl}}.
Using such decoupling addresses challenges (1)--(3) of previous application of intrinsic rewards. The exploration policy is optimised using the combined objective of extrinsic and intrinsic rewards as in typical intrinsically-motivated RL, but it is only trained to generate data for the training of the exploitation policy. The exploitation policy is thereby decoupled from the challenges of training with intrinsic rewards and optimised to be an effective policy in the given environment. Our experiments show that DeRL leverages the benefits of intrinsically-motivated exploration while stabilising its inherent sensitivity to scale and rate of decay of intrinsic rewards.

We implement and evaluate two versions of DeRL built upon on-policy actor-critic and off-policy Q-learning with five types of intrinsic rewards~\citep{raileanu2020ride,burda2018exploration,pathak2017curiosity,tang2017exploration} in two learning environments that focus on exploration. We analyse the sensitivity of DeRL and RL baselines to the scale and the rate of decay of intrinsic rewards to verify the general dependency of these methods on the hyperparameters of intrinsic rewards and show that DeRL is more robust to varying hyperparameters. Additionally, the exploitation policy of several DeRL algorithms is able to converge to higher evaluation returns using up to $\sim40\%$ fewer interactions and reaches higher returns in some tasks compared to intrinsically-motivated RL baselines. Such improved robustness and sample efficiency can justify the additional cost of training a second policy. However, we also observe that DeRL still suffers from variability in the off-policy optimisation of the exploitation policy \piexp in several tasks. We hypothesise that distribution shift caused by the divergence of \piexp and \pibeh leads to these instabilities, and show that regularisers can be applied to restrict divergence of both policies~\citep{wu2019behavior}, reducing deviations in returns of exploration and exploitation policies and further improving robustness to hyperparameters of intrinsic rewards.

\section{Background}
\subsection{Markov Decision Process}
We formulate an environment as a Markov Decision Process (MDP) \citep{howard1964dynamic} defined as a tuple $(\mathcal{S}, \mathcal{A}, \mathcal{P}, \mathcal{R}, \gamma)$. $\mathcal{S}$ and $\mathcal{A}$ denote the sets of states and actions, respectively, and $\mathcal{P}: \mathcal{S} \times \mathcal{A} \mapsto \Delta(\mathcal{S})$ represents the transition function defining a probability distribution over the next state given current state and applied action. The agent receives rewards for a given transition following $\mathcal{R}: \mathcal{S} \times \mathcal{A} \times \mathcal{S} \mapsto \mathbb{R}$. The objective is to learn a policy $\pi: \mathcal{S} \mapsto \Delta(\mathcal{A})$ which maximises the expected discounted returns $\mathbb{E}_\pi \left[ \sum_{t=0}^\infty \gamma^t \mathcal{R}(s_t, a_t, s_{t+1}) \mid a_t \sim \pi(s_t) \right]$ with discount factor $\gamma \in [0, 1)$.

\subsection{Intrinsically-Motivated Exploration}
\label{sec:background_int_exploration}
A variety of methods have been proposed to replicate the exploration incentive of curiosity in RL~\citep{barto2013intrinsic,oudeyer2009intrinsic,oudeyer2008can} guided by the idea that agents should be incentivised to explore novel or poorly understood parts of the environment. Therefore, intrinsic rewards are defined which reward the agent for such exploration. Over time, the agent should become less ``curious'' and exploitation will gradually take over. %
There are two common branches of intrinsic rewards for exploration: (1) count-based and (2) prediction-based intrinsic rewards. %

\subsubsection{Count-based Intrinsic Rewards}
\label{sec:exploration_intrinsic_count}
Count-based intrinsic rewards are inverse proportional to the visitations of encountered states.

\begin{equation}
    r_t^i \coloneqq \frac{1}{\sqrt{N(s_t)}}
    \label{eq:count_curiosity}
\end{equation}

Thereby, agents are incentivised to visit states within the environment which are less frequently encountered. Likewise, agents are discouraged from visiting frequently encountered states which are deemed less valuable for exploration.
While this approach is easily applicable in small, discrete state spaces, pseudo-counts have to be computed for large or continuous state spaces where encountering any state multiple times is rare. These pseudo-counts can be computed using density models predicting visitations of states~\citep{bellemare2016unifying,ostrovski2017count} or using locality-sensitive~\citep{andoni2008near} hash functions~\citep{tang2017exploration}.

\subsubsection{Prediction-based Intrinsic Rewards}
\label{sec:exploration_intrinsic_pred}
A separate approach defines intrinsic rewards using predictions in the environment. %
\citet{schmidhuber1991curious} proposed an intrinsic reward defined as the error of predicting the next state given the current state and action. However, stochastic and thereby unpredictable dynamics within the environment lead to the so-called ``noisy TV problem''~\citep{burda2018large}, i.e.\ the exploration signal remains high in the presence of unpredictability, which remains a major challenge of these approaches. %

\begin{figure*}[t]
    \centering
    \includegraphics[width=0.7\linewidth]{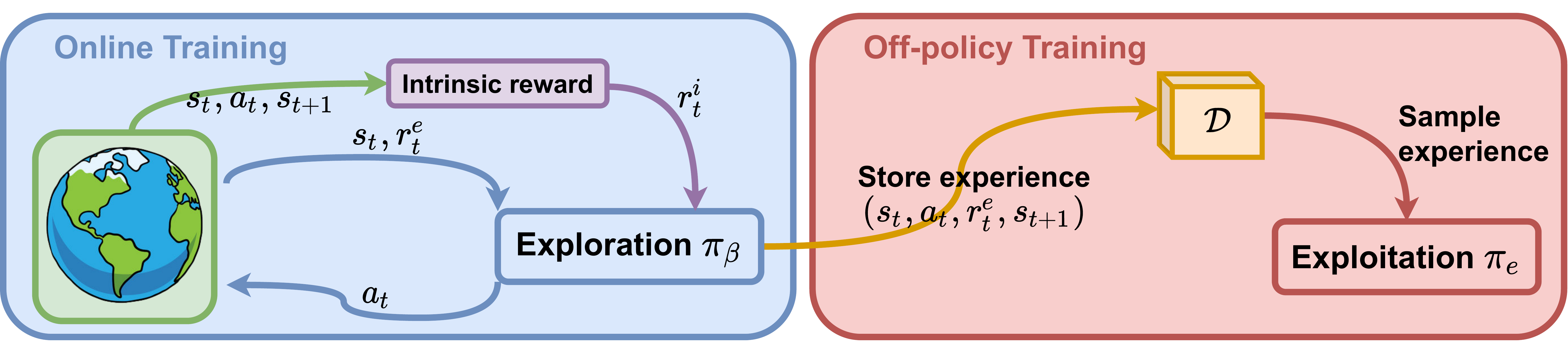}
    \caption{Visualisation of Decoupled Reinforcement Learning (DeRL) training loop.}
    \label{fig:derl_visualisation}
\end{figure*}

\textbf{Intrinsic curiosity module (ICM):} \citet{pathak2017curiosity} propose to learn efficient state representations $\phi(s)$ and assign an intrinsic reward for the prediction error of the next state
\begin{equation}
    \label{eq:icm_prediction}
    r^i_t \coloneqq \left( \widehat{\phi}(s_{t+1}) - \phi(s_{t+1}) \right)^2
\end{equation}
where $\phi$ is a learned self-supervised state-representation trained using an inverse-dynamics objective: given a representation of the current state $\phi(s_t)$ and next state $\phi(s_{t+1})$ predict the applied action $a_t$. Through this representation, the model learns to encode information which can be affected by the agent's actions. The intrinsic reward is given by the error of the prediction $\widehat{\phi}(s_{t+1})$ of the next state given current state $s_t$ and applied action $a_t$.

\textbf{Rewarding impact-driven exploration (RIDE):} \citet{raileanu2020ride} propose to reward the agent for applying actions which lead to significant change in the environment. Such change is defined as the difference between embeddings of consecutive states, where the embedding function $\phi$ is trained using an inverse dynamics model identical to ICM~\citep{pathak2017curiosity}. In order to avoid the agent going back and forth between a group of states, an episodic state-count $N_{ep}$ is added to the objective.
\begin{equation}
    \label{eq:ride_prediction}
    r^i_t \coloneqq \frac{\left( \phi(s_{t+1}) - \phi(s_t) \right)^2}{\sqrt{N_{ep}(s_{t+1})}}
\end{equation}

\textbf{Random network distillation (RND):}
\citet{burda2018exploration} propose a simplified prediction-based intrinsic reward which optimises a state representation function $\widehat{\phi}$ to mimic a randomly initialised, fixed target representation $\phi$.
\begin{equation}
    \label{eq:rnd_prediction}
    r^i_t \coloneqq \left( \widehat{\phi}(s_t) - \phi(s_t) \right)^2
\end{equation}

\section{Related Work}
\citet{liu2021decoupling} propose a new objective for Meta-RL optimisation to decouple exploration and exploitation. Using such an objective, they train separate exploration and exploitation policies guided by task-specific information for fast adaptation to novel scenarios. For multi-agent RL, \citet{liu2021cooperative} learn separate exploration and exploitation policies using off-policy RL to focus coordinated exploration across multiple agents towards underexplored parts within the state space. However, a mixture of both policies is applied to explore whereas our work fully decouples both policies and their training. Furthermore, both of these approaches consider the meta-learning and multi-agent settings, respectively, and do not address the challenge of single-agent exploration we focus on.%

Independently from our work, \citet{whitney2021rethinking} propose to concurrently train an exploration policy using only intrinsic rewards and train a task policy using off-policy soft Double-DQN~\citep{Hasselt:etal:DRL-16}. They apply a factored policy of both the task and exploration policies with optimisations focused on fast adaptation of the exploration policy. In contrast, we fully decouple both trained policies and find that training of the task policy using on-policy actor-critic algorithms with off-policy correction leads to higher returns and less sensitivity to hyperparameters in several tasks. Furthermore, we evaluate DeRL with several intrinsic rewards whereas \citet{whitney2021rethinking} use a single count-based intrinsic reward which is also used for optimisitic initialisation~\citep{rashid2020optimistic}.

\section{Decoupled Reinforcement Learning}
In this work, we propose to decouple exploration and exploitation into two separate policies to improve sample efficiency and reduce sensitivity to hyperparameters of intrinsic rewards. We train an exploration policy \pibeh with the intent to explore the environment. Using the data collected by the exploration policy, we train a separate exploitation policy \piexp, as visualised in \Cref{fig:derl_visualisation}. Separating exploration and exploitation in this way enables training of the exploration policy with intrinsic rewards without modifying the training objective of the exploitation policy.

Formally, an agent trains an exploration policy \pibeh to maximise the sum of intrinsic and extrinsic rewards,
\begin{align}
    \pi_\beta &\in \text{arg}\max_{\pi} \mathbb{E}\left[ \sum_{t=0}^\infty {\gamma}^t \left(r_t^e + \lambda r_{t}^{i} \right) \mid a_t \sim \pi(s_t) \right]\\
              &= \text{arg}\max_{\pi} \mathbb{E}\left[G_{t}^{e+i} \mid a_t \sim \pi(s_t) \right]
    \label{eq:derl_pibeh}
\end{align}
with $G^{e+i}$ denoting the discounted returns computed using the combination of extrinsic and intrinsic rewards with scaling factor $\lambda$ and discount factor $\gamma \in [0, 1)$. During training of \pibeh, experience samples $(s_t, a_t, r_t^e, s_{t+1})$ with extrinsic rewards are collected in $\mathcal{D}$. 

In addition to this typical intrinsically-motivated RL, we train a separate exploitation policy \piexp to maximise only expected cumulative extrinsic rewards using experience accumulated in $\mathcal{D}$ with $G^e_t$ denoting discounted extrinsic returns.
\begin{align}
    \pi_e &\in \text{arg}\max_{\pi} \mathbb{E}\left[ \sum_{t=0}^\infty {\gamma}^t r_{t}^{e} \mid (s_t, a_t, r^e_t, s_{t+1}) \sim \mathcal{D} \right]\\
              &= \text{arg}\max_{\pi} \mathbb{E}\left[G_{t}^{e} \mid (s_t, a_t, r^e_t, s_{t+1}) \sim \mathcal{D} \right]
\end{align}

Both exploration policy \pibeh and exploitation policy \piexp can be trained using any RL algorithm given the defined objectives. We optimise the exploration policy \pibeh as RL with intrinsic rewards~\citep{barto2013intrinsic,oudeyer2009intrinsic,oudeyer2008can}, whereas we train \piexp every $T_{Dec}$ timesteps on experience from $\mathcal{D}$ which is generated by \pibeh's interaction in the environment. Note that $\mathcal{D}$ only contains extrinsic rewards and is off-policy data for the optimisation of \piexp as it was generated by following \pibeh. Therefore, training the exploitation policy using experience generated by the exploration policy requires off-policy RL. Off-policy RL is concerned with the optimisation of a policy using experience generated within the environment by following a separate behaviour policy. %
Below, we propose two methods to apply such decoupled RL using an actor-critic and Q-learning framework.%

\subsection{Decoupled Actor-Critic}
In order to use on-policy RL algorithms to train \piexp using $\mathcal{D}$, off-policy correction must be applied to account for differences in trajectory distributions of both \piexp and \pibeh.%

One technique for off-policy correction %
is \textit{importance sampling} (IS). In the following, we train \piexp using an on-policy actor-critic RL algorithm with state value function $V$, parameterised by $\theta$, and policy \piexp, parameterised by $\phi$. We optimise the latter by minimising the actor loss given in \Cref{eq:dec_policyloss}

\begin{equation}
    \label{eq:dec_policyloss}
    \mathcal{L}(\phi) = \mathbb{E}\left[-\rho(a_t | s_t) \log \pi_e(a_t|s_t;\phi) \ A^e(s_t) \mid (s_t, a_t, r^e_t, s_{t+1}) \sim \mathcal{D}\right]
\end{equation}
with bootstrapped advantage estimates $A^e(s_t)$ and IS weights $\rho(a_t | s_t)$.
\begin{align}
    A^e(s_t) &= \left(r^e_t + \gamma V(s_{t+1};\theta)-V(s_t;\theta)\right)\\
    \rho(a_t | s_t) &= \frac{\pi_e(a_t|s_t; \phi)}{\pi_\beta(a_t|s_t)}
\end{align}

Similarly, the value loss for \piexp using IS weights can be defined as follows:

\begin{multline}
    \label{eq:dec_valueloss}
    \mathcal{L}(\theta) = \mathbb{E}\left[\rho(a_t | s_t) \ \left(V(s_t; \theta) - \left(r_t^e + \gamma V(s_{t+1};\theta)\right)\right)^2 \right.\\
    \left. \mid (s_t, a_t, r^e_t, s_{t+1}) \sim \mathcal{D}\right]
\end{multline}

The IS weights, $\rho$, can cause inconsistent returns during off-policy training through exploding weights when $\pi_e(a_t|s_t; \phi) \gg \pi_\beta(a_t|s_t)$ or vanishing weights for $\pi_e(a_t|s_t; \phi) \ll \pi_\beta(a_t|s_t)$~\citep{christianos2020shared}. In particular, such instabilities occur when one policy assigns approximately zero probability for some action. Various techniques have been proposed to address such exploding weights, %
including clipping of importance weights~\citep{munos2016safe,espeholt2018impala} to minimise vanishing or exploding gradients. The pseudocode for Decoupled Actor-Critic optimisation of \piexp can be found in \Cref{alg:dec_onpolicy}.

\begin{algorithm}[tb]
\caption{Decoupled Actor-Critic}
\label{alg:dec_onpolicy}
\begin{algorithmic}
    \STATE {\bfseries Initialise:} parameters $\phi$, $\theta$ and \pibeh
    \STATE $\mathcal{D} \gets \emptyset$
    \STATE $i \gets 0$
    \FOR {$\text{ep} = 0, \ldots, N_{ep}$}
        \STATE $a_t \sim \pi_\beta(s_t)$
        \STATE $s_{t+1}, r^e_t \gets$ \text{environment step with $a_t$}
        \STATE Update $\pi_\beta$ using RL on intrinsic rewards (\Cref{eq:derl_pibeh})
        \STATE $\mathcal{D} \gets \mathcal{D} \cup (s_t, a_t, r^e_t, s_{t+1})$
        \STATE $i \gets i + 1$
        \IF{$i \mod T_{Dec} = 0$}
            \STATE Update $\phi$ with \Cref{eq:dec_policyloss} and $\mathcal{D}$
            \STATE Update $\theta$ with \Cref{eq:dec_valueloss} and $\mathcal{D}$
            \STATE $\mathcal{D} \gets \emptyset$
        \ENDIF
    \ENDFOR
\end{algorithmic}
\end{algorithm}

\subsection{Decoupled Deep Q-Networks}
Instead of optimising \piexp using actor-critic algorithms with off-policy corrections, we can also apply off-policy algorithms such as Q-learning without the need for any correction. In this work, we consider optimising \piexp using Deep Q-Networks (DQN)~\citep{mnih:etal:atari-15}. For DQN optimisation, the following loss is minimised
\begin{align}
    \mathcal{L}(\theta) = \mathbb{E}\left[\left(Q(s_t, a_t; \theta) - \widebar{Q}(s_t)\right)^2 \mid (s_t, a_t, r^e_t, s_{t+1}) \sim \mathcal{D}\right]
    \label{eq:dec_dqnloss}
\end{align}
with target Q-values $\widebar{Q}(s_t)$ and $\widebar{\theta}$ denoting the parameters of the periodically updated target network.
\begin{equation}
    \widebar{Q}(s_t) = (r_t^e + \gamma \max_{a'}Q(s_{t+1}, a';\widebar{\theta}))
\end{equation}

Pseudocode for Decoupled Deep Q-Networks of \piexp can be found in \Cref{alg:dec_offpolicy}. Note that $\mathcal{D}$ is only used for a single update in Decoupled Actor-Critic, whereas in Decoupled Deep Q-Networks $\mathcal{D}$ represents a replay buffer~\citep{lin1992reinforcement} which is continually filled with experience.

\section{Evaluation}
\label{sec:eval_details}
We evaluate DeRL in two learning environments with a variety of RL algorithms and intrinsic rewards. In particular, we investigate the following three hypotheses: (1) Intrinsically-motivated RL is sensitive to varying scale $\lambda$ and rate of decay of intrinsic rewards, (2) DeRL is more robust than intrinsically-motivated RL baselines to varying scale and rate of decay, and (3) DeRL leads to similar or improved returns and sample efficiency compared to intrinsically-motivated RL baselines.

\subsection{Algorithms}
\label{sec:eval_algorithms}

\textbf{Baselines:} As baselines, we consider on-policy RL algorithms Advantage Actor-Critic (A2C)~\citep{mnih2016asynchronous} and Proximal Policy Optimisation (PPO)~\citep{schulman2017proximal}. Both algorithms are trained using the combined reward $r_t = r^e_t + \lambda r^i_t$ with some weighting factor $\lambda$ and various intrinsic reward definitions as stated below.

\textbf{DeRL:} For our decoupled RL optimisation, we consistently train \pibeh using A2C as we found it to be more robust than PPO. As intrinsic rewards, we use Count and ICM to train \pibeh. %
For the optimisation of \piexp, we consider A2C and PPO for Decoupled Actor-Critic and Decoupled Deep Q-Networks based on DQN. We refer to these algorithms as DeA2C, DePPO and DeDQN.

\subsection{Intrinsic Rewards}
\label{sec:eval_curiosity}

\textbf{Count-based:} We consider two count-based intrinsic rewards computing intrinsic rewards following \Cref{eq:count_curiosity}. \textbf{Count} directly stores and increments state occurrences in a table. \textbf{Hash-Count} first groups states using the SimHash function~\citep{tang2017exploration}.

\textbf{Prediction-based:} Besides count-based intrinsic exploration definitions, we consider \textbf{ICM}~\citep{pathak2017curiosity}, \textbf{RND}~\citep{burda2018exploration}, and \textbf{RIDE}~\citep{raileanu2020ride} as prediction-based approaches. %
For details on these intrinsic rewards, see \Cref{sec:exploration_intrinsic_pred}.

\begin{algorithm}[tb]
\caption{Decoupled Deep Q-Networks}
\label{alg:dec_offpolicy}
\begin{algorithmic}
    \STATE {\bfseries Initialise:} parameters $\theta$ and \pibeh
    \STATE $\mathcal{D} \gets \emptyset$
    \STATE $i \gets 0$
    \FOR {$\text{ep} = 0, \ldots, N_{ep}$}
        \STATE $a_t \sim \pi_\beta (s_t)$
        \STATE $s_{t+1}, r^e_t \gets$ \text{environment step with $a_t$}
        \STATE Update $\pi_\beta$ using RL on intrinsic rewards (\Cref{eq:derl_pibeh})
        \STATE $\mathcal{D} \gets \mathcal{D} \cup (s_t, a_t, r^e_t, s_{t+1})$
        \STATE $i \gets i + 1$
        \IF{$i \mod T_{Dec} = 0$}
            \STATE Update $\theta$ with \Cref{eq:dec_dqnloss} and $\mathcal{D}$
        \ENDIF
    \ENDFOR
\end{algorithmic}
\end{algorithm}

\subsection{Environments}
\label{sec:eval_environments}

\begin{figure}[t]
    \centering
    \includegraphics[width=0.5\linewidth]{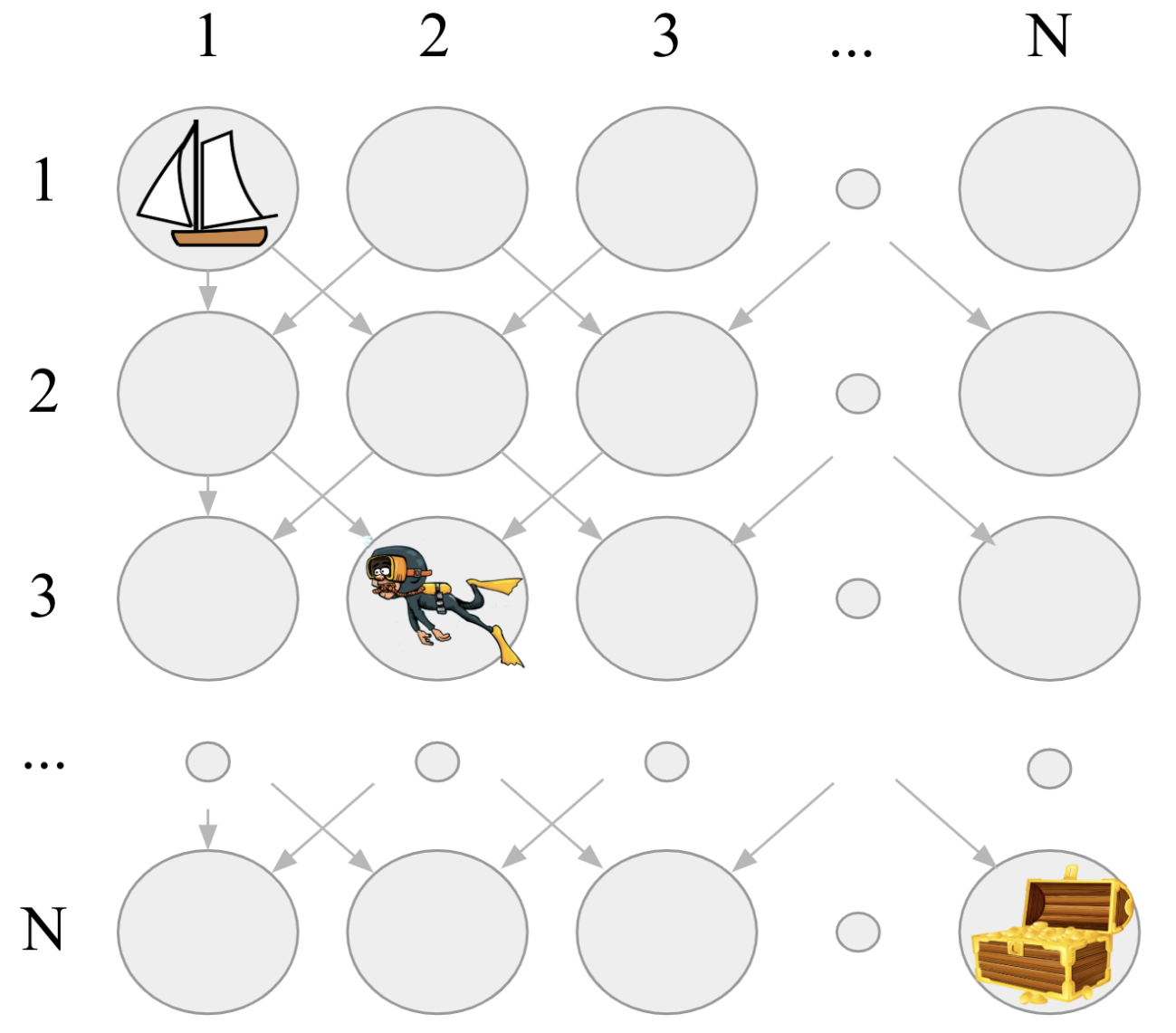}
    \caption{DeepSea environment, from \citet{osband2019behaviour}.}
    \label{fig:env_deepsea}
\end{figure}

\textbf{DeepSea} is an environment proposed as part of the Behaviour Suite (Bsuite) for RL~\citep{osband2019behaviour}, visualised in \Cref{fig:env_deepsea}. The environment targets the challenge of exploration and represents a $N \times N$ grid where the agent starts in the top left and has to reach a goal in the bottom right location. At each timestep, the agent moves one row down and can choose one out of two actions. For each row, both actions are randomly assigned to left and right movement. The agent observes the current location as a 2D one-hot encoding and receives a small negative reward of $\frac{-0.01}{N}$ for moving right and $0$ reward for moving left. Additionally, the agent receives a reward of $+1$ for reaching the goal and the episode ends after $N$ timesteps. The difficulty of the exploration in DeepSea can be adjusted using $N$: the larger $N$, the harder it becomes for the agent to reach the goal location for optimal returns of $0.99$. We evaluate all algorithms in the DeepSea task for $N \in \{10, 14, 20, 24, 30\}$.

\begin{figure}[t]
    \centering
    \includegraphics[width=0.6\linewidth]{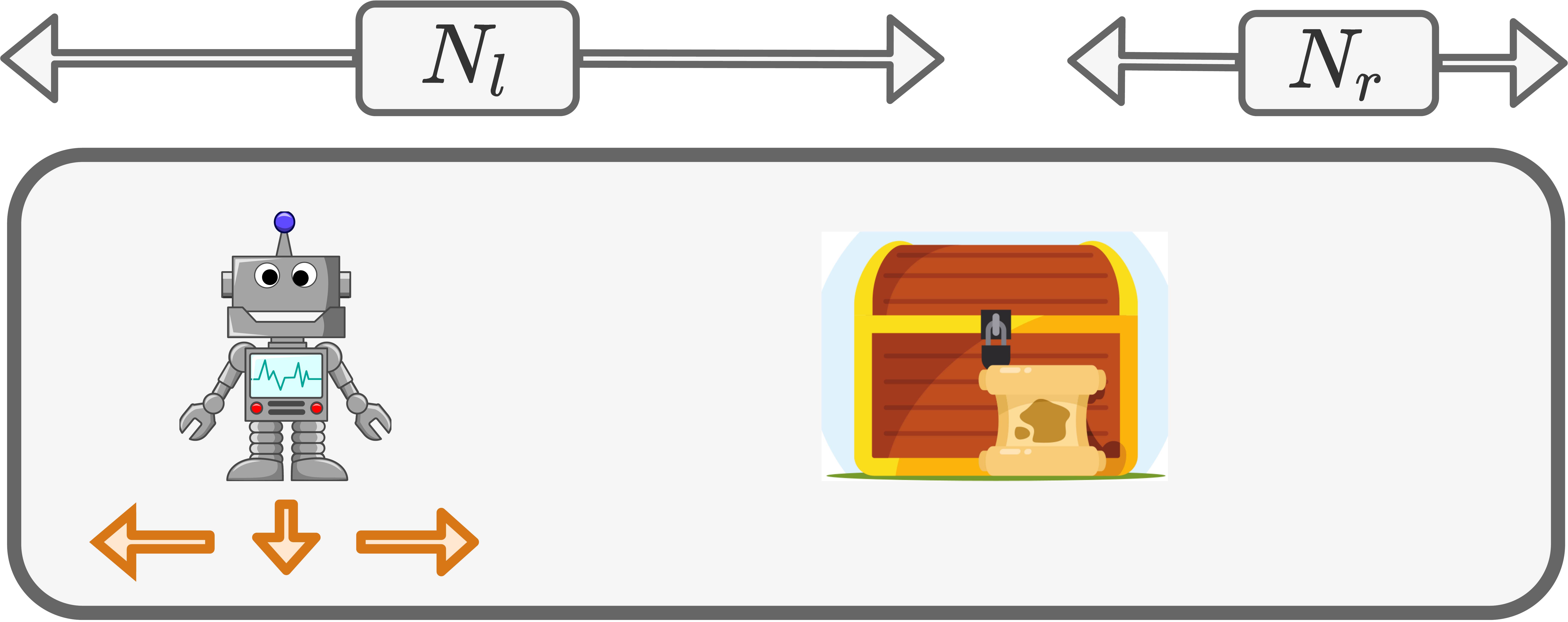}
    \caption{Hallway environment.}
    \label{fig:env_hallway}
\end{figure}

\textbf{Hallway} is a new environment proposed as part of this work to represent domains where exploration and exploitation are misaligned, 
visualised in \Cref{fig:env_hallway}. In DeepSea, agents receive reward by reaching states at the end of the environment, so intrinsic rewards for exploration strongly align with extrinsic rewards from the environment. We hypothesise that tasks in which intrinsic and extrinsic rewards are not well aligned require carefully balanced exploration through intrinsic rewards. Motivated by this hypothesis, we design the Hallway environment in which an agent is located in a hallway starting on the left. A goal can be reached by moving $N_l$ cells to the right. In contrast to DeepSea, the goal is not necessarily located at the right end of the hallway, but there might be further $N_r$ empty cells to the right of the goal location. At each timestep, the agent can choose between three actions: move left, stay or move right. The agent receives a reward of $+1$ for reaching the goal for the first time and every time it stays at the goal location for $10$ steps. Therefore, the agent needs to learn to move to the goal and stay there for the remaining timesteps of the episode to collect further reward. Episodes end after $2 N_l$ steps and small negative reward of $-0.01$ is assigned for moving right or stay. Hallway tasks, in particular with $N_r > 0$, require exploration through intrinsic rewards to be carefully balanced because staying at the goal for optimal returns and exploration are not aligned. %
We evaluate all algorithms in the Hallway environment with $N_l \in \{10, 20, 30\}$ and $N_r$ either being $0$ or equal to $N_l$.

\subsection{Implementation Details}
We compute n-step returns~\citep{sutton2018reinforcement} to reduce the bias of value estimates in all algorithms. On-policy training uses four parallel, synchronous environments and an additional entropy regularisation term in the policy loss~\citep{mnih2016asynchronous}. Double-DQN~\citep{Hasselt:etal:DRL-16} targets are computed for DQN. %
For details on the conducted hyperparameter search as well as all values used throughout experiments, see \Cref{app:hyperparameters}\footnote{All appendices are available online at \url{https://arxiv.org/pdf/2107.08966.pdf}.}.

We train all algorithms for 100,000 episodes and evaluate every 1,000 episodes for a total of 100 evaluations by applying the greedy (evaluation) policy in the respective task for 8 episodes. Following recent suggestions for evaluation in deep RL~\citep{agarwal2021deep}, we report averaged evaluation returns and stratified bootstrap $95\%$ confidence intervals using 5,000 samples for the bootstrapping across five random seeds. Optimal returns are indicated using a dashed horizontal line. A weighting factor of $\lambda=1$ is used for the combined reward signal unless stated otherwise.

\begin{figure*}[!ht]
    \centering
    \includegraphics[width=.5\linewidth]{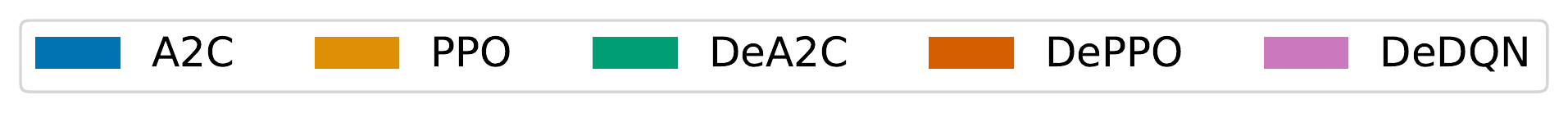}
    \vspace{-1em}
\end{figure*}

\begin{figure*}[!ht]
    \centering
    \begin{subfigure}{.51\linewidth}
        \centering
        \includegraphics[width=\textwidth]{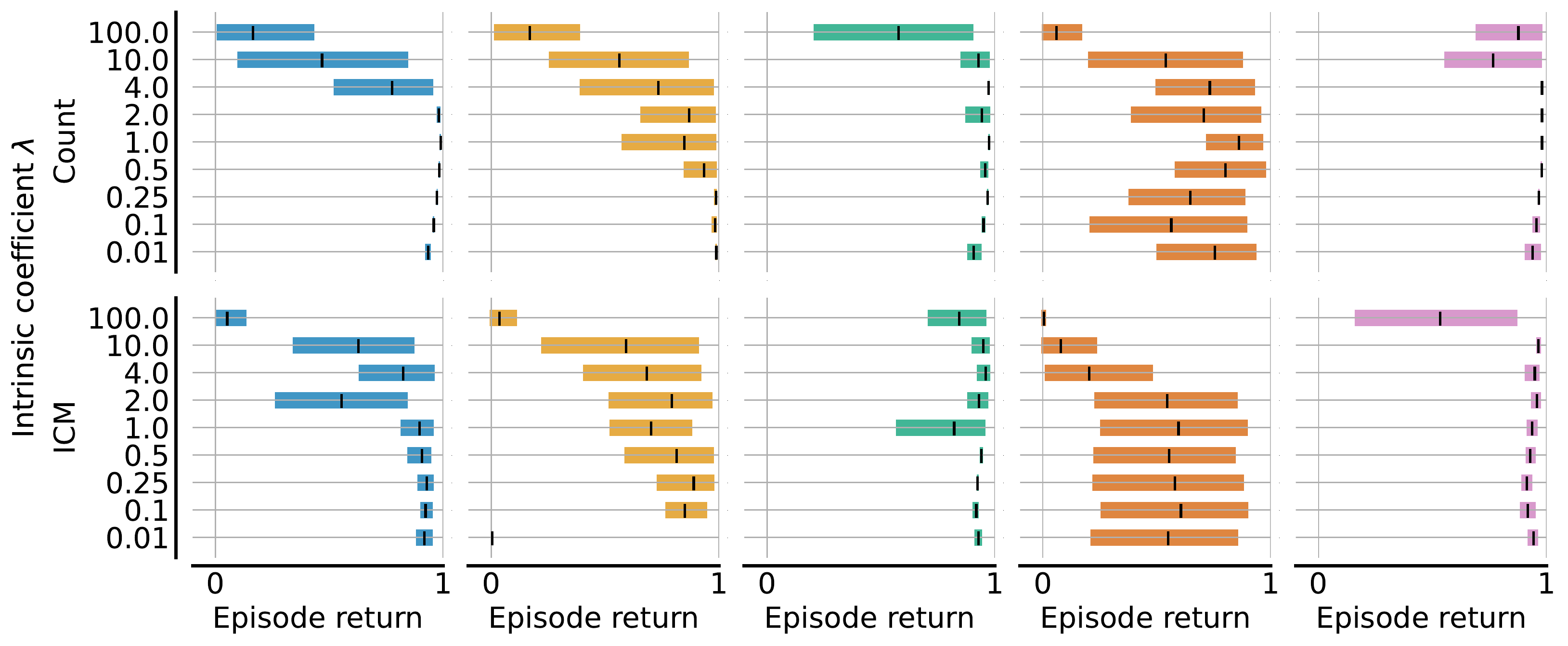}
        \caption{DeepSea 10}
        \label{fig:ds_sensitivity_intcoefs}
    \end{subfigure}
    \begin{subfigure}{.48\linewidth}
        \centering
        \includegraphics[width=\textwidth]{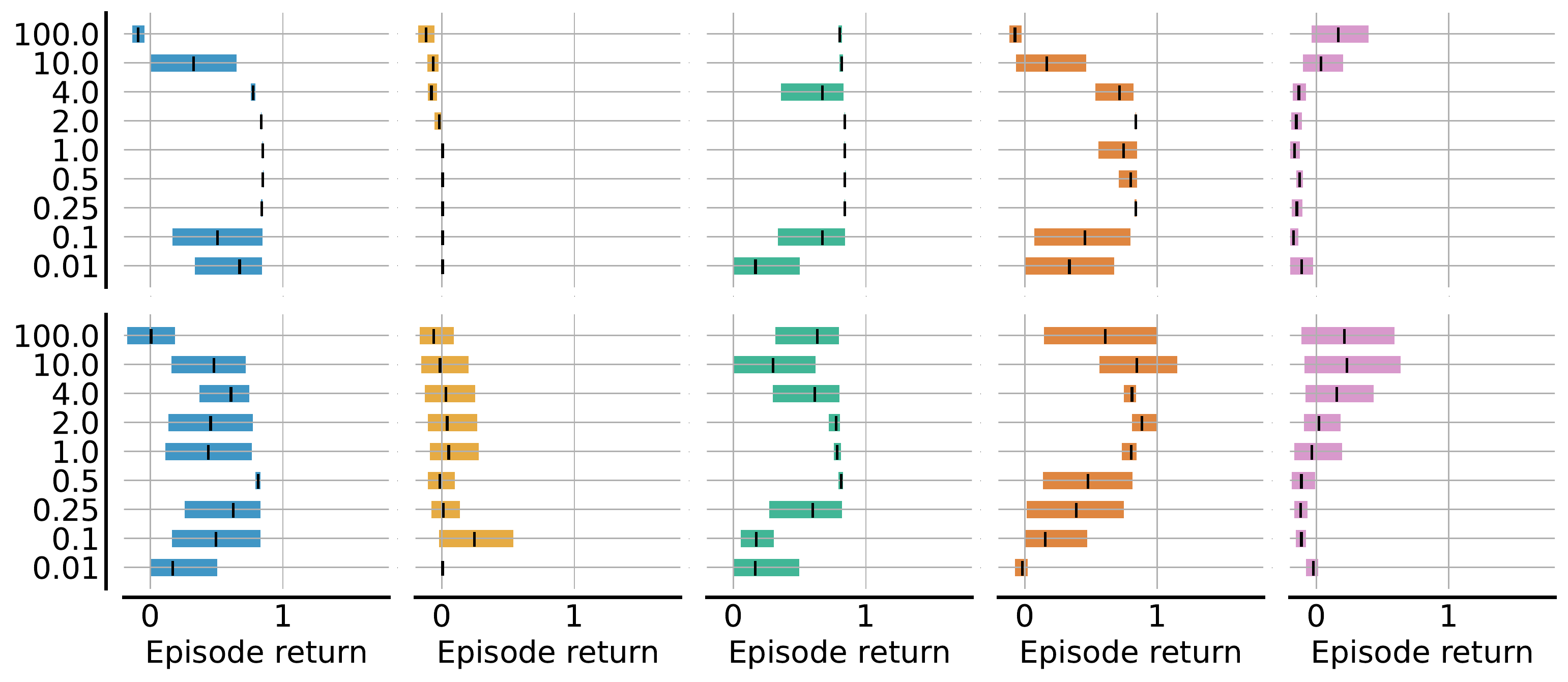}
        \caption{Hallway $N_l=N_r=10$}
        \label{fig:hw_sensitivity_intcoefs}
    \end{subfigure}
    \caption{Average evaluation returns in DeepSea 10 and Hallway $N_l=N_r=10$ with $\lambda \in \{0.01, 0.1, 0.25, 0.5, 1.0, 2.0, 4.0, 10.0, 100.0\}$. Shading indicates 95\% confidence intervals. A method that is insentive to hyperparameters will have final average episodic return concentrated to the right for all hyperparameter values.}
    \label{fig:sensitivity_intcoefs}
\end{figure*}
\begin{figure*}[!ht]
    \centering
    \begin{subfigure}{.51\linewidth}
        \centering
        \includegraphics[width=\textwidth]{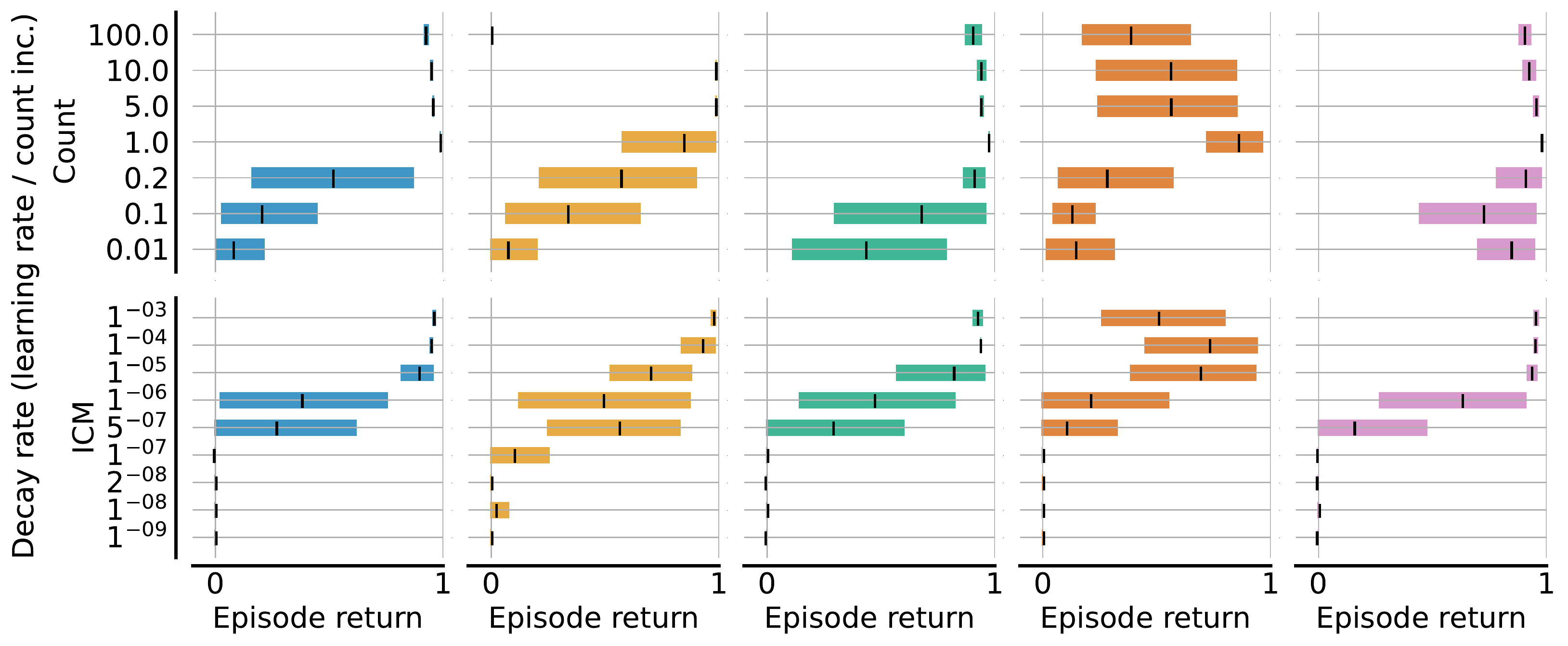}
        \caption{DeepSea 10}
        \label{fig:ds_sensitivity_decay}
    \end{subfigure}
    \begin{subfigure}{.48\linewidth}
        \centering
        \includegraphics[width=\textwidth]{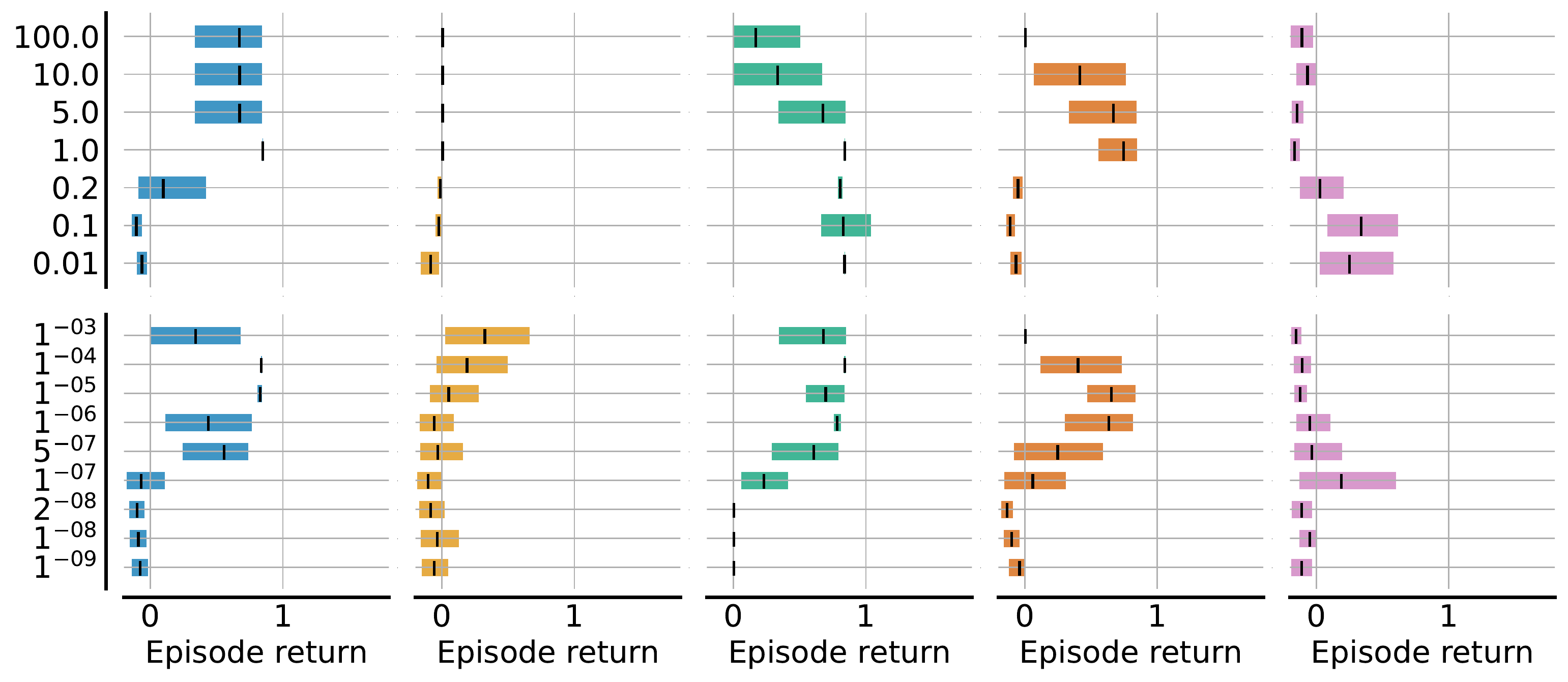}
        \caption{Hallway $N_l=N_r=10$}
        \label{fig:hw_sensitivity_decay}
    \end{subfigure}
    \caption{Average evaluation returns in DeepSea 10 and Hallway $N_l=N_r=10$ with varying rates of decay. Shading indicates 95\% confidence intervals.}
    \label{fig:sensitivity_decay}
\end{figure*}

\subsection{Hyperparameter Sensitivity}
\label{sec:results_sensitivity}
To investigate our first two hypotheses, we %
train all baselines and DeRL algorithms on the combined reward $r = r^e + \lambda r^i$ in DeepSea $N=10$ and Hallway $N_l = N_r = 10$ with varying $\lambda$ and rates of decay. We confirm our first hypothesis that intrinsically-motivated RL is indeed highly sensitive to scale and decay rate of intrinsic rewards, not learning at all or reaching significantly lower evaluation returns for many hyperparameter values. In particular in the Hallway environment where exploration and extrinsic rewards are misaligned, all algorithms exhibit significant dependency on carefully tuned scale and rate of decay. We further confirm our second hypothesis that DeRL algorithms are more robust to varying scale and decay rate of intrinsic rewards, reaching higher returns across a wider range of hyperparameters. \Cref{fig:sensitivity_intcoefs,fig:sensitivity_decay} show sensitivity of all baselines and DeRL algorithms with Count and ICM. Sensitivity analysis for all remaining intrinsic rewards can be found in \Cref{app:hyperparameter_sensitivity}.

\textbf{Scale of intrinsic rewards:}
We consider $\lambda \in \{0.01, 0.1, 0.25, 0.5,\allowbreak 1.0, 2.0, 4.0, 10.0, 100.0\}$ to analyse the sensitivity to varying scale of intrinsic rewards. \Cref{fig:sensitivity_intcoefs} shows average evaluation returns for all values of $\lambda$ for baselines and DeRL. Average returns and bootstrap confidence intervals are computed across five seeds before the average over all 100 evaluations is computed. In DeepSea $N=10$, DeA2C and DeDQN exhibit improved robustness by reaching close to optimal returns for almost all values of $\lambda$. In contrast, DePPO and the baselines are found to be more sensitive in particular to large values of $\lambda$. In Hallway, all algorithms exhibit larger variance for varying $\lambda$ compared to DeepSea with no significant learning being observed for large or small values of $\lambda$, with DeA2C and DePPO demonstrating slightly more robustness. These results indicate the sensitivity to values of $\lambda$. Even small deviations can make the difference between learning and not learning at all.

\textbf{Decay of intrinsic rewards:}
We also investigate the sensitivity of intrinsically-motivated baselines and DeRL algorithms to the rate of decay of intrinsic rewards. For count-based intrinsic rewards, the rate of decay can be determined by the increment of the state count $N(s)$. For a sensitivity analysis, we consider increments $\{0.01, 0.1, 0.2, 1.0, 5.0, 10.0, 100.0\}$. For deep prediction-based intrinsic rewards, we consider learning rates  $\{1e^{-9}, 1e^{-8}, 2e^{-8}, 1e^{-7}, 5e^{-7},\allowbreak 1e^{-6}, 1e^{-5}, 1e^{-4}, 1e^{-3}\}$ determining the rate of decay.
\Cref{fig:sensitivity_decay} shows average evaluation returns of baselines and DeRL with varying rates of decay in both DeepSea $N=10$ and Hallway $N_l=N_r=10$.
A2C is shown to be more robust to varying rates of decay in both environments compared to PPO. DePPO demonstrates larger sensitivity compared to A2C, but DeA2C and DeDQN are again shown to be the most robust algorithms, especially with Count intrinsic rewards, exhibiting high evaluation returns for most considered values in DeepSea $N=10$. Similar to $\lambda$ sensitivity, we observe very significant dependency on the rate of decay in the Hallway task with DeA2C exhibiting improved robustness to varying values. %

\begin{figure*}[ht]
    \centering
    \begin{minipage}{.65\textwidth}
        \centering
        \begin{subfigure}{.52\textwidth}
            \centering
            \includegraphics[width=\linewidth]{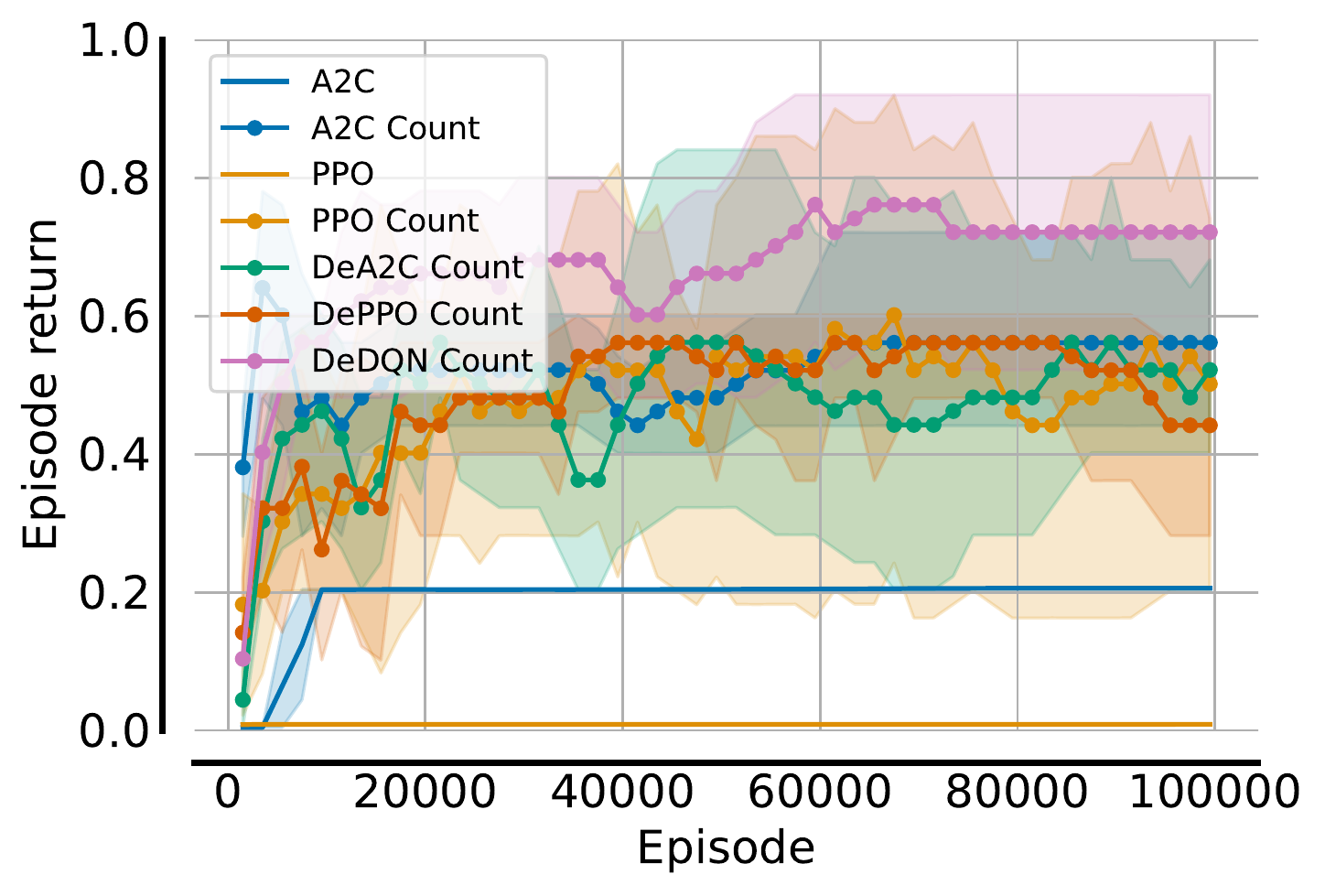}
        \end{subfigure}
        \hfill
        \begin{subfigure}{.47\textwidth}
            \centering
            \includegraphics[width=\linewidth]{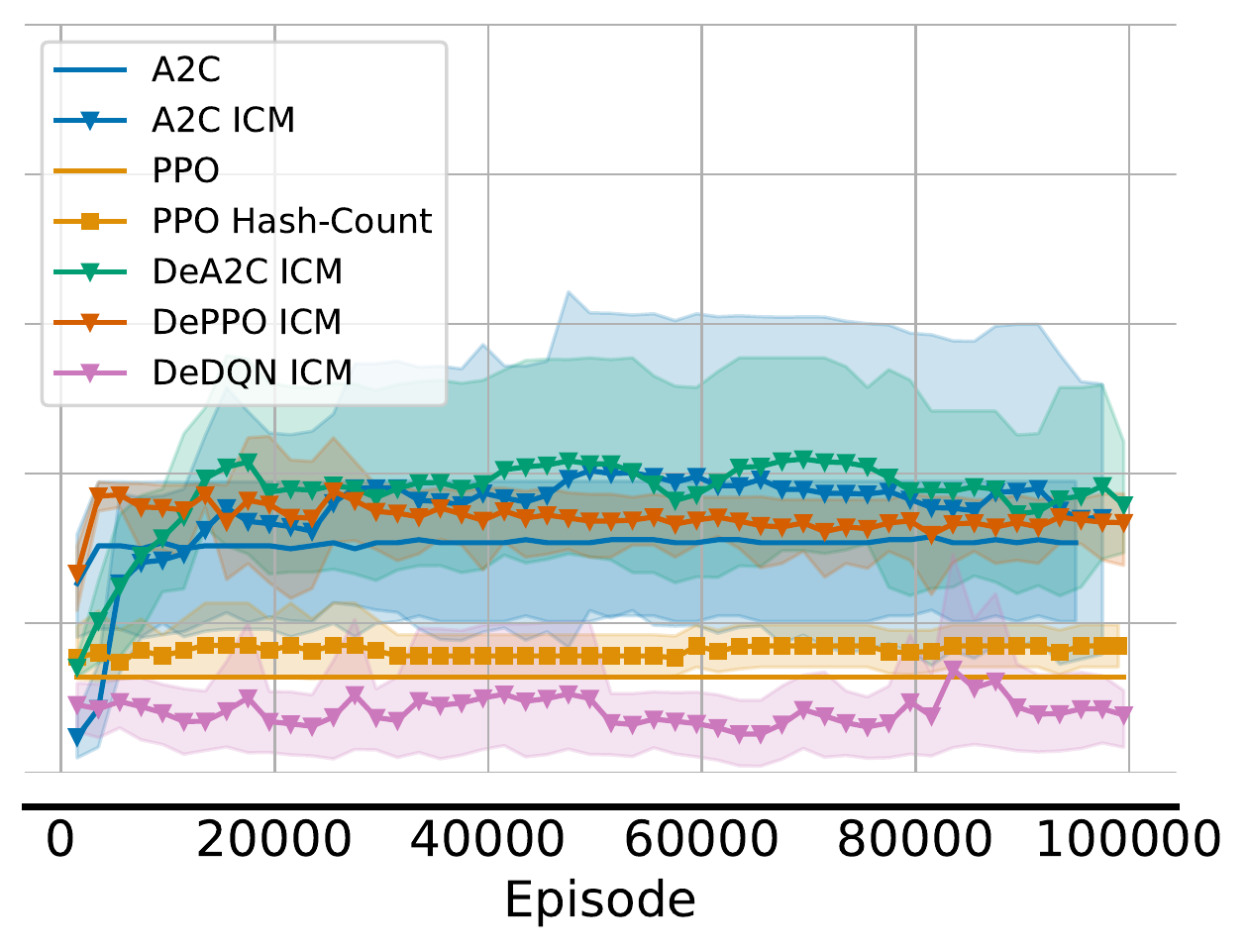}
        \end{subfigure}
        \caption{Normalised evaluation returns for DeepSea (left) and Hallway (right). Returns for each task are normalised to be within [0, 1] before averaged returns and 95\% confidence intervals are computed across all tasks and five random seeds.}
        \label{fig:returns_normalised}
    \end{minipage}%
    \hfill
    \begin{minipage}{.31\textwidth}
        \hspace*{-2em}
        \includegraphics[width=1.1\linewidth]{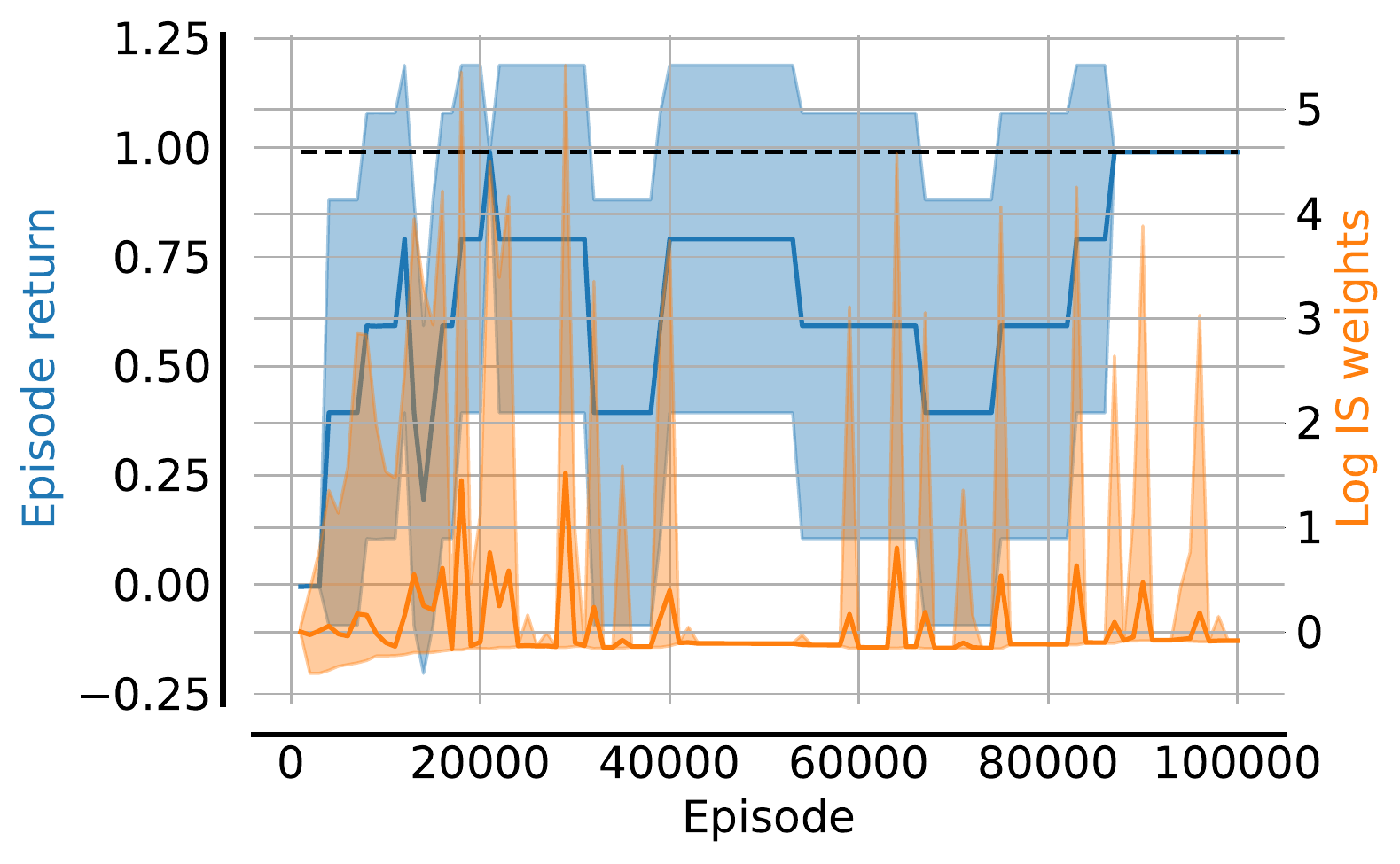}
        \vspace{0em}
        \caption{Evaluation returns and log IS weights for DeA2C Count in DeepSea $N=14$.}
        \label{fig:is_weights_deca2c_deepsea_14}
    \end{minipage}
\end{figure*}

\begin{table*}[t]
	\centering
	\caption{Average evaluation returns and a single standard deviation in all DeepSea and Hallway tasks over 100,000 episodes. The highest achieved returns in each task are highlighted in bold together with all returns within a single standard deviation. For DeRL algorithms, evaluations are executed using the exploitation policy.}
	\resizebox{\linewidth}{!}{
	\robustify\bf
	\begin{tabular}{l S S S S S | S S S S S S}
		\toprule
		{Alg} & {DeepSea 10} & {DeepSea 14} & {DeepSea 20} & {DeepSea 24} & {DeepSea 30} & {Hallway 10-0} & {Hallway 10-10} & {Hallway 20-0} & {Hallway 20-20} & {Hallway 30-0} & {Hallway 30-30}\\
		\midrule
		A2C & \bf 0.93(22) &  -0.00(0) &  -0.00(0) &  -0.00(0) &  -0.00(0) & 0.67(5) &  0.49(9) &  0.42(2) &  0.50(3) &  0.28(8) &  0.42(8) \\
		A2C Count & \bf 0.98(7) & \bf 0.94(16) & \bf 0.74(10) &  0.11(15) &  -0.01(0) & \bf 0.85(1) & \bf 0.85(2) &  0.61(3) & \bf 0.55(6) &  -0.33(15) &  -0.06(7) \\
		A2C Hash-Count & \bf 0.98(7) & \bf 0.96(15) &  0.39(14) & \bf 0.53(12) &  -0.01(0) & \bf 0.85(1) & \bf 0.85(3) &  0.56(3) & \bf 0.55(6) &  -0.34(15) &  -0.13(11) \\
		A2C ICM &  0.87(20) &  0.69(31) &  0.54(23) & \bf 0.46(30) & \bf 0.08(12) & 0.62(17) &  0.57(17) &  0.27(12) & \bf 0.78(27) & \bf 1.16(47) & \bf 0.64(38) \\
		A2C RND &  0.06(1) &  0.19(2) &  -0.01(0) &  -0.01(0) &  -0.01(0) & -0.12(2) &  -0.07(0) &  -0.20(1) &  -0.24(0) &  -0.24(1) &  -0.12(0) \\
		A2C RIDE &  -0.00(0) &  -0.00(0) &  -0.00(0) &  -0.00(0) &  -0.00(0) & \bf 0.85(4) & \bf 0.85(2) & \bf 0.70(0) & \bf 0.62(0) &  0.37(4) &  0.28(8) \\
		\midrule
		PPO &  0.00(0) &  -0.00(0) &  -0.00(0) &  -0.00(0) &  -0.00(0) & 0.00(0) &  0.00(0) &  0.00(0) &  0.00(0) &  0.00(0) &  0.00(0) \\
		PPO Count &  0.84(10) &  0.70(17) &  0.46(19) &  0.17(18) & \bf 0.20(15) & -0.00(1) &  -0.00(0) &  -0.00(1) &  -0.00(1) &  -0.00(0) &  -0.00(1) \\
		PPO Hash-Count &  0.86(8) &  0.77(13) &  0.34(14) &  0.28(20) & \bf 0.12(13) & 0.39(11) &  0.10(7) &  0.00(0) &  0.00(0) &  0.00(0) &  0.00(0) \\
		PPO ICM &  0.84(17) &  0.28(17) &  -0.00(3) &  0.12(17) &  -0.00(3) & 0.05(15) &  0.11(15) &  0.02(16) &  0.08(19) &  -0.04(8) &  -0.02(14) \\
		PPO RND &  0.26(12) &  0.15(8) &  -0.01(0) &  -0.00(0) &  -0.01(0) & -0.04(4) &  -0.04(11) &  -0.21(6) &  -0.17(9) &  -0.27(10) &  -0.27(11) \\
		PPO RIDE &  0.73(8) &  -0.00(0) &  -0.00(2) &  -0.01(0) &  -0.01(0) &  -0.10(3) &  0.02(8) &  -0.21(3) &  -0.08(8) &  -0.32(4) &  -0.29(8) \\
		\midrule
		DeA2C Count & \bf 0.98(10) &  0.65(23) &  0.42(16) &  0.07(10) & \bf 0.09(8) & \bf 0.84(7) & \bf 0.84(9) &  0.42(2) & \bf 0.70(1) &  0.55(0) &  0.22(2) \\
		DeA2C ICM &  0.86(19) &  0.52(28) &  0.27(24) &  0.08(14) & \bf 0.05(11) & 0.77(18) &  0.80(17) &  0.44(15) & \bf 0.53(20) &  0.52(34) & \bf 0.97(51) \\
		DePPO Count &  0.61(20) & \bf 0.92(18) &  -0.01(1) & \bf 0.63(27) &  -0.01(0) & 0.73(10) &  0.80(8) &  0.56(1) & \bf 0.55(4) &  -0.20(17) &  -0.06(7) \\
		DePPO ICM &  0.61(18) &  0.37(17) &  -0.00(1) &  -0.01(0) &  -0.00(0) &  0.82(11) &  0.81(11) &  0.64(16) & \bf 0.57(7) &  -0.01(25) &  0.26(6) \\
		DeDQN Count & \bf 0.98(9) & \bf 0.95(17) &  0.40(8) & \bf 0.53(27) & \bf 0.10(10) & -0.13(4) &  -0.15(4) &  -0.05(5) &  -0.12(8) &  -0.17(7) &  -0.10(6) \\
		DeDQN ICM & \bf 0.94(20) &  0.59(40) &  0.16(12) &  0.24(25) & \bf 0.05(12) & -0.09(9) &  0.02(16) &  -0.11(9) &  -0.19(8) &  -0.26(8) &  -0.19(8) \\
		\bottomrule
	\end{tabular}
	}
	\label{tab:mean_returns}
\end{table*}

\subsection{Evaluation Returns}
\label{sec:results}
Lastly, we report evaluation returns of all algorithms across all DeepSea and Hallway tasks in \Cref{tab:mean_returns}. Average returns and standard deviations are computed across all 100 evaluations after being averaged across five seeds to indicate achieved returns as well as sample efficiency. Additionally, we present normalised returns with 95\% confidence intervals across both environments in \Cref{fig:returns_normalised}, and tables with maximum achieved evaluation returns at any evaluation and learning curves for each individual task in \Cref{app:full_results}.

In DeepSea, DeDQN performs best out of all algorithms (\Cref{fig:returns_normalised}). DeDQN converges to returns comparable to or higher than the best performing baselines exhibiting highest average evaluation returns in all tasks but DeepSea $20$. DeA2C and DePPO demonstrate similar returns and sample efficiency in some of these tasks (see \Cref{fig:deepsea_10_best_app,fig:deepsea_14_best_app}). In DeepSea $24$ (\Cref{fig:deepsea_24_best_app}) and harder Hallway tasks with $N_l = 20, N_r = 0$ and $N_l = N_r = 30$ (\Cref{fig:hallway_results_20_0_app,fig:hallway_results_30_30_app}), the exploitation policies of DeA2C and DePPO converge to the highest returns and are shown to be more sample efficient reaching high returns after up to $40\%$ fewer episodes of training compared to the best performing baselines. %
Generally, we can see that DeA2C learns the optimal policy in the majority of Hallway tasks for some of the five executed runs, but fails to converge to such behaviour consistently. Instead, the majority of baselines and some DeRL runs learn to reach the goal but move back and forth between the goal and its left neighboured cell. Presumably, consistently staying at the goal is rarely discovered due to the small negative reward of staying at a cell.

However, we also observe some failure cases for DeRL algorithms.
DeDQN achieves low returns in the Hallway environment compared to both on-policy DeA2C and DePPO. Also, significant variance can be observed for baselines and DeRL algorithms in harder DeepSea and most Hallway tasks. Off-policy optimisation is theoretically independent of the policy generating training samples, and in DeA2C and DePPO IS weight correction is applied to correct for the off-policy training data. However, we believe distribution shift~\citep{fujimoto2018addressing} is causing inconsistent returns when optimising the exploitation policy from data generated by \pibeh. %
\Cref{fig:is_weights_deca2c_deepsea_14} visualises unstable IS weights for DeA2C in the DeepSea task with $N = 14$ averaged over five seeds. These appear to correlate with some of the noticeable drops in returns throughout training, indicating the negative impact of divergence of exploration and exploitation policies on RL training of \piexp. Even when applying Retrace($\lambda$)~\citep{munos2016safe} to clip IS weights, similar results are observed.%

\begin{figure*}[t]
    \centering
    \begin{subfigure}{.49\textwidth}
        \centering
        \includegraphics[width=1\linewidth]{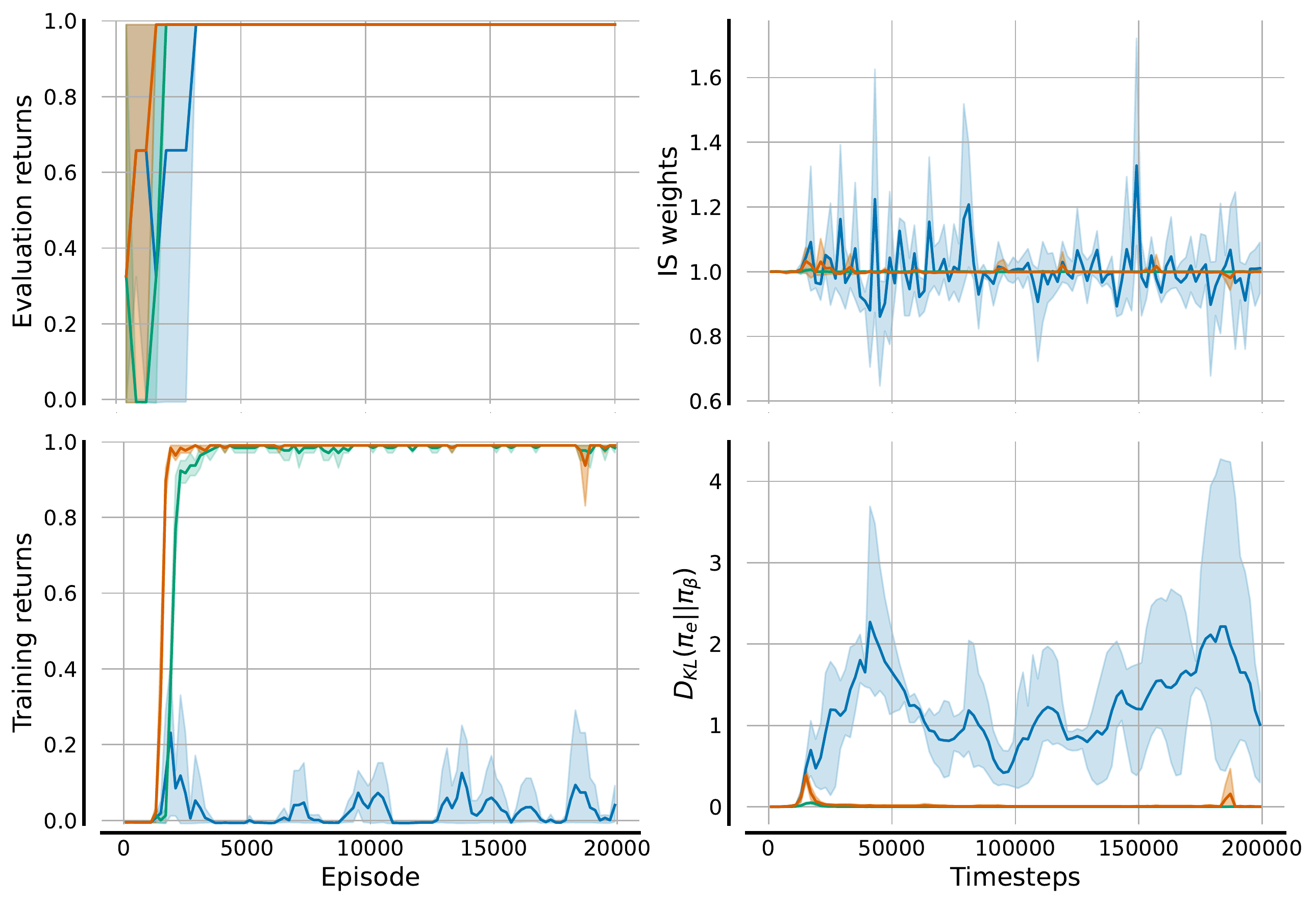}
        \includegraphics[width=.6\textwidth]{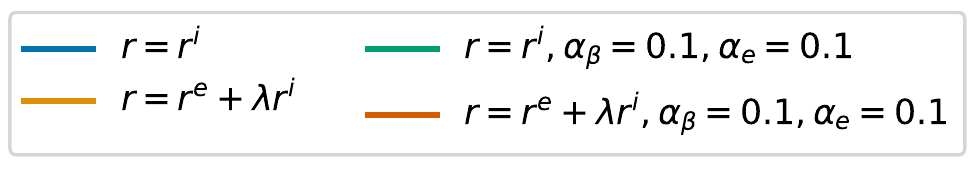}
        \caption{DeepSea $N=10$}
        \label{fig:deepsea10_kldivergence}
    \end{subfigure}
    \hfill
    \begin{subfigure}{.49\textwidth}
        \centering
        \includegraphics[width=1\linewidth]{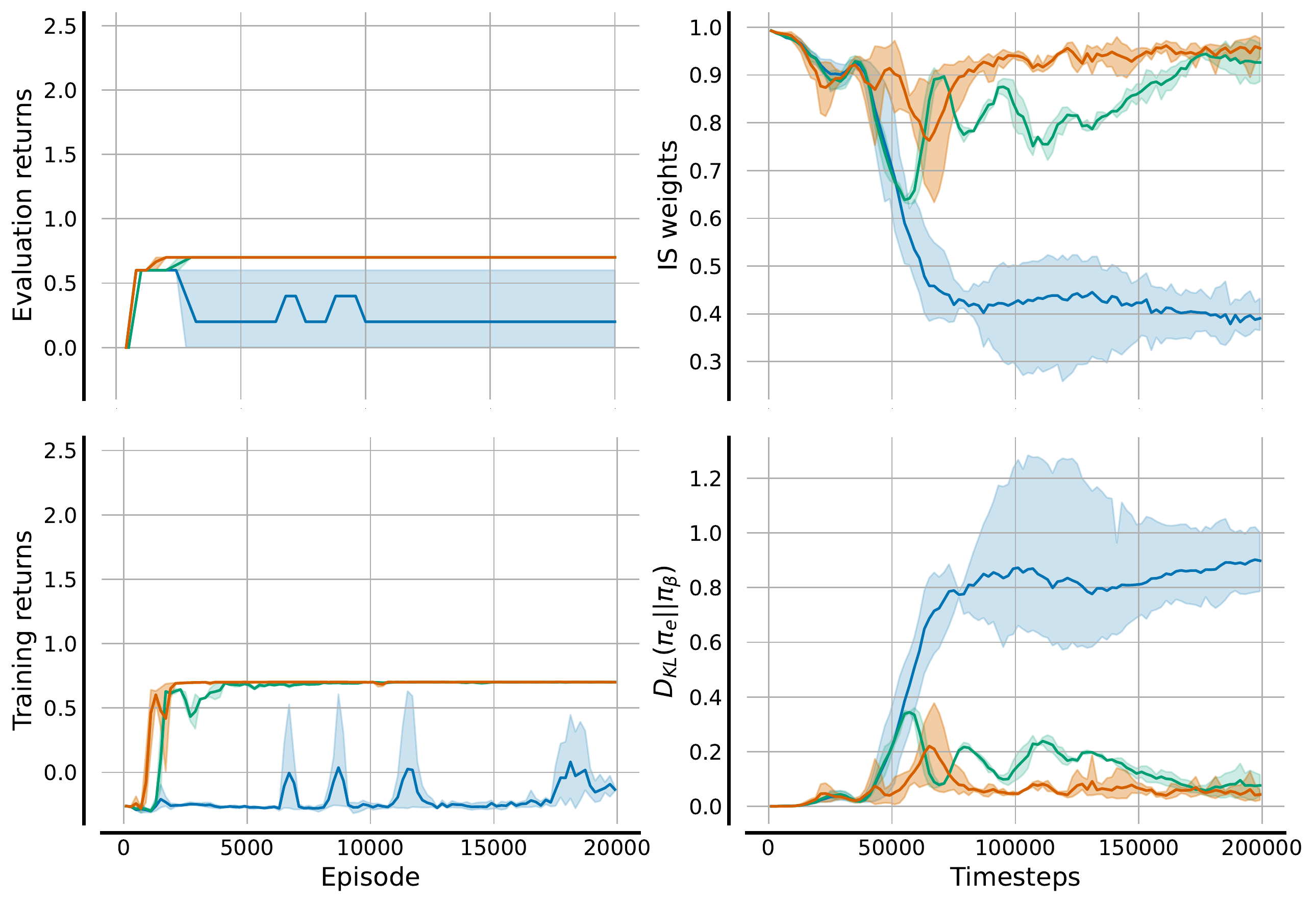}
        \includegraphics[width=.6\textwidth]{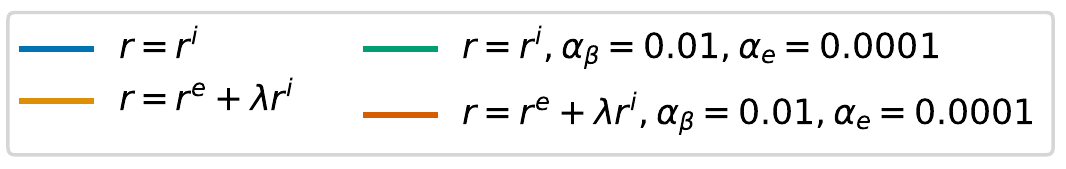}
        \caption{Hallway $N_l=N_r=20$}
        \label{fig:hallway20_20_kldivergence}
    \end{subfigure}
    \caption{Evaluation and training returns, IS weights and KL-divergence of exploration and exploitation policy for training of DeA2C with Count in (a) DeepSea 10 and (b) Hallway $N_l=N_r=20$ with intrinsic and extrinsic rewards or only intrinsic rewards as training signal for \pibeh, and with KL-divergence constraints with coefficients $\alpha_\beta$ and $\alpha_e$. Evaluation and training returns are achieved using the exploitation and exploration policies, respectively. Shading indicates 95\% confidence intervals.}
    \label{fig:kl_divergence}
\end{figure*}

\section{Exploration using only Intrinsic Rewards}
\label{sec:pure_exploration}
Prior work on intrinsic rewards for exploration investigated the effectiveness of training using only intrinsic rewards and no extrinsic rewards from the environment~\citep{flet-berliac2021adversarially,raileanu2020ride,burda2018exploration,pathak2017curiosity}.
Motivated by these experiments, we also investigate the possibility of optimising the exploration policy \pibeh using only intrinsic rewards. Such optimisation would likely lead to increased robustness to hyperparameters of intrinsic rewards as they would not be combined with extrinsic rewards of the environment. However, we also find that the optimisation of \pibeh without extrinsic rewards causes further divergence of \pibeh and \piexp. It should be noted that the evaluation policy is still trained using extrinsic rewards.

We conduct experiments in the DeepSea 10 and Hallway $N_l=N_r=20$ tasks training DeA2C with Count intrinsic rewards for 20,000 episodes. Results are averaged across three seeds and we directly compare optimising \pibeh using either the sum of extrinsic and intrinsic rewards (orange) or only using intrinsic rewards (blue) in \Cref{fig:kl_divergence}. We find that training the exploration policy with only intrinsic rewards does lead to increased divergence of both policies seen in IS weights (top right) and the KL divergence (bottom right). Training of the exploitation policy appears to suffer from such differences in the more challenging Hallway task, but in DeepSea the exploitation policy was successfully trained to solve the task despite the exploration policy never reaching high returns. Our results show the feasibility of training \pibeh using only intrinsic rewards but also the challenge of increased distribution shift.

\section{Divergence Constraints}
\label{sec:divergence_constraints}
In order to address distribution shift caused by diverging exploration and exploitation policies, we investigate the application of divergence constraints proposed in the literature of offline RL~\citep{levine2020offline}. 
These auxiliary objectives are introduced to the optimisation and enforce \piexp and \pibeh to not diverge significantly by introducing a term $\alpha D(\pi_e, \pi_\beta)$
to the optimisation loss. This term is based on a distance measure $D$ between the distribution of policies \pibeh and \piexp and some weighting hyperparameter $\alpha$. %
A common distance measure is the Kullback-Leibler (KL) divergence, which has been applied in offline RL~\citep{jaques2019way}, which can be written as following.
\begin{equation}
    D_{KL} (\pi_e(s_t), \pi_\beta(s_t)) = \mathbb{E}_{a \sim \pi_e(\cdot | S)} \left[ \log \pi_e(a|s_t) - \log \pi_\beta(a|s_t) \right]
    \label{eq:kl_divergence}
\end{equation}
For more distance measure candidates between two policies, see \citet{wu2019behavior} which found these metrics to perform comparably. %

In DeRL, divergence constraints can be directly applied to the optimisation of either the exploration \pibeh or exploitation policy \piexp, i.e.\ can choose to keep \pibeh close to \piexp and likewise can enforce \piexp to stay close to \pibeh. %
We consider either of these directions as well as a combination of both constraints.

We evaluate the application of KL constraints as regularisers in the policy loss of the exploration policy, $\alpha_\beta D_{KL}(\pi_\beta, \pi_e)$, and exploitation policy, $\alpha_e D_e(\pi_e, \pi_\beta)$, with varying weights $\alpha_\beta$ and $\alpha_e$, respectively. These constraints are applied in both settings introduced in \Cref{sec:pure_exploration} with \pibeh being optimised using only intrinsic (green) or intrinsic and extrinsic rewards (red) with results shown in \Cref{fig:kl_divergence} for selected constraint coefficients. We find KL divergence constraints successfully address distribution shift and thereby keep both policies close to each other, even if \pibeh is only trained using intrinsic rewards. Such minimised divergence also leads to reduced variability of returns in both tasks. These results indicate the feasibility of training \pibeh using only intrinsic rewards and the effectiveness of divergence constraints to minimise distribution shift. We further evaluate the sensitivity of DeA2C with KL divergence constraints and show that such regularisation can improve robustness. For figures showing distribution shift, training and evaluation returns for a range of KL constraint coefficients, $\alpha_\beta$ and $\alpha_e$ as well the conducted sensitivity analysis, see \Cref{app:kl_divergence_constraint}.

\section{Conclusion}
In this work, we proposed Decoupled RL (DeRL) which decouples exploration and exploitation into two separate policies. DeRL optimises the exploration policy with additional intrinsic rewards to incentivise exploration and trains the exploitation policy using only extrinsic rewards from data generated by the exploration policy. %
Based on this general framework, we formulate Decoupled Actor-Critic and Decoupled Deep Q-Networks and evaluated in two sparse-reward environments.
Our results demonstrate that intrinsically-motivated RL is highly dependent on careful hyperparameter tuning of intrinsic rewards, indicating the need for more robust solutions. We show that decoupling exploration and exploitation is possible and does lead to significant benefits in robustness to varying scale and rate of decay of intrinsic rewards. Furthermore, we identify distribution shift as a challenge in separating the RL optimisation into two policies with separate optimisation objectives and investigate the application of divergence constraints to minimise such divergence of both policies. Our results demonstrate the effectiveness of divergence constraint regularisation and indicate improved sample efficiency of DeRL in some tasks by reaching high returns in fewer interactions in the environment. %
Lastly, we demonstrated the feasibility of training the exploration policy using only intrinsic rewards. Alongside divergence constraints, such a training setting seems a promising directions for further research into decoupled exploitation and exploration.

\balance
\bibliographystyle{ACM-Reference-Format}
\bibliography{main}

\clearpage

\appendix
\onecolumn

\section{Hyperparameter Configuration}
\label{app:hyperparameters}

A gridsearch was conducted to identify the best hyperparameter configuration for each algorithm in both environments. In order to keep the search feasible, a hyperparameter search for DeepSea was conducted in DeepSea $N=20$ and the same identified configuration was applied in all other DeepSea tasks. Similarly, for the Hallway environment a hyperparameter search was conducted in Hallway $N_l = N_r = 10$. Every gridsearch configuration is evaluated using three different random seeds with results being averaged across all seeds.

\subsection{Baseline Hyperparameters}
First, the best hyperparameters for A2C and PPO baselines were identified by training each configuration in the respective training task of both environments with Count intrinsic rewards. We chose to conduct the hyperparameter search with intrinsic rewards as both baselines do perform poorly in most tasks without any intrinsic rewards which make different hyperparameter configurations hardly distinguishable. The Count intrinsic reward was chosen as it does not require any hyperparameter tuning in contrast to other intrinsic reward definitions. For all considered hyperparameter combinations, we conduct training in the same way as described in \Cref{sec:eval_details} with $\lambda = 1$. The best hyperparameter configuration is chosen as the one which leads to highest maximum evaluation returns. If multiple configurations reach the same maximum evaluation returns then the one with the highest average evaluation returns computed over all 100 evaluations throughout training is considered the best. Such latter metric considers achieved returns alongside sample efficiency. Identified hyperparameters for A2C and PPO in both environments can be found in \Cref{tab:hyper_a2c} and \Cref{tab:hyper_ppo}. All considered values of hyperparameters are listed with the best-identified combination highlighted in bold.

\begin{table}[H]
    \centering
    \small
    \caption{Hyperparameters for A2C baseline.}
    \begin{tabular}{l c c}
    \toprule
     Hyperparameter & DeepSea & Hallway \\
     \midrule
        Normalise observations      & False, \textbf{True}                              & \textbf{False}, True \\
        Normalise rewards           & False, \textbf{True}                              & \textbf{False}, True \\
        Learning rate               & $3e^{-4}, \mathbf{1e^{-3}}$                       & $\mathbf{3e^{-4}}, 1e^{-3}$ \\
        Nonlinearity                & Tanh, \textbf{ReLU}                               & \textbf{Tanh}, ReLU \\
        Maximum gradient norm       & $\mathbf{0.5}, 10.0, 40.0$                        & $\mathbf{0.5}, 10.0, 40.0$ \\
        Entropy loss coefficient    & $\mathbf{1e^{-4}}, 3e^{-4}, 7e^{-4}, 1e^{-3}$     & $\mathbf{1e^{-4}}, 3e^{-4}, 7e^{-4}, 1e^{-3}$ \\
        Actor architecture          & \textbf{Fully connected (64, 64)}                          & \textbf{Fully connected (64, 64)} \\
        Critic architecture         & \textbf{Fully connected (64, 64)}                          & \textbf{Fully connected (64, 64)} \\
        N-steps                     & \textbf{5}                                                 & \textbf{5} \\
        Adam $\epsilon$             & \textbf{0.001}                                             & \textbf{0.001} \\
        Value loss coefficient      & \textbf{0.5}                                               & \textbf{0.5} \\
     \bottomrule 
    \end{tabular}
    \label{tab:hyper_a2c}
\end{table}

\begin{table}[H]
    \centering
    \small
    \caption{Hyperparameters for PPO baseline.}
    \begin{tabular}{l c c}
    \toprule
     Hyperparameter & DeepSea & Hallway \\
     \midrule
        Normalise observations      & \textbf{False}, True                              & \textbf{False}, True \\
        Normalise rewards           & \textbf{False}, True                              & \textbf{False}, True \\
        Learning rate               & $3e^{-4}, \mathbf{1e^{-3}}$                       & $\mathbf{3e^{-4}}, 1e^{-3}$ \\
        Nonlinearity                & \textbf{Tanh}, ReLU                               & Tanh, \textbf{ReLU} \\
        Maximum gradient norm       & $\mathbf{0.5}, 10.0, 40.0$                        & $\mathbf{0.5}, 10.0, 40.0$ \\
        Entropy loss coefficient    & $\mathbf{1e^{-4}}, 3e^{-4}, 7e^{-4}, 1e^{-3}$     & $1e^{-4}, 3e^{-4}, \mathbf{7e^{-4}}, 1e^{-3}$ \\
        Actor architecture          & \textbf{Fully connected (64, 64)}                 & \textbf{Fully connected (64, 64)} \\
        Critic architecture         & \textbf{Fully connected (64, 64)}                 & \textbf{Fully connected (64, 64)} \\
        N-steps                     & \textbf{10}                                       & \textbf{10} \\
        Adam $\epsilon$             & \textbf{0.001}                                    & \textbf{0.001} \\
        Value loss coefficient      & \textbf{0.5}                                      & \textbf{0.5} \\
        Number of epochs            & \textbf{10}                                       & \textbf{10} \\
        Number of minibatches       & \textbf{4}                                        & \textbf{4} \\
        Clipping ratio              & \textbf{0.1}                                      & \textbf{0.1} \\
        Clip value loss             & \textbf{True}                                     & \textbf{True} \\
     \bottomrule 
    \end{tabular}
    \label{tab:hyper_ppo}
\end{table}

\subsection{Intrinsic Reward Hyperparameters}
Following the hyperparameter search of A2C and PPO with Count intrinsic rewards, a search over hyperparameters of all parameterised intrinsic reward definitions was conducted. The setup of the hyperparameter search is identical to the one described above and the best identified hyperparameter configuration for the baseline algorithms are used in the gridsearch for intrinsic rewards. Best identified hyperparameters and all considered hyperparameter configurations can be found in \Cref{tab:hyper_hashcount,tab:hyper_icm,tab:hyper_rnd,tab:hyper_ride}.

\begin{table}[H]
    \centering
    \small
    \caption{Hyperparameters for Hash-Count.}
    \begin{tabular}{p{1em} l c c}
    \toprule
     & Hyperparameter & DeepSea & Hallway \\
     \midrule
        A2C & Hash key dimensionality             & \textbf{16}, 32, 64, 128      & \textbf{16}, 32, 64, 128 \\
     \midrule
        PPO & Hash key dimensionality             & \textbf{16}, 32, 64, 128      & \textbf{16}, 32, 64, 128 \\
     \bottomrule 
    \end{tabular}
    \label{tab:hyper_hashcount}
\end{table}

\begin{table}[H]
    \centering
    \small
    \caption{Hyperparameters for ICM.}
    \begin{tabular}{p{1em} l c c}
    \toprule
     & Hyperparameter & DeepSea & Hallway \\
     \midrule
        \multirow{4}{*}{\rotatebox[origin=c]{90}{General}} & $\phi$ architecture & \textbf{Fully connected (64, 64)} & \textbf{Fully connected (64, 64)} \\
        & $\phi(s)$ dimensionality & \textbf{16} & \textbf{16} \\
        & Forward prediction architecture & \textbf{Fully connected (64)} & \textbf{Fully connected (64)} \\
        & Inverse prediction architecture & \textbf{Fully connected (64)} & \textbf{Fully connected (64)} \\
     \midrule
        \multirow{3}{*}{\rotatebox[origin=c]{90}{A2C}} & Learning rate               & $1e^{-7}, 5e^{-7}, 1e^{-6}, 5e^{-6}, \mathbf{1e^{-5}}$ & $1e^{-7}, 5e^{-7}, \mathbf{1e^{-6}}, 5e^{-6}, 1e^{-5}$ \\
        & Forward loss coefficient    & $0.5, 1.0, \mathbf{5.0}, 10.0$                         & $0.5, 1.0, \mathbf{5.0}, 10.0$ \\
        & Inverse loss coefficient    & $0.5, \mathbf{1.0}, 5.0, 10.0$                         & $\mathbf{0.5}, 1.0, 5.0, 10.0$ \\
     \midrule
        \multirow{3}{*}{\rotatebox[origin=c]{90}{PPO}} & Learning rate               & $1e^{-7}, 5e^{-7}, 1e^{-6}, 5e^{-6}, \mathbf{1e^{-5}}$ & $1e^{-7}, 5e^{-7}, 1e^{-6}, 5e^{-6}, \mathbf{1e^{-5}}$ \\
        & Forward loss coefficient    & $0.5, 1.0, \mathbf{5.0}, 10.0$                         & $\mathbf{0.5}, 1.0, 5.0, 10.0$ \\
        & Inverse loss coefficient    & $0.5, \mathbf{1.0}, 5.0, 10.0$                         & $0.5, 1.0, 5.0, \mathbf{10.0}$ \\
     \bottomrule 
    \end{tabular}
    \label{tab:hyper_icm}
\end{table}

\begin{table}[H]
    \centering
    \small
    \caption{Hyperparameters for RND.}
    \begin{tabular}{l l c c}
    \toprule
     & Hyperparameter & DeepSea & Hallway \\
     \midrule
        General & $\phi$ architecture & \textbf{Fully connected (64, 64)} & \textbf{Fully connected (64, 64)} \\
        & $\phi(s)$ dimensionality & \textbf{16} & \textbf{16} \\
     \midrule
        A2C & Learning rate               & $\mathbf{1e^{-7}}, 5e^{-7}, 1e^{-6}, 5e^{-6}, 1e^{-5}$ & $1e^{-7}, 5e^{-7}, 1e^{-6}, 5e^{-6}, \mathbf{1e^{-5}}$ \\
     \midrule
        PPO & Learning rate               & $\mathbf{1e^{-7}}, 5e^{-7}, 1e^{-6}, 5e^{-6}, 1e^{-5}$ & $1e^{-7}, \mathbf{5e^{-7}}, 1e^{-6}, 5e^{-6}, 1e^{-5}$ \\
     \bottomrule 
    \end{tabular}
    \label{tab:hyper_rnd}
\end{table}

\begin{table}[H]
    \centering
    \small
    \caption{Hyperparameters for RIDE.}
    \begin{tabular}{p{1em} l c c}
    \toprule
     & Hyperparameter & DeepSea & Hallway \\
     \midrule
        \multirow{5}{*}{\rotatebox[origin=c]{90}{General}} & $\phi$ architecture & \textbf{Fully connected (64, 64)} & \textbf{Fully connected (64, 64)} \\
        & $\phi(s)$ dimensionality & \textbf{16} & \textbf{16} \\
        & Forward prediction architecture & \textbf{Fully connected (64)} & \textbf{Fully connected (64)} \\
        & Inverse prediction architecture & \textbf{Fully connected (64)} & \textbf{Fully connected (64)} \\
        & State count & \textbf{Count} & \textbf{Count} \\
     \midrule
        \multirow{3}{*}{\rotatebox[origin=c]{90}{A2C}} & Learning rate               & $1e^{-7}, 5e^{-7}, 1e^{-6}, 5e^{-6}, \mathbf{1e^{-5}}$ & $1e^{-7}, 5e^{-7}, 1e^{-6}, 5e^{-6}, \mathbf{1e^{-5}}$ \\
        & Forward loss coefficient    & $\mathbf{0.5}, 1.0, 5.0, 10.0$                         & $0.5, 1.0, 5.0, \mathbf{10.0}$ \\
        & Inverse loss coefficient    & $0.5, 1.0, 5.0, \mathbf{10.0}$                         & $\mathbf{0.5}, 1.0, 5.0, 10.0$ \\
     \midrule
        \multirow{3}{*}{\rotatebox[origin=c]{90}{PPO}} & Learning rate               & $1e^{-7}, 5e^{-7}, 1e^{-6}, \mathbf{5e^{-6}}, 1e^{-5}$ & $\mathbf{1e^{-7}}, 5e^{-7}, 1e^{-6}, 5e^{-6}, 1e^{-5}$ \\
        & Forward loss coefficient    & $0.5, 1.0, 5.0, \mathbf{10.0}$                         & $0.5, \mathbf{1.0}, 5.0, 10.0$ \\
        & Inverse loss coefficient    & $0.5, \mathbf{1.0}, 5.0, 10.0$                         & $0.5, \mathbf{1.0}, 5.0, 10.0$ \\
     \bottomrule 
    \end{tabular}
    \label{tab:hyper_ride}
\end{table}

\clearpage
\subsection{Decoupled Reinforcement Learning Hyperparameters}
Based on the aforementioned hyperparameters, a gridsearch for all DeRL algorithms, DeA2C, DePPO and DeDQN, was conducted using the best identified A2C and Count intrinsic rewards to train the exploration policy. Identified configurations are listed in \Cref{tab:hyper_dea2c,tab:hyper_deppo,tab:hyper_dedqn}. \Cref{tab:hyper_derl_icm} shows hyperparameters for DeRL algorithms with ICM intrinsic rewards used to train the A2C exploration policy. The same general architecture is used for ICM when applied alongside DeRL as reported in \Cref{tab:hyper_icm}.

\begin{table}[H]
    \centering
    \small
    \caption{Hyperparameters for DeA2C.}
    \begin{tabular}{l c c}
    \toprule
     Hyperparameter & DeepSea & Hallway \\
     \midrule
        Importance sampling         & \textbf{Default IS weights}, Retrace($\lambda$)            & Default IS weights, \textbf{Retrace($\lambda$)} \\
        Learning rate               & $3e^{-4}, \mathbf{1e^{-3}}$                                & $\mathbf{3e^{-4}}, 1e^{-3}$ \\
        Nonlinearity                & Tanh, \textbf{ReLU}                                        & \textbf{Tanh}, ReLU \\
        Maximum gradient norm       & $\mathbf{0.5}$                                             & $\mathbf{0.5}$ \\
        Entropy loss coefficient    & $0.0, \mathbf{1e^{-6}}, 1e^{-5}, 1e^{-4}$                  & $0.0, 1e^{-6}, \mathbf{1e^{-5}}, 1e^{-4}$ \\
        Actor architecture          & \textbf{Fully connected (64, 64)}                 & \textbf{Fully connected (64, 64)} \\
        Critic architecture         & \textbf{Fully connected (64, 64)}                 & \textbf{Fully connected (64, 64)} \\
        N-steps                     & \textbf{5}                                        & \textbf{5} \\
        Adam $\epsilon$             & \textbf{0.001}                                    & \textbf{0.001} \\
        Value loss coefficient      & \textbf{0.5}                                      & \textbf{0.5} \\
        $T_{Dec}$                   & \textbf{1}                                        & \textbf{1} \\
     \bottomrule 
    \end{tabular}
    \label{tab:hyper_dea2c}
\end{table}

\begin{table}[H]
    \centering
    \small
    \caption{Hyperparameters for DePPO.}
    \begin{tabular}{l c c}
    \toprule
     Hyperparameter & DeepSea & Hallway \\
     \midrule
        Importance sampling         & \textbf{Default IS weights}                       & \textbf{Default IS weights} \\
        Learning rate               & $3e^{-4}, \mathbf{1e^{-3}}$                       & $\mathbf{3e^{-4}}, 1e^{-3}$ \\
        Nonlinearity                & Tanh, \textbf{ReLU}                               & Tanh, \textbf{ReLU} \\
        Maximum gradient norm       & $\mathbf{0.5}$                                    & $\mathbf{0.5}$ \\
        Entropy loss coefficient    & $0.0, 1e^{-6}, 1e^{-5}, \mathbf{1e^{-4}}$         & $0.0, \mathbf{1e^{-6}}, 1e^{-5}, 1e^{-4}$ \\
        Actor architecture          & \textbf{Fully connected (64, 64)}                 & \textbf{Fully connected (64, 64)} \\
        Critic architecture         & \textbf{Fully connected (64, 64)}                 & \textbf{Fully connected (64, 64)} \\
        N-steps                     & \textbf{10}                                       & \textbf{10} \\
        Adam $\epsilon$             & \textbf{0.001}                                    & \textbf{0.001} \\
        Value loss coefficient      & \textbf{0.5}                                      & \textbf{0.5} \\
        Number of epochs            & \textbf{10}                                       & \textbf{10} \\
        Number of minibatches       & \textbf{4}                                        & \textbf{4} \\
        Clipping ratio              & \textbf{0.1}                                      & \textbf{0.1} \\
        Clip value loss             & \textbf{True}                                     & \textbf{True} \\
        $T_{Dec}$                   & \textbf{1}                                        & \textbf{1} \\
     \bottomrule 
    \end{tabular}
    \label{tab:hyper_deppo}
\end{table}

\begin{table}[H]
    \centering
    \small
    \caption{Hyperparameters for DeDQN.}
    \begin{tabular}{l c c}
    \toprule
     Hyperparameter & DeepSea & Hallway \\
     \midrule
        Learning rate               & $1e^{-4}, 3e^{-4}, \mathbf{1e^{-3}}$                       & $\mathbf{1e^{-4}}, 3e^{-4}, 1e^{-3}$ \\
        Soft target update $\tau$   & \textbf{0.01}, 0.001                                       & 0.01, \textbf{0.001} \\
        Batch size                  & 128, \textbf{256}, 512                                     & 128, 256, \textbf{512} \\
        N-steps                     & \textbf{5}                                                 & \textbf{5} \\
        Nonlinearity                & \textbf{Tanh}, ReLU                                        & Tanh, \textbf{ReLU} \\
        Replay buffer               & \textbf{Default}, Prioritised                              & \textbf{Default}, Prioritised \\
        Replay buffer capacity      & \textbf{100,000}                                           & \textbf{100,000} \\
        Maximum gradient norm       & $\mathbf{0.5}$                                             & $\mathbf{0.5}$ \\
        DQN architecture            & \textbf{Fully connected (64, 64)}                          & \textbf{Fully connected (64, 64)} \\
        Adam $\epsilon$             & \textbf{0.001}                                             & \textbf{0.001} \\
        $T_{Dec}$                   & \textbf{1}                                                 & \textbf{1} \\
     \bottomrule 
    \end{tabular}
    \label{tab:hyper_dedqn}
\end{table}

\begin{table}[H]
    \centering
    \small
    \caption{Hyperparameters for DeRL with ICM.}
    \begin{tabular}{p{1em} l c c}
    \toprule
     & Hyperparameter & DeepSea & Hallway \\
     \midrule
        \multirow{3}{*}{\rotatebox[origin=c]{90}{DeA2C}} & Learning rate               & $1e^{-7}, 5e^{-7}, 1e^{-6}, 5e^{-6}, \mathbf{1e^{-5}}$ & $1e^{-7}, 5e^{-7}, \mathbf{1e^{-6}}, 5e^{-6}, 1e^{-5}$ \\
        & Forward loss coefficient    & $0.5, 1.0, \mathbf{5.0}, 10.0$                         & $0.5, 1.0, \mathbf{5.0}, 10.0$ \\
        & Inverse loss coefficient    & $0.5, \mathbf{1.0}, 5.0, 10.0$                         & $\mathbf{0.5}, 1.0, 5.0, 10.0$ \\
        \midrule
        \multirow{3}{*}{\rotatebox[origin=c]{90}{DePPO}} & Learning rate               & $1e^{-7}, 5e^{-7}, 1e^{-6}, 5e^{-6}, \mathbf{1e^{-5}}$ & $1e^{-7}, 5e^{-7}, 1e^{-6}, 5e^{-6}, \mathbf{1e^{-5}}$ \\
        & Forward loss coefficient    & $0.5, 1.0, \mathbf{5.0}, 10.0$                         & $\mathbf{0.5}, 1.0, 5.0, 10.0$ \\
        & Inverse loss coefficient    & $0.5, \mathbf{1.0}, 5.0, 10.0$                         & $0.5, 1.0, 5.0, \mathbf{10.0}$ \\
        \midrule
        \multirow{3}{*}{\rotatebox[origin=c]{90}{DeDQN}} & Learning rate               & $1e^{-7}, 5e^{-7}, 1e^{-6}, 5e^{-6}, \mathbf{1e^{-5}}$ & $1e^{-7}, 5e^{-7}, 1e^{-6}, 5e^{-6}, \mathbf{1e^{-5}}$ \\
        & Forward loss coefficient    & $0.5, 1.0, \mathbf{5.0}, 10.0$                         & $\mathbf{0.5}, 1.0, 5.0, 10.0$ \\
        & Inverse loss coefficient    & $0.5, \mathbf{1.0}, 5.0, 10.0$                        & $0.5, 1.0, 5.0, \mathbf{10.0}$ \\
     \bottomrule 
    \end{tabular}
    \label{tab:hyper_derl_icm}
\end{table}

\clearpage

\section{Evaluation Results}
\label{app:full_results}
In this section, we provide tables containing maximum evaluation returns for all baselines and DeRL algorithms in every task. As described in \Cref{sec:eval_details}, average returns and stratified bootstrap 95\% confidence intervals are computed using 5,000 samples across five random seeds for the best identified hyperparameter configuration as reported in \Cref{app:hyperparameters}. For maximum evaluation returns, we identify the single evaluation out of all 100 conducted evaluations with the maximum evaluation returns averaged across samples and report its value with the standard deviation across all samples. Within tables, the highest performing algorithms for each task are highlighted in bold together with every algorithm within a single standard deviation of the highest return. Besides these tables, we also provide learning curves for all baselines and DeRL algorithms in every evaluated task. In order to maintain visual clarity, we average two following evaluations (of the total of 100 evaluations throughout training) for a total of 50 plotted evaluation points for each algorithm.

\subsection{DeepSea}

\begin{table*}[h]
	\centering
	\caption{Maximum evaluation returns in the DeepSea environment with a single standard deviation.}
	\small
	\resizebox{\linewidth}{!}{
	\robustify\bf
	\begin{tabular}{l S S S S S}
		\toprule
		{Algorithm \textbackslash \ Task} & {DeepSea 10} & {DeepSea 14} & {DeepSea 20} & {DeepSea 24} & {DeepSea 30} \\
		\midrule
		 A2C & \bf 0.99(0) &  0.00(0) &  0.00(0) &  -0.00(0) &  -0.00(0) \\
		 A2C Count & \bf 0.99(0) & \bf 0.99(0) & \bf 0.99(0) & \bf 0.79(40) &  -0.01(0) \\
		 A2C Hash-Count & \bf 0.99(0) & \bf 0.99(0) &  0.64(48) & \bf 0.59(49) &  -0.00(0) \\
		 A2C ICM & \bf 0.99(0) & \bf 0.99(0) &  0.70(45) & \bf 0.79(40) & \bf 0.39(49) \\
		 A2C RND &  0.06(24) &  0.19(40) &  -0.00(0) &  -0.00(0) &  -0.00(0) \\
		 A2C RIDE &  -0.00(0) &  -0.00(0) &  -0.00(0) &  -0.00(0) &  -0.00(0) \\
		 PPO &  0.00(0) &  -0.00(0) &  -0.00(0) &  -0.00(0) &  -0.00(0) \\
		 PPO Count & \bf 0.99(0) &  0.79(40) &  0.73(44) & \bf 0.59(49) & \bf 0.59(49) \\
		 PPO Hash-Count & \bf 0.99(0) & \bf 0.99(0) &  0.59(49) & \bf 0.79(40) & \bf 0.59(49) \\
		 PPO ICM & \bf 0.99(0) &  0.39(49) &  0.31(46) & \bf 0.79(40) & \bf 0.19(40) \\
		 PPO RND &  0.38(48) &  0.19(40) &  -0.00(0) &  -0.00(0) &  -0.00(0) \\
		 PPO RIDE &  0.77(41) &  -0.00(0) &  0.15(36) &  -0.00(0) &  -0.00(0) \\
		 DeA2C Count & \bf 0.99(0) & \bf 0.99(0) &  0.90(28) &  0.19(40) & \bf 0.16(37) \\
		 DeA2C ICM & \bf 0.99(0) &  0.82(37) &  0.66(47) &  0.39(49) & \bf 0.33(47) \\
		 DePPO Count & \bf 0.99(0) & \bf 0.99(0) &  0.08(28) & \bf 0.82(37) &  -0.00(0) \\
		 DePPO ICM & \bf 0.99(0) &  0.66(47) &  0.09(30) &  -0.00(0) &  -0.00(0) \\
		 DeDQN Count & \bf 0.99(0) & \bf 0.99(0) &  0.44(49) & \bf 0.79(40) & \bf 0.19(40) \\
		 DeDQN ICM & \bf 0.99(0) & \bf 0.99(0) &  0.30(46) & \bf 0.59(49) & \bf 0.44(49) \\
		\bottomrule
	\end{tabular}
	}
	\label{tab:deepsea_max_returns}
\end{table*}

\begin{figure*}[ht!]
    \centering
    \begin{subfigure}{.33\textwidth}
        \centering
        \includegraphics[width=\linewidth]{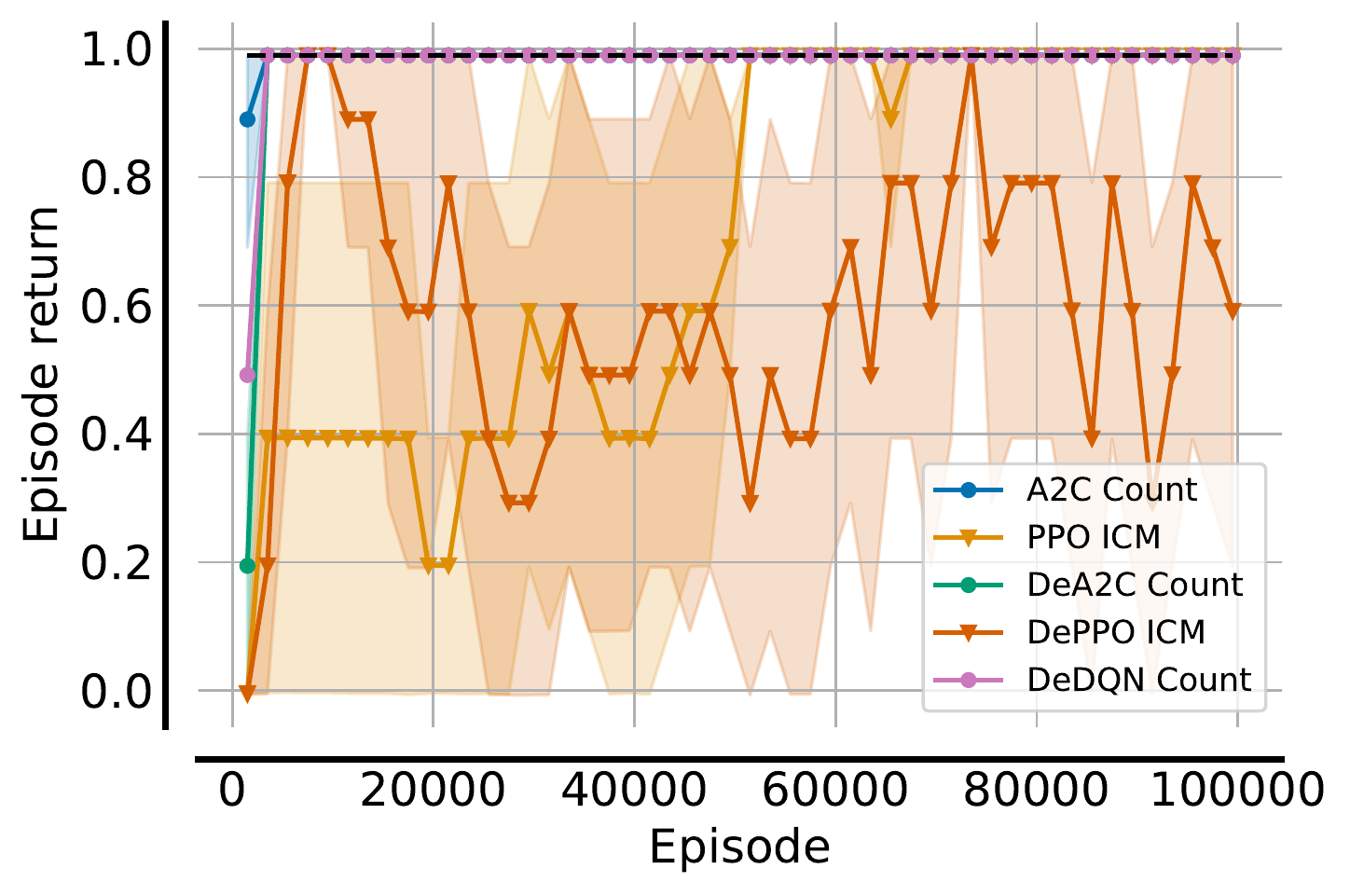}
        \caption{DeepSea $N=10$}
        \label{fig:deepsea_10_best_app}
    \end{subfigure}
    \hfill
    \begin{subfigure}{.33\textwidth}
        \centering
        \includegraphics[width=\linewidth]{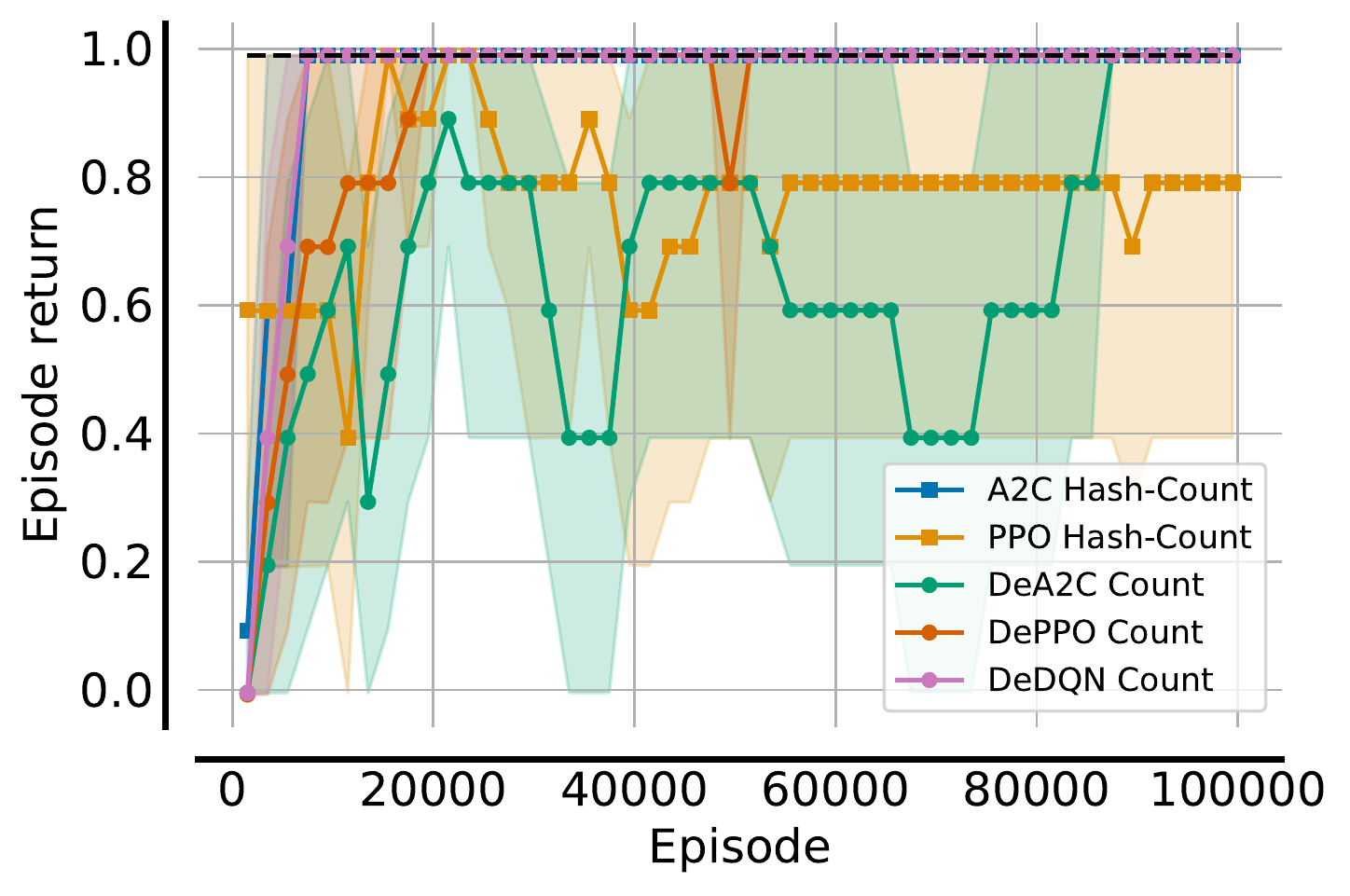}
        \caption{DeepSea $N=14$}
        \label{fig:deepsea_14_best_app}
    \end{subfigure}
    \hfill
    \begin{subfigure}{.33\textwidth}
        \centering
        \includegraphics[width=\linewidth]{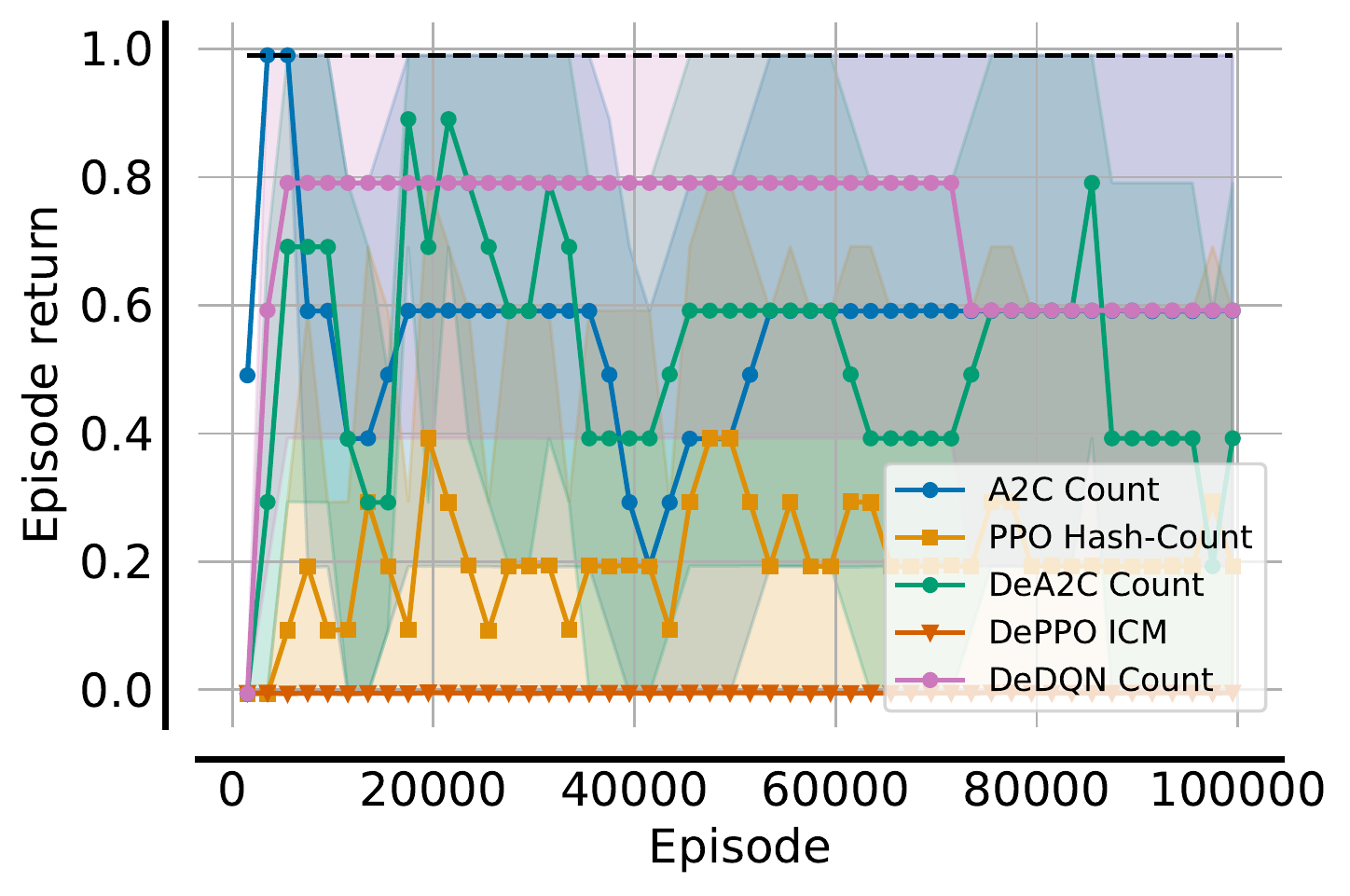}
        \caption{DeepSea $N=20$}
        \label{fig:deepsea_20_best_app}
    \end{subfigure}
    
    \begin{subfigure}{.33\textwidth}
        \centering
        \includegraphics[width=\linewidth]{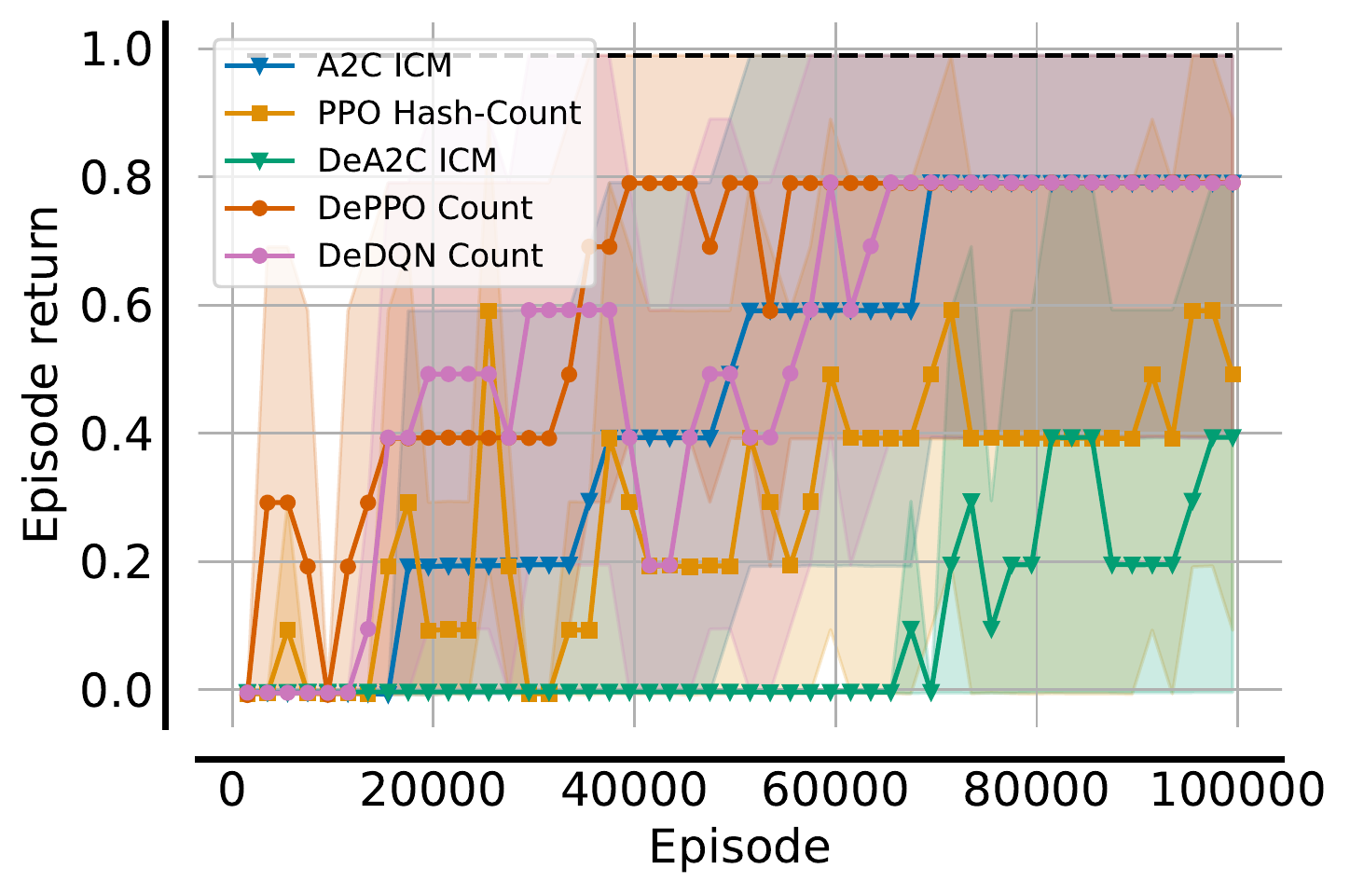}
        \caption{DeepSea $N=24$}
        \label{fig:deepsea_24_best_app}
    \end{subfigure}
    \hspace{5em}
    \begin{subfigure}{.33\textwidth}
        \centering
        \includegraphics[width=\linewidth]{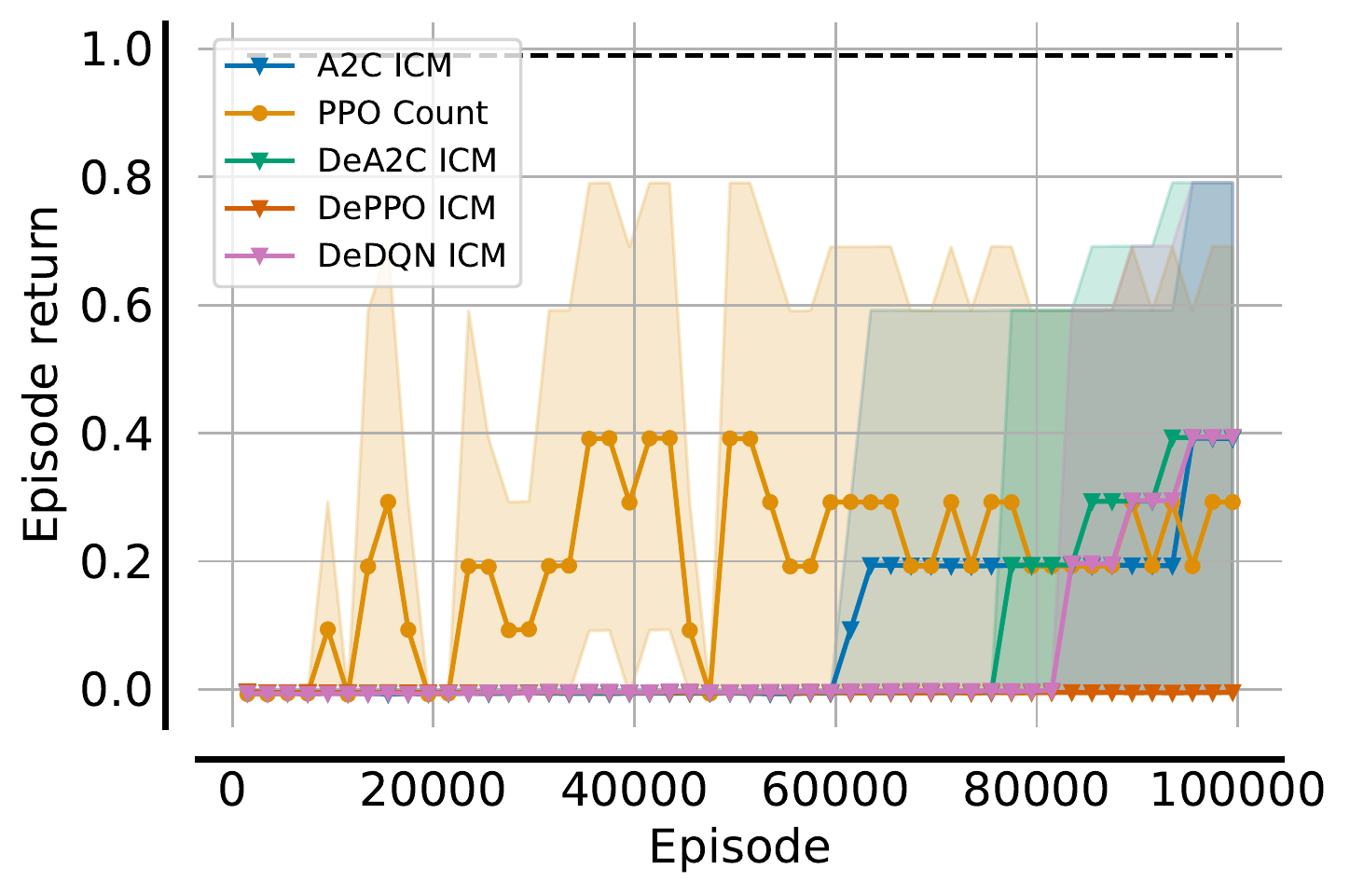}
        \caption{DeepSea $N=30$}
        \label{fig:deepsea_30_best_app}
    \end{subfigure}
    \caption{Average Evaluation returns for A2C, PPO and DeRL with the highest achieving intrinsic reward in all DeepSea tasks. Shading indicates 95\% confidence intervals.}
    \label{fig:deepsea_results_all_best}
\end{figure*}

\begin{figure*}[t]
    \centering
    \begin{subfigure}{.33\textwidth}
        \centering
        \includegraphics[width=\linewidth]{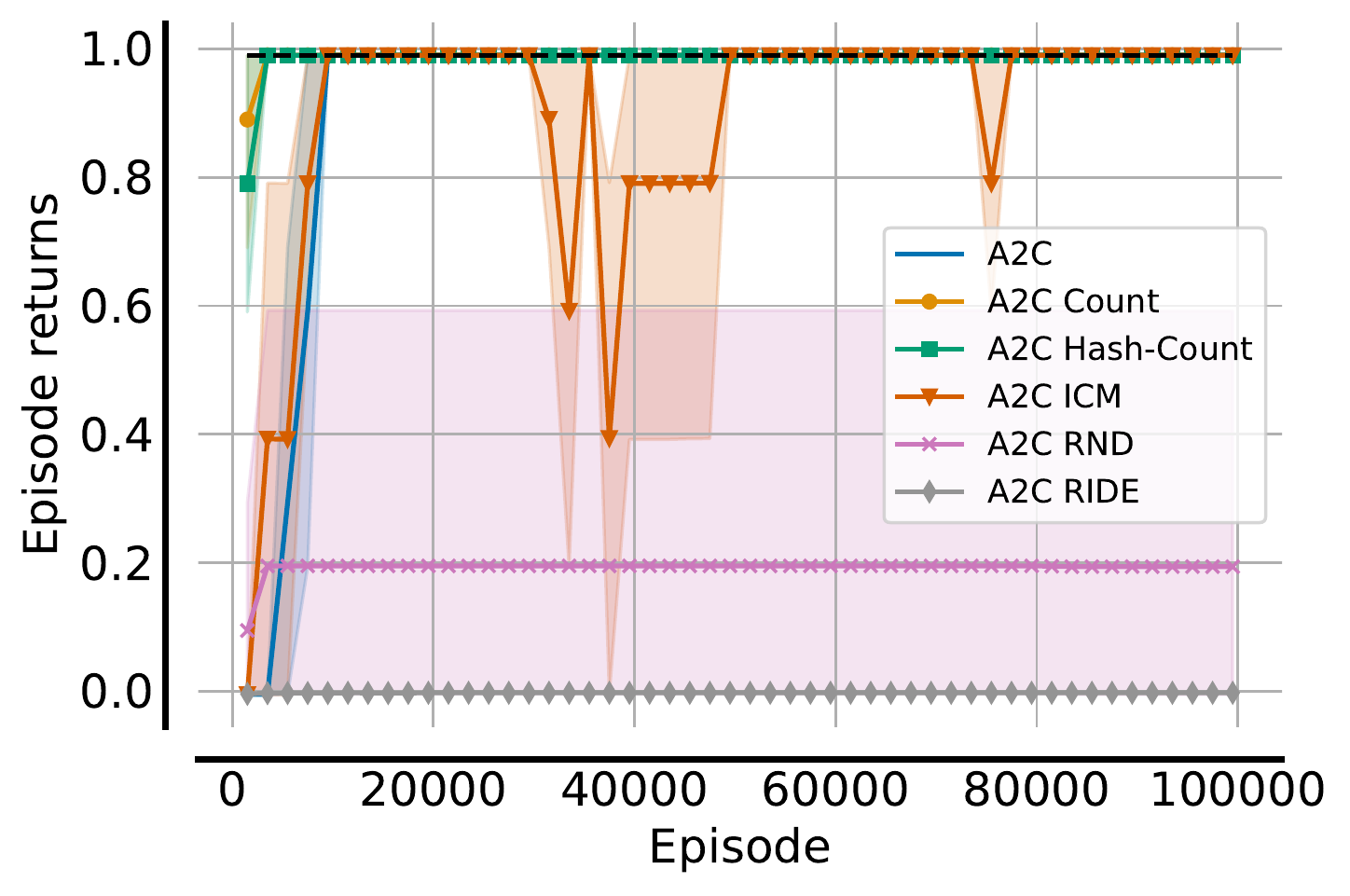}
        \caption{DeepSea $N=10$ A2C}
        \label{fig:deepsea_10_a2c_app}
    \end{subfigure}
    \hfill
    \begin{subfigure}{.33\textwidth}
        \centering
        \includegraphics[width=\linewidth]{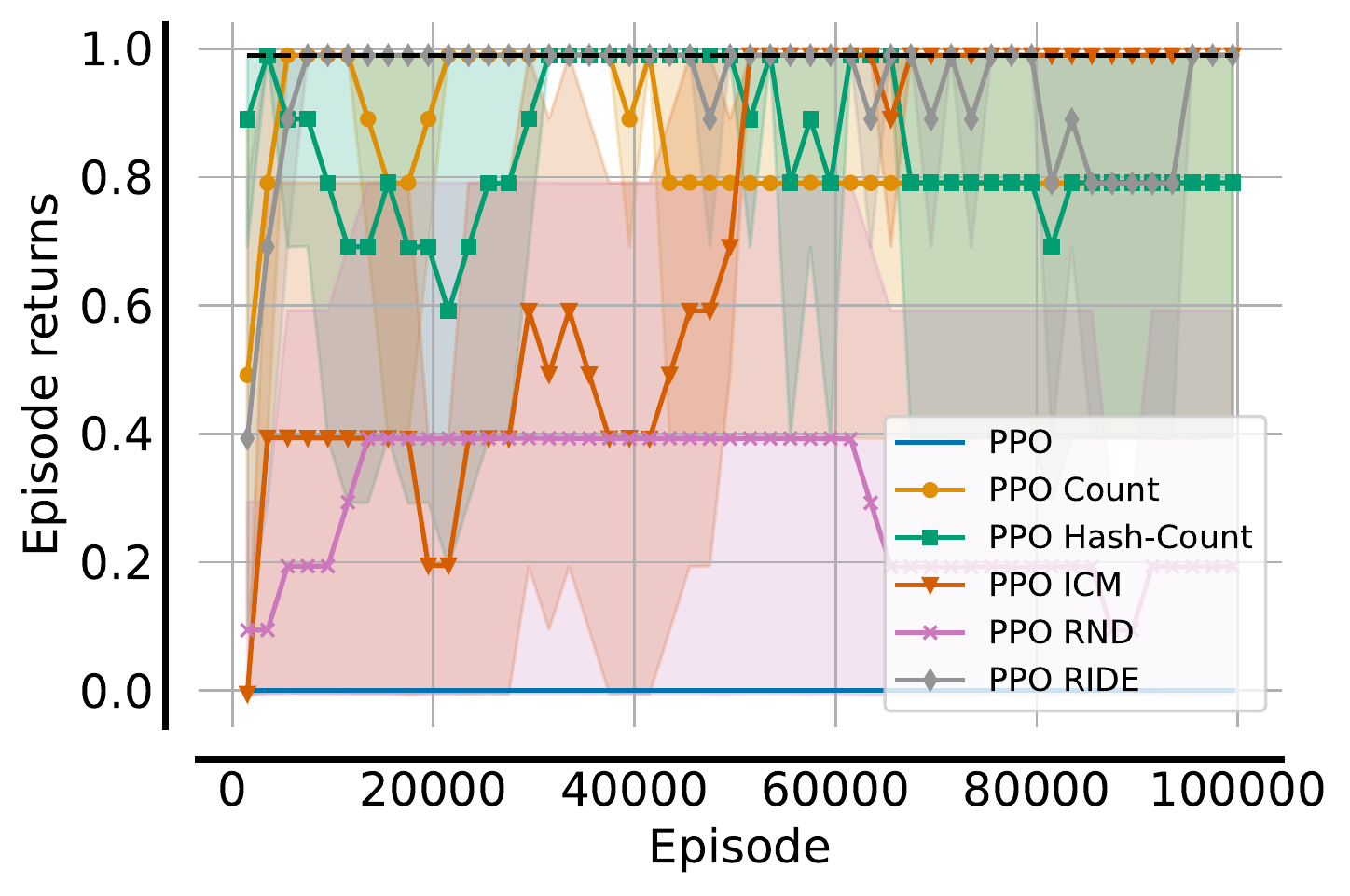}
        \caption{DeepSea $N=10$ PPO}
        \label{fig:deepsea_10_ppo_app}
    \end{subfigure}
    \hfill
    \begin{subfigure}{.33\textwidth}
        \centering
        \includegraphics[width=\linewidth]{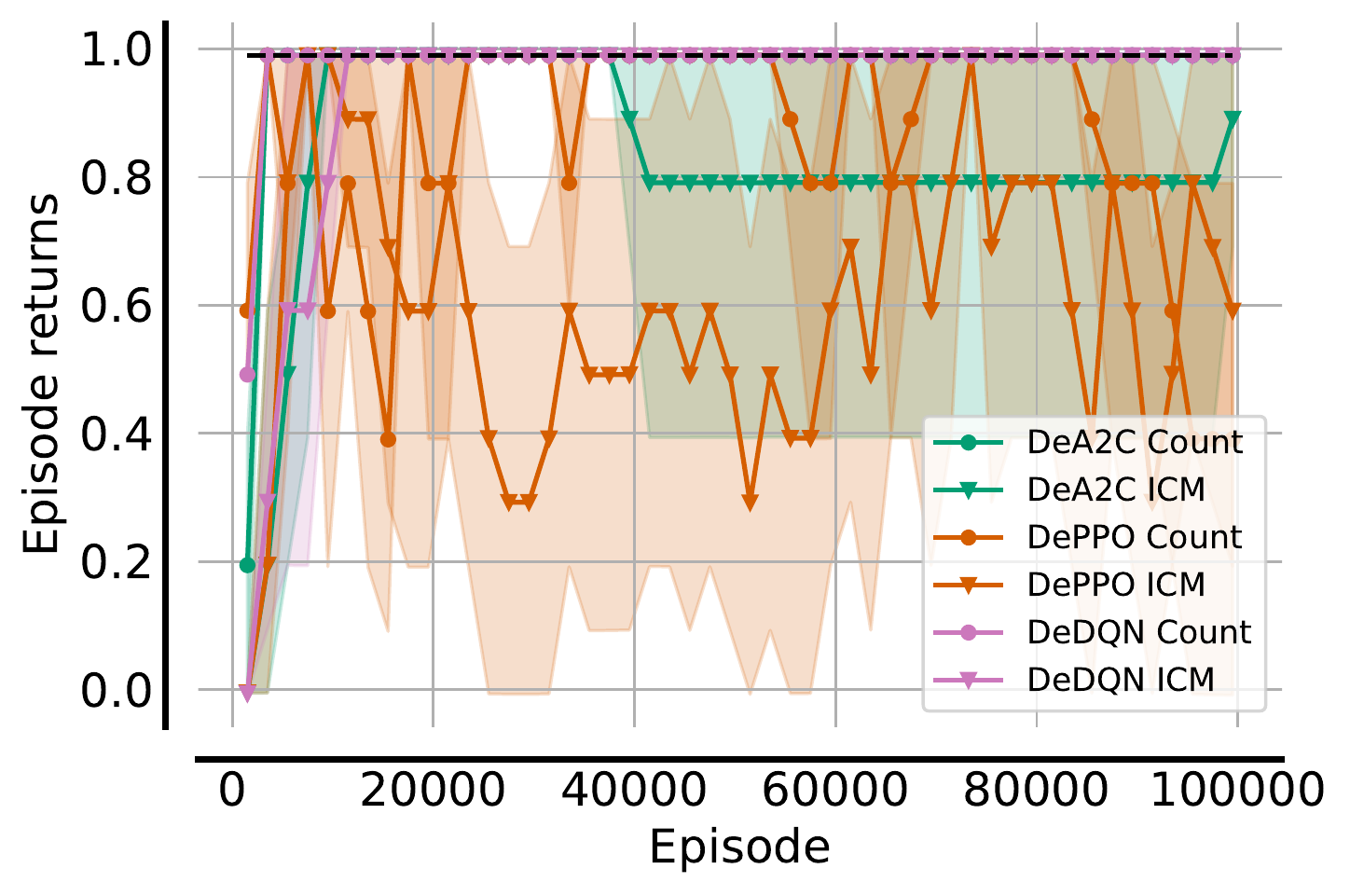}
        \caption{DeepSea $N=10$ DeRL}
        \label{fig:deepsea_10_derl_app}
    \end{subfigure}
    
    \begin{subfigure}{.33\textwidth}
        \centering
        \includegraphics[width=\linewidth]{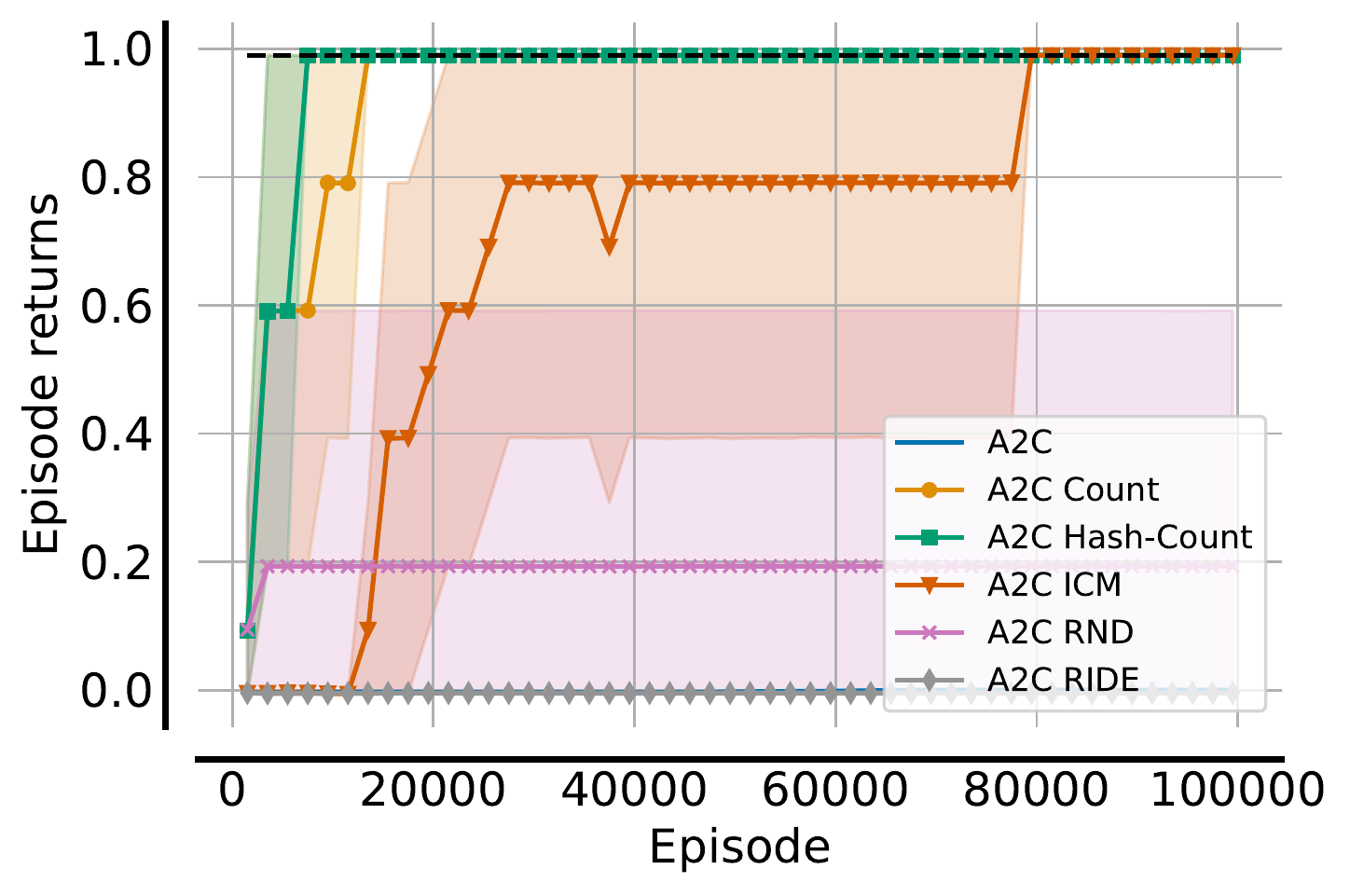}
        \caption{DeepSea $N=14$ A2C}
        \label{fig:deepsea_14_a2c_app}
    \end{subfigure}
    \hfill
    \begin{subfigure}{.33\textwidth}
        \centering
        \includegraphics[width=\linewidth]{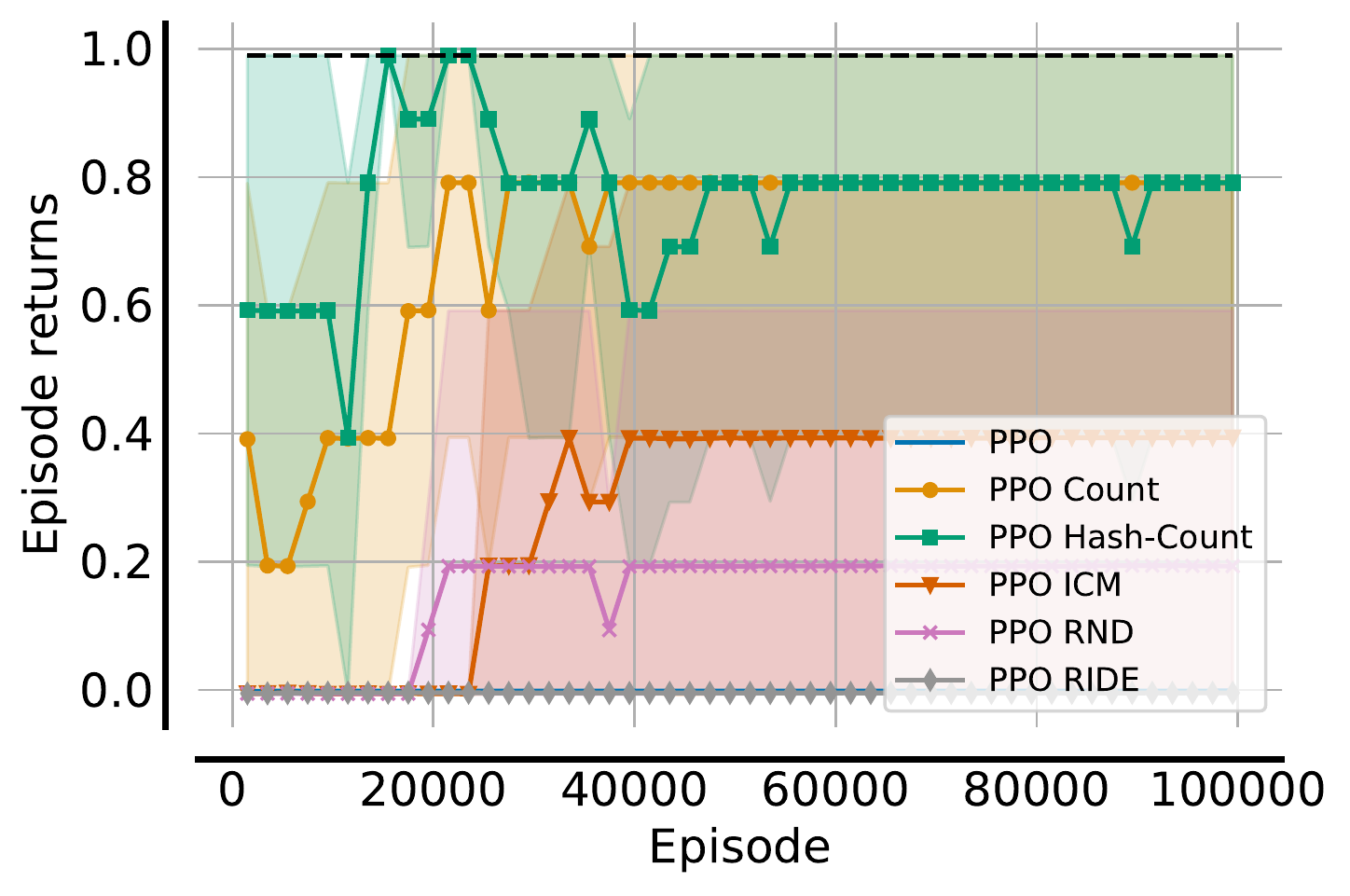}
        \caption{DeepSea $N=14$ PPO}
        \label{fig:deepsea_14_ppo_app}
    \end{subfigure}
    \hfill
    \begin{subfigure}{.33\textwidth}
        \centering
        \includegraphics[width=\linewidth]{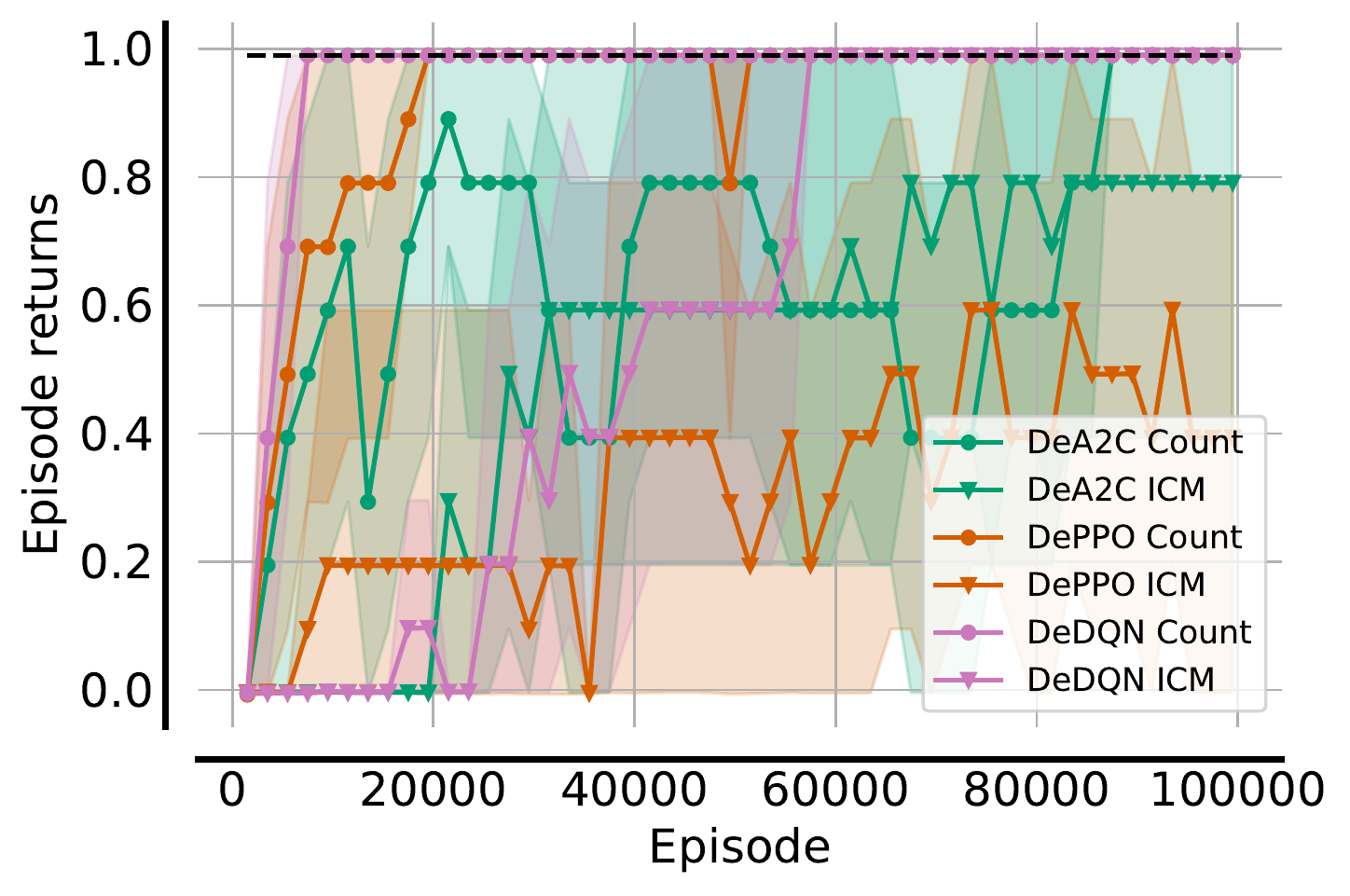}
        \caption{DeepSea $N=14$ DeRL}
        \label{fig:deepsea_14_derl_app}
    \end{subfigure}
    
    \begin{subfigure}{.33\textwidth}
        \centering
        \includegraphics[width=\linewidth]{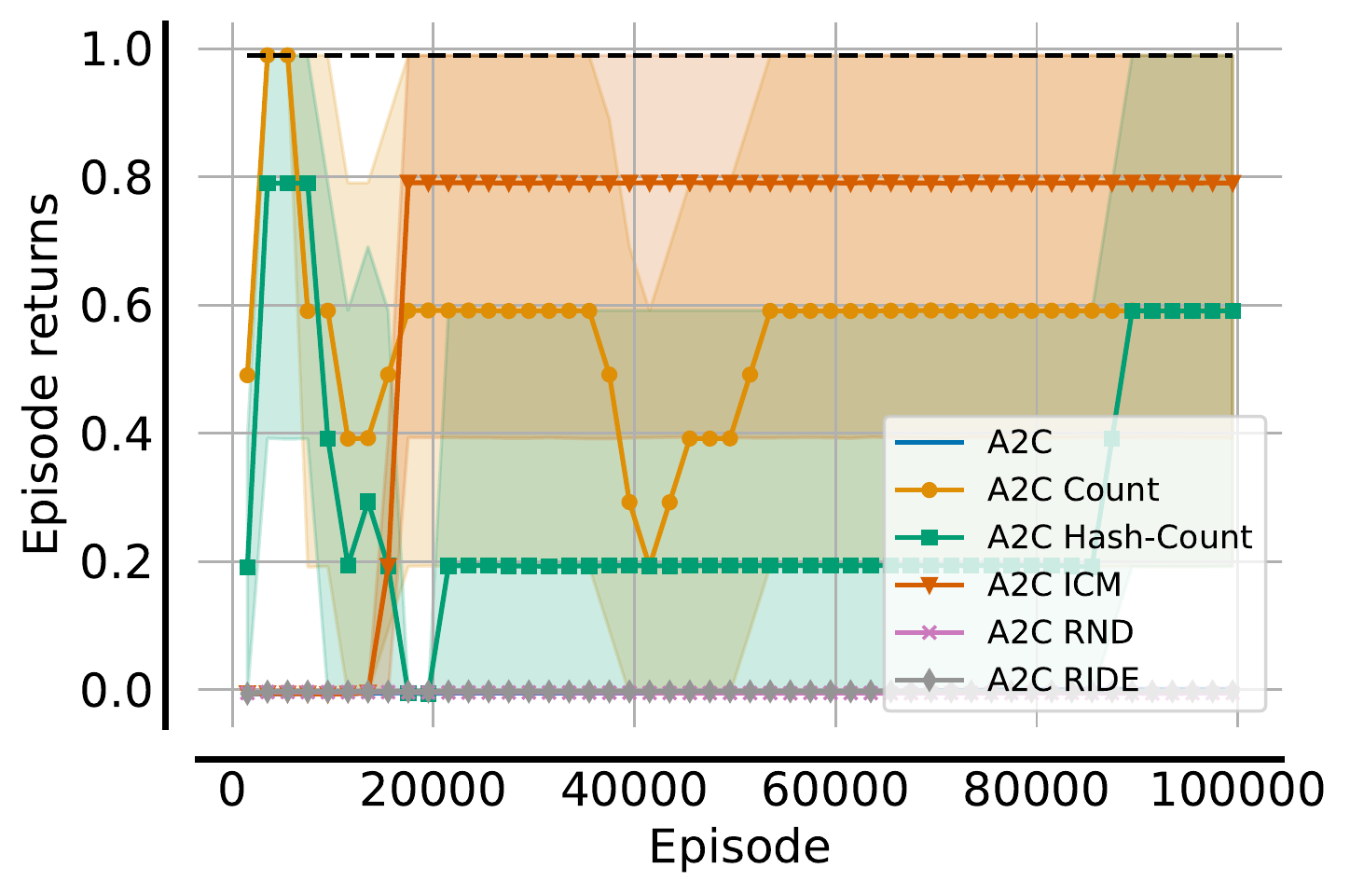}
        \caption{DeepSea $N=20$ A2C}
        \label{fig:deepsea_20_a2c_app}
    \end{subfigure}
    \hfill
    \begin{subfigure}{.33\textwidth}
        \centering
        \includegraphics[width=\linewidth]{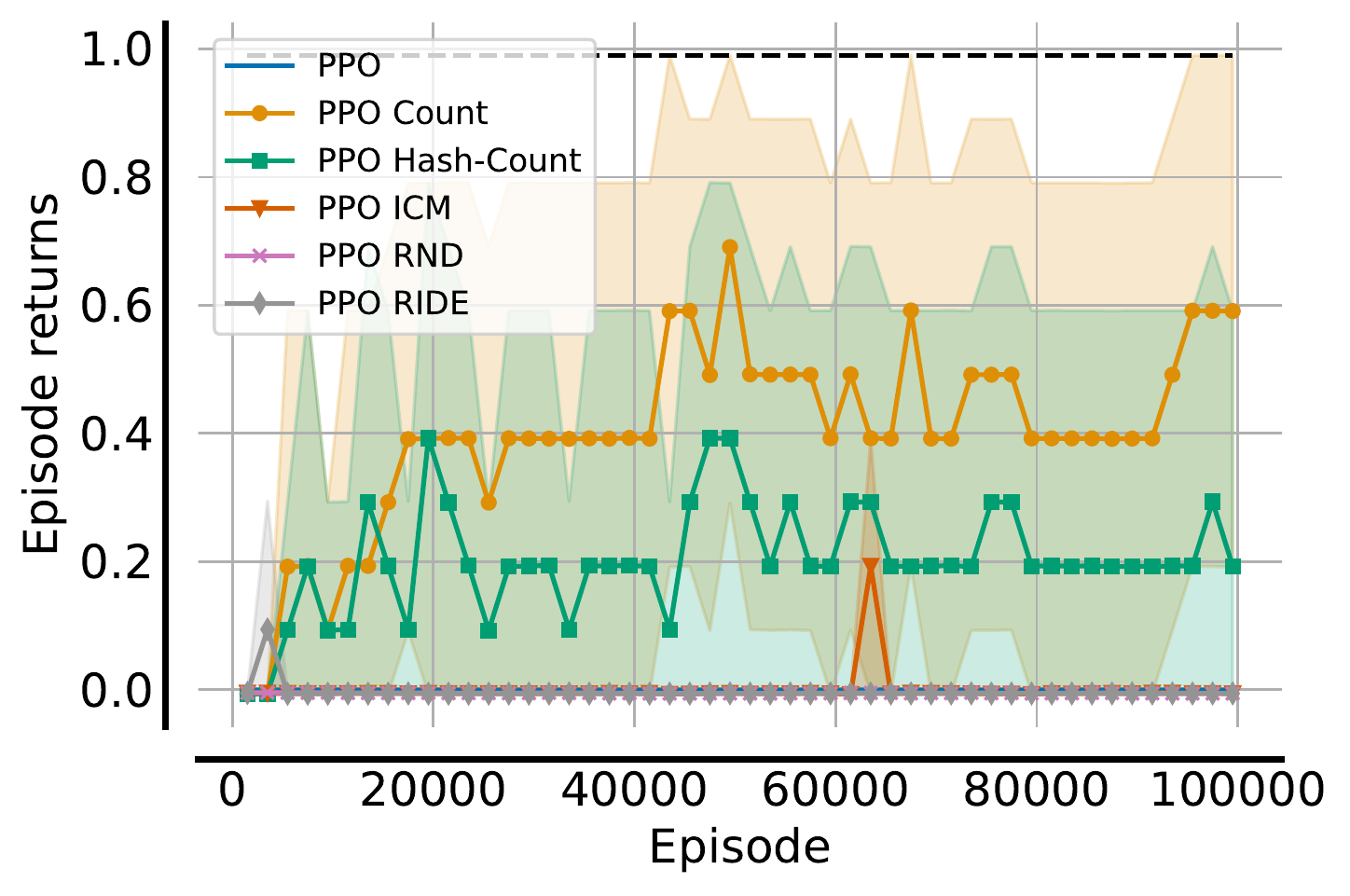}
        \caption{DeepSea $N=20$ PPO}
        \label{fig:deepsea_20_ppo_app}
    \end{subfigure}
    \hfill
    \begin{subfigure}{.33\textwidth}
        \centering
        \includegraphics[width=\linewidth]{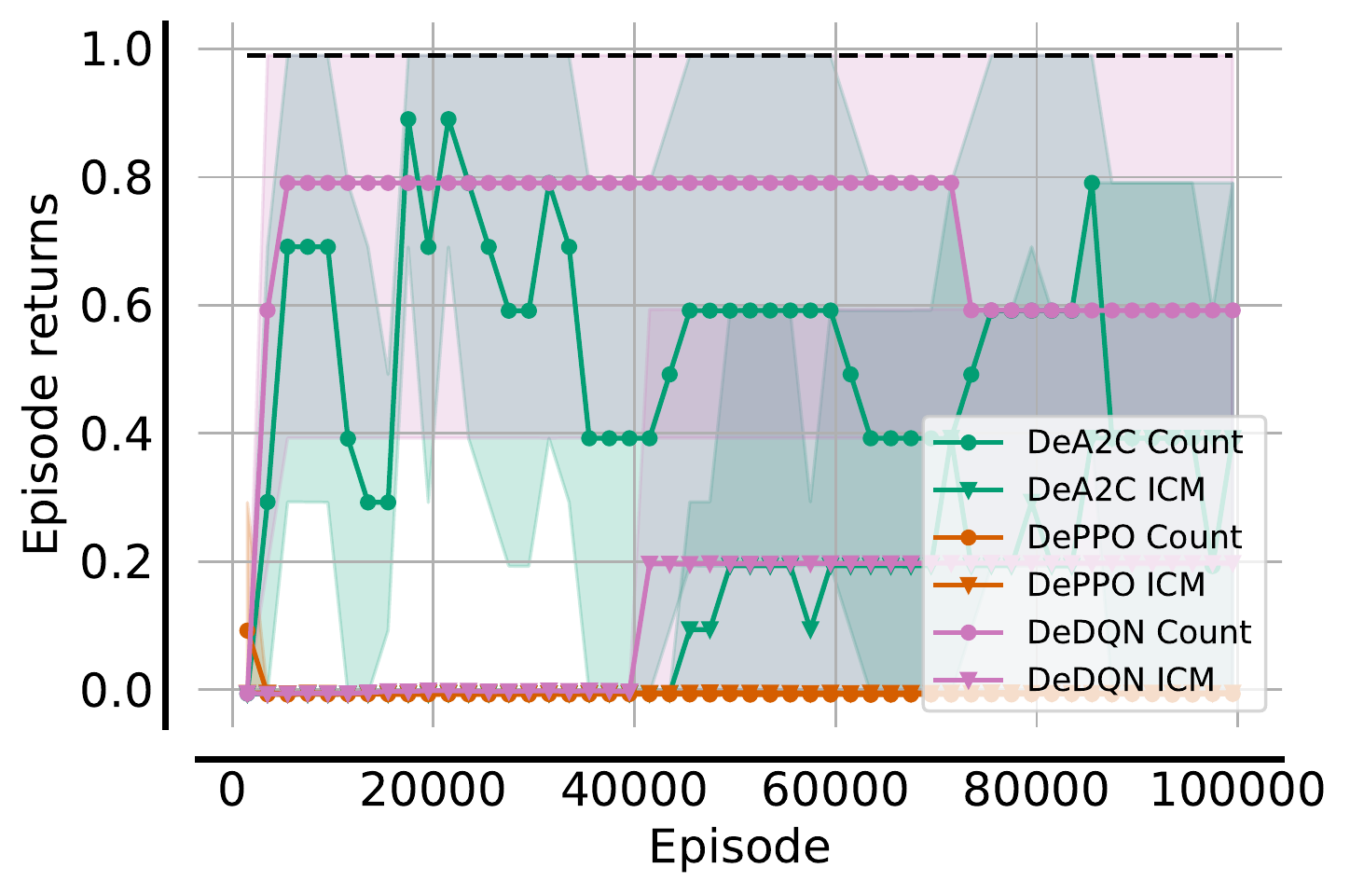}
        \caption{DeepSea $N=20$ DeRL}
        \label{fig:deepsea_20_derl_app}
    \end{subfigure}
    
    \begin{subfigure}{.33\textwidth}
        \centering
        \includegraphics[width=\linewidth]{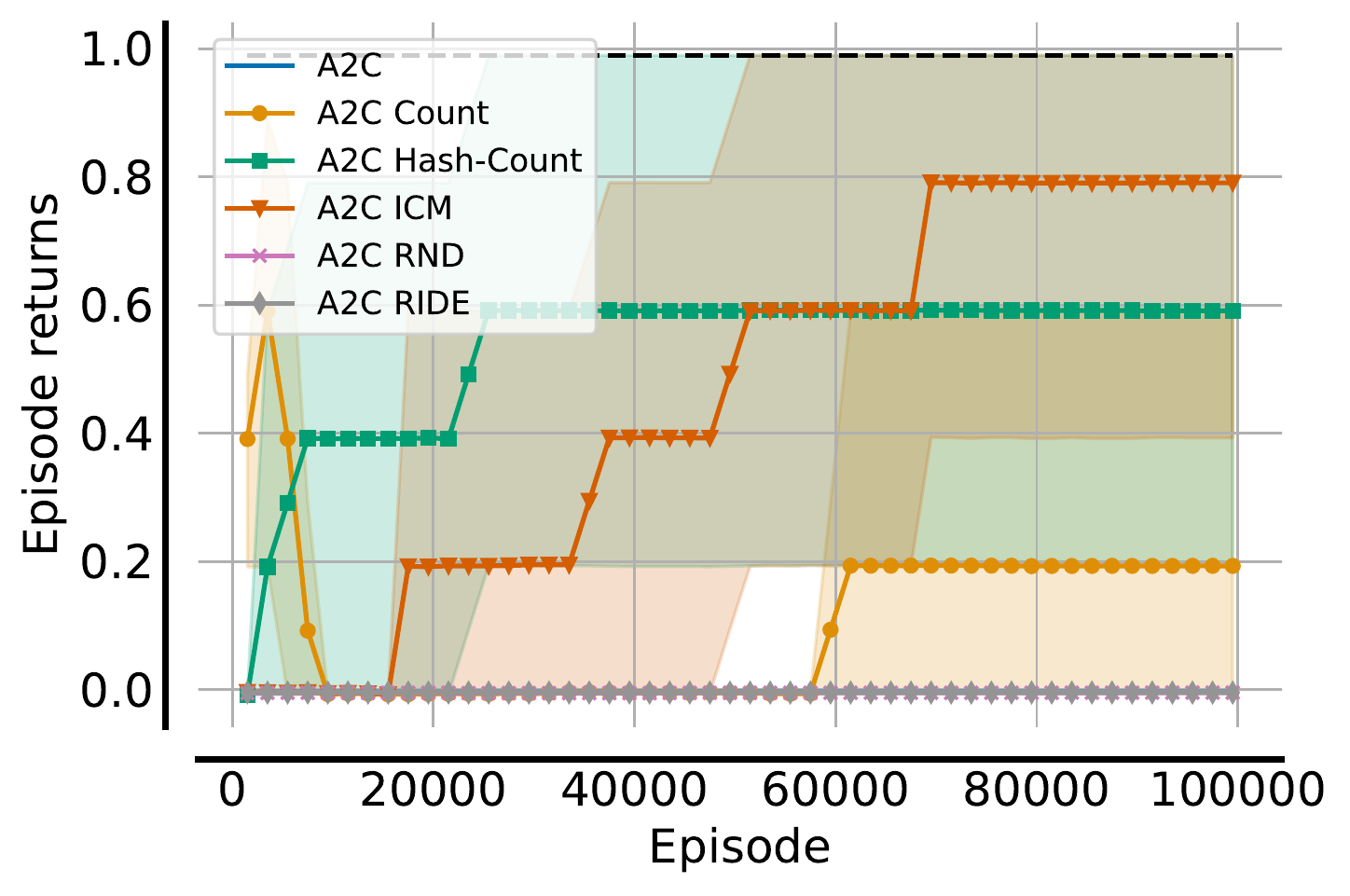}
        \caption{DeepSea $N=24$ A2C}
        \label{fig:deepsea_24_a2c_app}
    \end{subfigure}
    \hfill
    \begin{subfigure}{.33\textwidth}
        \centering
        \includegraphics[width=\linewidth]{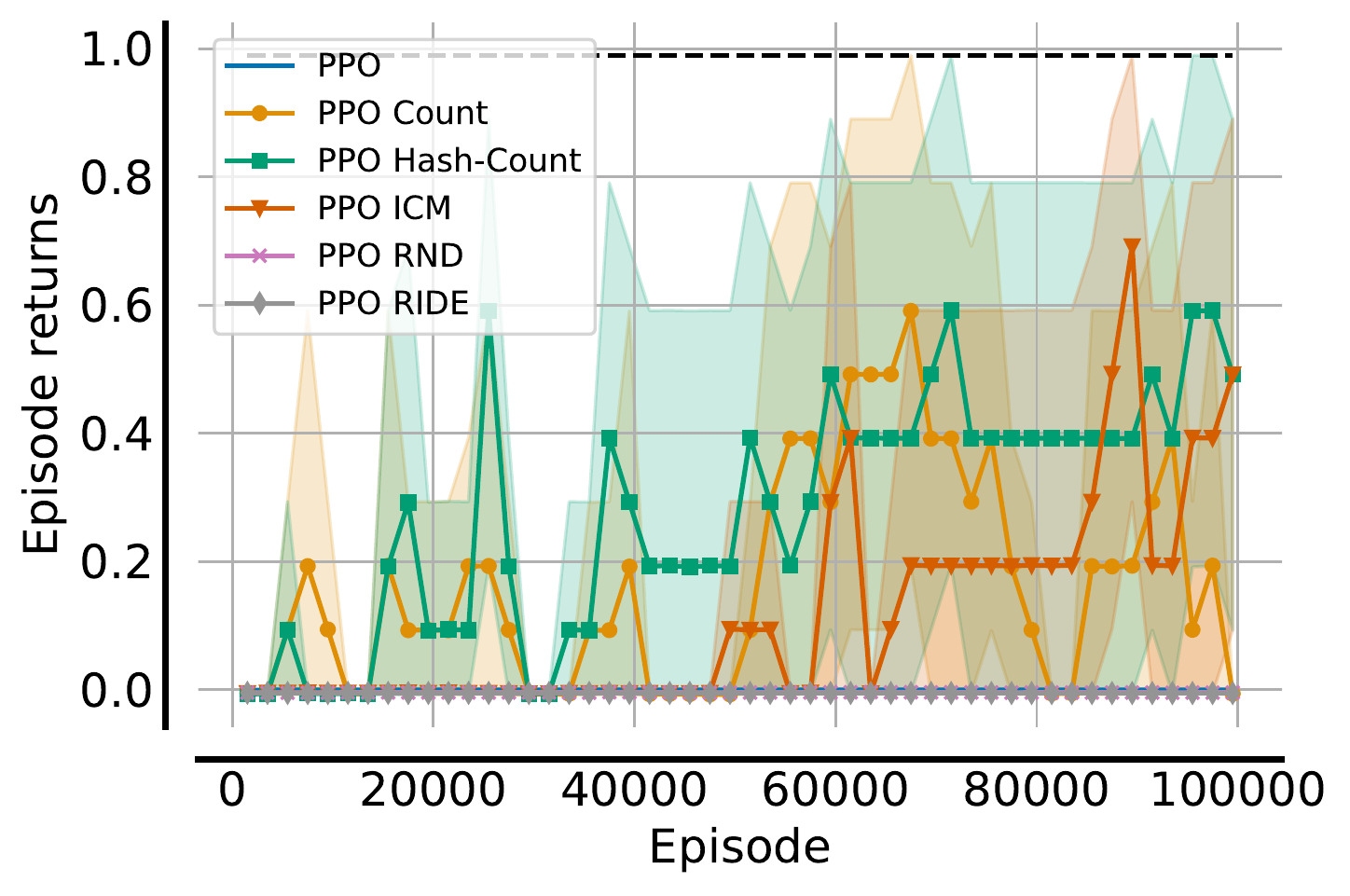}
        \caption{DeepSea $N=24$ PPO}
        \label{fig:deepsea_24_ppo_app}
    \end{subfigure}
    \hfill
    \begin{subfigure}{.33\textwidth}
        \centering
        \includegraphics[width=\linewidth]{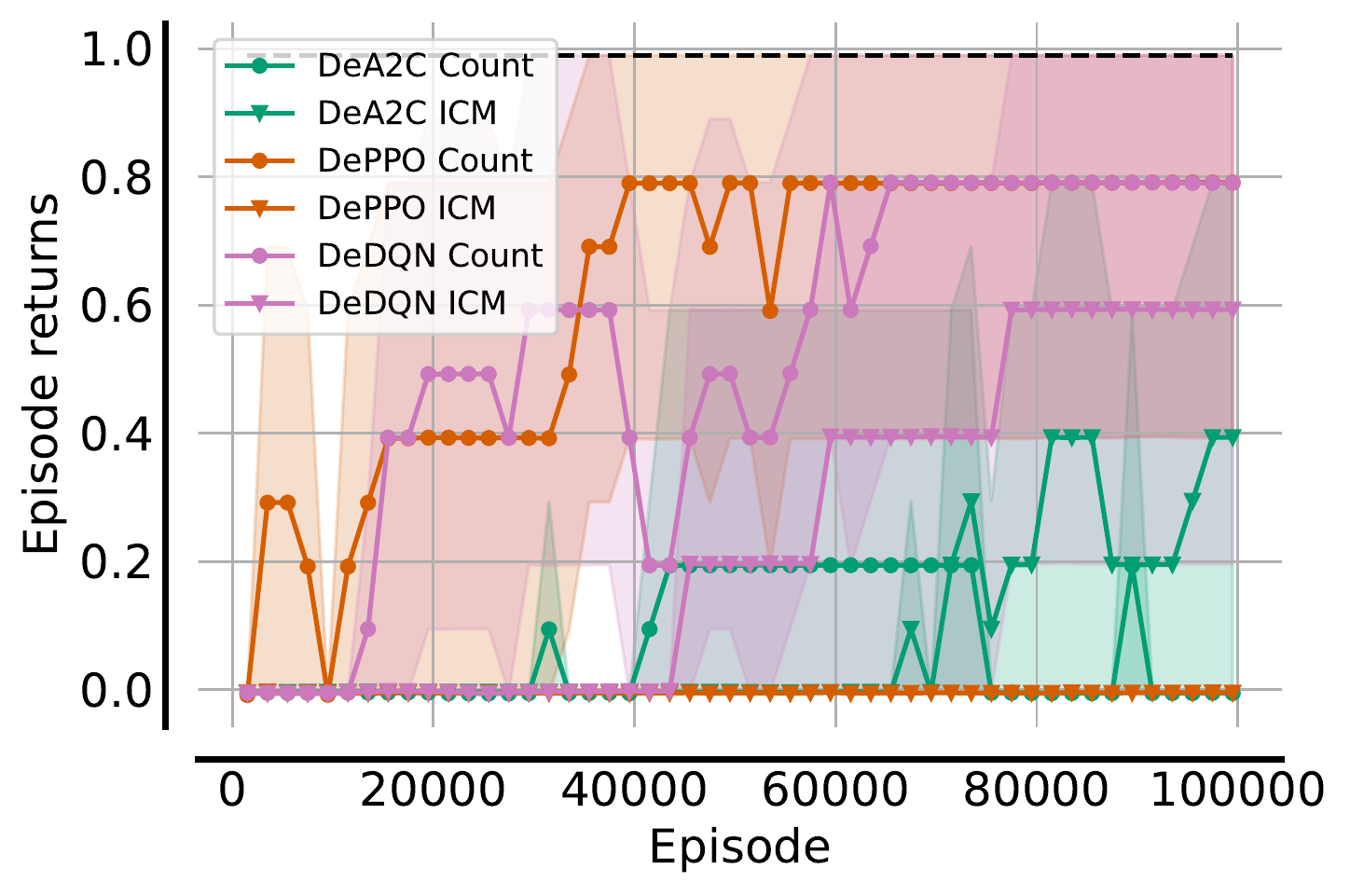}
        \caption{DeepSea $N=24$ DeRL}
        \label{fig:deepsea_24_derl_app}
    \end{subfigure}
    
    \begin{subfigure}{.33\textwidth}
        \centering
        \includegraphics[width=\linewidth]{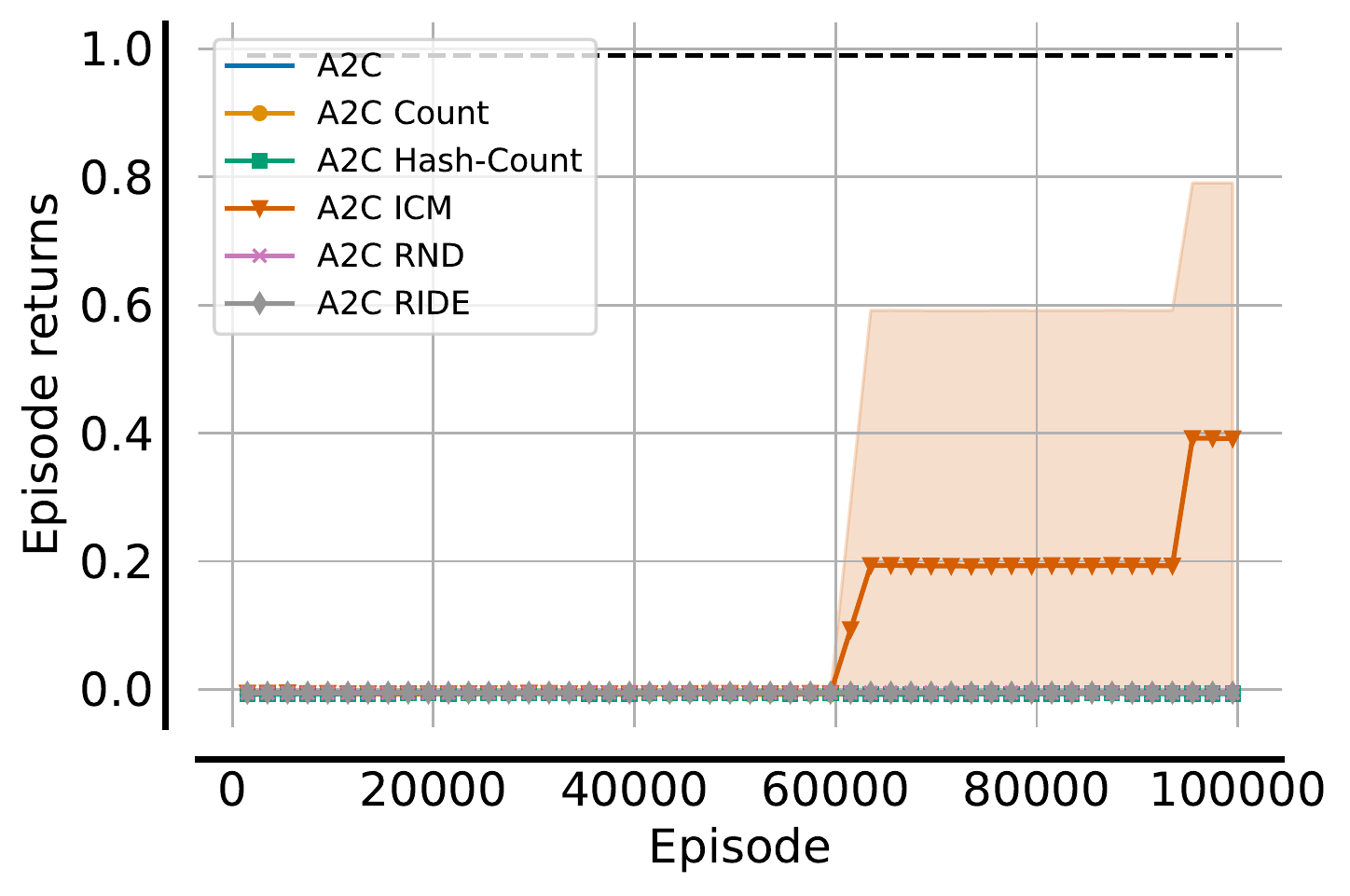}
        \caption{DeepSea $N=30$ A2C}
        \label{fig:deepsea_30_a2c_app}
    \end{subfigure}
    \hfill
    \begin{subfigure}{.33\textwidth}
        \centering
        \includegraphics[width=\linewidth]{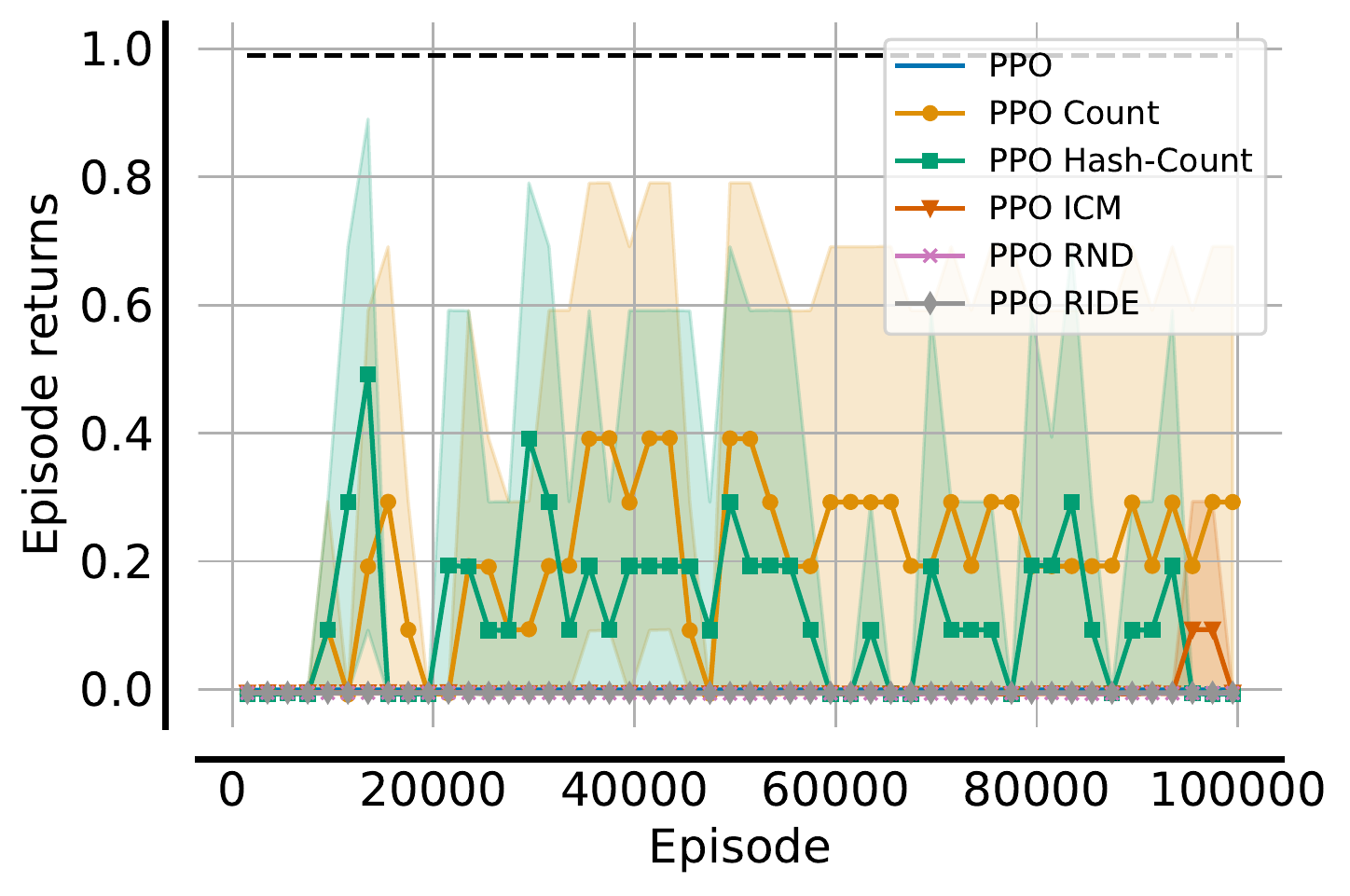}
        \caption{DeepSea $N=30$ PPO}
        \label{fig:deepsea_30_ppo_app}
    \end{subfigure}
    \hfill
    \begin{subfigure}{.33\textwidth}
        \centering
        \includegraphics[width=\linewidth]{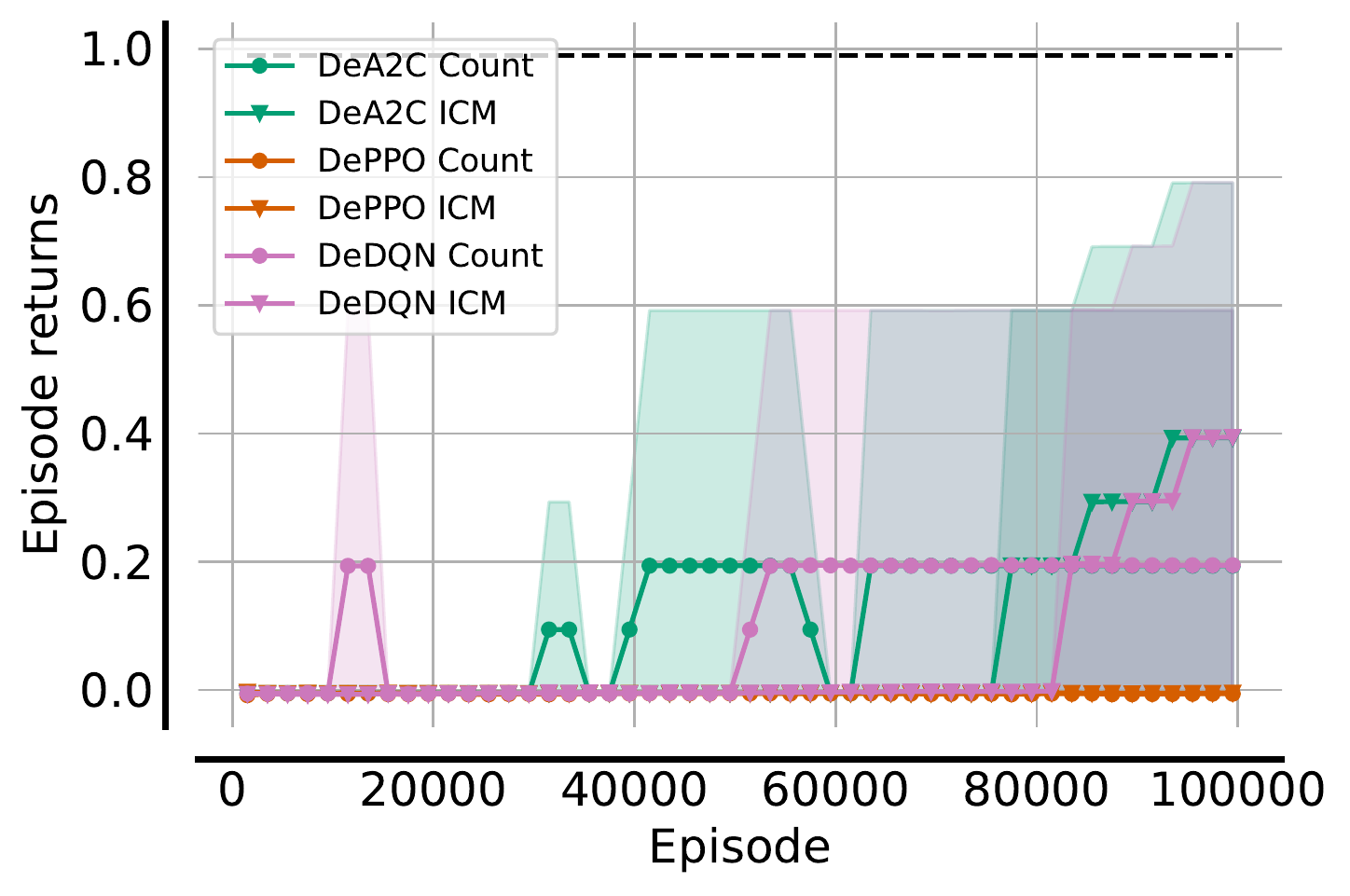}
        \caption{DeepSea $N=30$ DeRL}
        \label{fig:deepsea_30_derl_app}
    \end{subfigure}
    \caption{Average evaluation returns for A2C (first column), PPO (second column) and DeRL (third column) with all intrinsic rewards for all DeepSea tasks. Shading indicates 95\% confidence intervals.}
    \label{fig:deepsea_results_all}
\end{figure*}

\clearpage
\subsection{Hallway}

\begin{table*}[ht!]
	\centering
	\caption{Maximum evaluation returns in the Hallway environment with a single standard deviation.}
	\resizebox{\linewidth}{!}{
	\robustify\bf
	\begin{tabular}{l S S S S S S}
		\toprule
		{Algorithm \textbackslash \ Task} & {Hallway 10-0} & {Hallway 10-10} & {Hallway 20-0} & {Hallway 20-20} & {Hallway 30-0} & {Hallway 30-30} \\
		\midrule
		 A2C &  0.68(34) &  0.51(42) & \bf 0.42(34) & \bf 0.50(25) & \bf 0.44(22) & \bf 0.46(7) \\
		 A2C Count & \bf 0.85(0) & \bf 0.85(0) & \bf 0.70(0) & \bf 0.60(0) & \bf 0.52(6) &  0.14(24) \\
		 A2C Hash-Count & \bf 0.85(0) & \bf 0.85(0) & \bf 0.70(0) & \bf 0.60(0) & \bf 0.52(6) &  0.22(27) \\
		 A2C ICM &  0.68(34) &  0.68(34) & \bf 0.66(101) & \bf 1.08(76) & \bf 1.58(150) & \bf 1.09(115) \\
		 A2C RND &  0.12(35) &  -0.07(9) &  -0.17(15) &  -0.24(20) & \bf -0.24(28) &  -0.12(24) \\
		 A2C RIDE & \bf 0.85(0) & \bf 0.85(0) & \bf 0.70(0) & \bf 0.62(4) & \bf 0.55(0) & \bf 0.43(6) \\
		 PPO &  0.00(0) &  0.00(0) &  0.00(0) &  0.00(0) &  0.00(0) &  0.00(0) \\
		 PPO Count &  0.00(0) &  0.00(0) &  0.00(0) &  0.00(0) &  0.00(0) &  0.00(0) \\
		 PPO Hash-Count &  0.50(41) &  0.27(39) &  0.00(0) &  0.00(0) &  0.00(0) &  0.00(0) \\
		 PPO ICM &  0.48(39) &  0.40(41) &  0.28(34) & \bf 0.39(32) & \bf 0.08(16) &  0.10(20) \\
		 PPO RND &  0.13(37) &  0.47(44) &  0.02(31) &  0.17(40) &  -0.04(32) &  0.00(0) \\
		 PPO RIDE &  0.10(35) &  0.26(44) &  -0.03(37) &  0.13(41) &  -0.19(28) &  -0.01(40) \\
		 DeA2C Count & \bf 0.85(0) & \bf 0.85(0) & \bf 0.60(0) & \bf 0.70(0) & \bf 0.55(0) & \bf 0.43(6) \\
		 DeA2C ICM & \bf 0.85(0) & \bf 0.85(0) & \bf 1.08(76) & \bf 1.08(76) & \bf 1.38(168) & \bf 1.69(140) \\
		 DePPO Count & \bf 0.85(0) & \bf 0.85(0) & \bf 0.56(28) & \bf 0.60(0) & \bf 0.52(6) & \bf 0.43(6) \\
		 DePPO ICM & \bf 0.85(0) & \bf 0.85(0) & \bf 1.08(76) & \bf 0.62(4) & \bf 0.55(0) & \bf 0.43(6) \\
		 DeDQN Count &  0.13(37) &  0.11(37) &  0.00(0) &  0.00(0) &  0.00(0) &  0.00(0) \\
		 DeDQN ICM &  0.27(77) &  0.39(82) &  0.06(36) & \bf 0.35(112) &  0.00(0) &  0.00(0) \\
		\bottomrule
	\end{tabular}
	}
	\label{tab:hallway_max_returns}
\end{table*}

\begin{figure*}[ht!]
    \centering
    \begin{subfigure}{.33\textwidth}
        \centering
        \includegraphics[width=\linewidth]{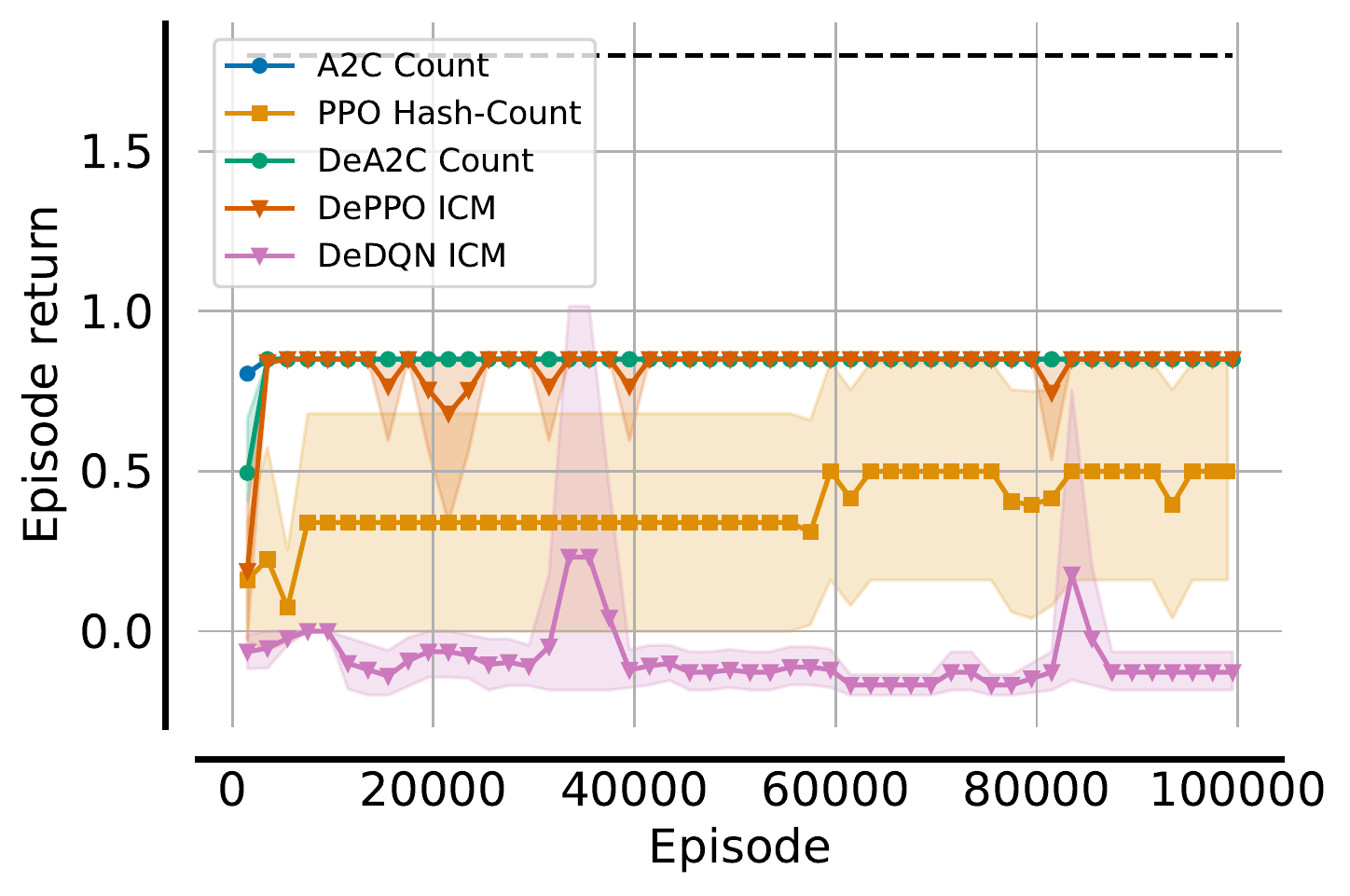}
        \caption{Hallway $N_l=10, N_r=0$}
        \label{fig:hallway_results_10_0_app}
    \end{subfigure}
    \hfill
    \begin{subfigure}{.33\textwidth}
        \centering
        \includegraphics[width=\linewidth]{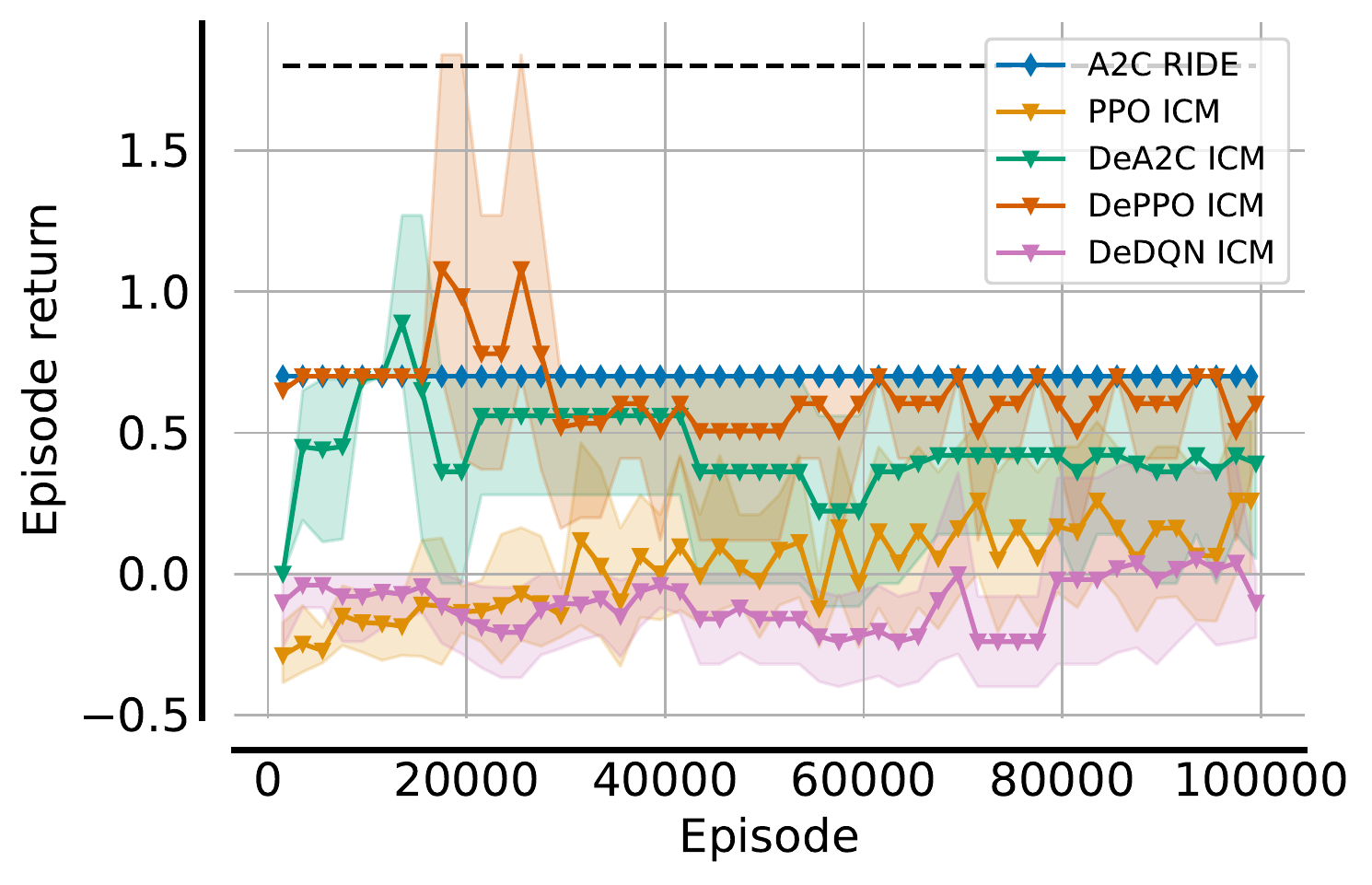}
        \caption{Hallway $N_l=20, N_r=0$}
        \label{fig:hallway_results_20_0_app}
    \end{subfigure}
    \hfill
    \begin{subfigure}{.33\textwidth}
        \centering
        \includegraphics[width=\linewidth]{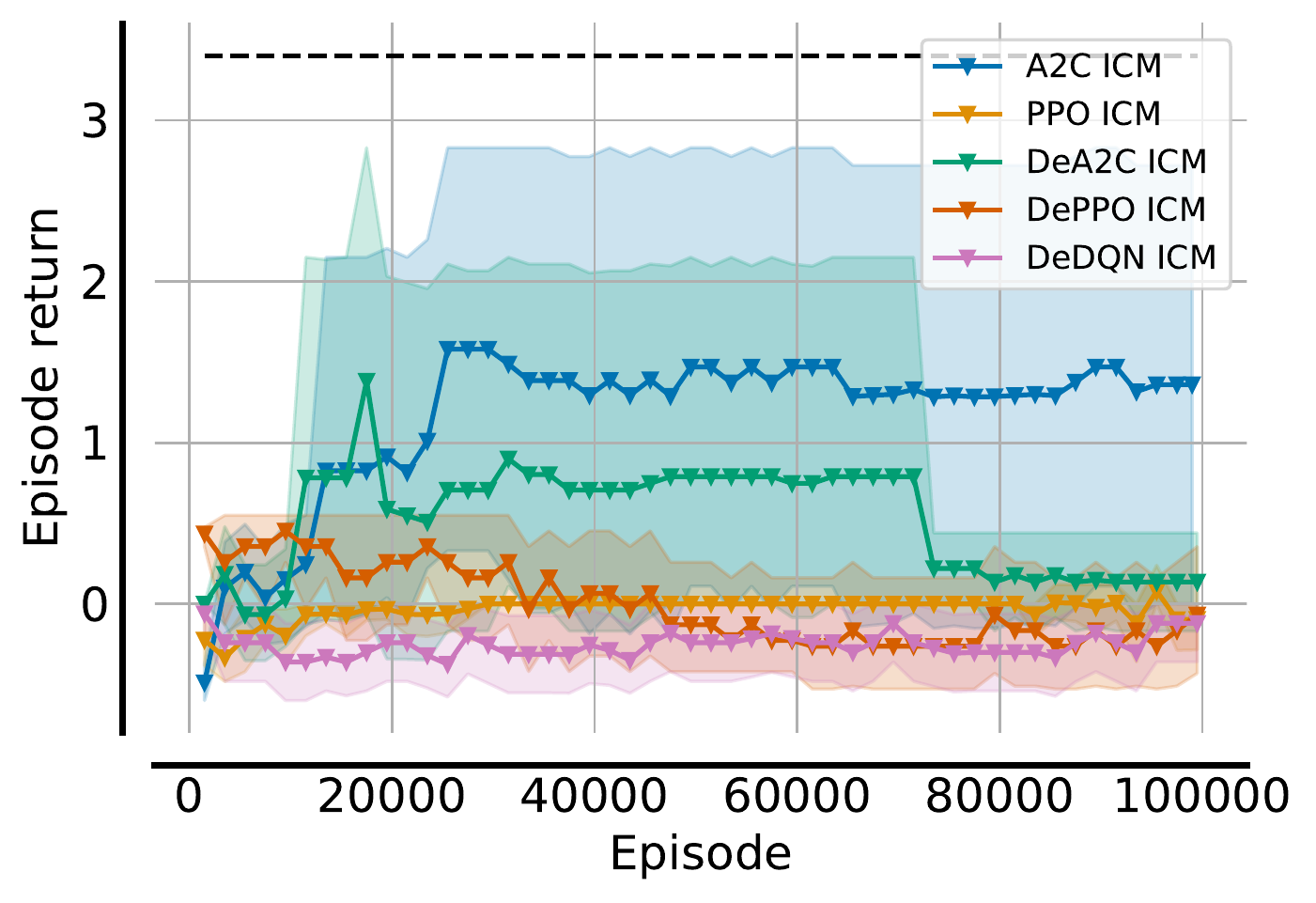}
        \caption{Hallway $N_l=30, N_r=0$}
        \label{fig:hallway_results_30_0_app}
    \end{subfigure}
    
    \begin{subfigure}{.33\textwidth}
        \centering
        \includegraphics[width=\linewidth]{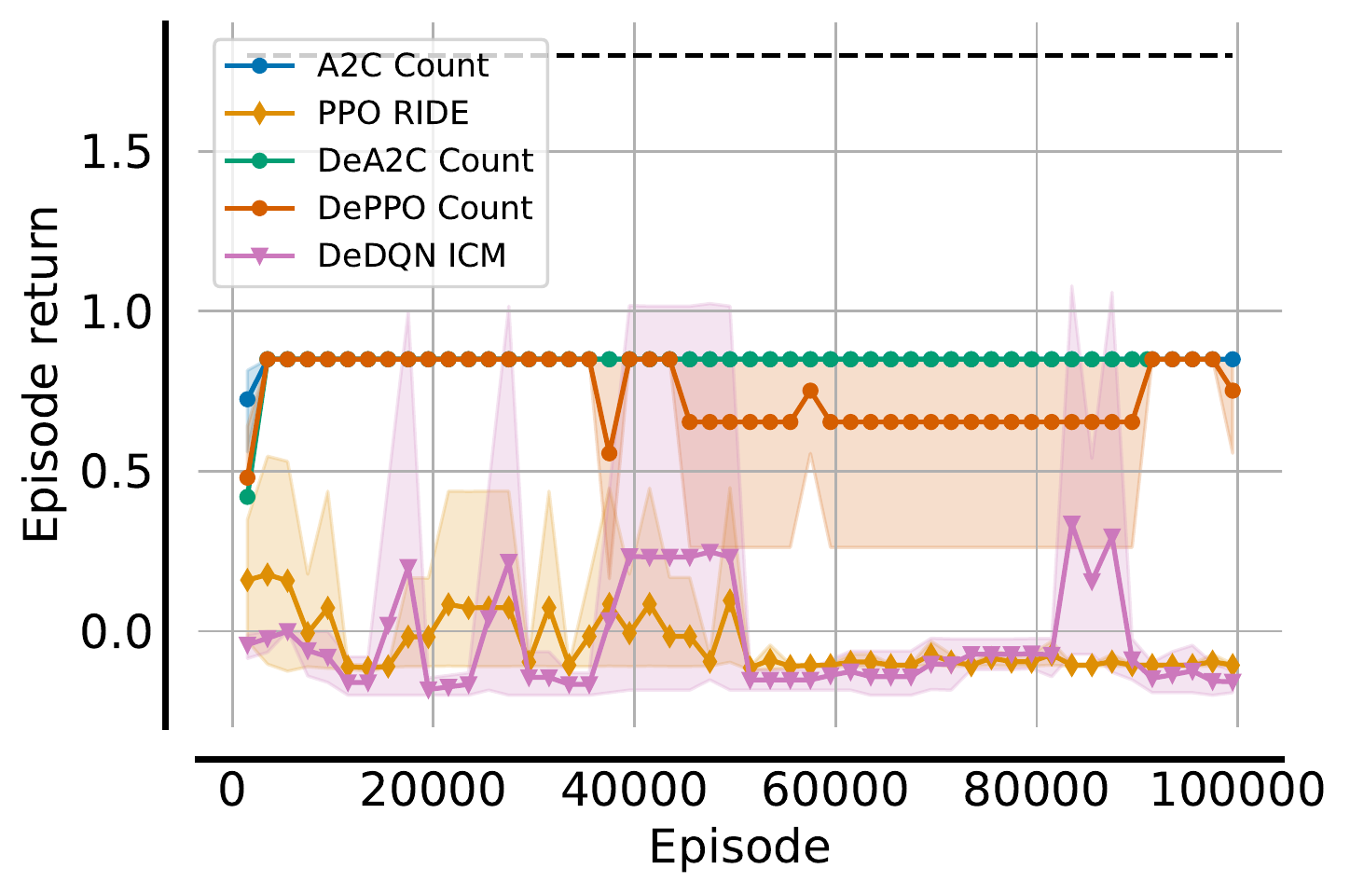}
        \caption{Hallway $N_l=10, N_r=10$}
        \label{fig:hallway_results_10_10_app}
    \end{subfigure}
    \hfill
    \begin{subfigure}{.33\textwidth}
        \centering
        \includegraphics[width=\linewidth]{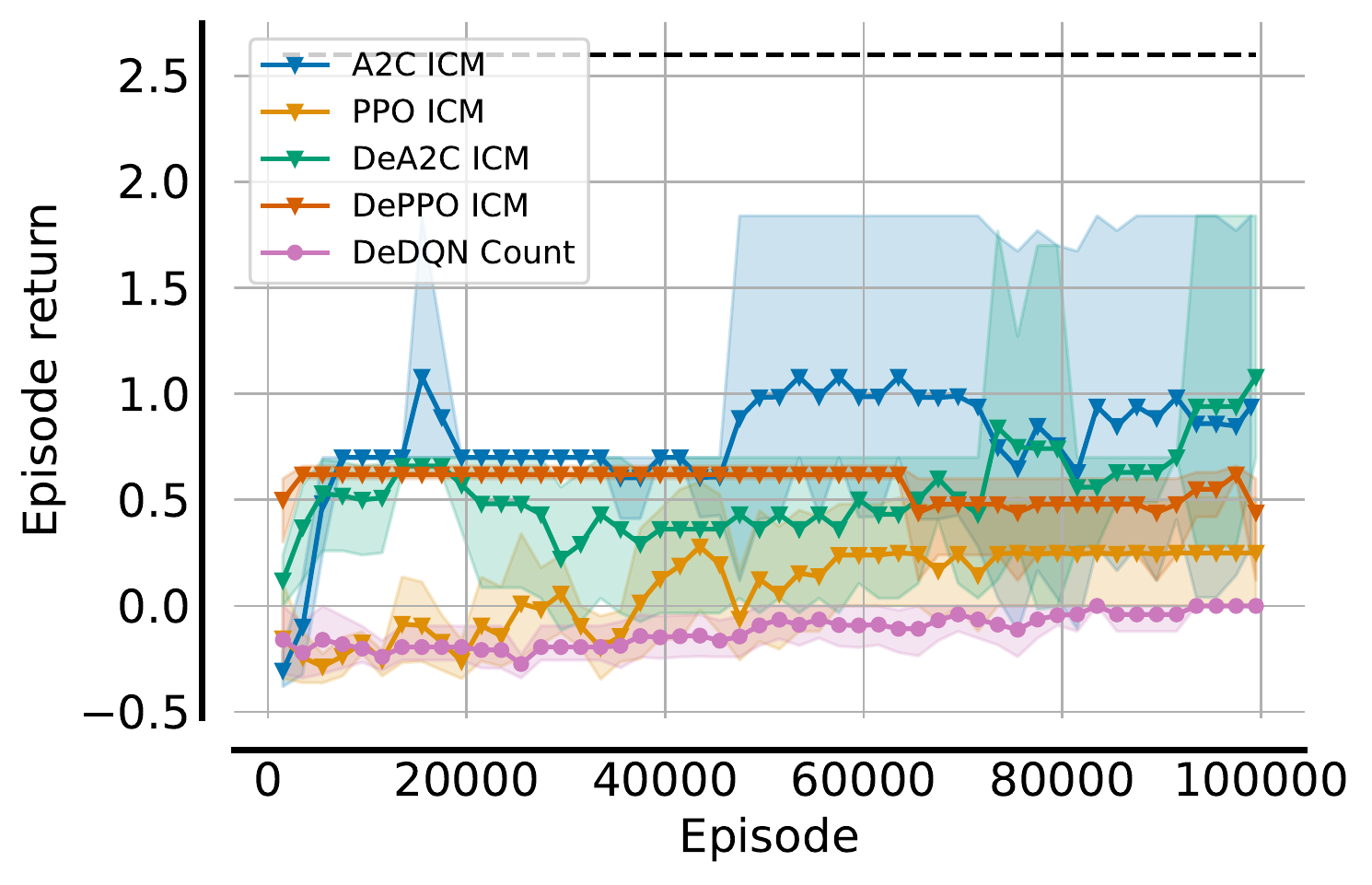}
        \caption{Hallway $N_l=20, N_r=20$}
        \label{fig:hallway_results_20_20_app}
    \end{subfigure}
    \hfill
    \begin{subfigure}{.33\textwidth}
        \centering
        \includegraphics[width=\linewidth]{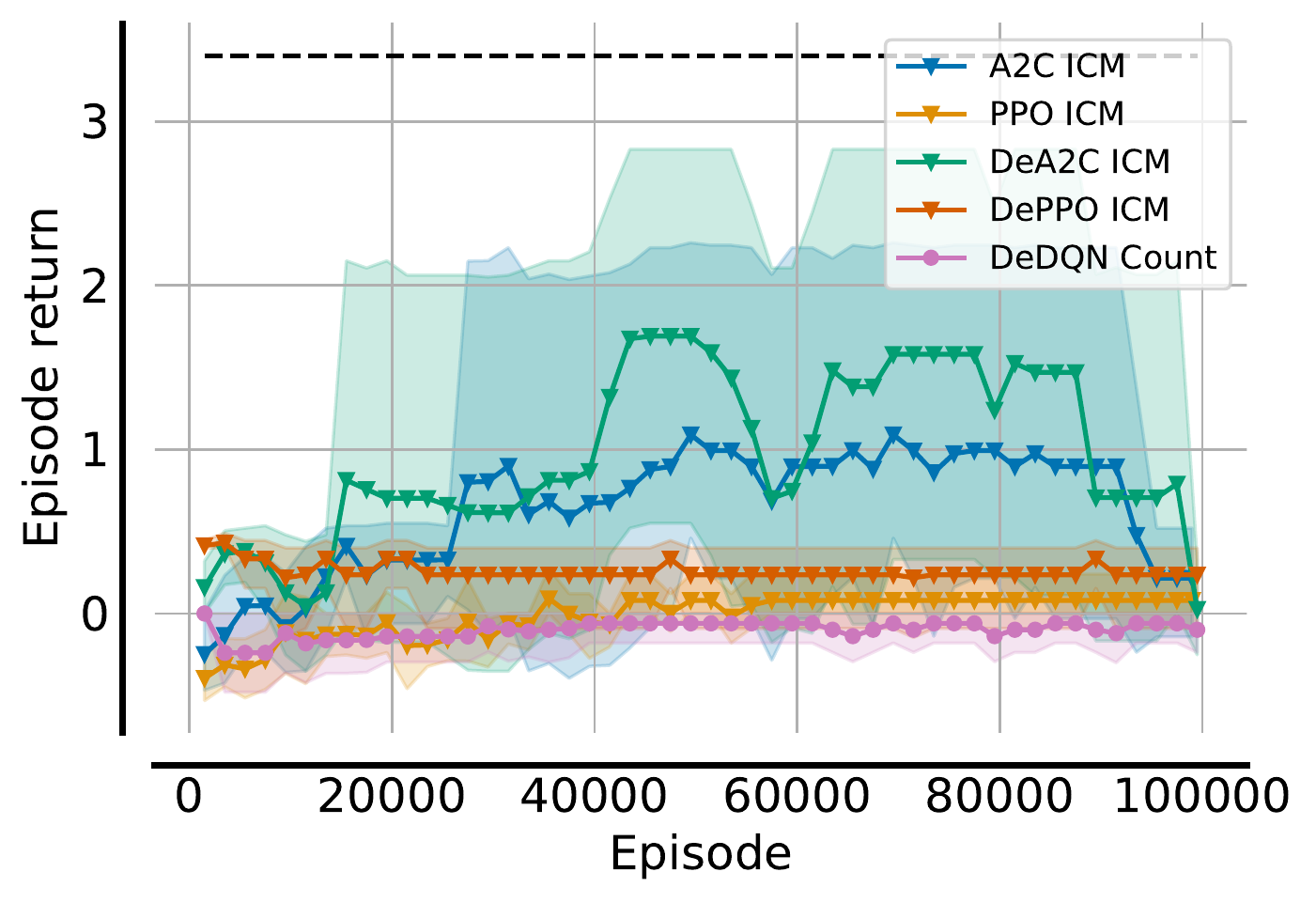}
        \caption{Hallway $N_l=30, N_r=30$}
        \label{fig:hallway_results_30_30_app}
    \end{subfigure}
    \caption{Average evaluation returns for A2C, PPO and DeRL with the highest achieving intrinsic reward in all Hallway tasks. Shading indicates 95\% confidence intervals.}
    \label{fig:hallway_results_all_best}
\end{figure*}

\begin{figure*}[t]
    \centering
    \begin{subfigure}{.33\textwidth}
        \centering
        \includegraphics[width=\linewidth]{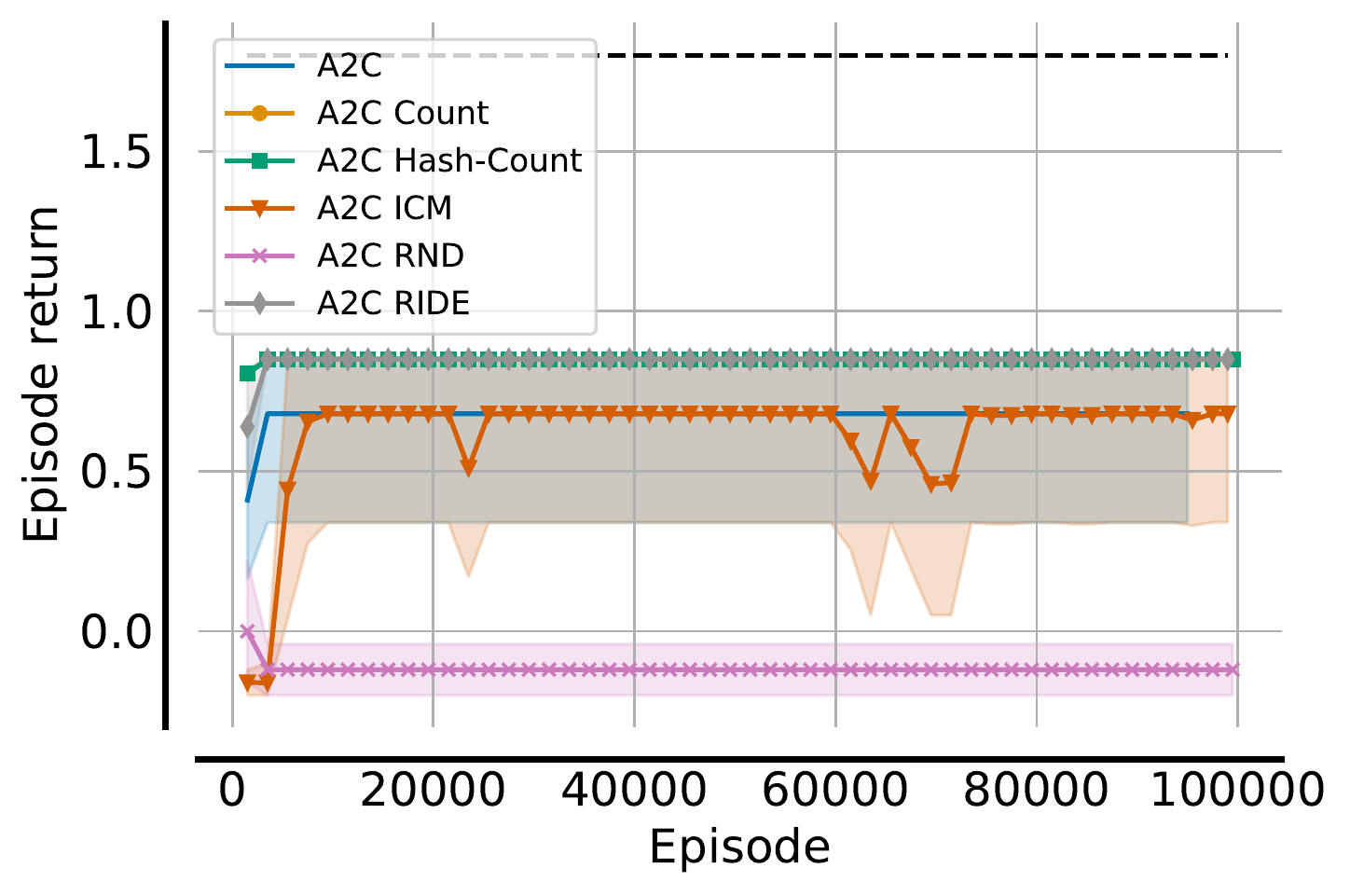}
        \caption{Hallway $N_l=10, N_r=0$ A2C}
        \label{fig:hallway_results_10_0_a2c_app}
    \end{subfigure}
    \hfill
    \begin{subfigure}{.33\textwidth}
        \centering
        \includegraphics[width=\linewidth]{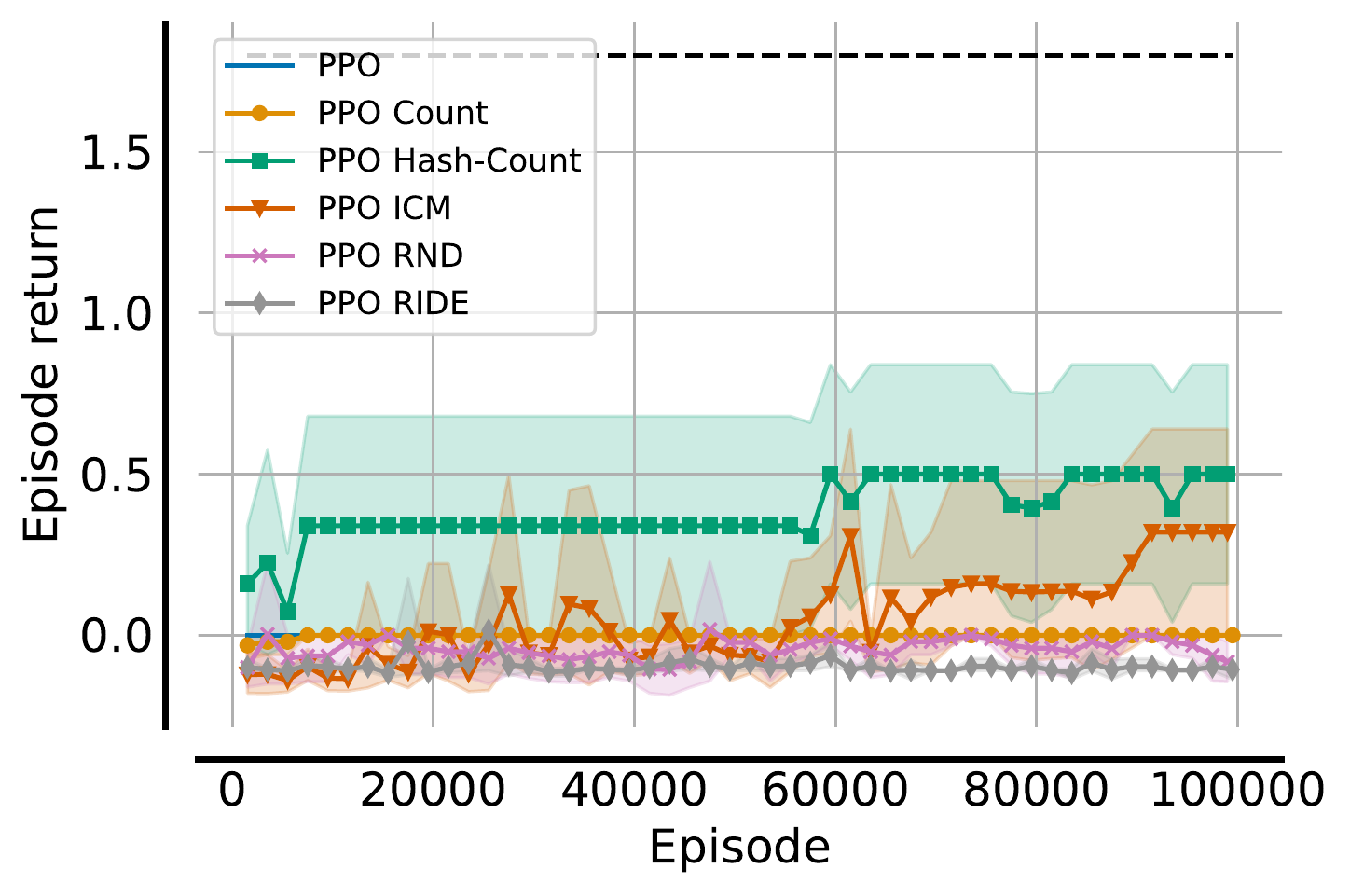}
        \caption{Hallway $N_l=10, N_r=0$ PPO}
        \label{fig:hallway_results_10_0_ppo_app}
    \end{subfigure}
    \hfill
    \begin{subfigure}{.33\textwidth}
        \centering
        \includegraphics[width=\linewidth]{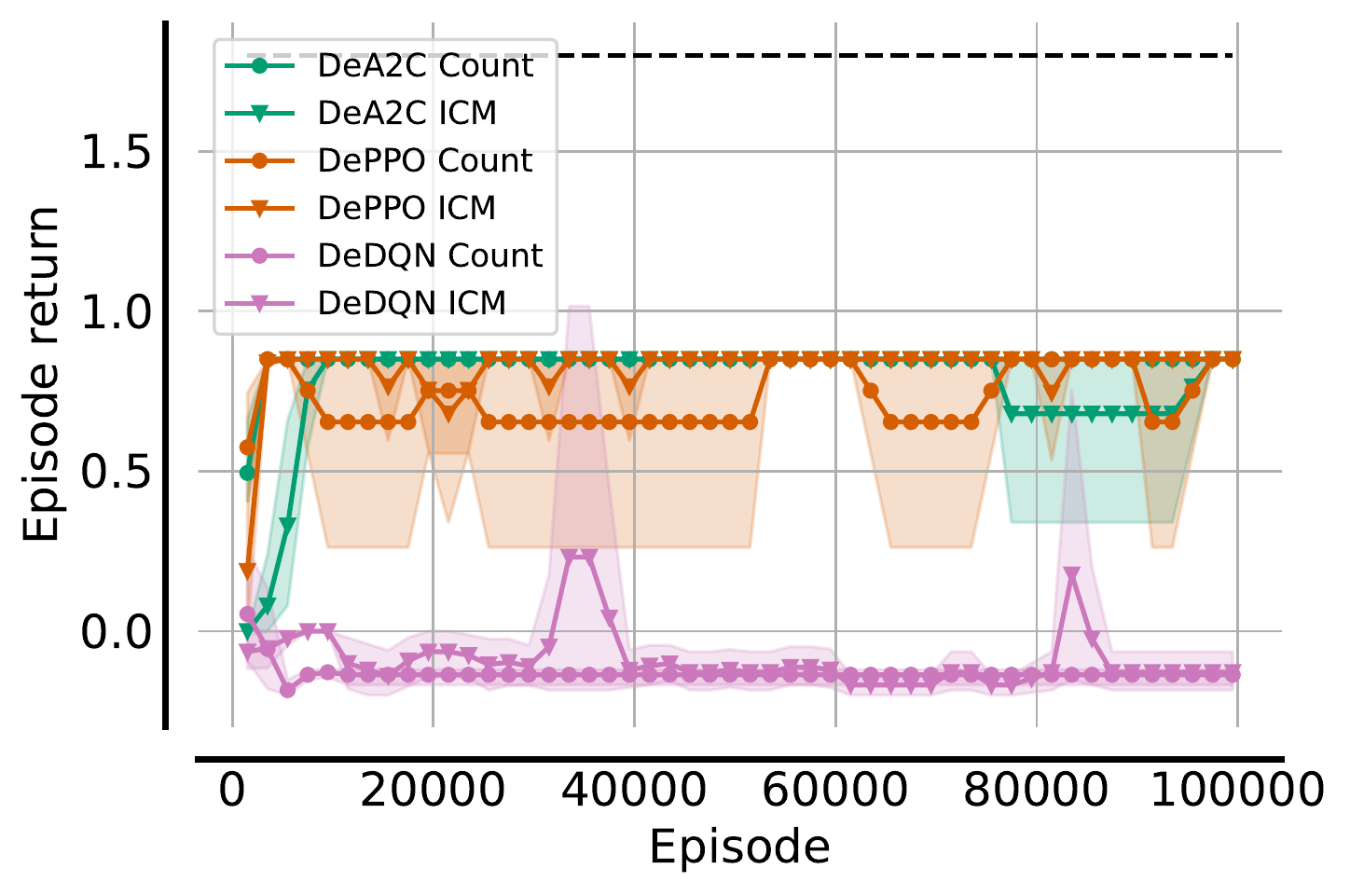}
        \caption{Hallway $N_l=10, N_r=0$ DeRL}
        \label{fig:hallway_results_10_0_derl_app}
    \end{subfigure}
    
    \begin{subfigure}{.33\textwidth}
        \centering
        \includegraphics[width=\linewidth]{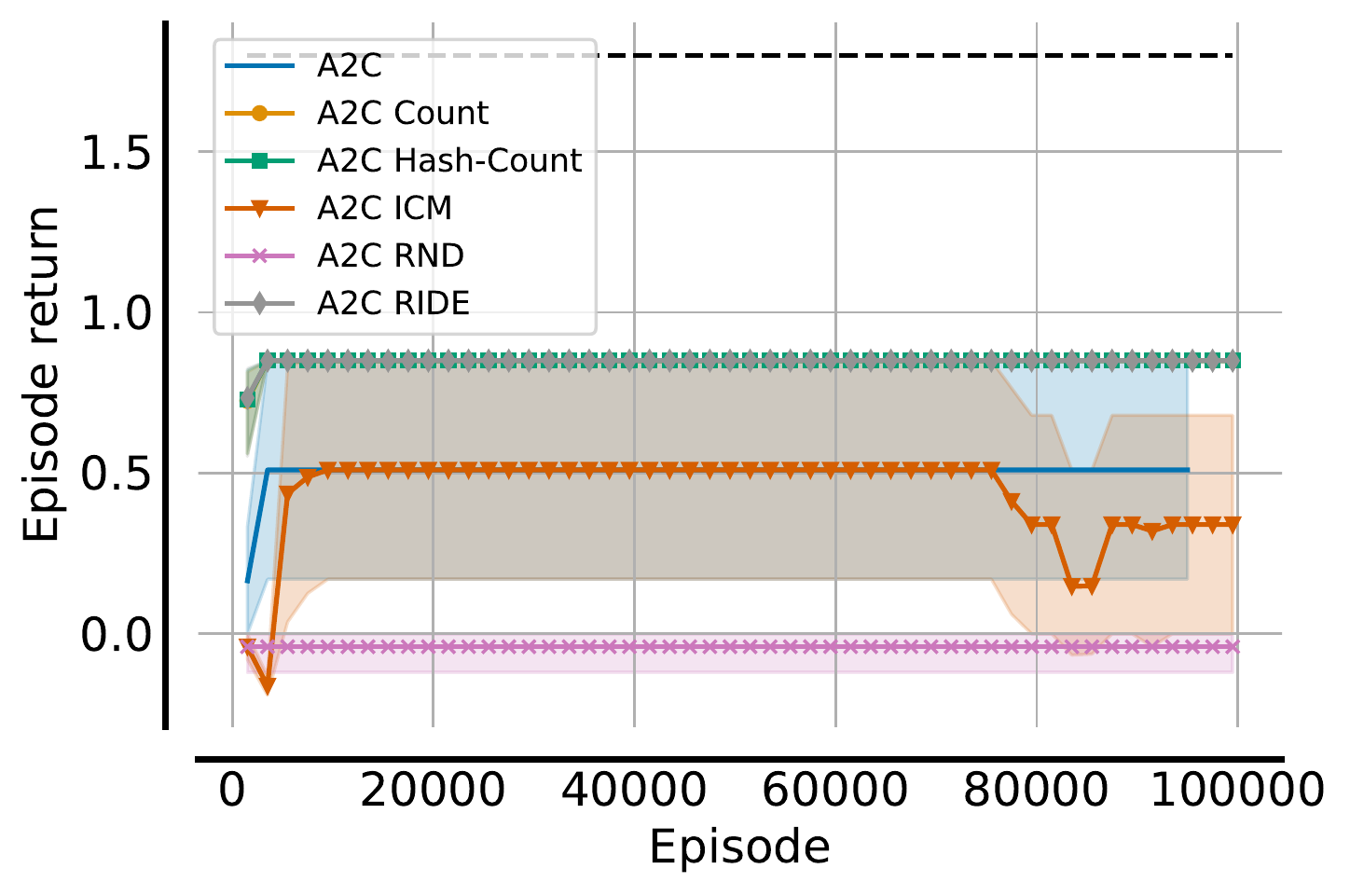}
        \caption{Hallway $N_l=10, N_r=10$ A2C}
        \label{fig:hallway_results_10_10_a2c_app}
    \end{subfigure}
    \hfill
    \begin{subfigure}{.33\textwidth}
        \centering
        \includegraphics[width=\linewidth]{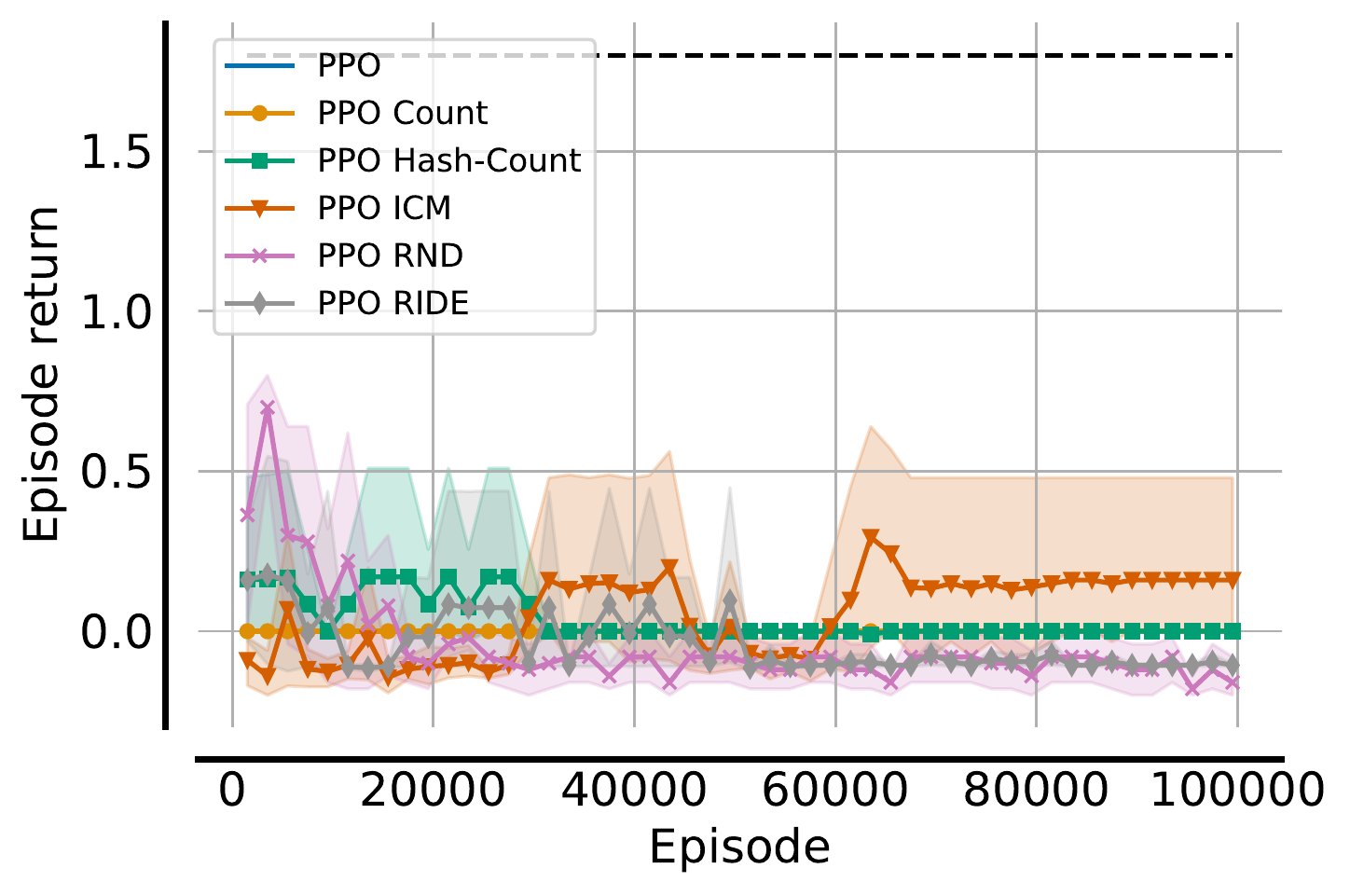}
        \caption{Hallway $N_l=10, N_r=10$ PPO}
        \label{fig:hallway_results_10_10_ppo_app}
    \end{subfigure}
    \hfill
    \begin{subfigure}{.33\textwidth}
        \centering
        \includegraphics[width=\linewidth]{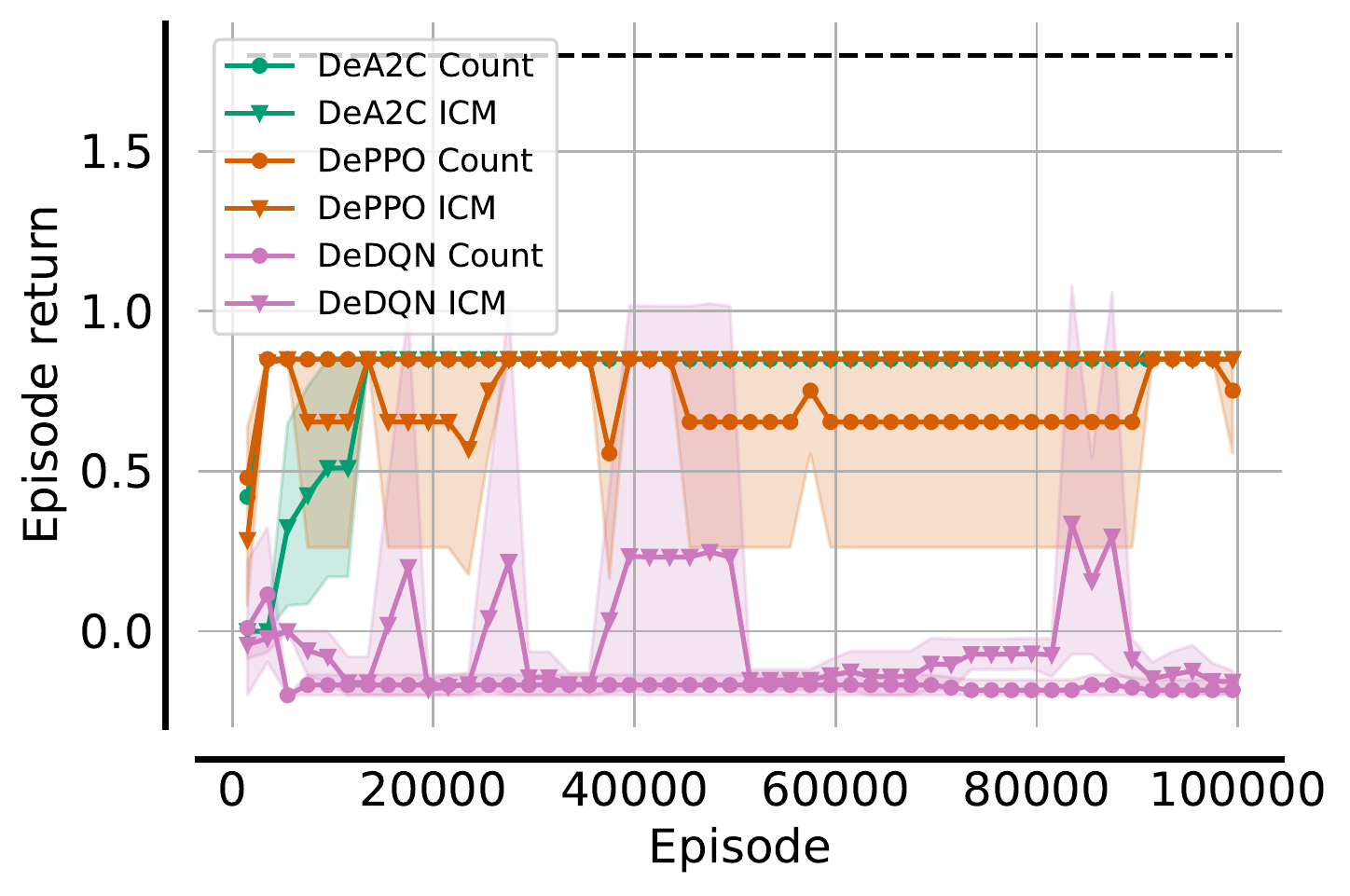}
        \caption{Hallway $N_l=10, N_r=10$ DeRL}
        \label{fig:hallway_results_10_10_derl_app}
    \end{subfigure}
    
    \begin{subfigure}{.33\textwidth}
        \centering
        \includegraphics[width=\linewidth]{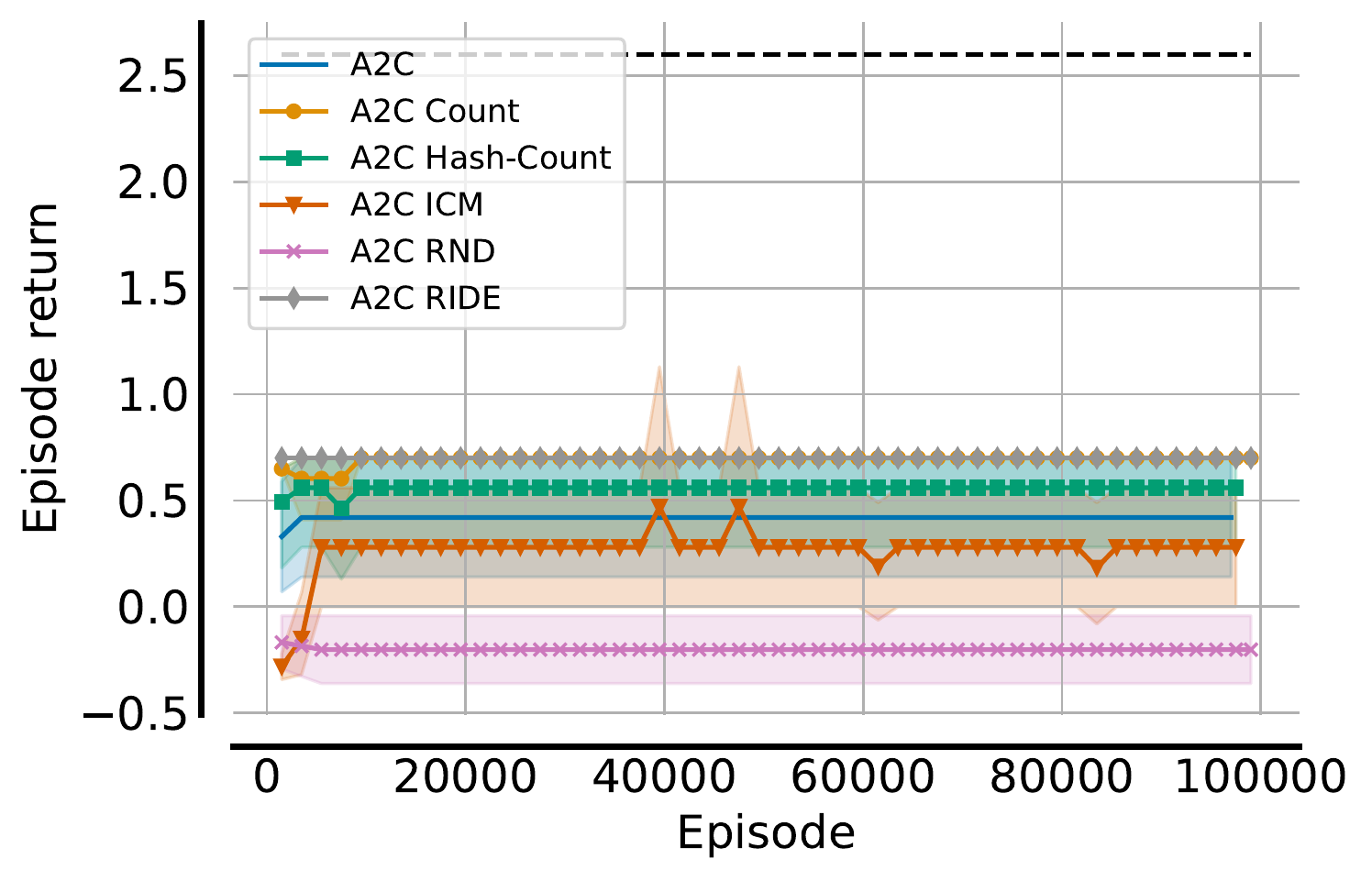}
        \caption{Hallway $N_l=20, N_r=0$ A2C}
        \label{fig:hallway_results_20_0_a2c_app}
    \end{subfigure}
    \hfill
    \begin{subfigure}{.33\textwidth}
        \centering
        \includegraphics[width=\linewidth]{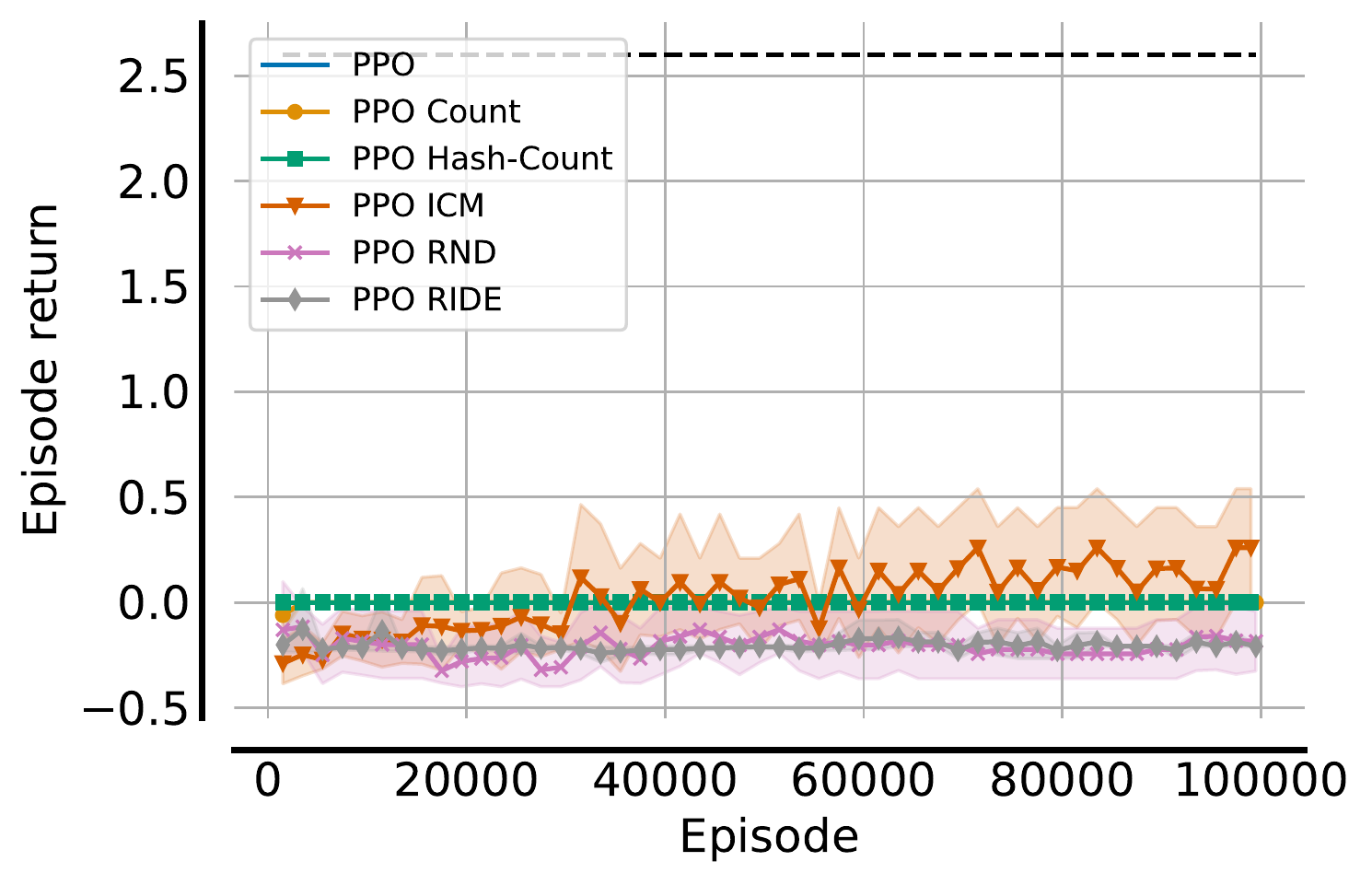}
        \caption{Hallway $N_l=20, N_r=0$ PPO}
        \label{fig:hallway_results_20_0_ppo_app}
    \end{subfigure}
    \hfill
    \begin{subfigure}{.33\textwidth}
        \centering
        \includegraphics[width=\linewidth]{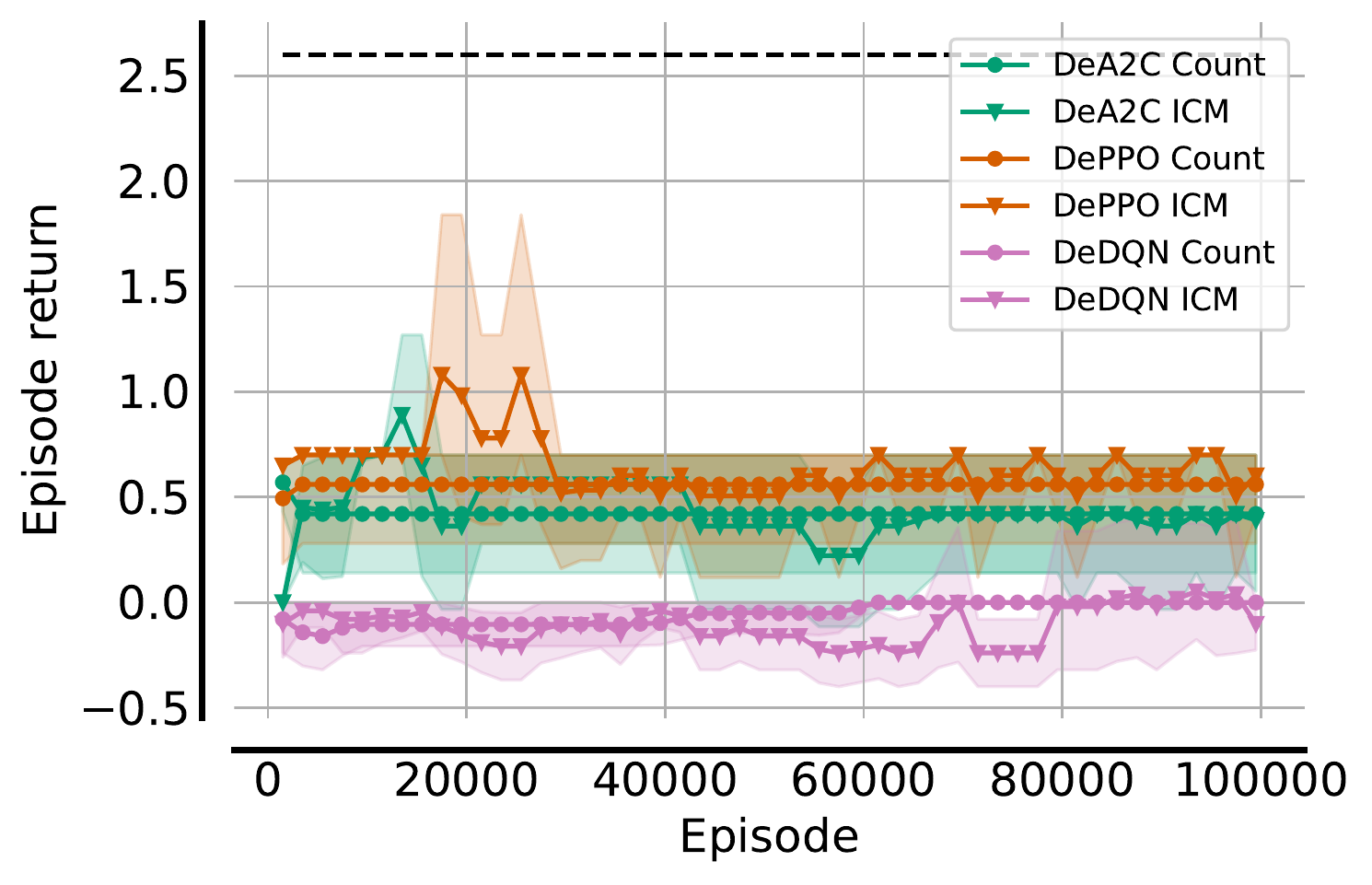}
        \caption{Hallway $N_l=20, N_r=0$ DeRL}
        \label{fig:hallway_results_20_0_derl_app}
    \end{subfigure}
    
    \begin{subfigure}{.33\textwidth}
        \centering
        \includegraphics[width=\linewidth]{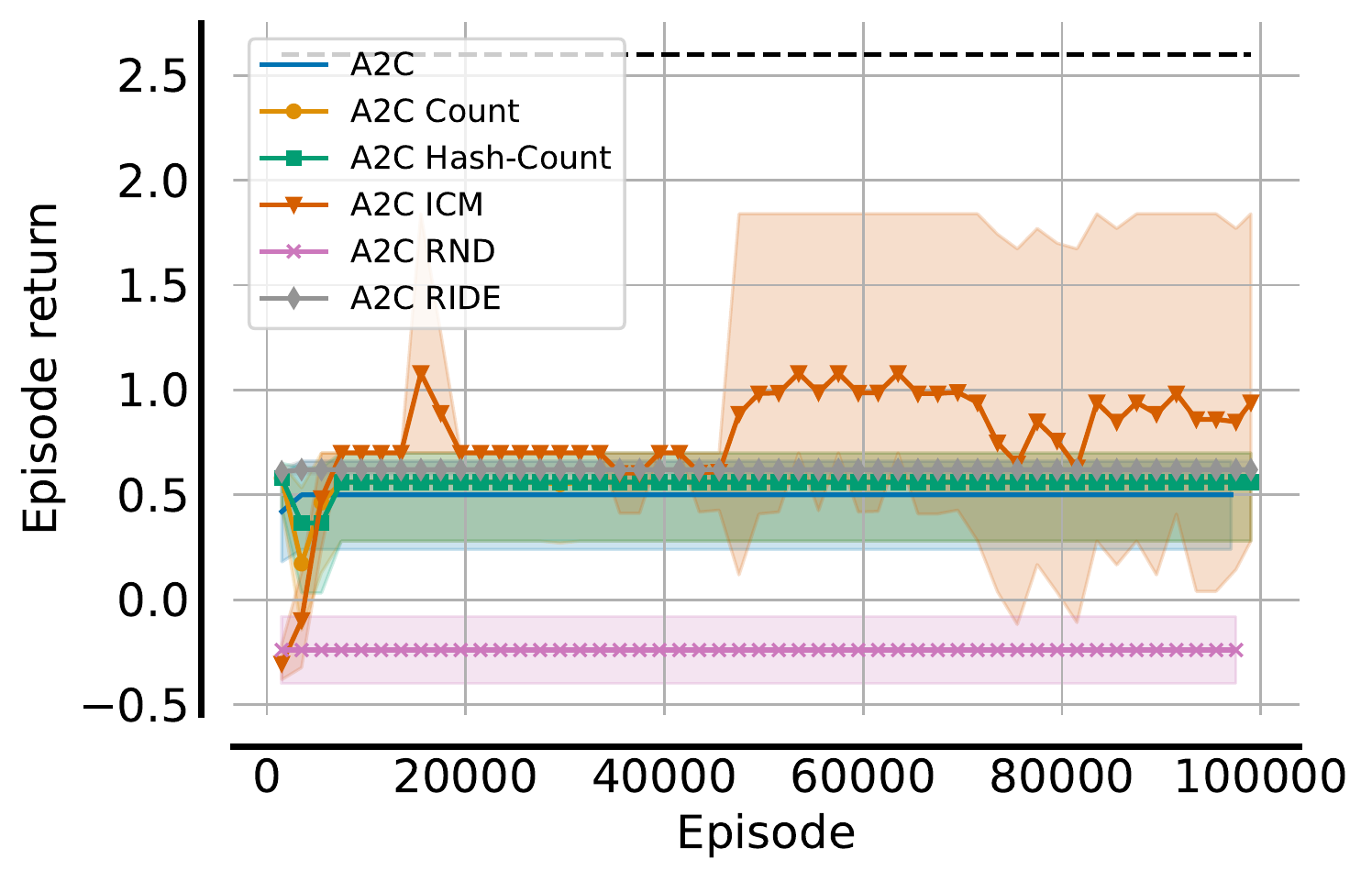}
        \caption{Hallway $N_l=20, N_r=20$ A2C}
        \label{fig:hallway_results_20_20_a2c_app}
    \end{subfigure}
    \hfill
    \begin{subfigure}{.33\textwidth}
        \centering
        \includegraphics[width=\linewidth]{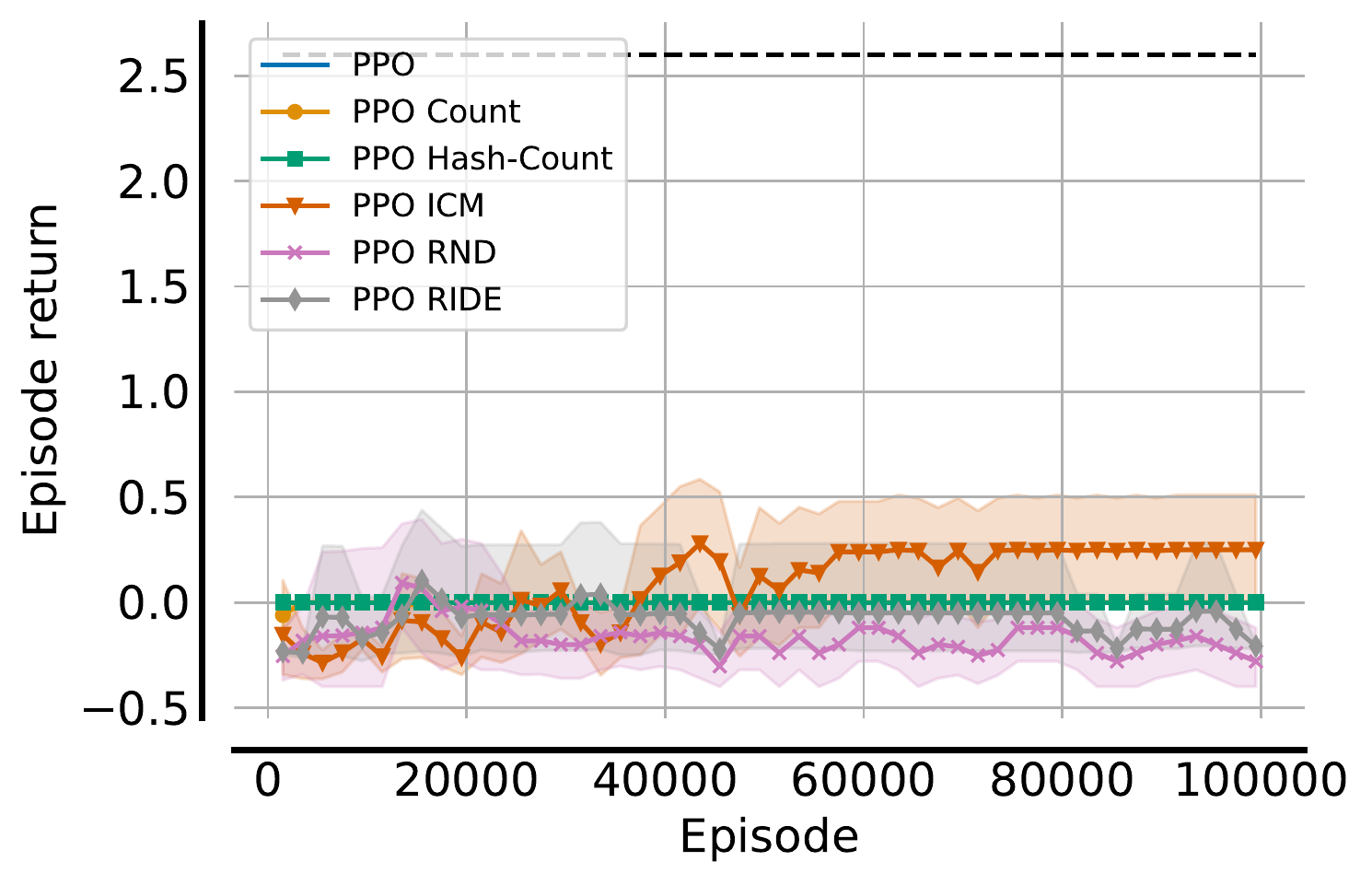}
        \caption{Hallway $N_l=20, N_r=20$ PPO}
        \label{fig:hallway_results_20_20_ppo_app}
    \end{subfigure}
    \hfill
    \begin{subfigure}{.33\textwidth}
        \centering
        \includegraphics[width=\linewidth]{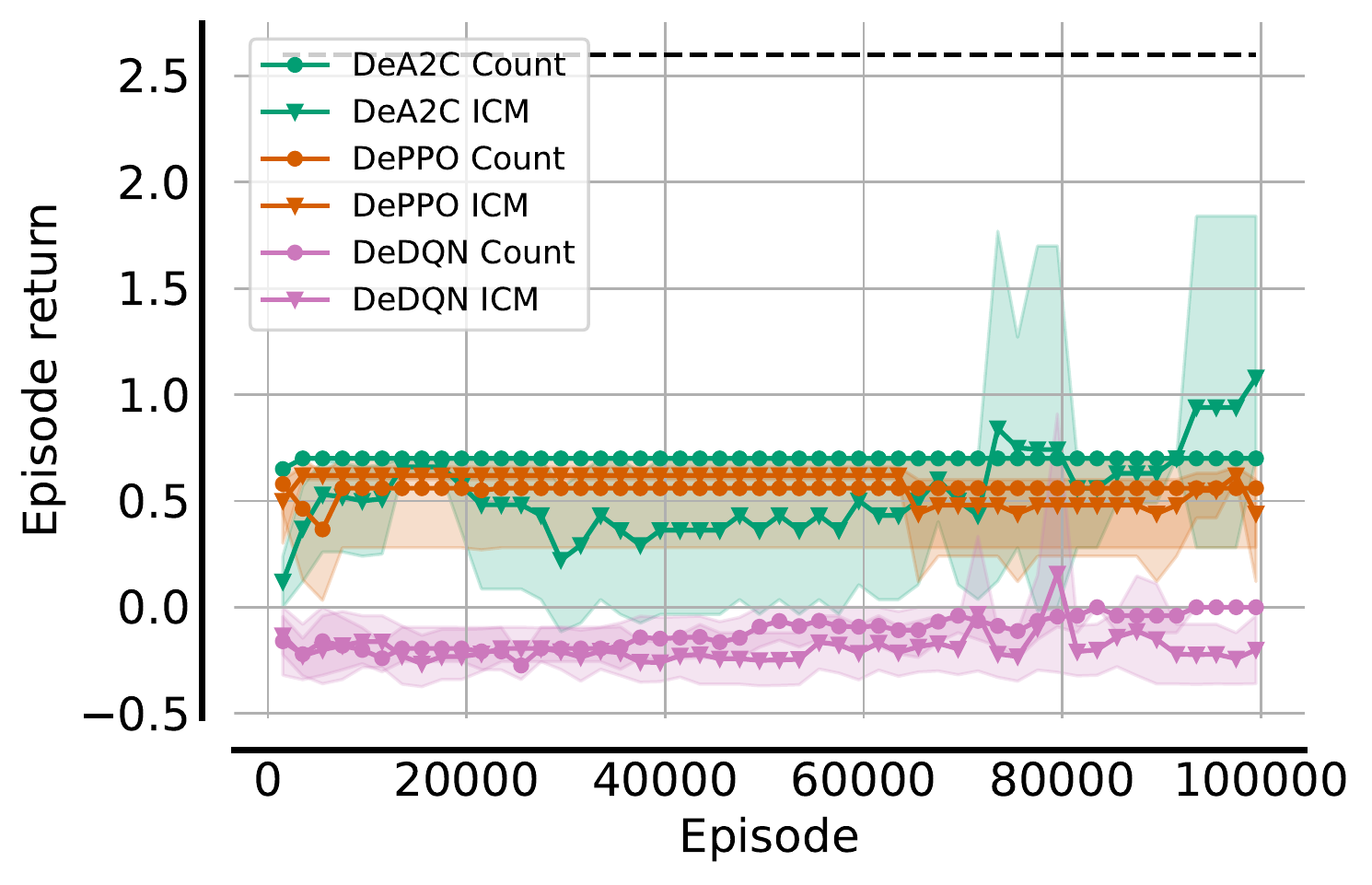}
        \caption{Hallway $N_l=20, N_r=20$ DeRL}
        \label{fig:hallway_results_20_20_derl_app}
    \end{subfigure}
    \caption{Average evaluation returns for A2C (first column), PPO (second column) and DeRL (third column) with all intrinsic rewards for all Hallway tasks with $N_l \in \{10, 20\}$. Shading indicates 95\% confidence intervals.}
    \label{fig:hallway_results_all}
\end{figure*}

\begin{figure*}[t]
    \centering
    \begin{subfigure}{.33\textwidth}
        \centering
        \includegraphics[width=\linewidth]{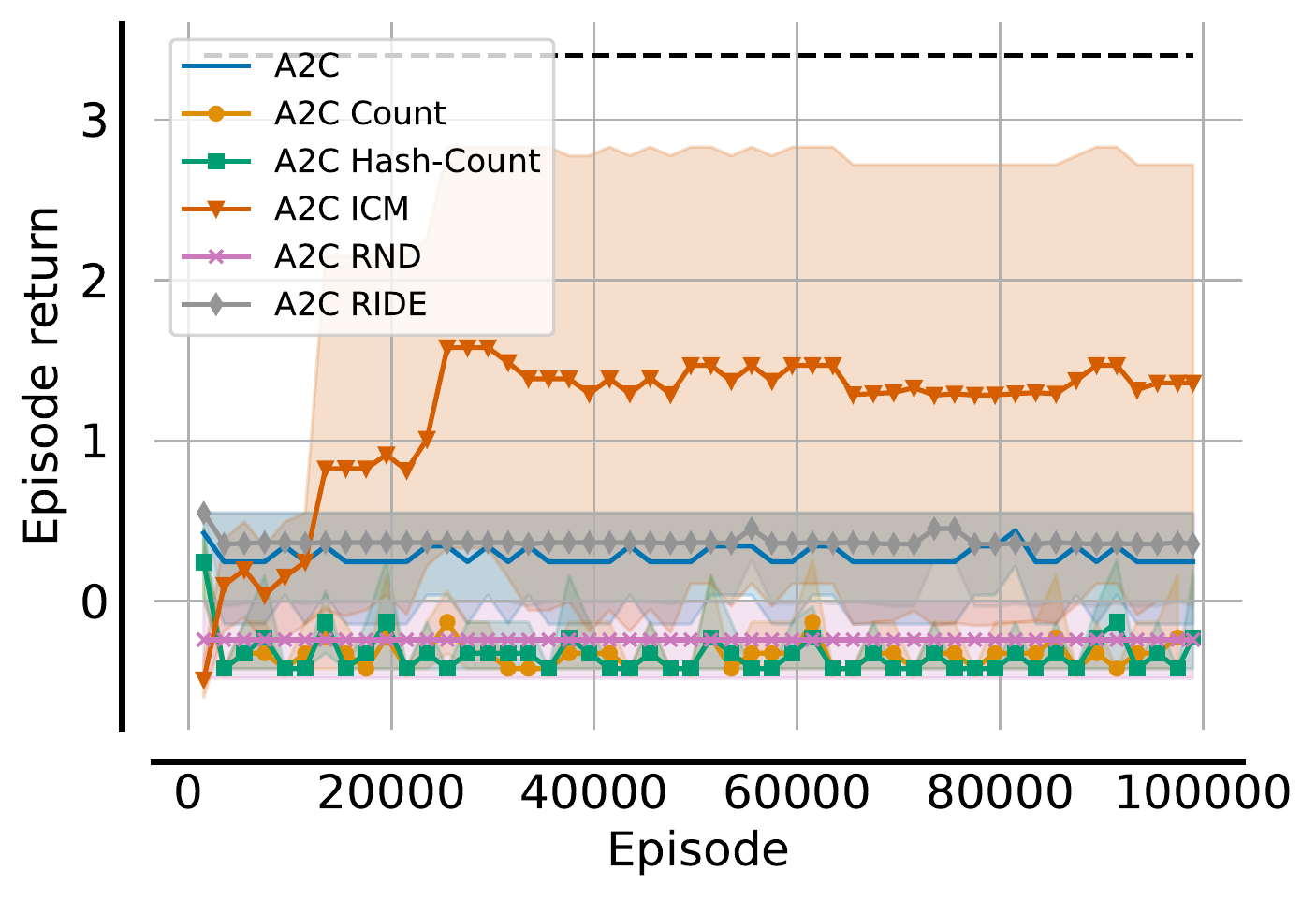}
        \caption{Hallway $N_l=30, N_r=0$ A2C}
        \label{fig:hallway_results_30_0_a2c_app}
    \end{subfigure}
    \hfill
    \begin{subfigure}{.33\textwidth}
        \centering
        \includegraphics[width=\linewidth]{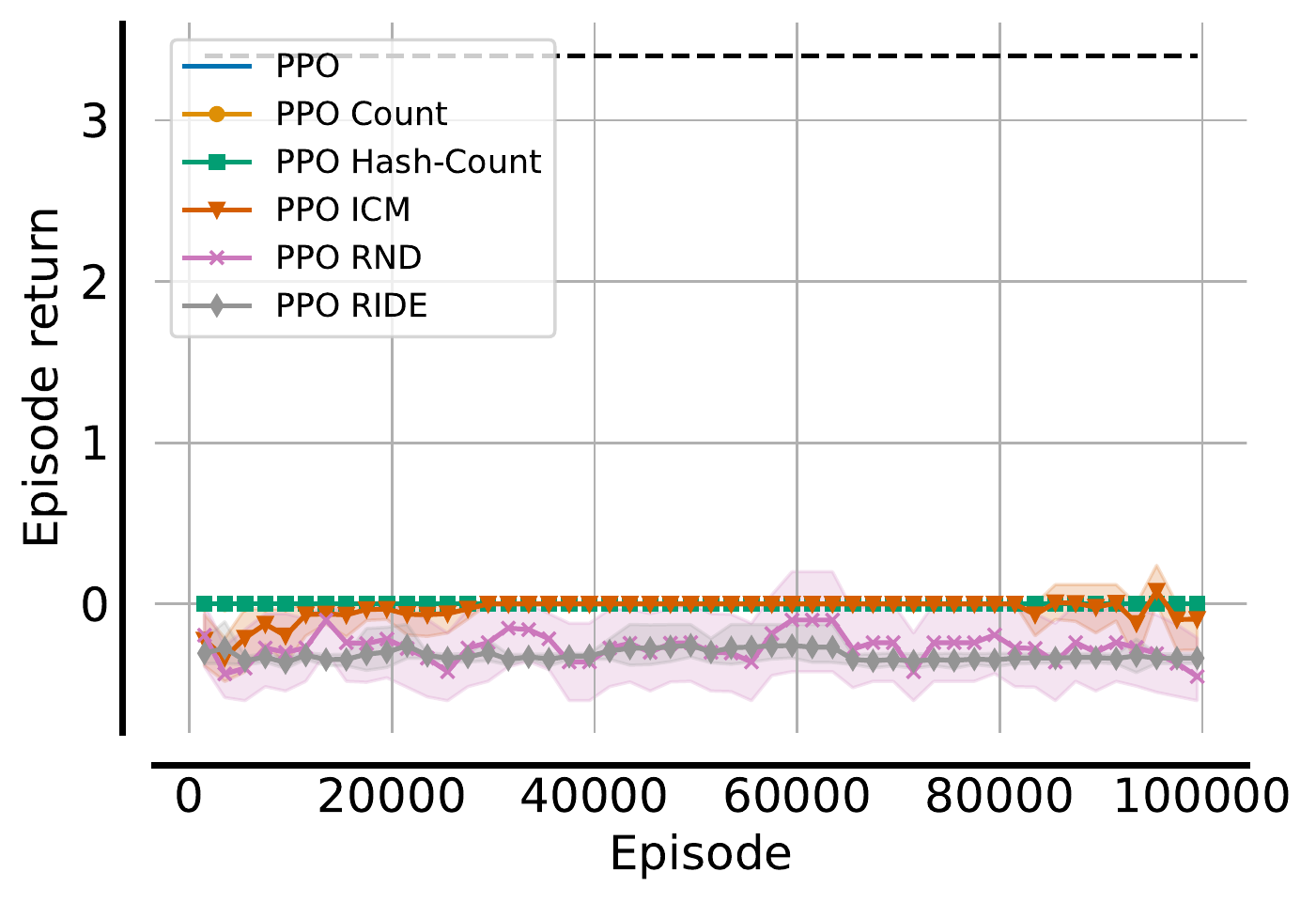}
        \caption{Hallway $N_l=30, N_r=0$ PPO}
        \label{fig:hallway_results_30_0_ppo_app}
    \end{subfigure}
    \hfill
    \begin{subfigure}{.33\textwidth}
        \centering
        \includegraphics[width=\linewidth]{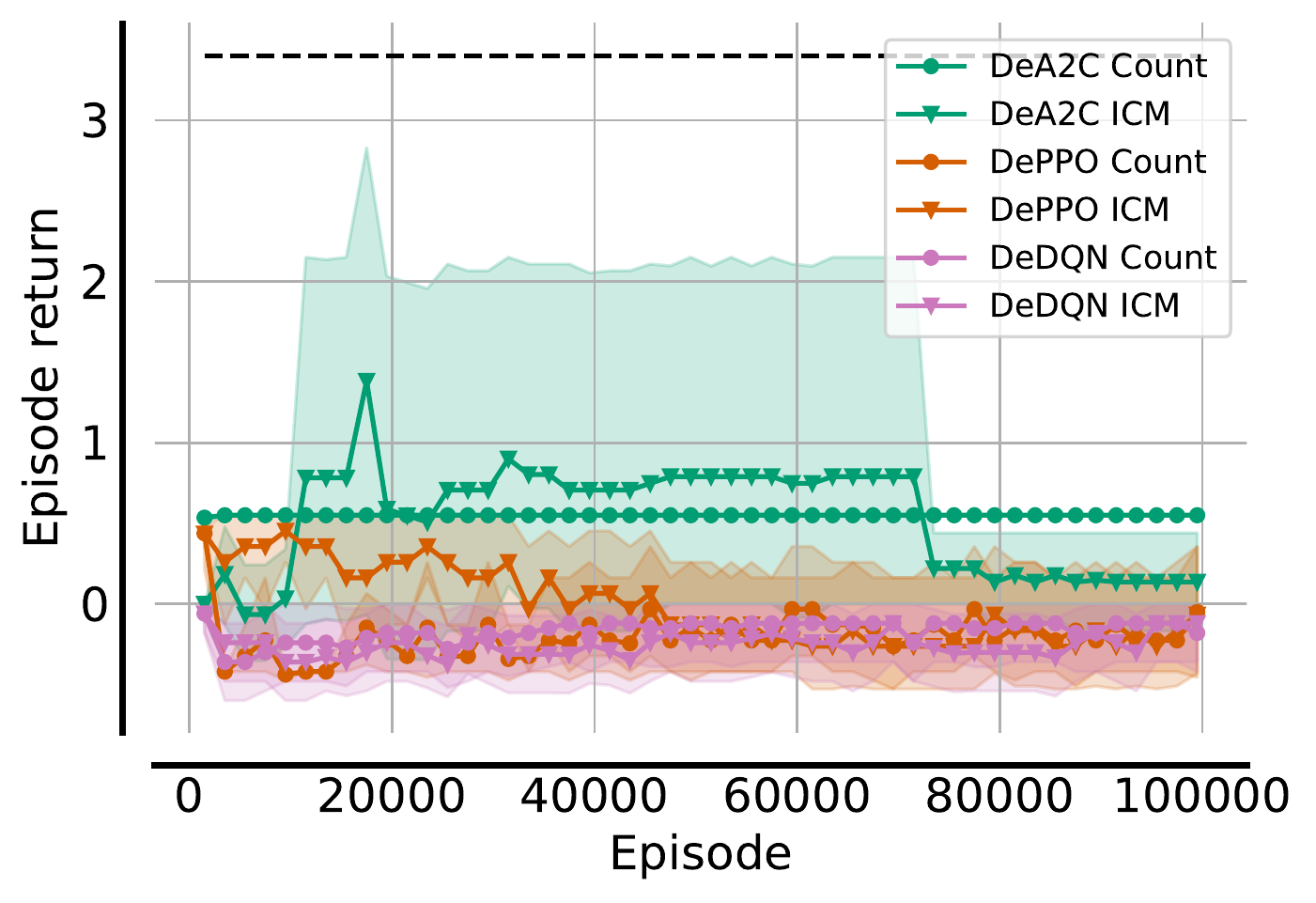}
        \caption{Hallway $N_l=30, N_r=0$ DeRL}
        \label{fig:hallway_results_30_0_derl_app}
    \end{subfigure}
    
    \begin{subfigure}{.33\textwidth}
        \centering
        \includegraphics[width=\linewidth]{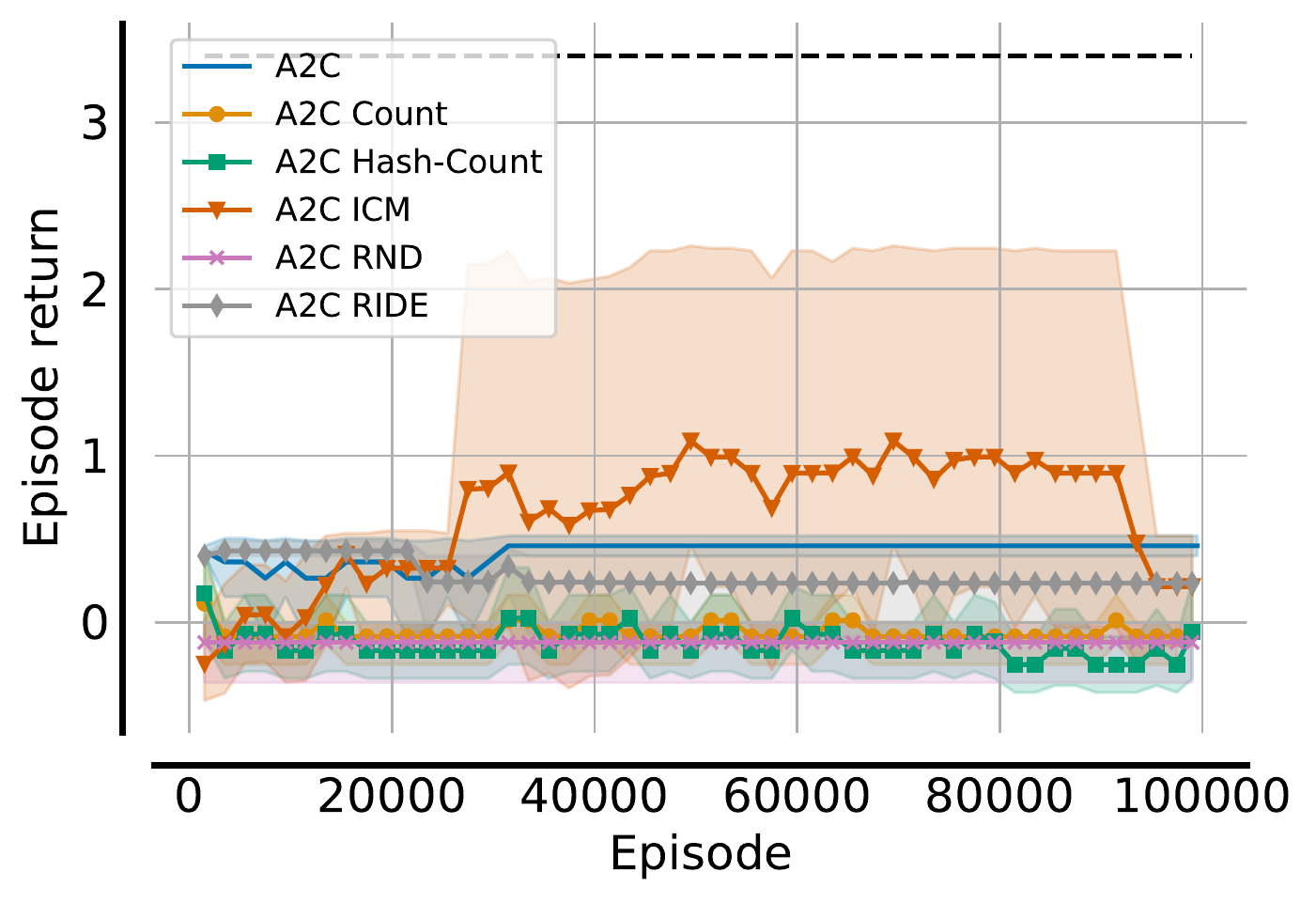}
        \caption{Hallway $N_l=30, N_r=30$ A2C}
        \label{fig:hallway_results_30_30_a2c_app}
    \end{subfigure}
    \hfill
    \begin{subfigure}{.33\textwidth}
        \centering
        \includegraphics[width=\linewidth]{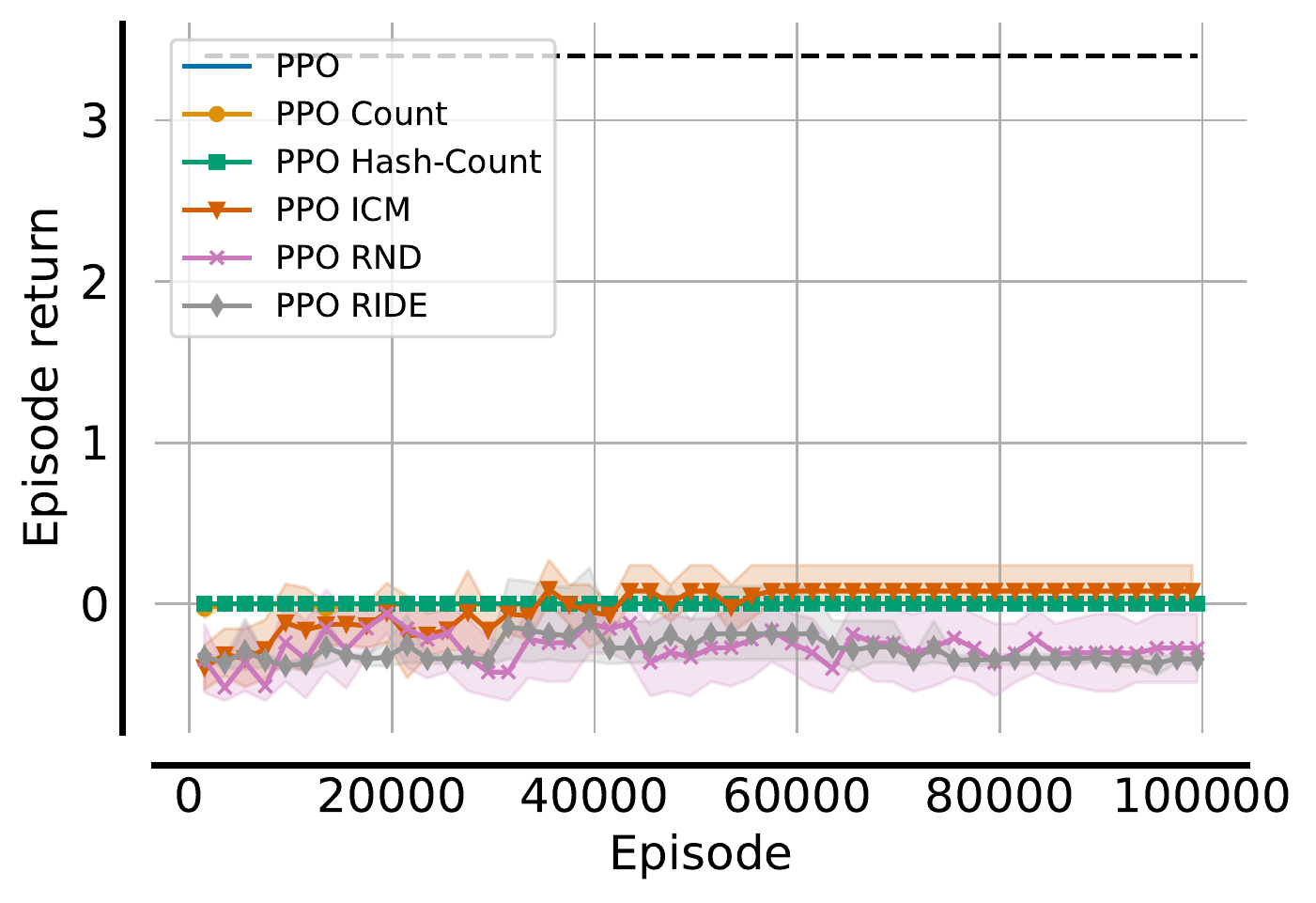}
        \caption{Hallway $N_l=30, N_r=30$ PPO}
        \label{fig:hallway_results_30_30_ppo_app}
    \end{subfigure}
    \hfill
    \begin{subfigure}{.33\textwidth}
        \centering
        \includegraphics[width=\linewidth]{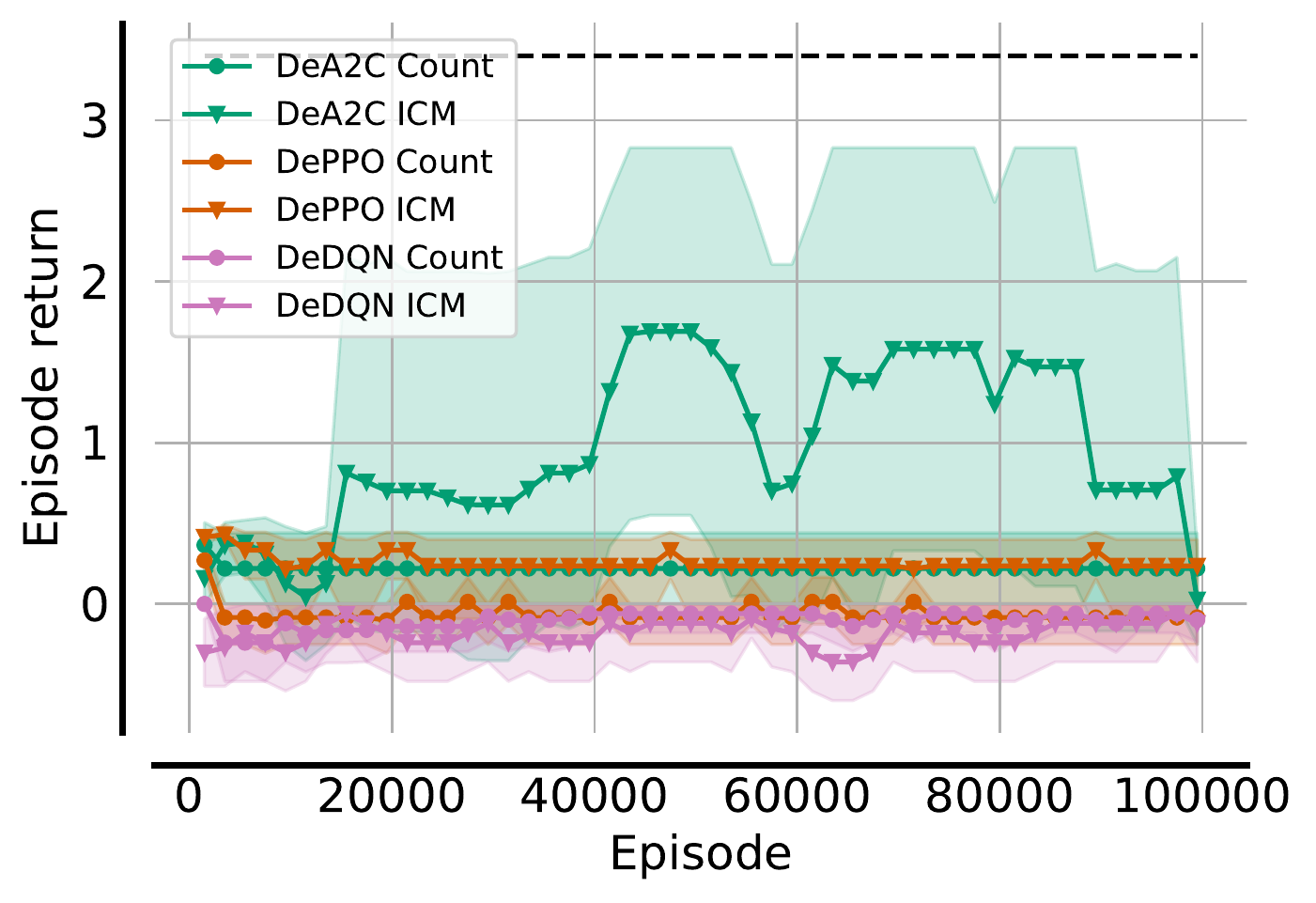}
        \caption{Hallway $N_l=30, N_r=30$ DeRL}
        \label{fig:hallway_results_30_30_derl_app}
    \end{subfigure}
    \caption{Average evaluation returns for A2C (first column), PPO (second column) and DeRL (third column) with all intrinsic rewards for all Hallway tasks with $N_l=30$. Shading indicates 95\% confidence intervals.}
    \label{fig:hallway_30_results_all}
\end{figure*}

\clearpage
\section{Hyperparameter Sensitivity}
\label{app:hyperparameter_sensitivity}
In this section, we provide further plots and tables containing maximum and average achieved evaluation returns for both sets of hyerparameter sensitivity experiments. As described in \Cref{sec:results_sensitivity}, we evaluate both baselines and DeRL algorithms with varying intrinsic reward coefficients, $\lambda$, and rate of decay of intrinsic rewards. Both experiments were conducted in DeepSea 10 and Hallway $N_l = N_r = 10$. We provide tables showing the maximum and average evaluation returns as in \Cref{app:full_results} and bar plots indicating the varying average evaluation returns for various parameterisation of intrinsic rewards. Within tables, the highest performing configuration for a single algorithm is highlighted in bold within each row with all configurations within a single standard deviation of the highest return.

\subsection{Intrinsic Reward Scale}
\label{app:exploration_bonus_scale}
First, we evaluate all baselines and DeRL algorithms for various intrinsic reward coefficients $\lambda \in \{0.01, 0.1, 0.25, 0.5, 1.0, 2.0, 4.0, 10.0, 100.0\}$ in DeepSea and Hallway.

\begin{figure*}[h]
    \centering
    \includegraphics[width=.8\linewidth]{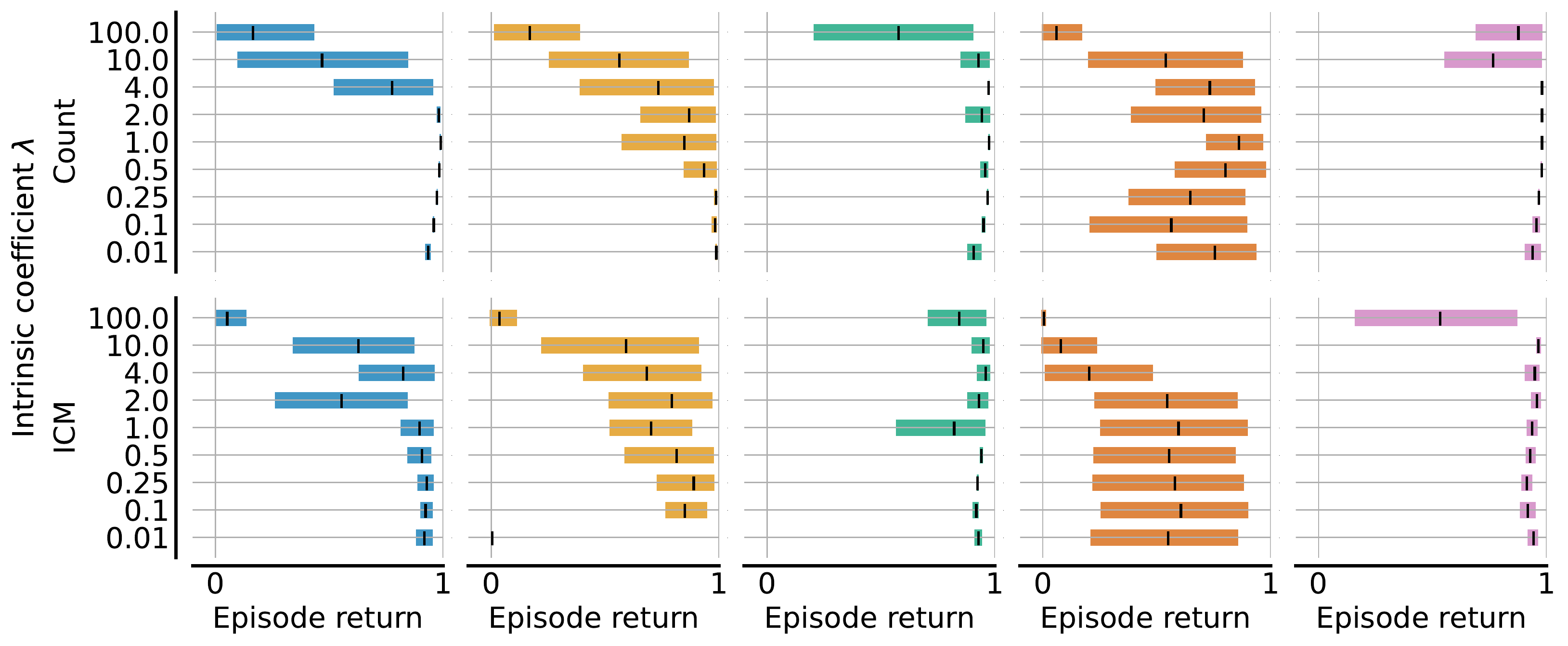}
    \includegraphics[width=.5\linewidth]{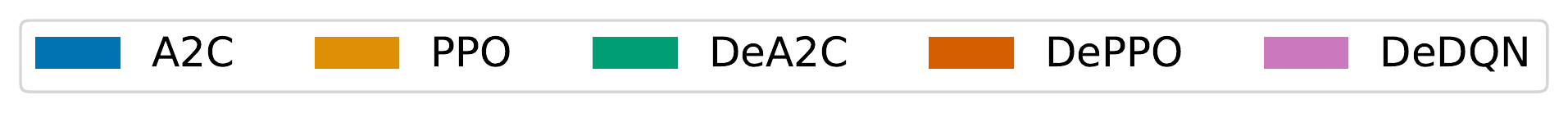}
    
    \caption{Average evaluation returns for baselines and DeRL with Count (upper row) and ICM (lower row) intrinsic rewards in DeepSea 10 with $\lambda \in \{0.01, 0.1, 0.25, 0.5, 1.0, 2.0, 4.0, 10.0, 100.0\}$. Shading indicates 95\% confidence intervals.}
\end{figure*}

\begin{figure*}[h]
    \centering
    \begin{subfigure}{.33\textwidth}
        \centering
        \includegraphics[width=.85\linewidth]{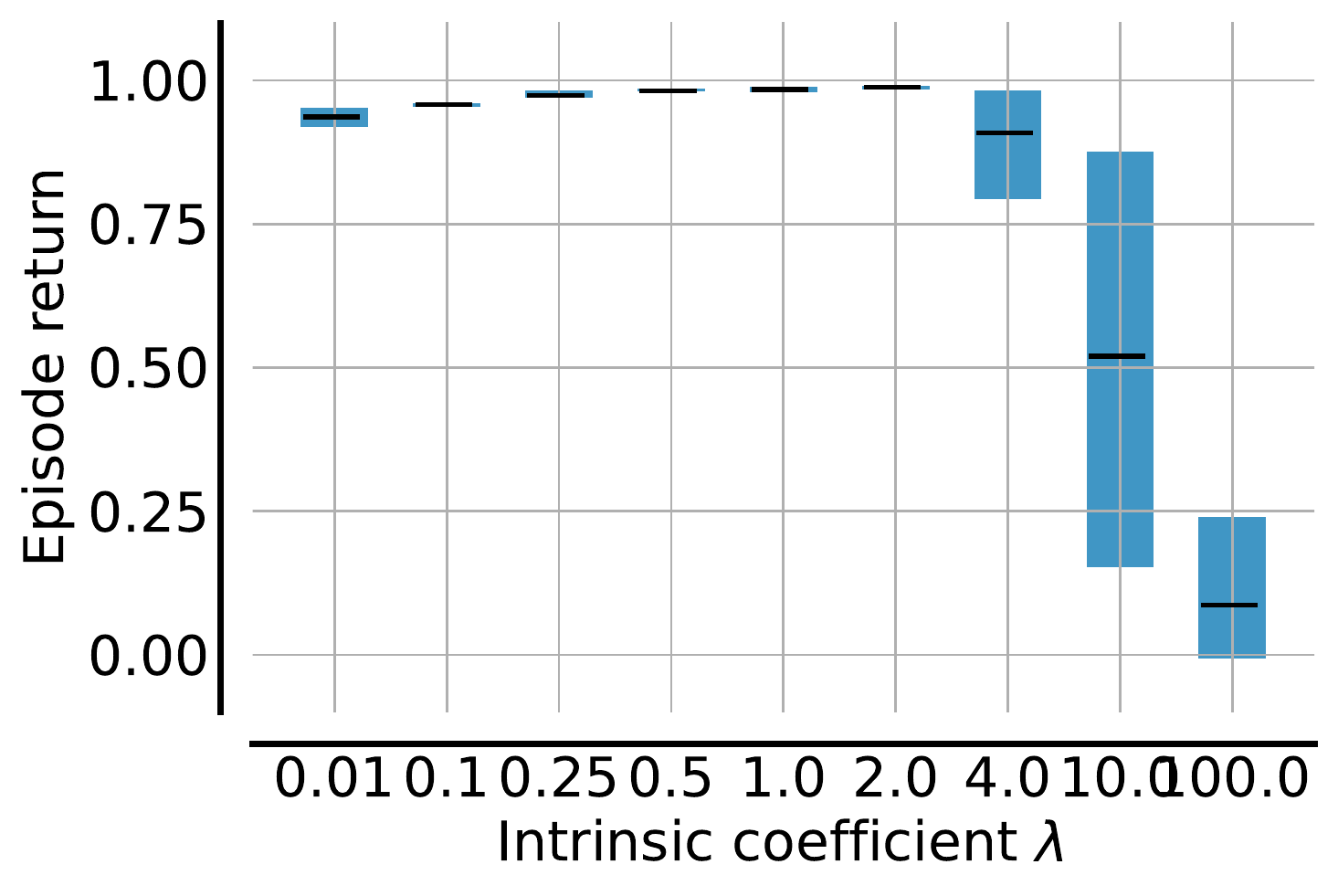}
        \caption{A2C Hash-Count}
    \end{subfigure}
    \hfill
    \begin{subfigure}{.33\textwidth}
        \centering
        \includegraphics[width=.85\linewidth]{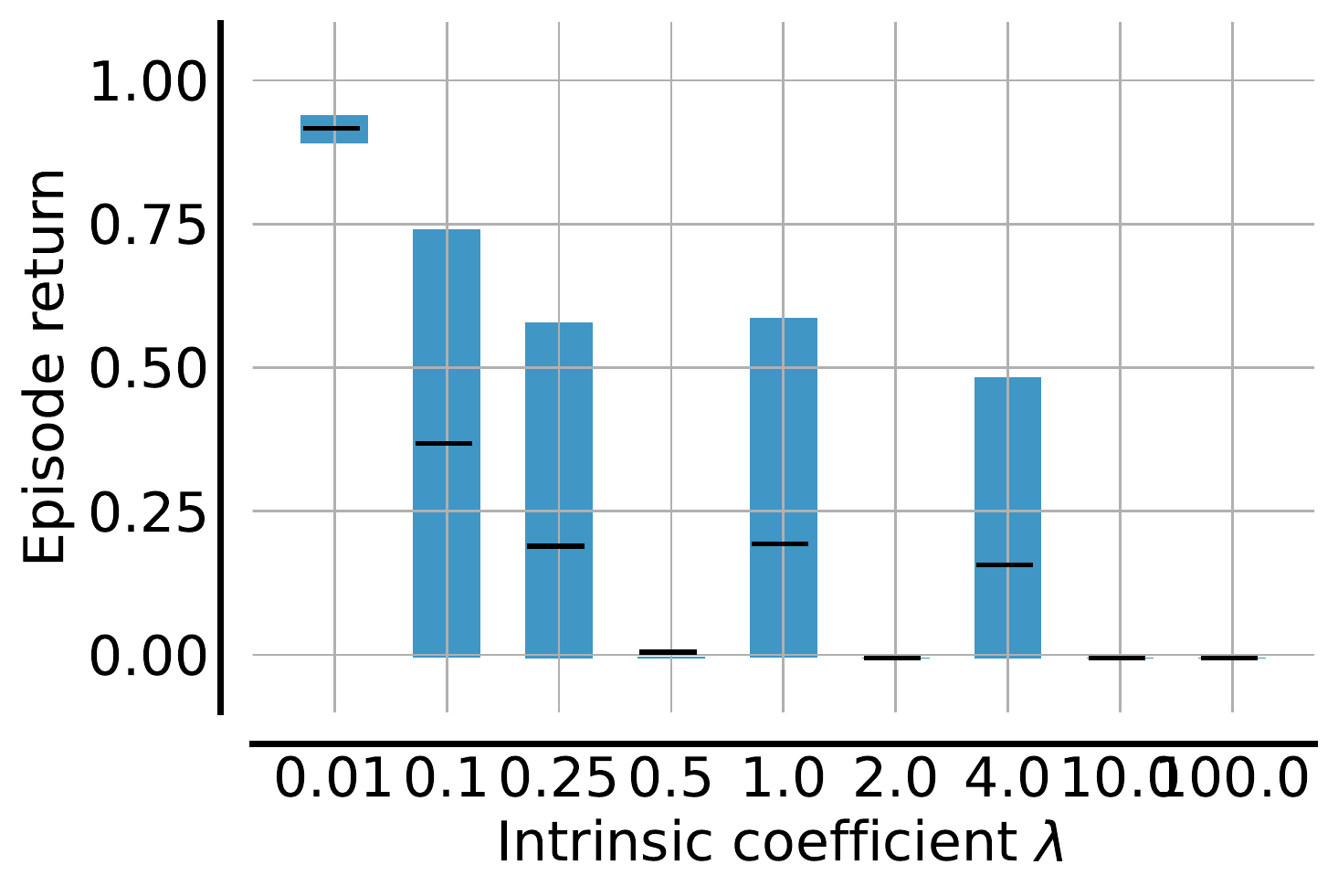}
        \caption{A2C RND}
    \end{subfigure}
    \hfill
    \begin{subfigure}{.33\textwidth}
        \centering
        \includegraphics[width=.85\linewidth]{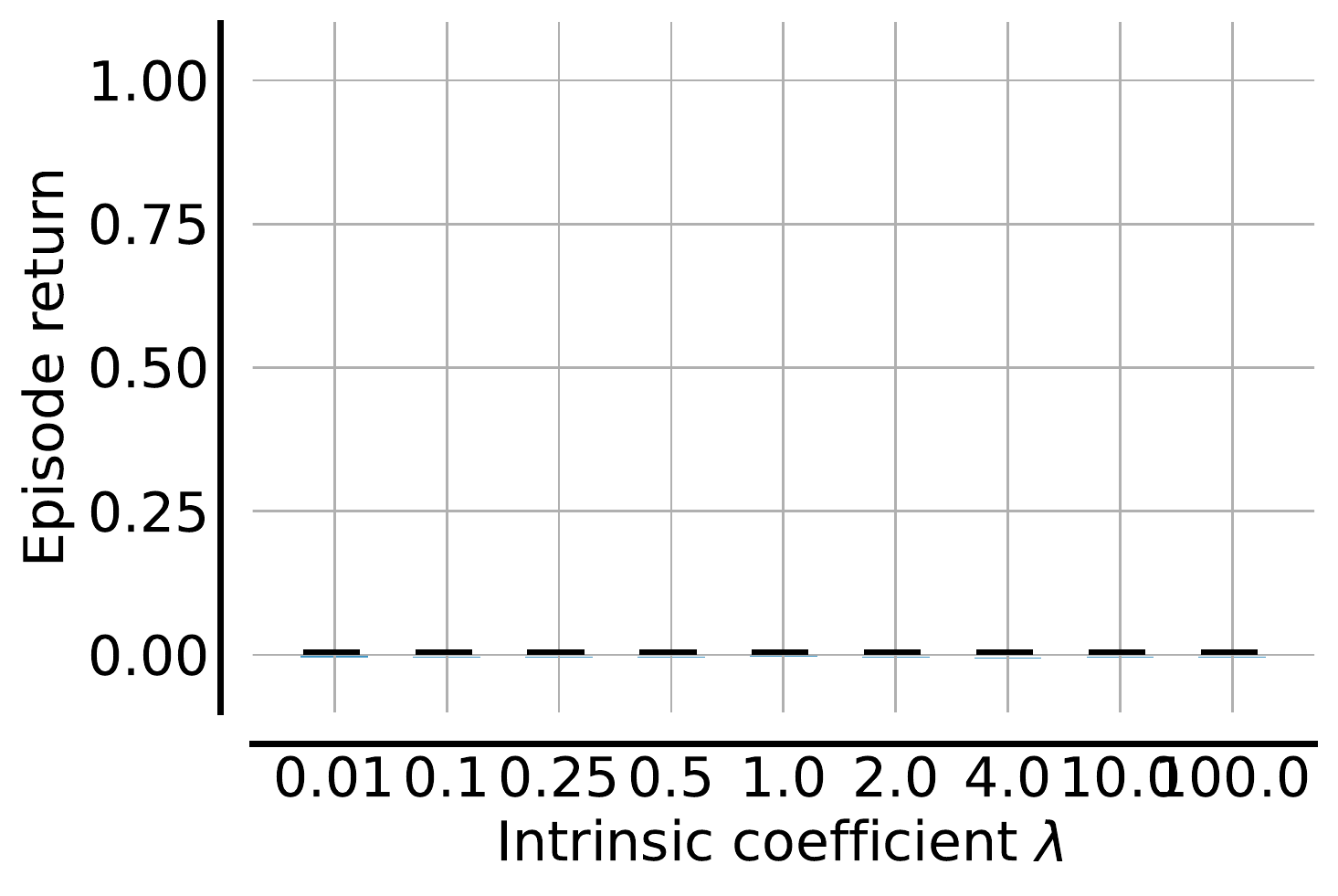}
        \caption{A2C RIDE}
    \end{subfigure}
    
    \begin{subfigure}{.33\textwidth}
        \centering
        \includegraphics[width=.85\linewidth]{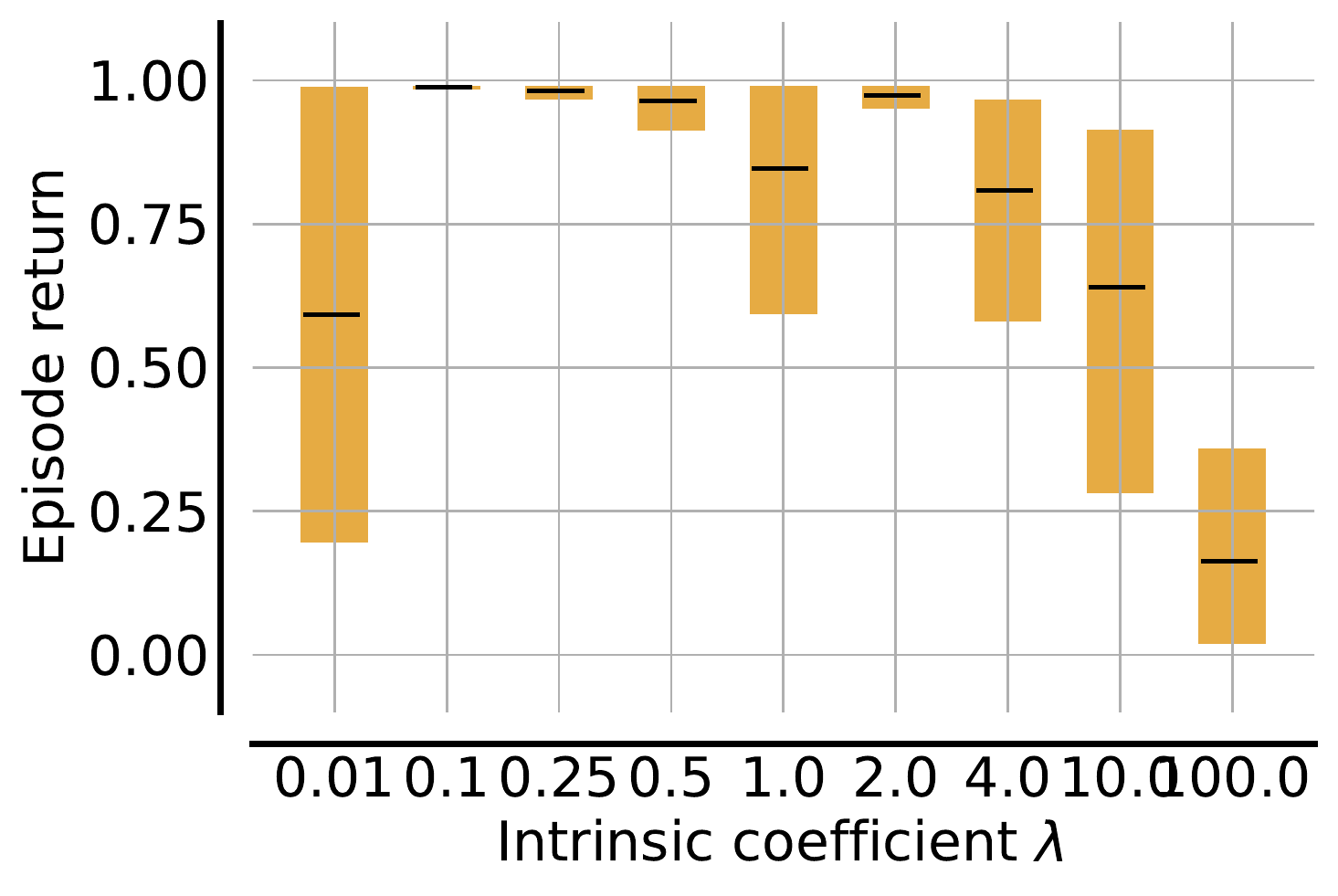}
        \caption{PPO Hash-Count}
    \end{subfigure}
    \hfill
    \begin{subfigure}{.33\textwidth}
        \centering
        \includegraphics[width=.85\linewidth]{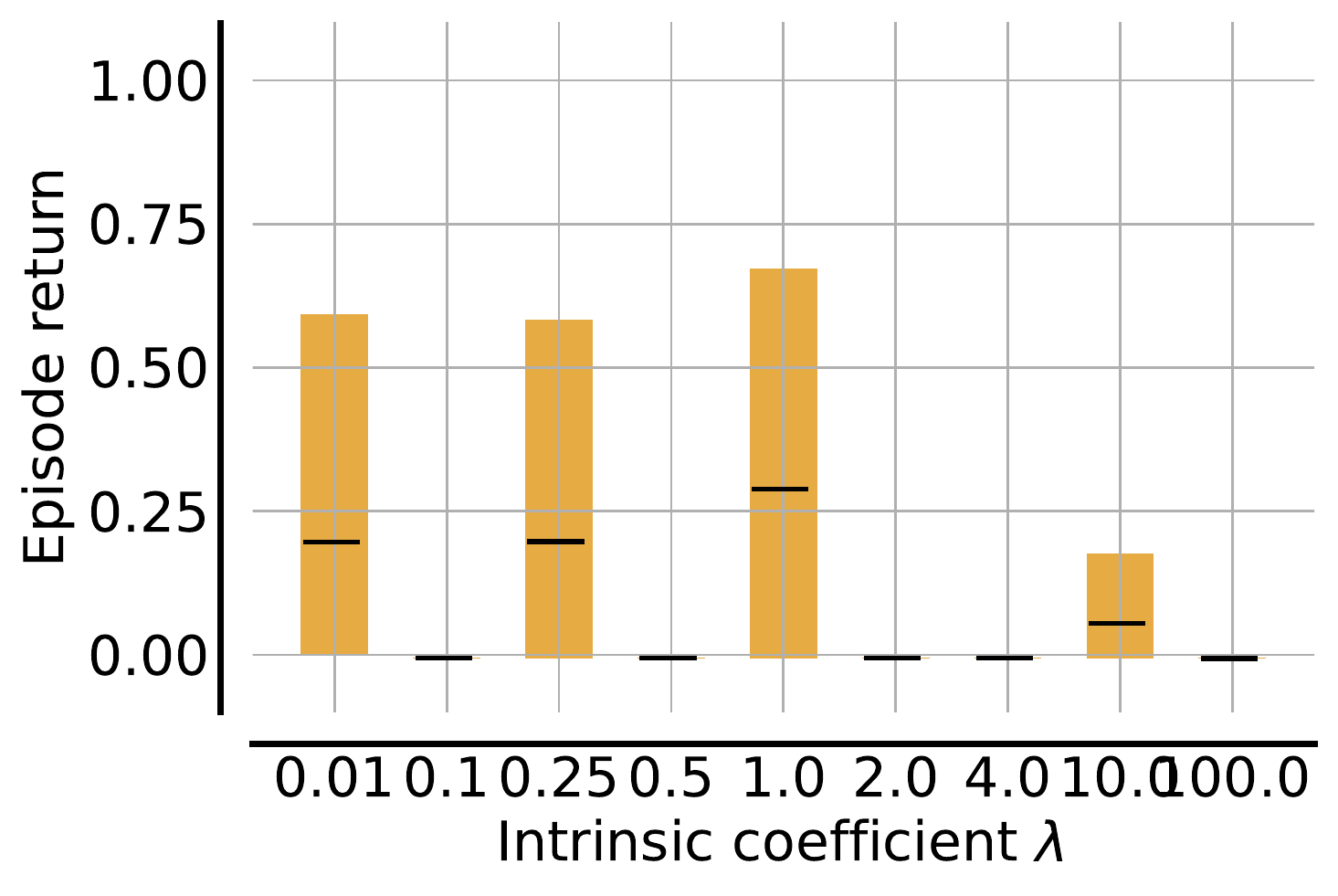}
        \caption{PPO RND}
    \end{subfigure}
    \hfill
    \begin{subfigure}{.33\textwidth}
        \centering
        \includegraphics[width=.85\linewidth]{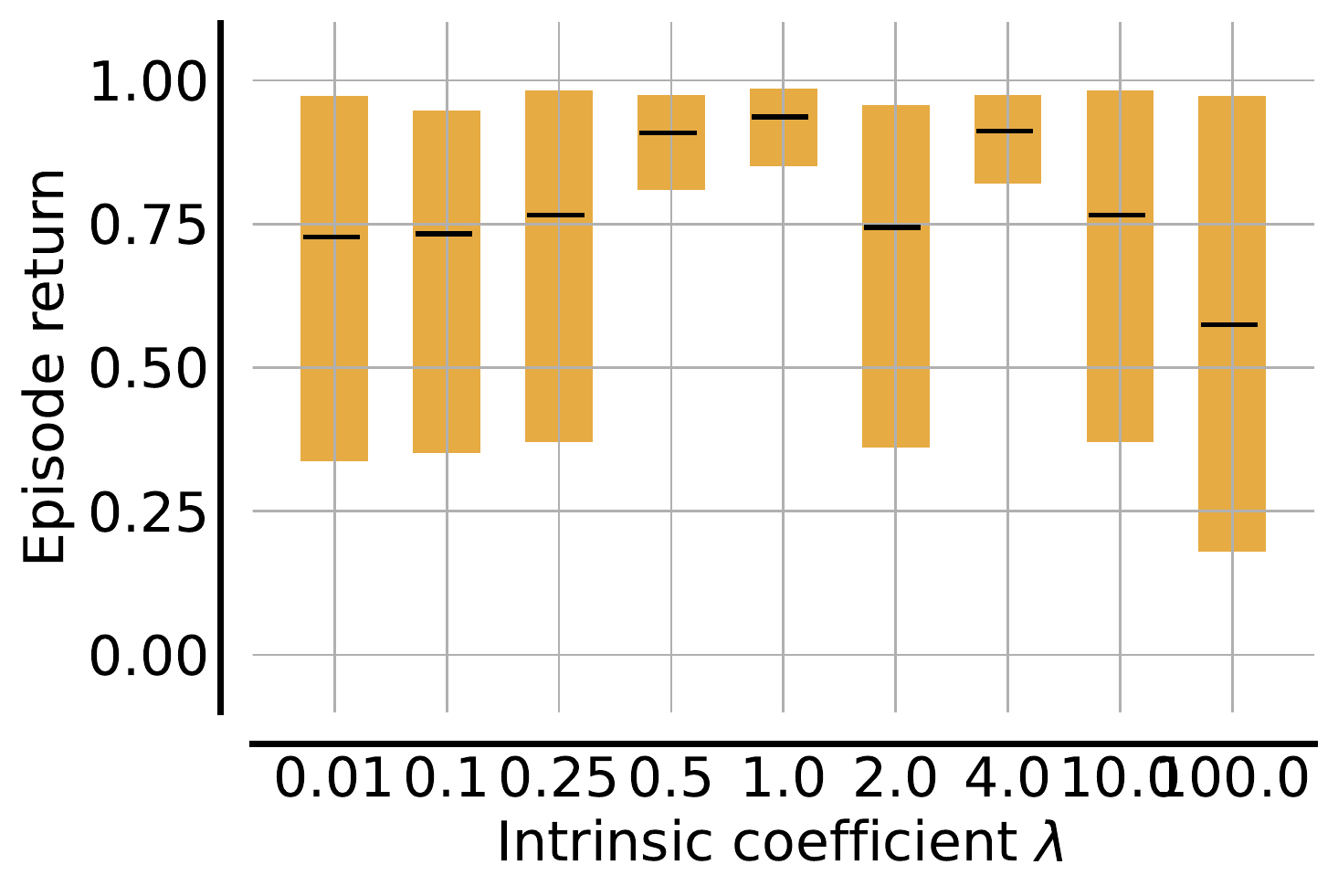}
        \caption{PPO RIDE}
    \end{subfigure}
    \caption{Average evaluation returns for A2C and PPO with Hash-Count, RND and RIDE intrinsic rewards in DeepSea 10 with $\lambda \in \{0.01, 0.1, 0.25, 0.5, 1.0, 2.0, 4.0, 10.0, 100.0\}$. Shading indicates 95\% confidence intervals.}
\end{figure*}

\begin{table}[!ht]
	\centering
	\caption{Maximum evaluation returns with a single standard deviation in DeepSea 10 for $\lambda \in \{0.01, 0.1, 0.25, 0.5, 1.0, 2.0, 4.0, 10.0, 100.0\}$.}
	\resizebox{\textwidth}{!}{
	\robustify\bf
	\begin{tabular}{l S S S S S S S S S}
		\toprule
		{Algorithm \textbackslash \ $\lambda$} & {0.01} & {0.1} & {0.25} & {0.5} & {1.0} & {2.0} & {4.0} & {10.0} & {100.0} \\
		\midrule
		 A2C Count & \bf 0.99(0) & \bf 0.99(0) & \bf 0.99(0) & \bf 0.99(0) & \bf 0.99(0) & \bf 0.99(0) & \bf 0.99(0) &  0.79(30) &  0.79(30) \\
		 A2C Hash-Count & \bf 0.99(0) & \bf 0.99(0) & \bf 0.99(0) & \bf 0.99(0) & \bf 0.99(0) & \bf 0.99(0) & \bf 0.99(0) &  0.79(30) &  0.39(40) \\
		 A2C ICM & \bf 0.99(0) & \bf 0.99(0) & \bf 0.99(0) & \bf 0.99(0) & \bf 0.99(0) & \bf 0.99(0) & \bf 0.99(0) & \bf 0.99(0) &  0.59(40) \\
		 A2C RND & \bf 0.99(0) &  0.39(40) &  0.19(30) &  -0.00(0) &  0.20(30) &  -0.00(0) &  0.19(30) &  -0.00(0) &  -0.01(0) \\
		 A2C RIDE & -0.00(0) & -0.00(0) & -0.00(0) & -0.00(0) & -0.00(0) & -0.00(0) & -0.00(0) & -0.00(0) & -0.00(0) \\
		 \midrule
		 PPO Count & \bf 0.99(0) & \bf 0.99(0) & \bf 0.99(0) & \bf 0.99(0) & \bf 0.99(0) & \bf 0.99(0) & \bf 0.99(0) & \bf 0.99(0) & \bf 0.99(0) \\
		 PPO Hash-Count &  0.59(40) & \bf 0.99(0) & \bf 0.99(0) & \bf 0.99(0) & \bf 0.99(0) & \bf 0.99(0) & \bf 0.99(0) & \bf 0.99(0) &  0.79(30) \\
		 PPO ICM &  0.00(0) & \bf 0.99(0) & \bf 0.99(0) & \bf 0.99(0) & \bf 0.99(0) & \bf 0.99(0) & \bf 0.99(0) & \bf 0.99(0) &  0.39(40) \\
		 PPO RND &  0.20(30) &  -0.00(0) & 0.39(40) &  -0.00(0) & 0.39(40) &  -0.00(0) &  -0.01(0) &  0.19(30) &  -0.01(0) \\
		 PPO RIDE &  0.79(30) &  0.79(30) &  0.79(30) & \bf 0.99(0) & \bf 0.99(0) &  0.79(30) & \bf 0.99(0) &  0.79(30) &  0.59(40) \\
		 \midrule
		 DeA2C Count & \bf 0.99(0) & \bf 0.99(0) & \bf 0.99(0) & \bf 0.99(0) & \bf 0.99(0) & \bf 0.99(0) & \bf 0.99(0) & \bf 0.99(0) & \bf 0.99(0) \\
		 DeA2C ICM & \bf 0.99(0) & \bf 0.99(0) & \bf 0.99(0) & \bf 0.99(0) & \bf 0.99(0) & \bf 0.99(0) & \bf 0.99(0) & \bf 0.99(0) & \bf 0.99(0) \\
		 DePPO Count & \bf 0.99(0) & \bf 0.99(0) & \bf 0.99(0) & \bf 0.99(0) & \bf 0.99(0) & \bf 0.99(0) & \bf 0.99(0) & \bf 0.99(0) &  0.59(40) \\
		 DePPO ICM & \bf 0.99(0) & \bf 0.99(0) & \bf 0.99(0) & \bf 0.99(0) & \bf 0.99(0) & \bf 0.99(0) &  0.59(40) &  0.39(40) &  0.19(30) \\
		 DeDQN Count & \bf 0.99(0) & \bf 0.99(0) & \bf 0.99(0) & \bf 0.99(0) & \bf 0.99(0) & \bf 0.99(0) & \bf 0.99(0) & \bf 0.99(0) & \bf 0.99(0) \\
		 DeDQN ICM & \bf 0.99(0) & \bf 0.99(0) & \bf 0.99(0) & \bf 0.99(0) & \bf 0.99(0) & \bf 0.99(0) & \bf 0.99(0) & \bf 0.99(0) &  0.79(30) \\
		\bottomrule
	\end{tabular}
	}
	\label{tab:deepsea_10_intrinsic_coef_max}
\end{table}

\begin{table}[!ht]
	\centering
    \caption{Average evaluation returns with a single standard deviation in DeepSea 10 for $\lambda \in \{0.01, 0.1, 0.25, 0.5, 1.0, 2.0, 4.0, 10.0, 100.0\}$.}
    \resizebox{\textwidth}{!}{
	\robustify\bf
	\begin{tabular}{l S S S S S S S S S}
		\toprule
		{Algorithm \textbackslash \ $\lambda$} & {0.01} & {0.1} & {0.25} & {0.5} & {1.0} & {2.0} & {4.0} & {10.0} & {100.0} \\
		\midrule
		 A2C Count &  0.93(21) &  0.96(17) & \bf 0.97(13) & \bf 0.98(8) & \bf 0.99(2) & \bf 0.98(7) &  0.77(21) &  0.47(16) &  0.17(16) \\
		 A2C Hash-Count &  0.94(20) &  0.96(17) & \bf 0.97(11) & \bf 0.98(8) & \bf 0.98(6) & \bf 0.99(2) & \bf 0.91(15) &  0.52(19) &  0.09(13) \\
		 A2C ICM &  0.92(22) &  0.92(22) & 0.93(20) & \bf 0.91(22) &  0.90(21) &  0.55(27) &  0.82(21) &  0.63(28) &  0.05(13) \\
		 A2C RND & 0.92(23) &  0.37(10) &  0.19(3) &  -0.00(0) &  0.19(2) &  -0.01(0) &  0.16(8) &  -0.01(0) &  -0.01(0) \\
		 A2C RIDE & -0.00(0) & -0.00(0) & -0.00(0) & -0.00(0) & -0.00(0) & -0.00(0) & -0.00(0) & -0.00(0) & -0.00(0) \\
		 \midrule
		 PPO Count & \bf 0.99(2) & \bf 0.98(4) & \bf 0.99(3) & \bf 0.93(11) &  0.85(11) &  0.87(13) &  0.73(16) &  0.56(25) &  0.17(20) \\
		 PPO Hash-Count &  0.59(2) & \bf 0.99(2) & \bf 0.98(4) & \bf 0.96(7) &  0.85(12) & \bf 0.97(7) &  0.81(20) &  0.64(22) &  0.16(21) \\
		 PPO ICM &  -0.00(0) &  0.85(26) & 0.89(15) &  0.81(18) &  0.70(31) &  0.79(18) &  0.68(25) &  0.59(20) &  0.04(9) \\
		 PPO RND &  0.20(0) &  -0.01(0) &  0.20(6) &  -0.01(0) & 0.29(11) &  -0.01(0) &  -0.01(0) &  0.05(9) &  -0.01(0) \\
		 PPO RIDE &  0.73(14) &  0.73(17) &  0.77(9) & \bf 0.91(17) & 0.94(12) &  0.74(15) & \bf 0.91(17) &  0.77(9) &  0.57(6) \\
		 \midrule
		 DeA2C Count &  0.91(24) &  0.95(18) & \bf 0.97(14) & \bf 0.96(15) & \bf 0.97(11) &  0.94(13) & \bf 0.97(13) & \bf 0.93(15) &  0.58(19) \\
		 DeA2C ICM &  0.93(22) &  0.92(24) &  0.92(24) & \bf 0.94(21) &  0.82(19) &  0.93(17) & \bf 0.96(11) & \bf 0.95(13) &  \bf0.84(23) \\
		 DePPO Count &  0.76(24) &  0.56(20) &  0.65(27) &  0.80(19) & 0.86(20) &  0.71(20) &  0.73(25) &  0.54(22) &  0.06(12) \\
		 DePPO ICM &  0.55(24) & 0.61(22) &  0.58(22) &  0.55(25) &  0.60(23) &  0.55(25) &  0.20(18) &  0.08(11) &  0.00(3) \\
		 DeDQN Count &  0.94(17) &  0.96(15) & \bf 0.97(14) & \bf 0.98(10) & \bf 0.98(10) & \bf 0.98(10) & \bf 0.98(10) &  0.77(21) & \bf 0.88(15) \\
		 DeDQN ICM &  0.94(18) &  0.92(21) &  0.91(24) & \bf 0.93(21) &  0.94(19) & 0.96(14) & \bf 0.95(15) & \bf 0.96(14) &  0.53(20) \\
		\bottomrule
	\end{tabular}
	}
	\label{tab:deepsea_10_intrinsic_coef_mean}
\end{table}

\clearpage

\begin{figure*}[h]
    \centering
    \includegraphics[width=\linewidth]{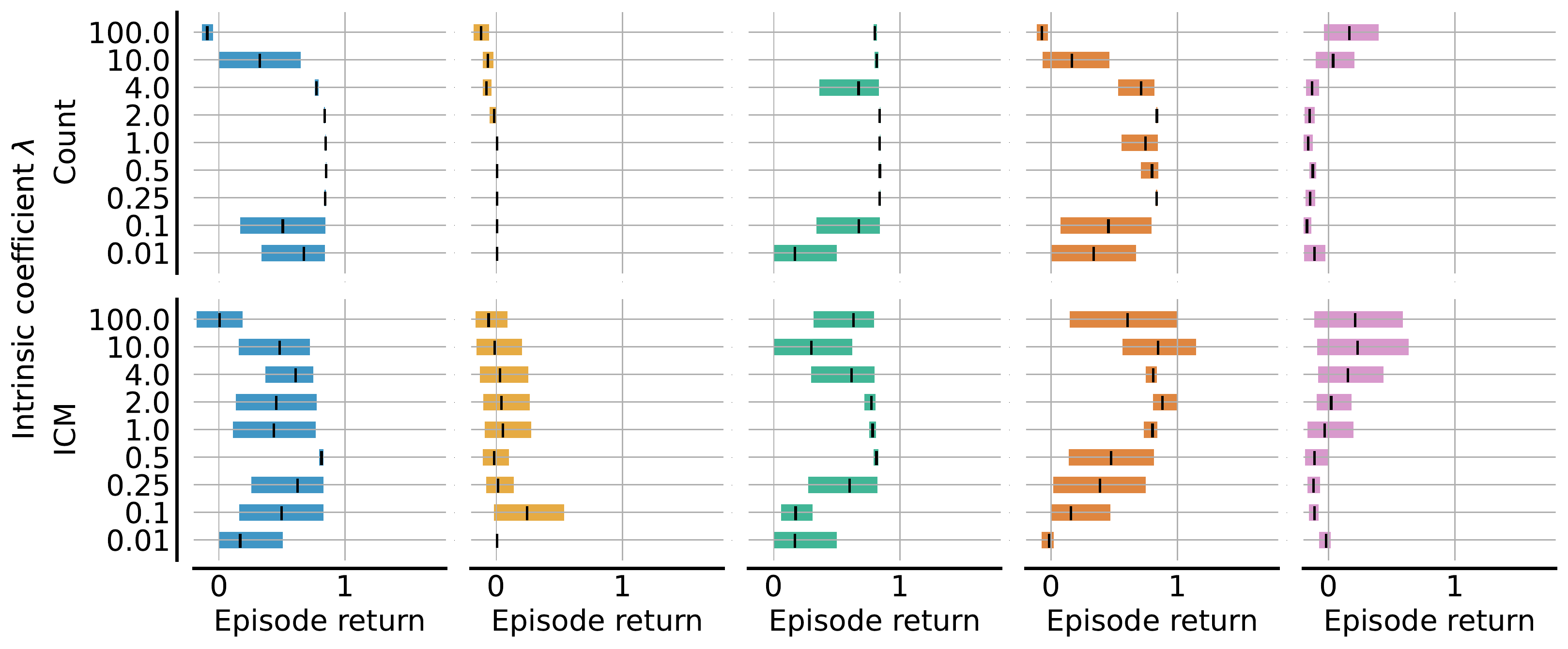}
    \includegraphics[width=.5\linewidth]{media/legend.pdf}
    
    \caption{Average evaluation returns for baselines and DeRL with Count (upper row) and ICM (lower row) intrinsic rewards in Hallway $N_l=N_r=10$ with $\lambda \in \{0.01, 0.1, 0.25, 0.5, 1.0, 2.0, 4.0, 10.0, 100.0\}$. Shading indicates 95\% confidence intervals.}
\end{figure*}

\begin{figure*}[h]
    \centering
    \begin{subfigure}{.33\textwidth}
        \centering
        \includegraphics[width=1\linewidth]{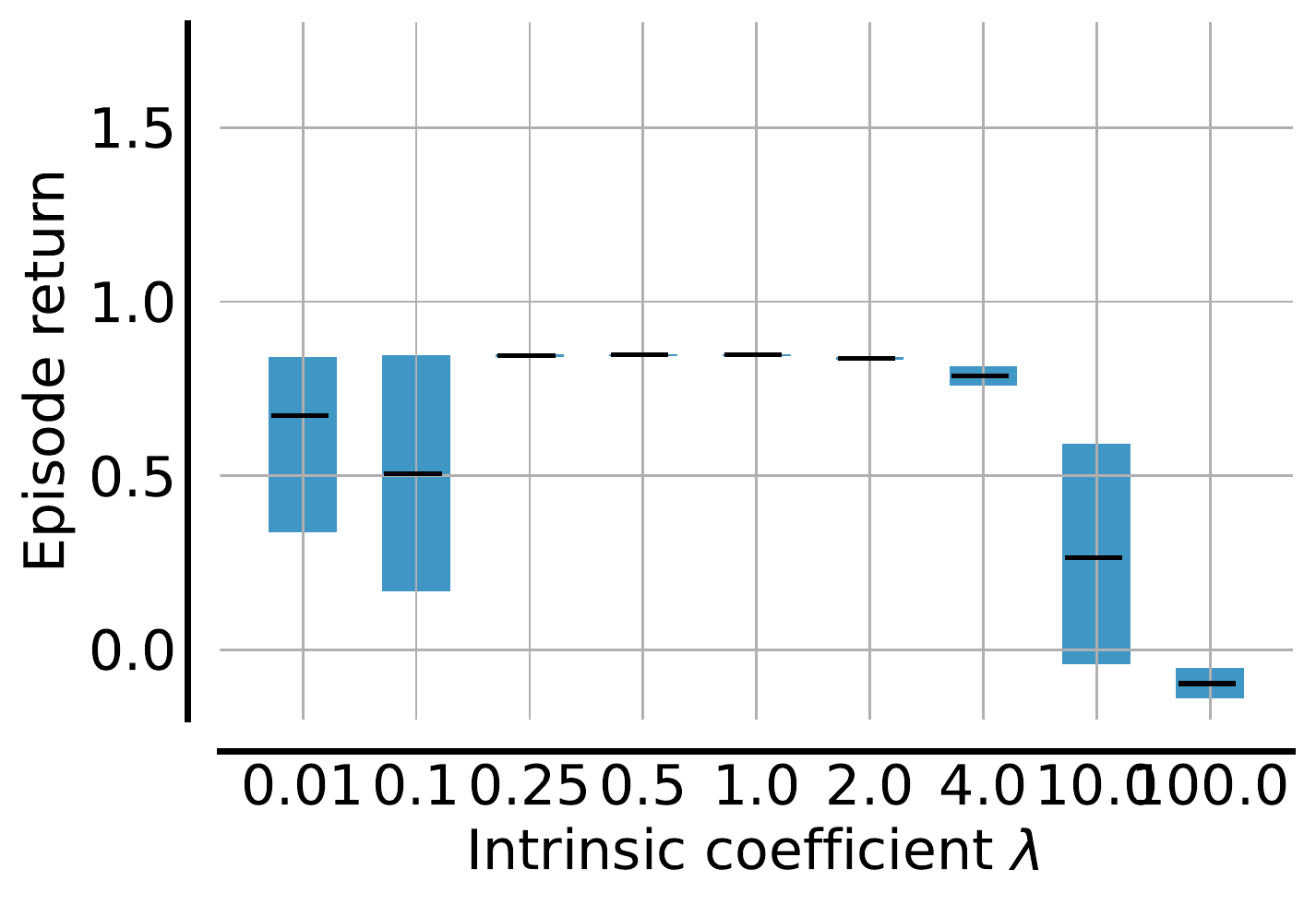}
        \caption{A2C Hash-Count}
    \end{subfigure}
    \hfill
    \begin{subfigure}{.33\textwidth}
        \centering
        \includegraphics[width=1\linewidth]{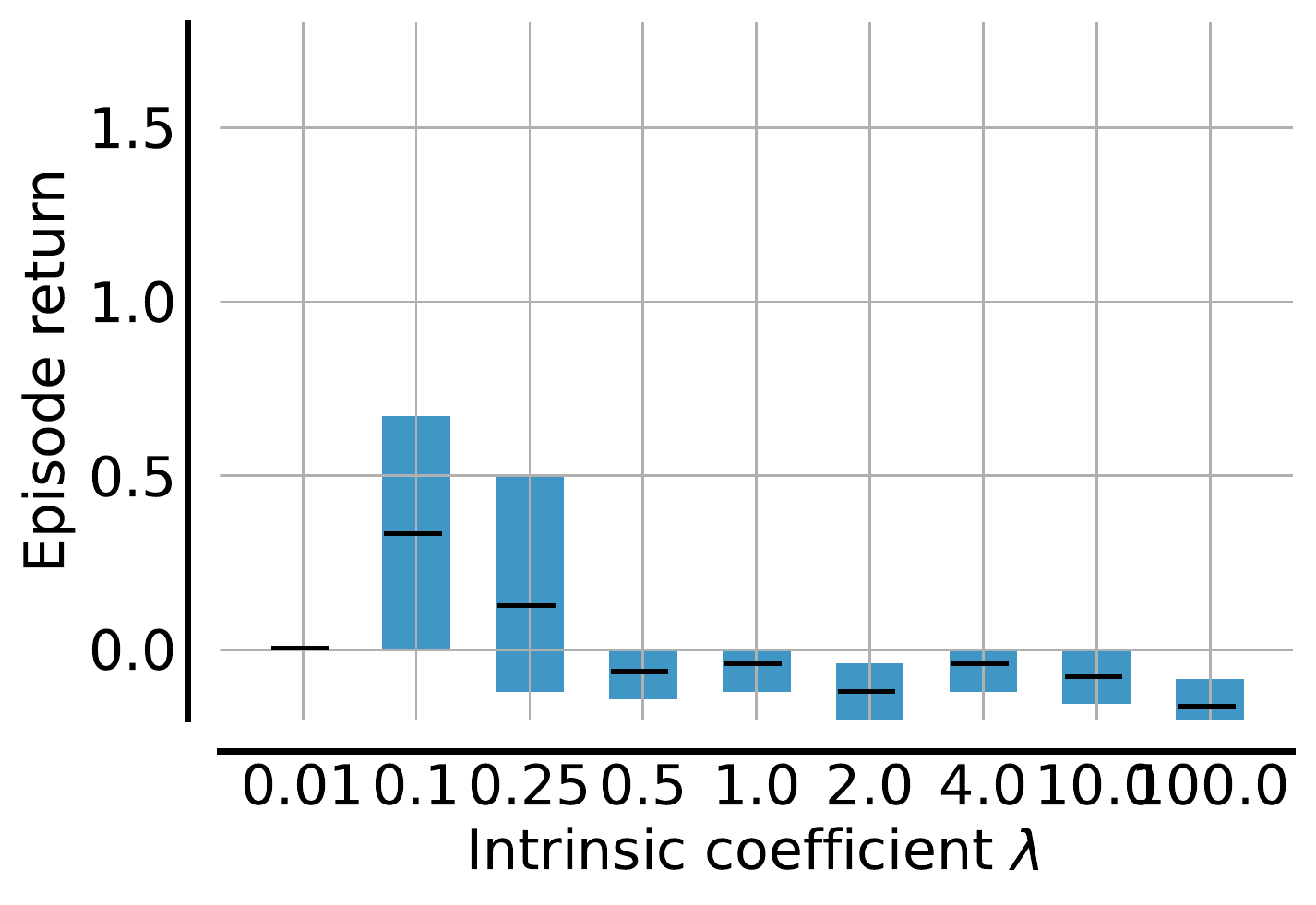}
        \caption{A2C RND}
    \end{subfigure}
    \hfill
    \begin{subfigure}{.33\textwidth}
        \centering
        \includegraphics[width=1\linewidth]{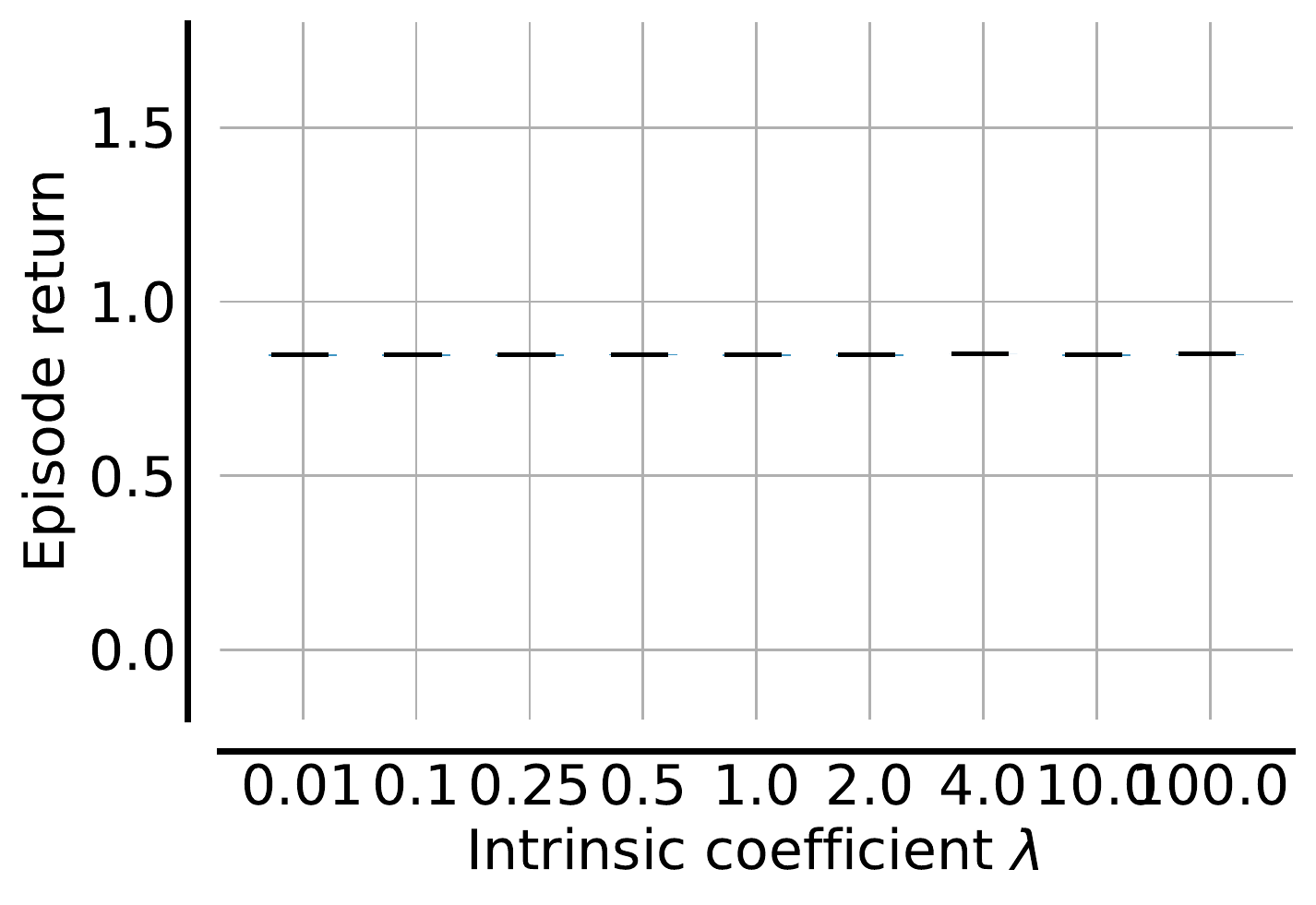}
        \caption{A2C RIDE}
    \end{subfigure}
    
    \begin{subfigure}{.33\textwidth}
        \centering
        \includegraphics[width=1\linewidth]{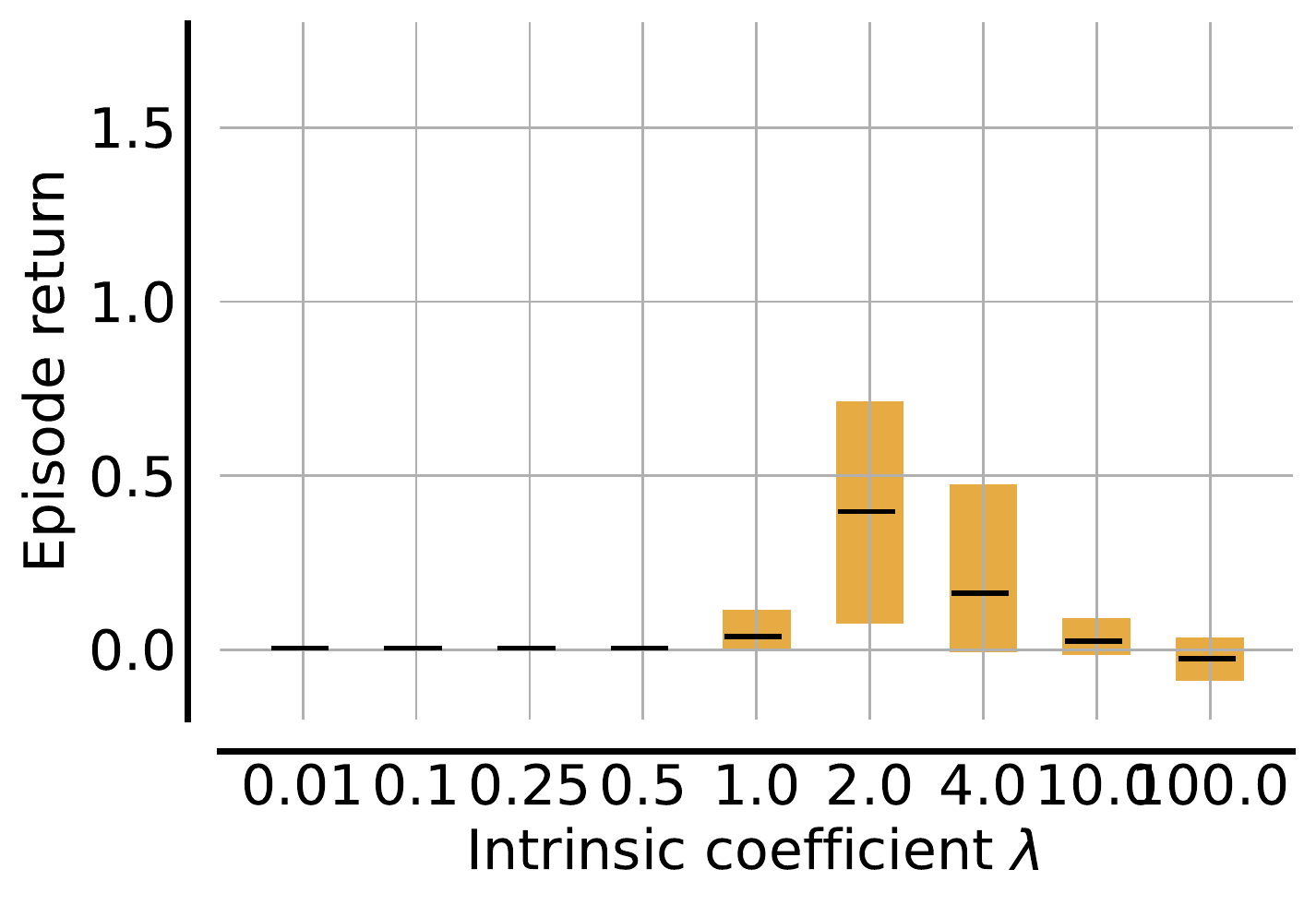}
        \caption{PPO Hash-Count}
    \end{subfigure}
    \hfill
    \begin{subfigure}{.33\textwidth}
        \centering
        \includegraphics[width=1\linewidth]{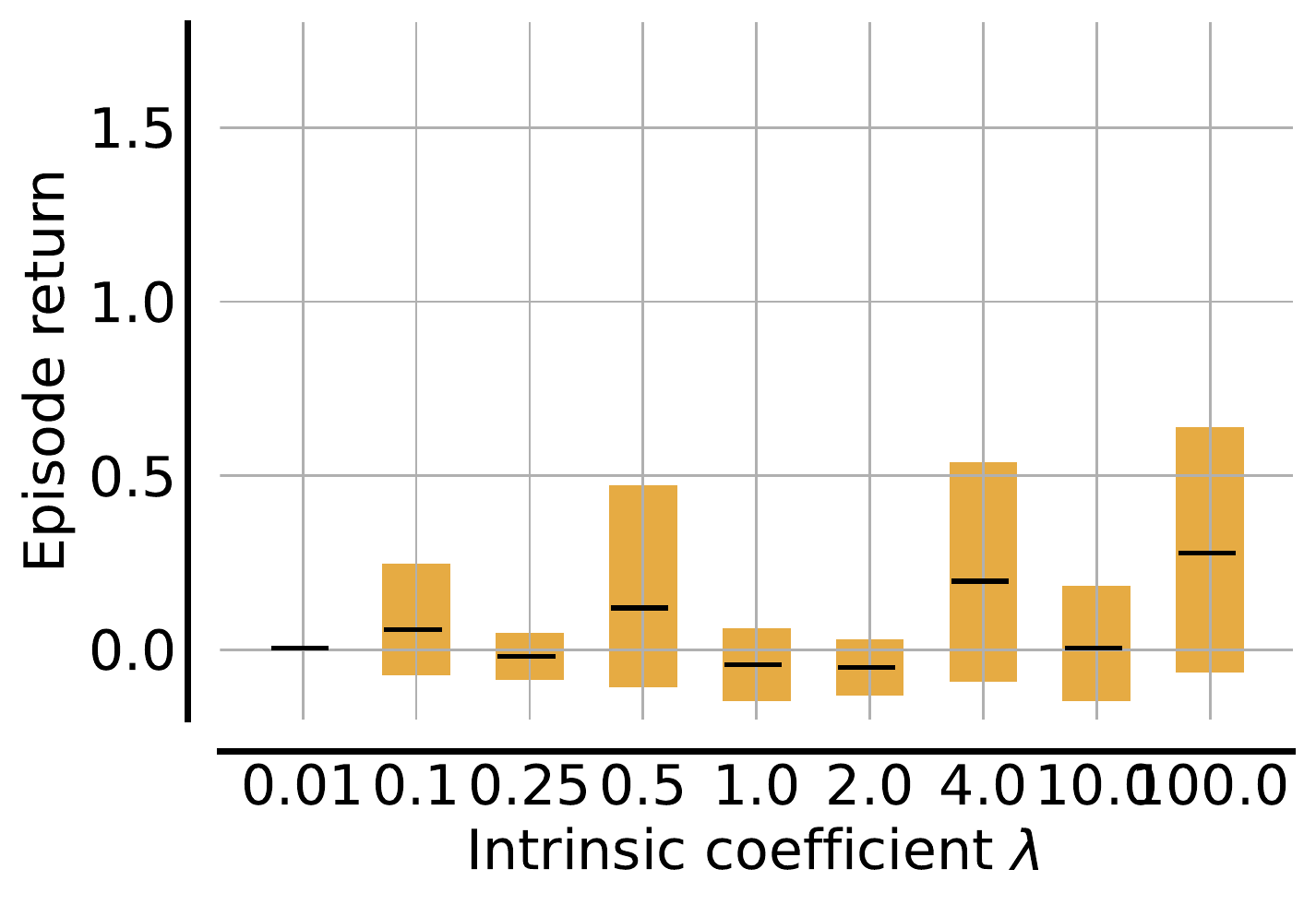}
        \caption{PPO RND}
    \end{subfigure}
    \hfill
    \begin{subfigure}{.33\textwidth}
        \centering
        \includegraphics[width=1\linewidth]{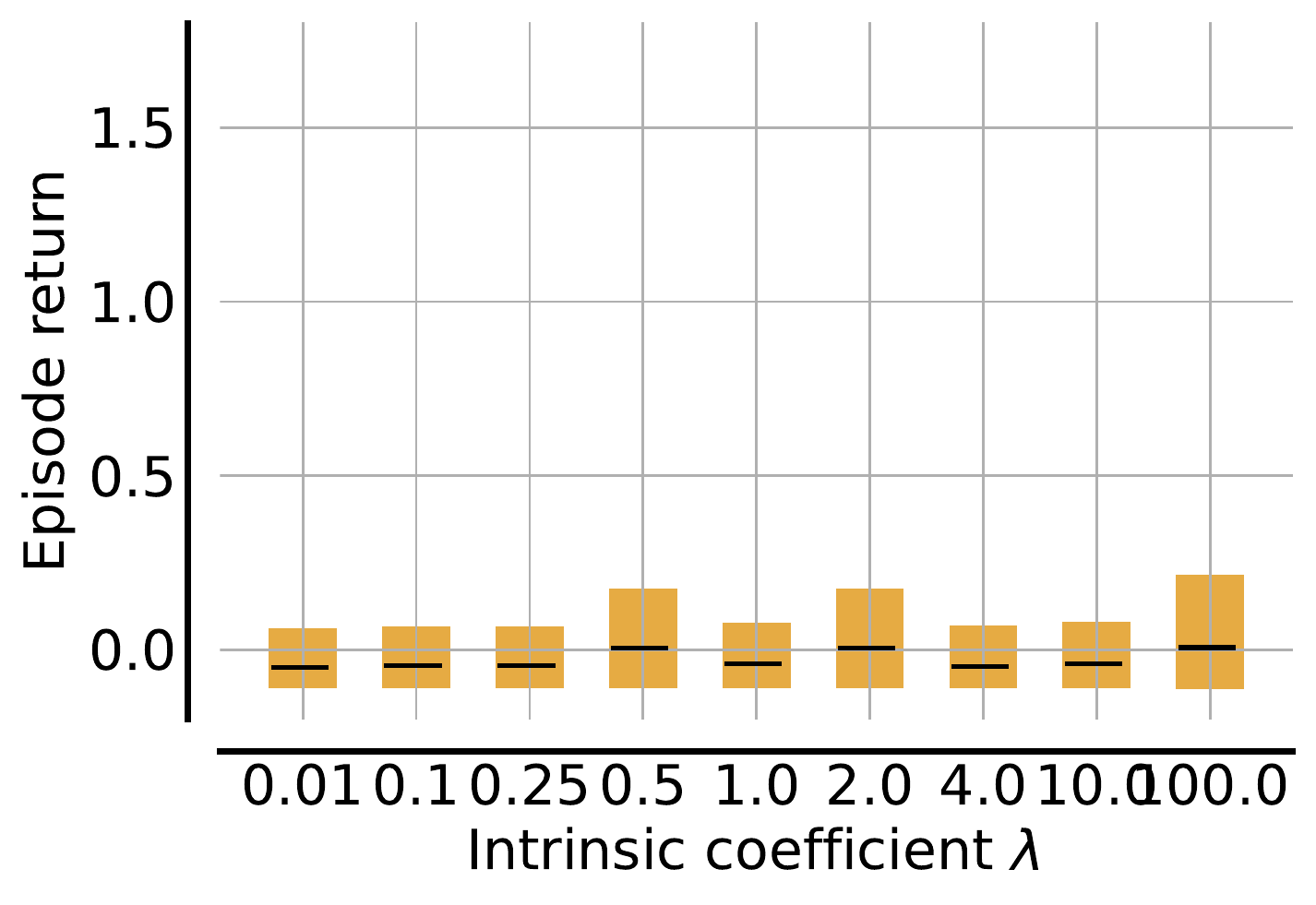}
        \caption{PPO RIDE}
    \end{subfigure}
    \caption{Average evaluation returns for A2C and PPO with Hash-Count, RND and RIDE intrinsic rewards in Hallway $N_l=N_r=10$ with $\lambda \in \{0.01, 0.1, 0.25, 0.5, 1.0, 2.0, 4.0, 10.0, 100.0\}$. Shading indicates 95\% confidence intervals.}
\end{figure*}

\begin{table}[!ht]
	\centering
	\caption{Maximum evaluation returns with a single standard deviation in Hallway $N_l=N_r=10$ for $\lambda \in \{0.01, 0.1, 0.25, 0.5, 1.0, 2.0, 4.0, 10.0, 100.0\}$.}
	\resizebox{\textwidth}{!}{
	\robustify\bf
	\begin{tabular}{l S S S S S S S S S}
		\toprule
		{Algorithm \textbackslash \ $\lambda$} & {0.01} & {0.1} & {0.25} & {0.5} & {1.0} & {2.0} & {4.0} & {10.0} & {100.0} \\
		\midrule
		 A2C Count &  0.68(26) &  0.51(34) & \bf 0.85(0) & \bf 0.85(0) & \bf 0.85(0) & \bf 0.85(0) & \bf 0.85(0) &  0.51(34) &  0.80(0) \\
		 A2C Hash-Count &  0.68(26) &  0.51(34) & \bf 0.85(0) & \bf 0.85(0) & \bf 0.85(0) & \bf 0.85(0) & \bf 0.85(0) &  0.51(34) &  0.80(0) \\
		 A2C ICM &  0.17(26) &  0.68(26) & \bf 0.85(0) & \bf 0.85(0) &  0.51(34) &  0.51(34) & \bf 0.85(0) & 0.89(72) &  0.28(40) \\
		 A2C RND &  0.00(0) & 0.34(34) &  0.13(32) &  -0.06(7) &  -0.04(6) &  -0.08(8) &  -0.04(6) &  -0.06(7) &  -0.16(6) \\
		 A2C RIDE & \bf 0.85(0) & \bf 0.85(0) & \bf 0.85(0) & \bf 0.85(0) & \bf 0.85(0) & \bf 0.85(0) & \bf 0.85(0) & 0.85(0) & \bf 0.85(0) \\
		 \midrule
		 PPO Count & 0.00(0) & 0.00(0) & 0.00(0) & 0.00(0) & 0.00(0) & 0.00(0) &  -0.06(4) & 0.00(0) &  -0.06(7) \\
		 PPO Hash-Count &  0.00(0) &  0.00(0) &  0.00(0) &  0.00(0) &  0.17(26) & 0.64(24) &  0.32(32) &  0.17(25) &  0.14(27) \\
		 PPO ICM &  0.00(0) & 0.51(34) &  0.26(38) &  0.32(32) &  0.33(33) &  0.32(36) &  0.33(34) &  0.42(40) &  0.24(40) \\
		 PPO RND &  0.16(24) &  0.28(37) &  0.32(32) &  0.48(32) &  0.80(0) &  0.61(31) &  0.64(24) &  0.60(30) & 0.81(1) \\
		 PPO RIDE &  0.10(28) &  0.26(37) &  0.26(37) &  0.30(57) & 0.43(37) &  0.30(57) &  0.29(57) &  0.30(57) &  0.26(37) \\
		 \midrule
		 DeA2C Count &  0.17(26) &  0.68(26) & \bf 0.85(0) & \bf 0.85(0) & \bf 0.85(0) & \bf 0.85(0) &  0.82(2) &  0.84(2) &  0.82(2) \\
		 DeA2C ICM &  0.17(26) & \bf 0.85(0) &  0.68(26) & \bf 0.85(0) & \bf 0.85(0) & \bf 0.85(0) &  0.67(26) &  0.34(34) &  0.81(1) \\
		 DePPO Count &  0.34(34) &  0.68(26) & \bf 0.85(0) & \bf 0.85(0) & \bf 0.85(0) & \bf 0.85(0) & \bf 0.85(0) &  0.34(34) &  0.80(0) \\
		 DePPO ICM &  0.17(26) &  0.17(26) &  0.68(26) &  0.51(34) & \bf 0.85(0) & \bf 1.04(28) & \bf 0.85(1) & 1.04(28) & \bf 0.85(56) \\
		 DeDQN Count &  0.26(42) &  0.28(40) &  0.32(60) &  0.43(42) &  0.43(42) &  0.68(76) &  0.47(40) & \bf 1.60(30) & \bf 1.23(38) \\
		 DeDQN ICM &  0.32(60) &  0.11(31) &  0.25(60) &  0.22(61) &  0.34(57) &  0.50(63) &  0.66(26) &  0.64(57) & 0.70(54) \\
		\bottomrule
	\end{tabular}
	}
	\label{tab:hallway10_10_intrinsic_coef_max}
\end{table}

\begin{table}[!ht]
	\centering
    \caption{Average evaluation returns with a single standard deviation in Hallway $N_l=N_r=10$ for $\lambda \in \{0.01, 0.1, 0.25, 0.5, 1.0, 2.0, 4.0, 10.0, 100.0\}$.}
    \resizebox{\textwidth}{!}{
	\robustify\bf
	\begin{tabular}{l S S S S S S S S S}
		\toprule
		{Algorithm \textbackslash \ $\lambda$} & {0.01} & {0.1} & {0.25} & {0.5} & {1.0} & {2.0} & {4.0} & {10.0} & {100.0} \\
		\midrule
		 A2C Count &  0.67(7) &  0.51(3) & \bf 0.84(7) & \bf 0.85(1) & \bf 0.85(2) & \bf 0.84(10) &  0.77(24) &  0.32(18) &  -0.09(18) \\
		 A2C Hash-Count &  0.67(7) &  0.51(2) & \bf 0.84(5) & \bf 0.85(2) & \bf 0.85(2) & \bf 0.84(10) &  0.79(21) &  0.27(18) &  -0.10(19) \\
		 A2C ICM &  0.17(2) &  0.50(9) &  0.62(13) & 0.81(15) &  0.44(14) &  0.45(17) &  0.61(28) &  0.48(27) &  -0.00(18) \\
		 A2C RND &  0.00(0) & 0.33(4) &  0.13(2) &  -0.06(0) &  -0.04(0) &  -0.12(0) &  -0.04(0) &  -0.08(1) &  -0.16(1) \\
		 A2C RIDE & \bf 0.85(2) & \bf 0.85(2) & \bf 0.85(2) & \bf 0.85(1) & \bf 0.85(2) & \bf 0.85(2) & \bf 0.85(0) & \bf 0.85(2) & \bf 0.85(1) \\
		 \midrule
		 PPO Count & 0.00(0) & 0.00(0) & 0.00(0) & 0.00(0) & 0.00(0) &  -0.02(1) &  -0.08(1) &  -0.07(3) &  -0.12(2) \\
		 PPO Hash-Count &  0.00(0) &  0.00(0) &  0.00(0) &  0.00(0) &  0.04(7) & 0.40(9) &  0.16(6) &  0.03(6) &  -0.03(5) \\
		 PPO ICM &  0.00(0) & 0.24(20) &  0.01(9) &  -0.02(9) &  0.05(13) &  0.04(12) &  0.03(13) &  -0.01(13) &  -0.06(11) \\
		 PPO RND &  0.00(2) &  0.06(9) &  -0.02(7) &  0.12(7) &  -0.04(17) &  -0.05(9) &  0.20(20) &  0.00(17) & 0.28(17) \\
		 PPO RIDE &  -0.05(8) &  -0.05(9) &  -0.05(9) &  0.00(14) &  -0.04(10) &  0.00(14) &  -0.05(9) &  -0.04(11) & 0.01(11) \\
		 \midrule
		 DeA2C Count &  0.17(2) &  0.67(7) & \bf 0.84(9) & \bf 0.84(8) & \bf 0.84(9) & \bf 0.84(8) &  0.67(10) &  0.82(12) &  0.80(6) \\
		 DeA2C ICM &  0.17(3) &  0.17(24) &  0.60(14) & 0.81(14) &  0.78(20) & \bf 0.77(20) &  0.62(16) &  0.30(9) &  0.63(6) \\
		 DePPO Count &  0.34(3) &  0.45(13) & \bf 0.84(9) &  0.80(10) &  0.75(12) & \bf 0.84(9) &  0.71(19) &  0.16(16) &  -0.07(19) \\
		 DePPO ICM &  -0.02(4) &  0.15(5) &  0.39(13) &  0.48(9) &  0.80(12) & \bf 0.88(14) &  0.81(12) & \bf 0.85(20) &  0.61(17) \\
		 DeDQN Count &  -0.11(4) &  -0.17(5) &  -0.15(5) &  -0.13(8) &  -0.16(7) &  -0.15(11) &  -0.13(10) &  0.04(35) & 0.17(33) \\
		 DeDQN ICM &  -0.02(4) &  -0.11(5) &  -0.12(6) &  -0.11(10) &  -0.03(16) &  0.02(14) &  0.15(20) & 0.23(18) &  0.21(19) \\
		\bottomrule
	\end{tabular}
	}
	\label{tab:hallway10_10_intrinsic_coef_mean}
\end{table}

\clearpage
\subsection{Intrinsic Reward Decay}
\label{app:intrinsic_reward_decay}
Second, we evaluate all baselines and DeRL algorithms with varying rate of decay of intrinsic rewards in DeepSea and Hallway. For count-based intrinsic rewards, the rate of decay can be determined by the increment of the state count $N(s)$. By default, the count is incremented by $1$ for each state occurrence. For this analysis, we consider increments $\{0.01, 0.1, 0.2, 1.0, 5.0, 10.0, 100.0\}$. For deep prediction-based intrinsic rewards, ICM, RND and RIDE, the rate of decay is determined by their learning rates. We consider learning rates  $\{1e^{-9}, 1e^{-8}, 2e^{-8}, 1e^{-7}, 5e^{-7}, 1e^{-6}, 1e^{-5}, 1e^{-4}, 1e^{-3}\}$.

\begin{figure*}[h]
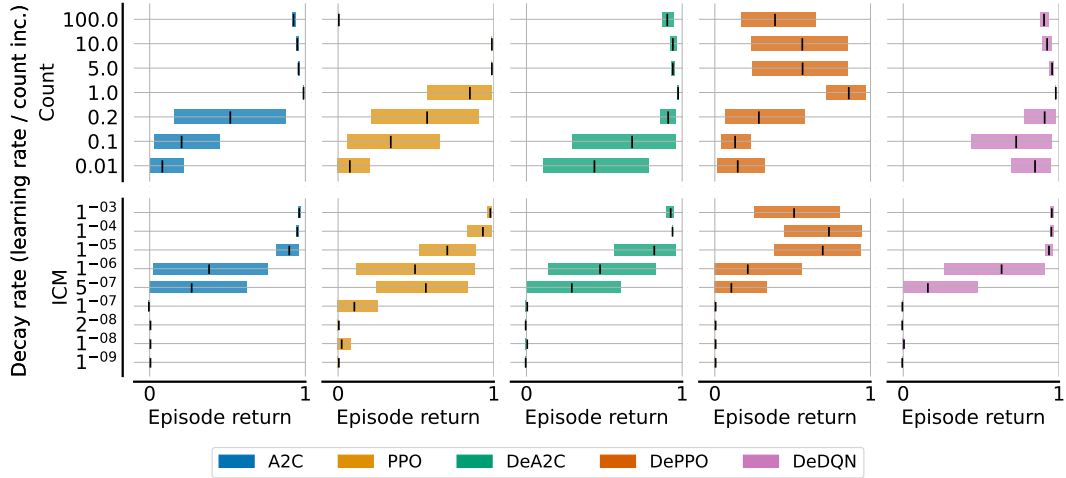

    \centering
    \includegraphics[width=.8\linewidth]{media/results_deepsea/decay_speed/deepsea10_agg_decay_bar.pdf}
    \includegraphics[width=.5\linewidth]{media/legend.pdf}
    
    \caption{Average evaluation returns for baselines and DeRL with Count (upper row) and ICM (lower row) intrinsic rewards in DeepSea 10 with varying count increments and learning rates. Shading indicates 95\% confidence intervals.}
\end{figure*}

\begin{figure*}[h]
    \centering
    \begin{subfigure}{.33\textwidth}
        \centering
        \includegraphics[width=1\linewidth]{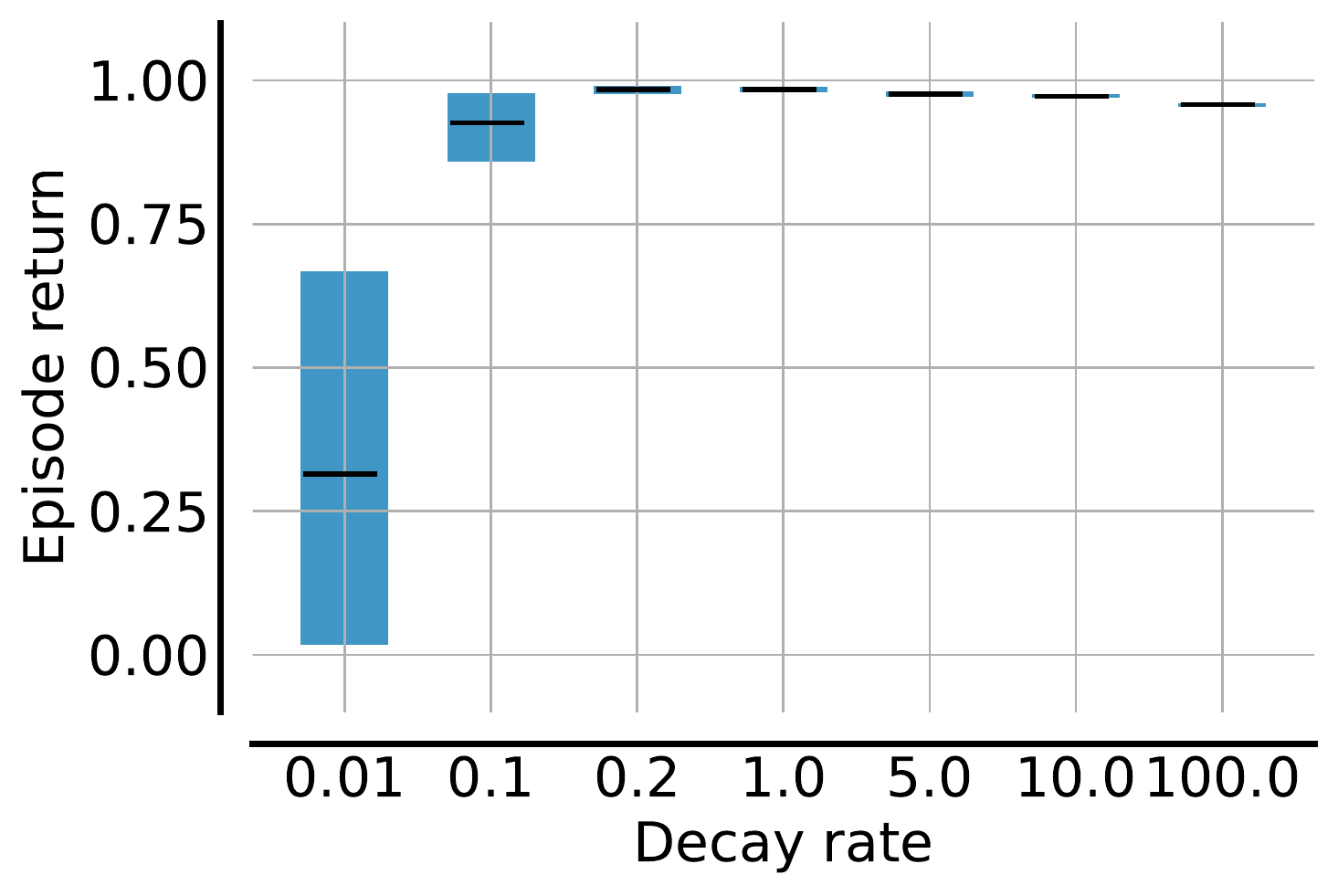}
        \caption{A2C Hash-Count}
    \end{subfigure}
    \hfill
    \begin{subfigure}{.33\textwidth}
        \centering
        \includegraphics[width=1\linewidth]{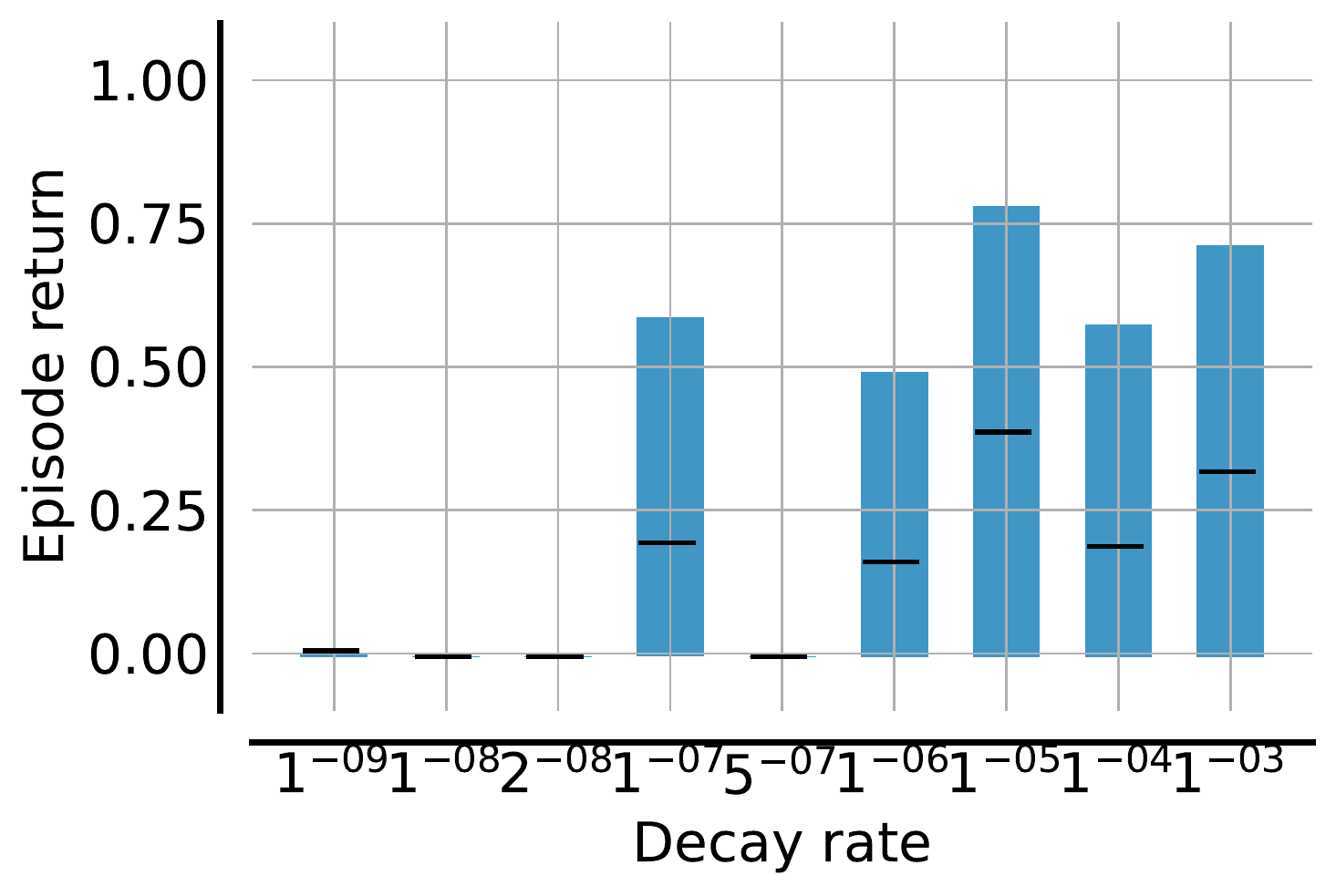}
        \caption{A2C RND}
    \end{subfigure}
    \hfill
    \begin{subfigure}{.33\textwidth}
        \centering
        \includegraphics[width=1\linewidth]{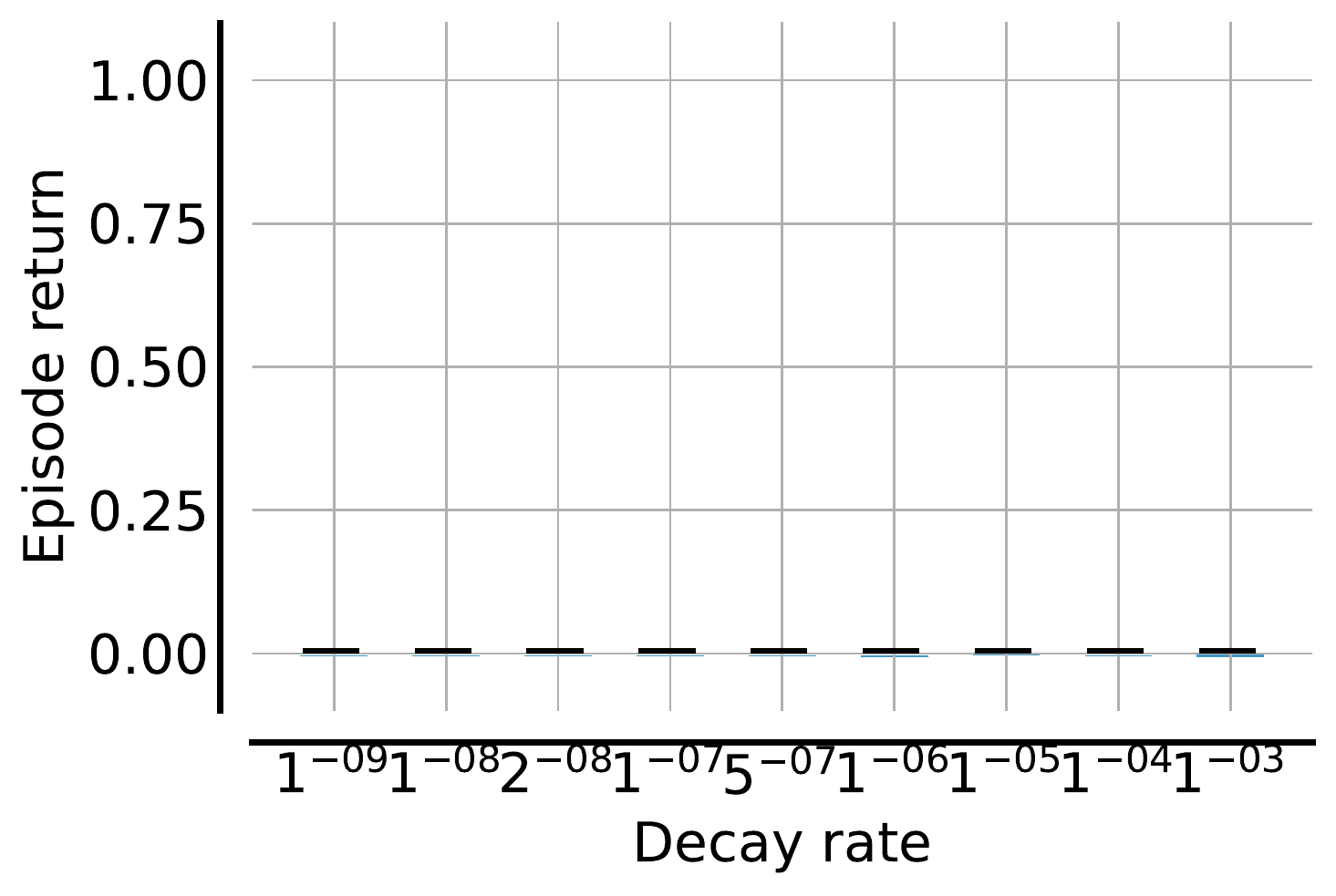}
        \caption{A2C RIDE}
    \end{subfigure}
    
    \begin{subfigure}{.33\textwidth}
        \centering
        \includegraphics[width=1\linewidth]{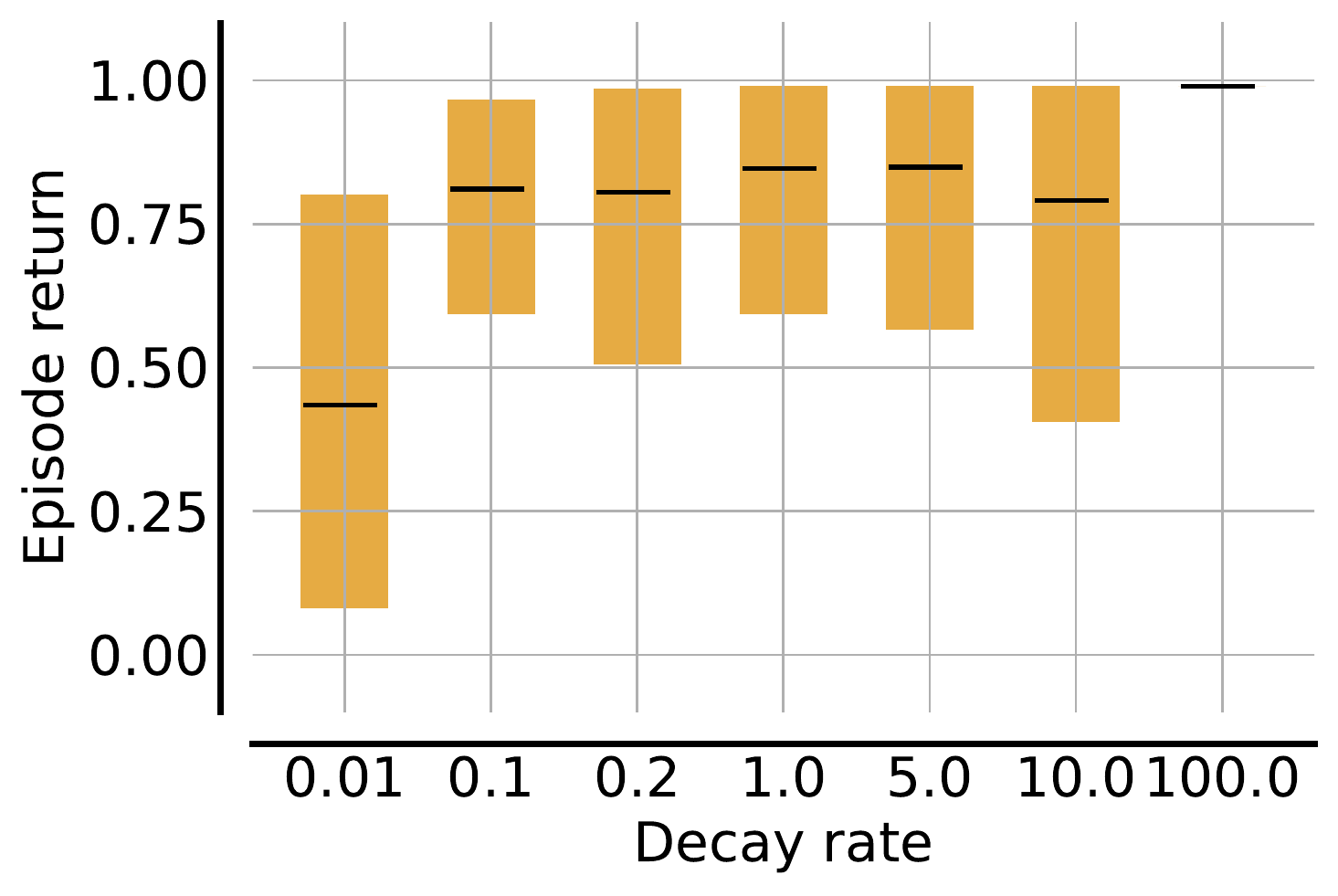}
        \caption{PPO Hash-Count}
    \end{subfigure}
    \hfill
    \begin{subfigure}{.33\textwidth}
        \centering
        \includegraphics[width=1\linewidth]{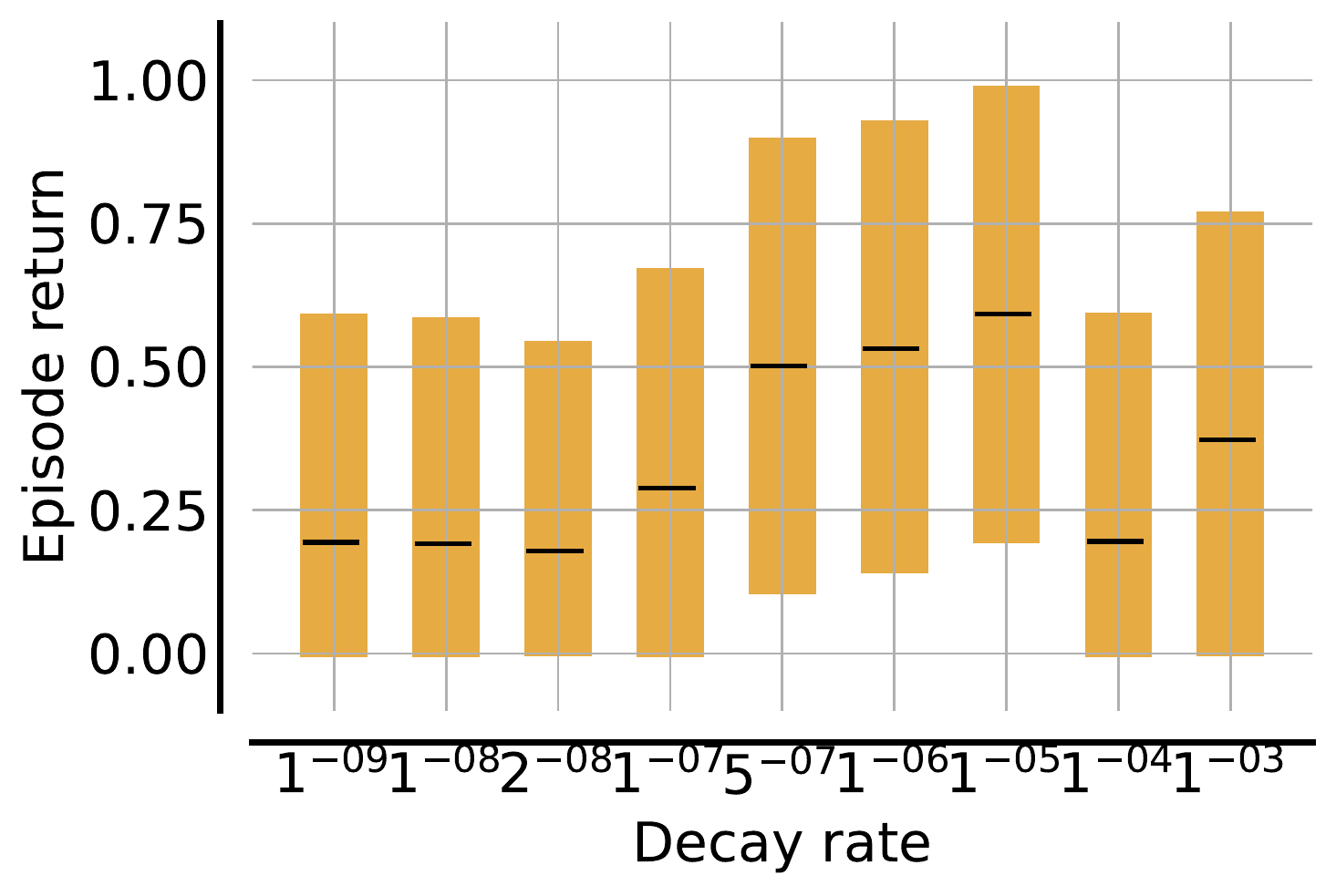}
        \caption{PPO RND}
    \end{subfigure}
    \hfill
    \begin{subfigure}{.33\textwidth}
        \centering
        \includegraphics[width=1\linewidth]{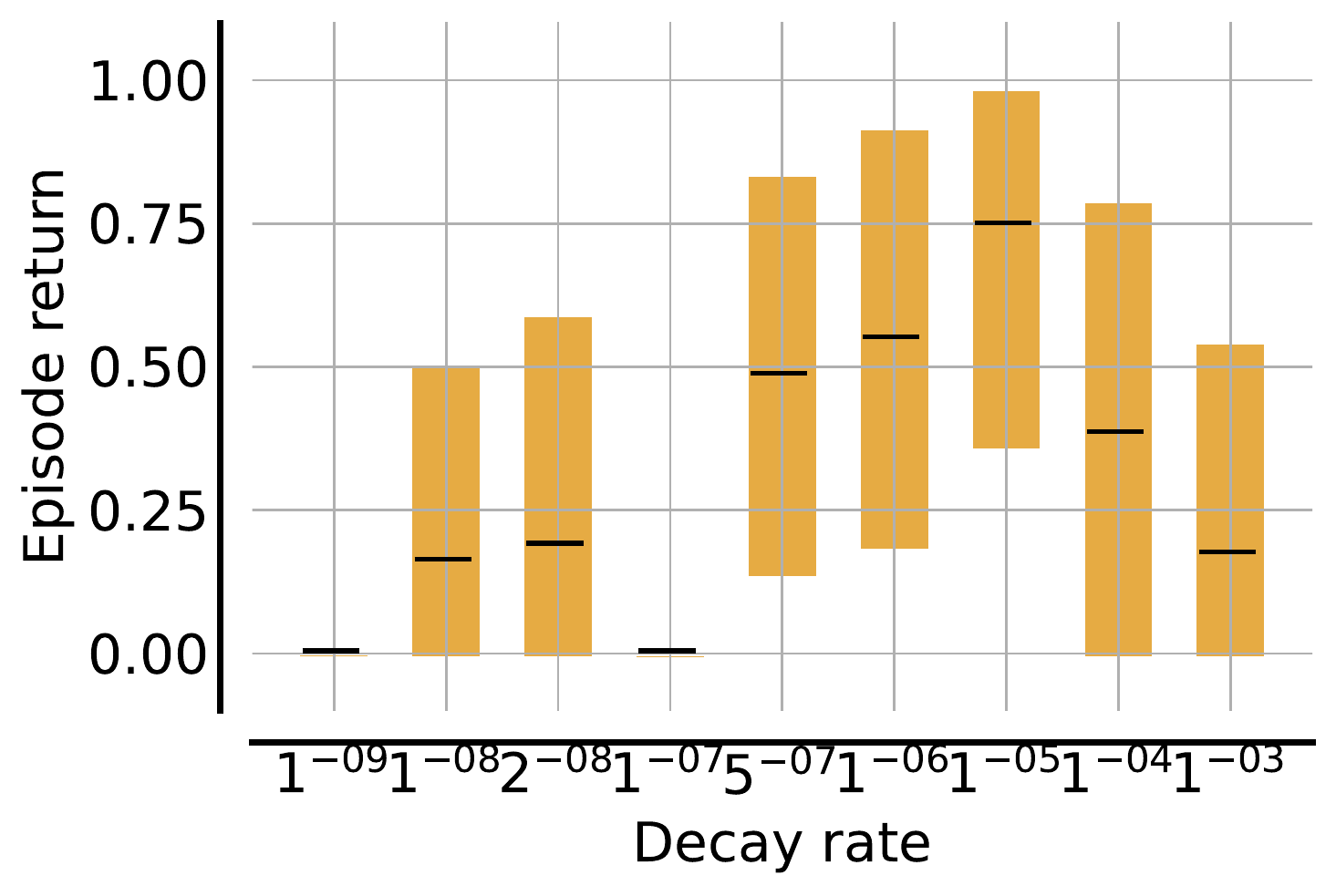}
        \caption{PPO RIDE}
    \end{subfigure}
    \caption{Average evaluation returns for A2C and PPO with Hash-Count, RND and RIDE intrinsic rewards in DeepSea 10 with varying count increments and learning rates. Shading indicates 95\% confidence intervals.}
\end{figure*}

\begin{table}[hbt!]
	\centering
	\caption{Maximum evaluation returns with a single standard deviation in DeepSea 10 with count-based intrinsic rewards and varying count increments.}
	\resizebox{\linewidth}{!}{
	\robustify\bf
	\begin{tabular}{l S S S S S S S}
		\toprule
		{Algorithm \textbackslash \ Count increment} & {0.01} & {0.1} & {0.2} & {1.0} & {5.0} & {10.0} & {100.0} \\
		\midrule
		 A2C Count & \bf 0.99(0) & \bf 0.99(0) &  0.79(30) & \bf 0.99(0) & \bf 0.99(0) & \bf 0.99(0) & \bf 0.99(0) \\
		 A2C Hash-Count & \bf 0.99(0) & \bf 0.99(0) & \bf 0.99(0) & \bf 0.99(0) & \bf 0.99(0) & \bf 0.99(0) & \bf 0.99(0) \\
		 \midrule
		 PPO Count &  0.59(40) & \bf 0.99(0) & \bf 0.99(0) & \bf 0.99(0) & \bf 0.99(0) & \bf 0.99(0) &  0.00(0) \\
		 PPO Hash-Count &  0.79(30) & \bf 0.99(0) & \bf 0.99(0) & \bf 0.99(0) & \bf 0.99(0) & \bf 0.99(0) & \bf 0.99(0) \\
		 \midrule
		 DeA2C Count & \bf 0.99(0) & \bf 0.99(0) & \bf 0.99(0) & \bf 0.99(0) & \bf 0.99(0) & \bf 0.99(0) & \bf 0.99(0) \\
		 DePPO Count &  0.79(30) & \bf 0.99(0) & \bf 0.99(0) & \bf 0.99(0) & \bf 0.99(0) & \bf 0.99(0) & \bf 0.99(0) \\
		 DeDQN Count & \bf 0.99(0) & \bf 0.99(0) & \bf 0.99(0) & \bf 0.99(0) & \bf 0.99(0) & \bf 0.99(0) & \bf 0.99(0) \\
		\bottomrule
	\end{tabular}
	}
	\label{tab:decay_sensitivity_count_deepsea_max}
\end{table}

\begin{table}[hbt!]
	\centering
	\caption{Average evaluation returns with a single standard deviation in DeepSea 10 with count-based intrinsic rewards and varying count increments.}
	\resizebox{\linewidth}{!}{
	\robustify\bf
	\begin{tabular}{l S S S S S S S}
		\toprule
		{Algorithm \textbackslash \ Count increment} & {0.01} & {0.1} & {0.2} & {1.0} & {5.0} & {10.0} & {100.0} \\
		\midrule
		 A2C Count &  0.08(15) &  0.20(22) &  0.52(19) & \bf 0.99(2) &  0.96(17) &  0.95(19) &  0.92(23) \\
		 A2C Hash-Count &  0.31(17) & \bf 0.93(16) & \bf 0.98(4) & \bf 0.98(6) & \bf 0.98(11) & \bf 0.97(13) &  0.96(17) \\
		 \midrule
		 PPO Count &  0.08(13) &  0.34(23) &  0.57(20) &  0.85(11) & \bf 0.99(2) & \bf 0.99(2) &  0.00(0) \\
		 PPO Hash-Count &  0.43(19) & \bf 0.81(21) &  0.80(14) &  0.85(12) &  0.85(9) &  0.79(5) & \bf 0.99(0) \\
		 \midrule
		 DeA2C Count &  0.44(22) &  0.68(15) &  0.91(21) & \bf 0.97(11) &  0.94(20) &  0.94(19) &  0.90(24) \\
		 DePPO Count &  0.15(21) &  0.13(25) &  0.28(25) & 0.86(20) &  0.56(26) &  0.56(26) &  0.39(32) \\
		 DeDQN Count & \bf 0.85(23) &  0.73(21) &  0.91(14) & \bf 0.98(10) &  0.96(16) &  0.92(21) &  0.90(25) \\
		\bottomrule
	\end{tabular}
	}
	\label{tab:decay_sensitivity_count_deepsea_mean}
\end{table}

\begin{table}[hbt!]
	\centering
	\caption{Maximum evaluation returns with a single standard deviation in DeepSea 10 with prediction-based intrinsic rewards and varying learning rates.}
	\resizebox{\linewidth}{!}{
	\robustify\bf
	\begin{tabular}{l S S S S S S S S S}
		\toprule
		{Algorithm \textbackslash \ Learning rate} & {1e-09} & {1e-08} & {2e-08} & {1e-07} & {5e-07} & {1e-06} & {1e-05} & {0.0001} & {0.001} \\
		\midrule
		 A2C ICM & \bf -0.00(0) & \bf -0.00(0) &  \bf-0.00(0) & \bf -0.00(0) &  0.39(40) & \bf 0.79(30) & \bf 0.99(0) & \bf 0.99(0) & \bf 0.99(0) \\
		 A2C RND & \bf 0.19(30) & \bf -0.00(0) & \bf -0.01(0) & \bf 0.20(30) &  -0.00(0) &  0.19(30) & 0.39(40) &  0.19(30) & 0.39(40) \\
		 A2C RIDE & \bf -0.00(0) & \bf -0.00(0) & \bf -0.00(0) & \bf -0.00(0) & -0.00(0) & -0.00(0) & -0.00(0) & -0.00(0) & -0.00(0) \\
		 \midrule
		 PPO ICM & \bf -0.01(0) & \bf 0.19(30) & \bf -0.00(0) & \bf 0.39(40) & \bf 0.99(0) & \bf 0.59(40) & \bf 0.99(0) & \bf 0.99(0) & \bf 0.99(0) \\
		 PPO RND & \bf 0.19(30) & \bf 0.19(30) & \bf 0.19(30) & \bf 0.39(40) & 0.59(40) & \bf 0.59(40) & 0.59(40) &  0.39(40) &  0.39(40) \\
		 PPO RIDE & \bf -0.00(0) & \bf 0.19(30) & \bf 0.20(30) & \bf -0.00(0) & 0.79(30) & \bf 0.79(30) & 0.79(30) &  0.39(40) &  0.20(30) \\
		 \midrule
		 DeA2C ICM & \bf -0.00(0) & \bf -0.00(0) & \bf -0.00(0) & \bf -0.00(0) &  0.39(40) & \bf 0.59(40) & \bf 0.99(0) & \bf 0.99(0) & \bf 0.99(0) \\
		 DePPO ICM & \bf -0.00(0) & \bf -0.00(0) & \bf -0.00(0) & \bf -0.00(0) &  0.19(30) &  0.39(40) & \bf 0.99(0) & \bf 0.99(0) & \bf 0.99(0) \\
		 DeDQN ICM & \bf -0.00(0) & \bf 0.19(30) & \bf -0.00(0) & \bf -0.00(0) &  0.39(40) & \bf 0.79(30) & \bf 0.99(0) & \bf 0.99(0) & \bf 0.99(0) \\
		\bottomrule
	\end{tabular}
	}
	\label{tab:decay_sensitivity_deep_deepsea_max}
\end{table}

\begin{table}[hbt!]
	\centering
	\caption{Average evaluation returns with a single standard deviation in DeepSea 10 with prediction-based intrinsic rewards and varying learning rates.}
	\resizebox{\linewidth}{!}{
	\robustify\bf
	\begin{tabular}{l S S S S S S S S S}
		\toprule
		{Algorithm \textbackslash \ Learning rate} & {1e-09} & {1e-08} & {2e-08} & {1e-07} & {5e-07} & {1e-06} & {1e-05} & {0.0001} & {0.001} \\
		\midrule
		 A2C ICM &  -0.00(0) &  -0.00(0) &  -0.00(0) &  -0.01(0) &  0.27(14) &  0.38(13) & \bf 0.90(21) & \bf 0.95(19) & \bf 0.96(16) \\
		 A2C RND &  -0.00(2) &  -0.01(0) &  -0.01(0) & \bf 0.19(2) &  -0.01(0) &  0.16(7) & 0.39(4) &  0.19(3) &  0.32(10) \\
		 A2C RIDE & -0.00(0) & -0.00(0) & -0.00(0) & -0.00(0) & -0.00(0) & -0.00(0) & -0.00(0) & -0.00(0) & -0.00(0) \\
		 \midrule
		 PPO ICM &  -0.00(0) &  0.02(7) &  -0.00(0) &  0.10(15) & \bf 0.56(28) & \bf 0.49(15) &  0.70(31) & \bf 0.93(12) & \bf 0.98(7) \\
		 PPO RND &  \bf 0.19(0) & \bf 0.19(2) & \bf 0.18(5) & \bf 0.29(11) & \bf 0.50(10) & \bf 0.53(10) & 0.59(0) &  0.20(2) &  0.37(6) \\
		 PPO RIDE &  -0.00(0) &  0.16(7) & \bf 0.19(2) &  -0.00(0) & \bf 0.49(22) & \bf 0.55(18) & \bf 0.75(11) &  0.39(3) &  0.18(6) \\
		 \midrule
		 DeA2C ICM &  -0.01(0) &  -0.00(0) &  -0.01(0) &  -0.00(0) & \bf 0.29(16) & \bf 0.47(21) & \bf 0.82(19) & \bf 0.94(22) & \bf 0.93(22) \\
		 DePPO ICM &  -0.00(0) &  -0.00(0) &  -0.00(0) &  -0.00(0) &  0.11(10) &  0.21(12) &  0.69(22) & 0.73(22) &  0.51(31) \\
		 DeDQN ICM &  -0.01(0) &  -0.00(2) &  -0.01(0) &  -0.01(0) &  0.16(8) & \bf 0.63(21) & \bf 0.94(19) & \bf 0.95(17) & \bf 0.95(16) \\
		\bottomrule
	\end{tabular}
	}
	\label{tab:decay_sensitivity_deep_deepsea_mean}
\end{table}

\clearpage

\begin{figure*}[h]
    \centering
    \includegraphics[width=\linewidth]{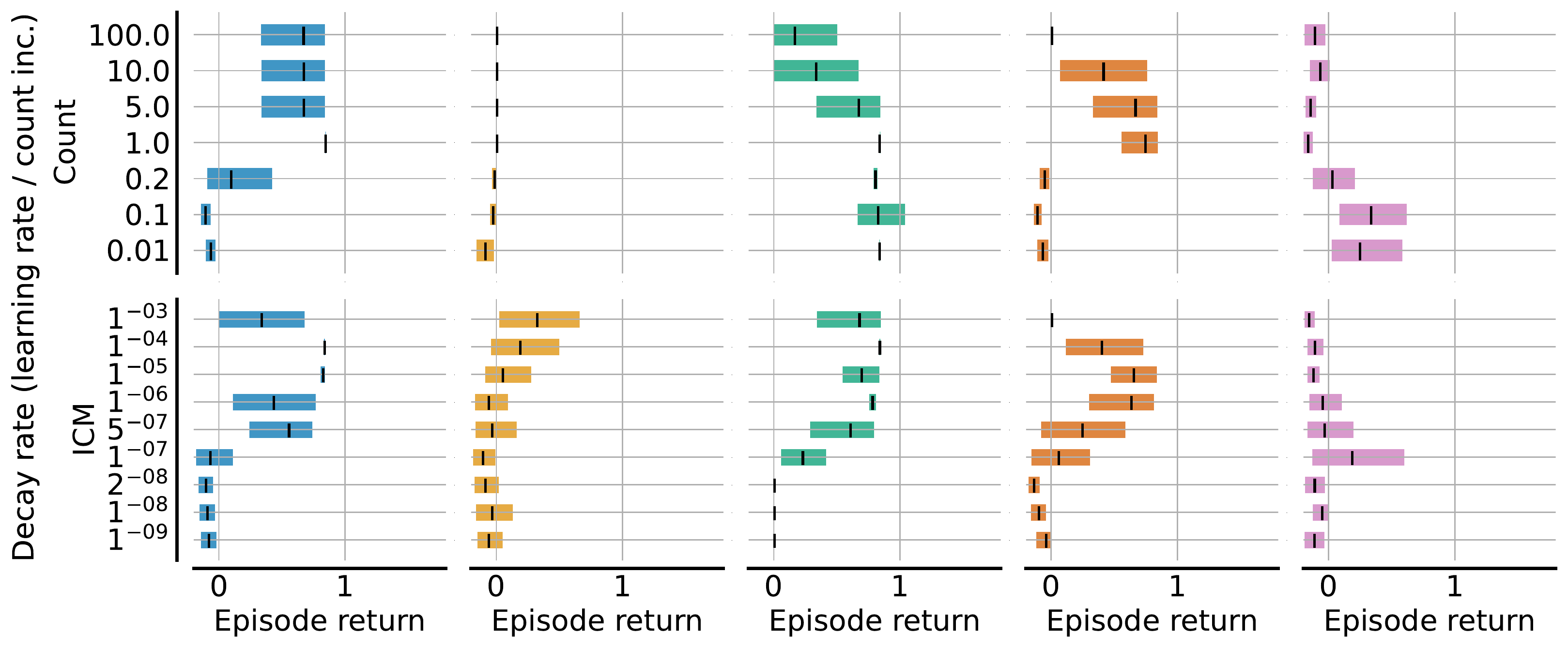}
    \includegraphics[width=.5\linewidth]{media/legend.pdf}
    
    \caption{Average evaluation returns for baselines and DeRL with Count (upper row) and ICM (lower row) intrinsic rewards in Hallway $N_l=N_r=10$ with varying count increments and learning rates. Shading indicates 95\% confidence intervals.}
\end{figure*}

\begin{figure*}[h]
    \centering
    \begin{subfigure}{.33\textwidth}
        \centering
        \includegraphics[width=1\linewidth]{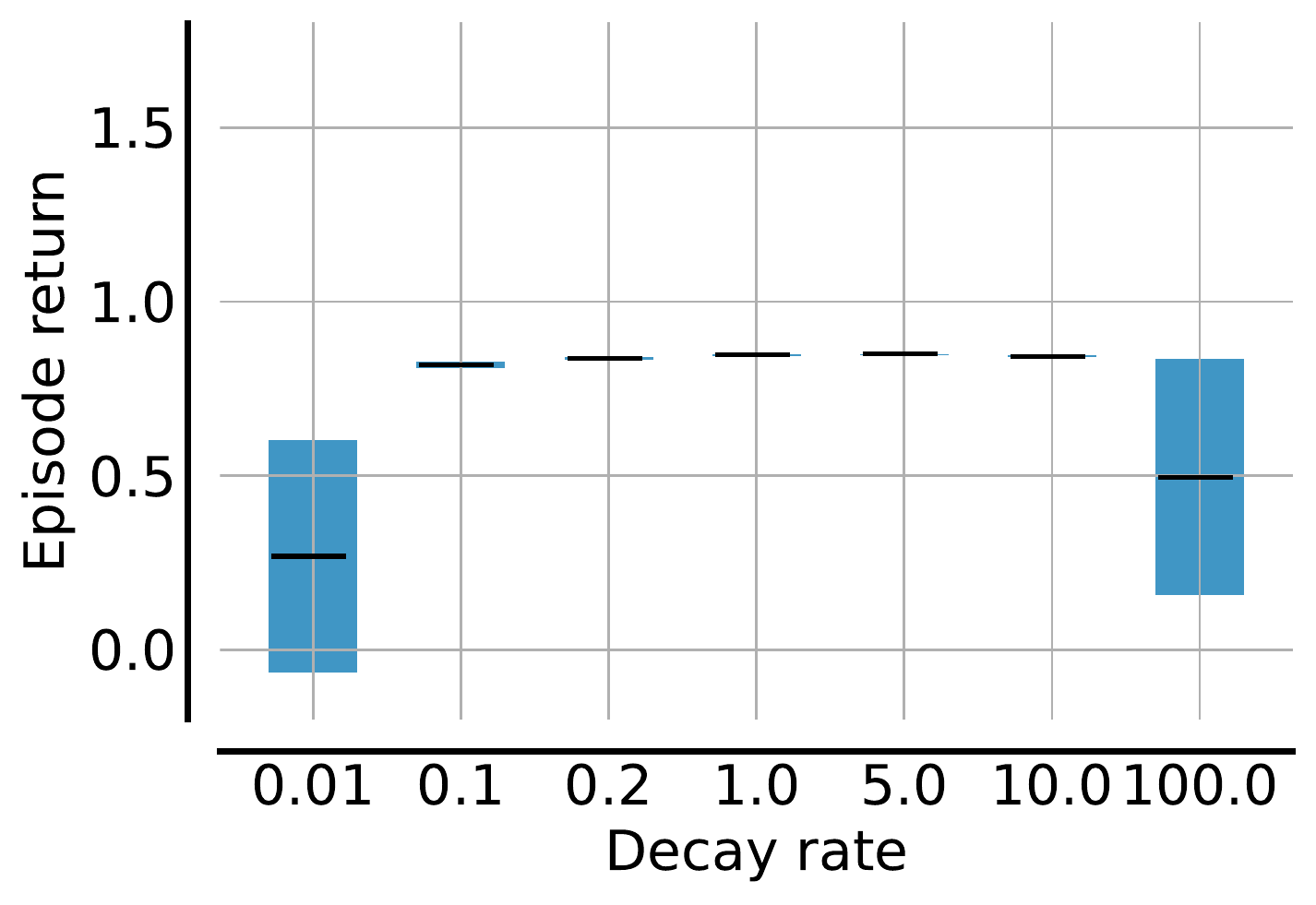}
        \caption{A2C Hash-Count}
    \end{subfigure}
    \hfill
    \begin{subfigure}{.33\textwidth}
        \centering
        \includegraphics[width=1\linewidth]{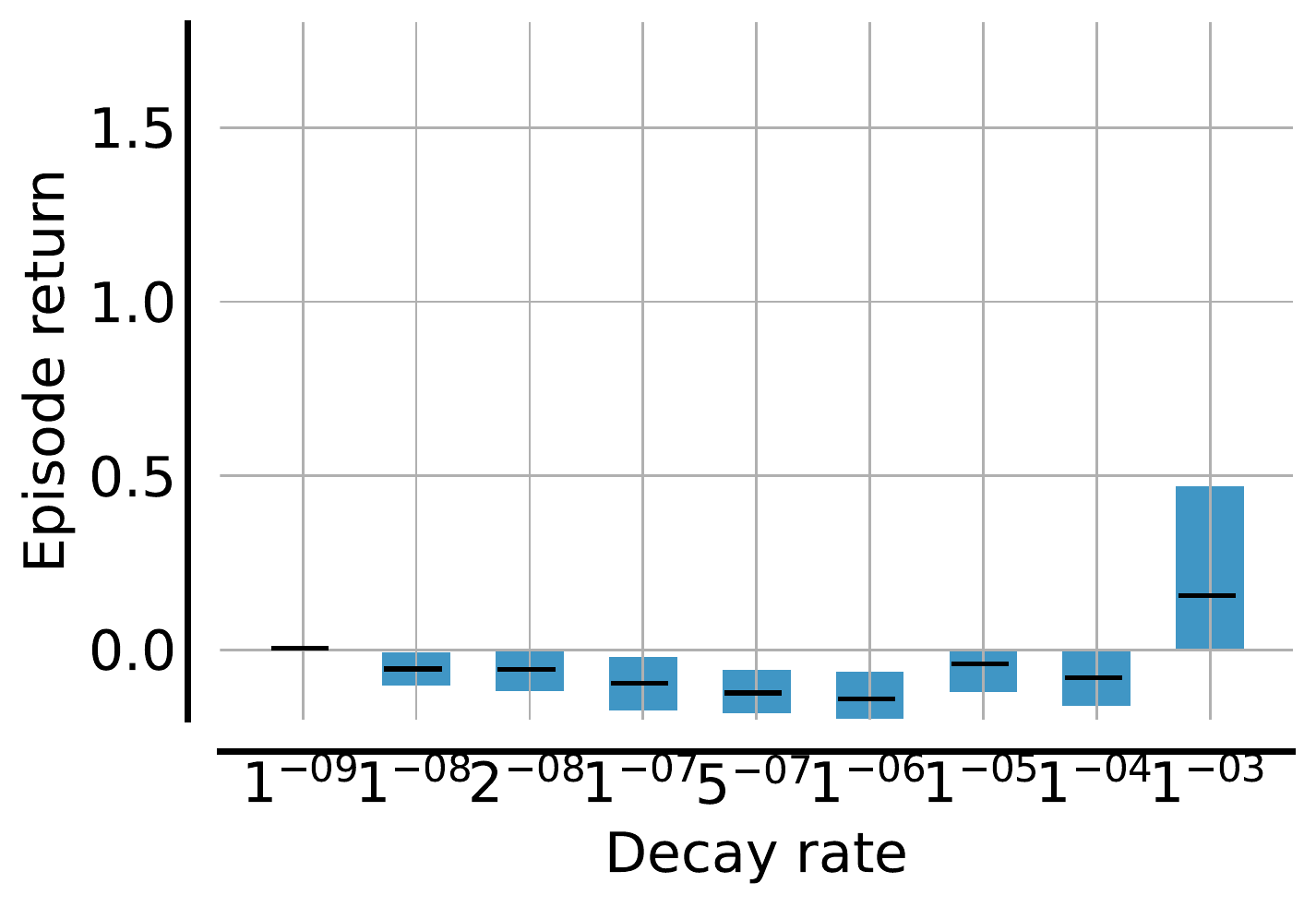}
        \caption{A2C RND}
    \end{subfigure}
    \hfill
    \begin{subfigure}{.33\textwidth}
        \centering
        \includegraphics[width=1\linewidth]{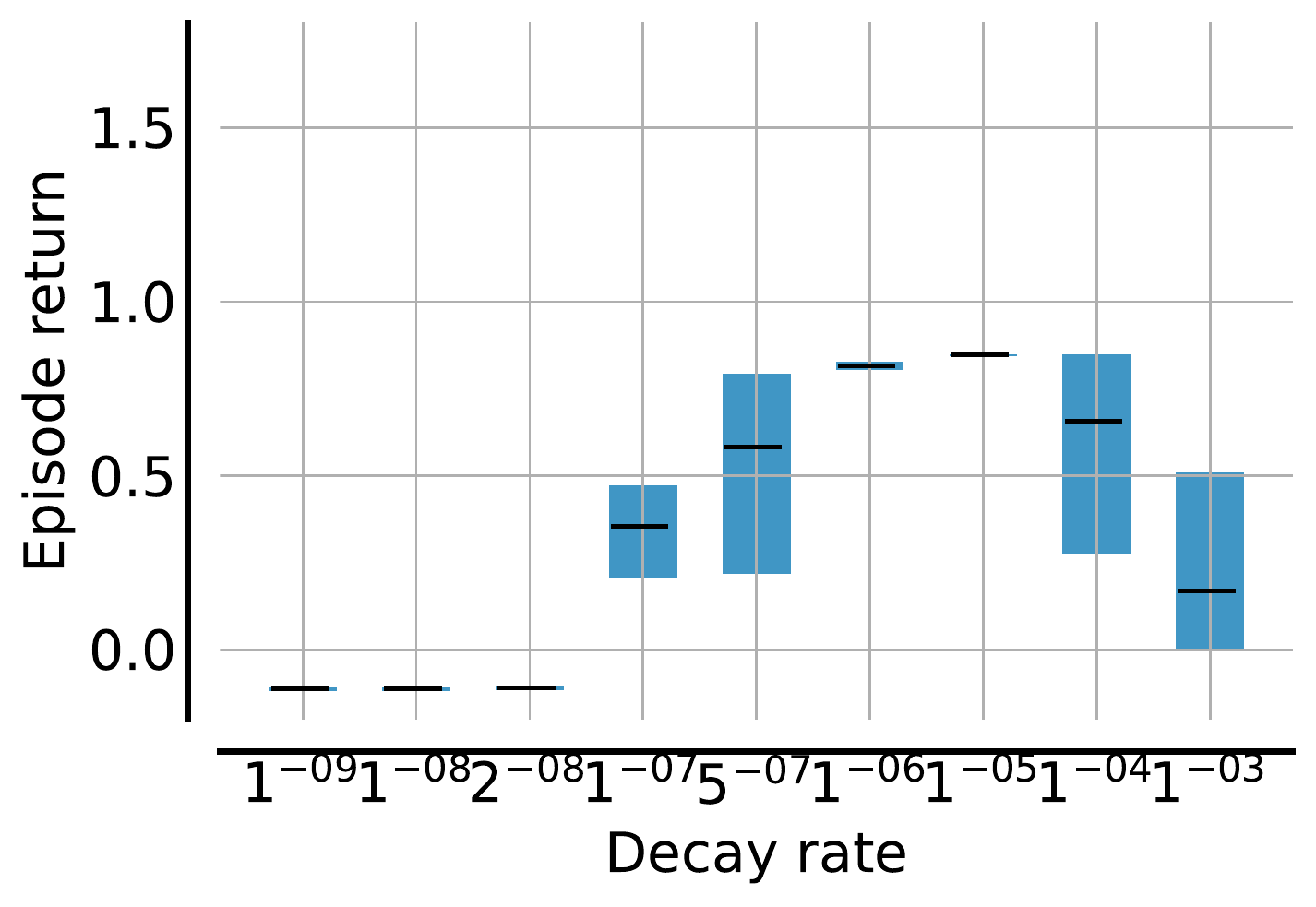}
        \caption{A2C RIDE}
    \end{subfigure}
    
    \begin{subfigure}{.33\textwidth}
        \centering
        \includegraphics[width=1\linewidth]{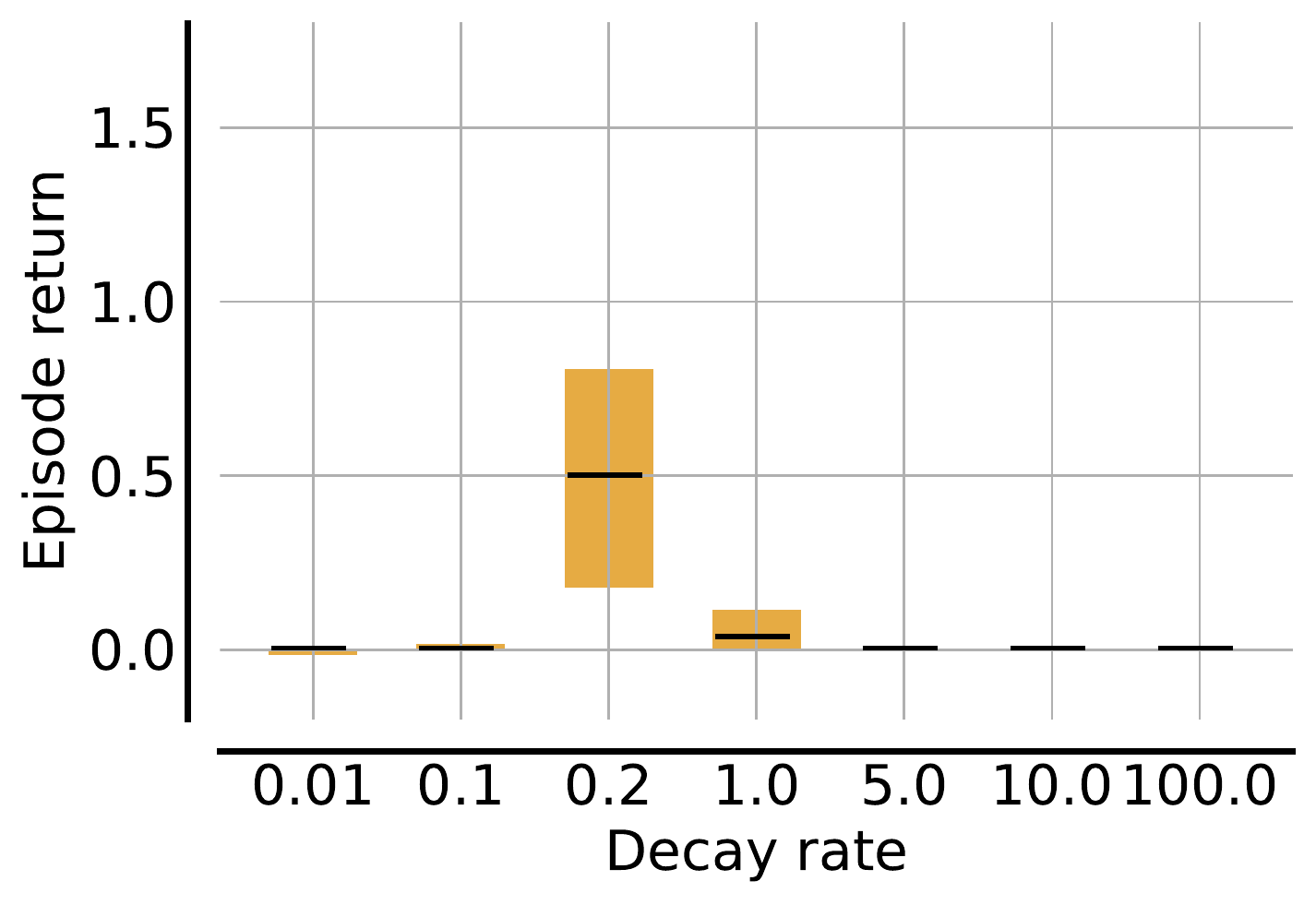}
        \caption{PPO Hash-Count}
    \end{subfigure}
    \hfill
    \begin{subfigure}{.33\textwidth}
        \centering
        \includegraphics[width=1\linewidth]{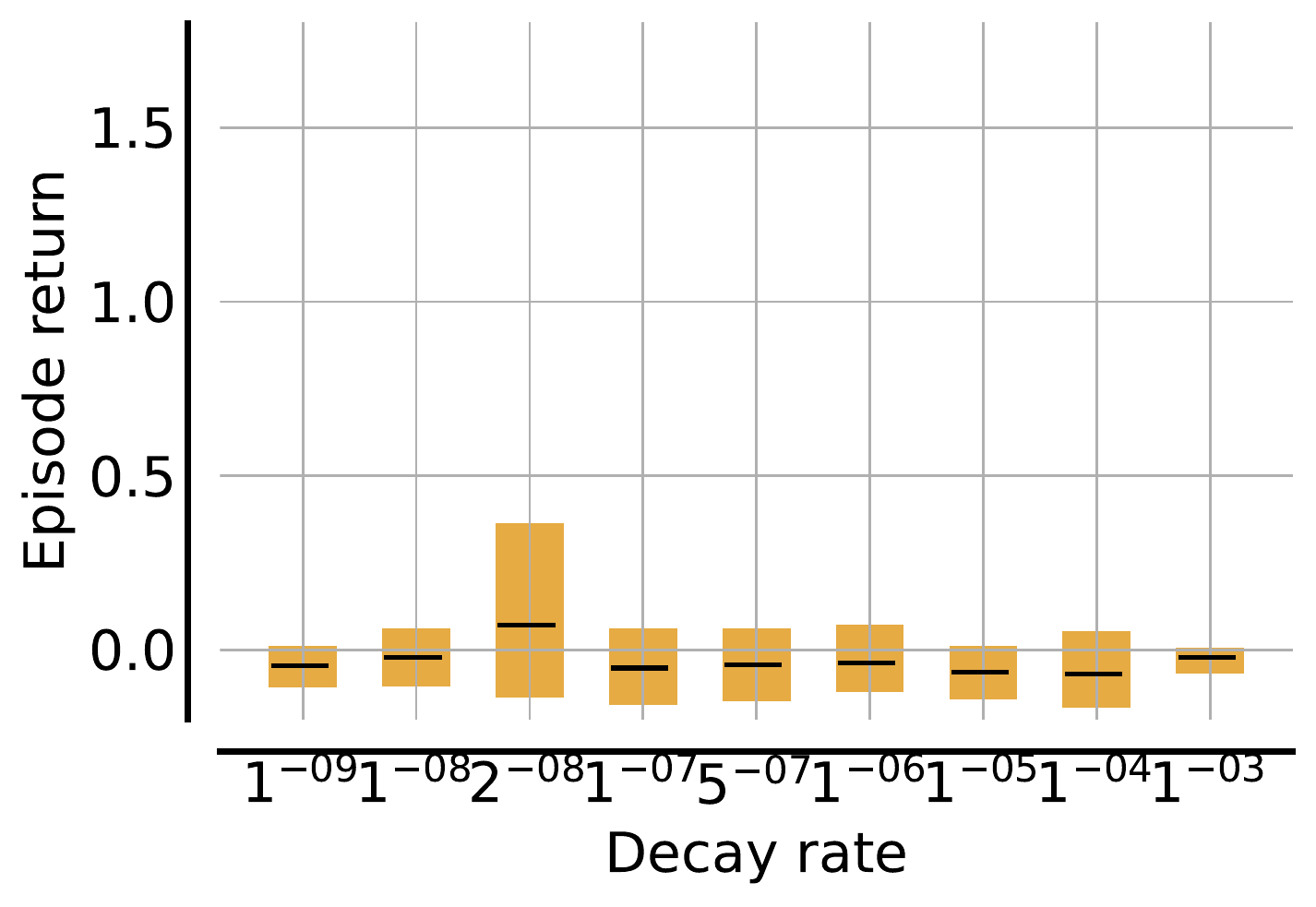}
        \caption{PPO RND}
    \end{subfigure}
    \hfill
    \begin{subfigure}{.33\textwidth}
        \centering
        \includegraphics[width=1\linewidth]{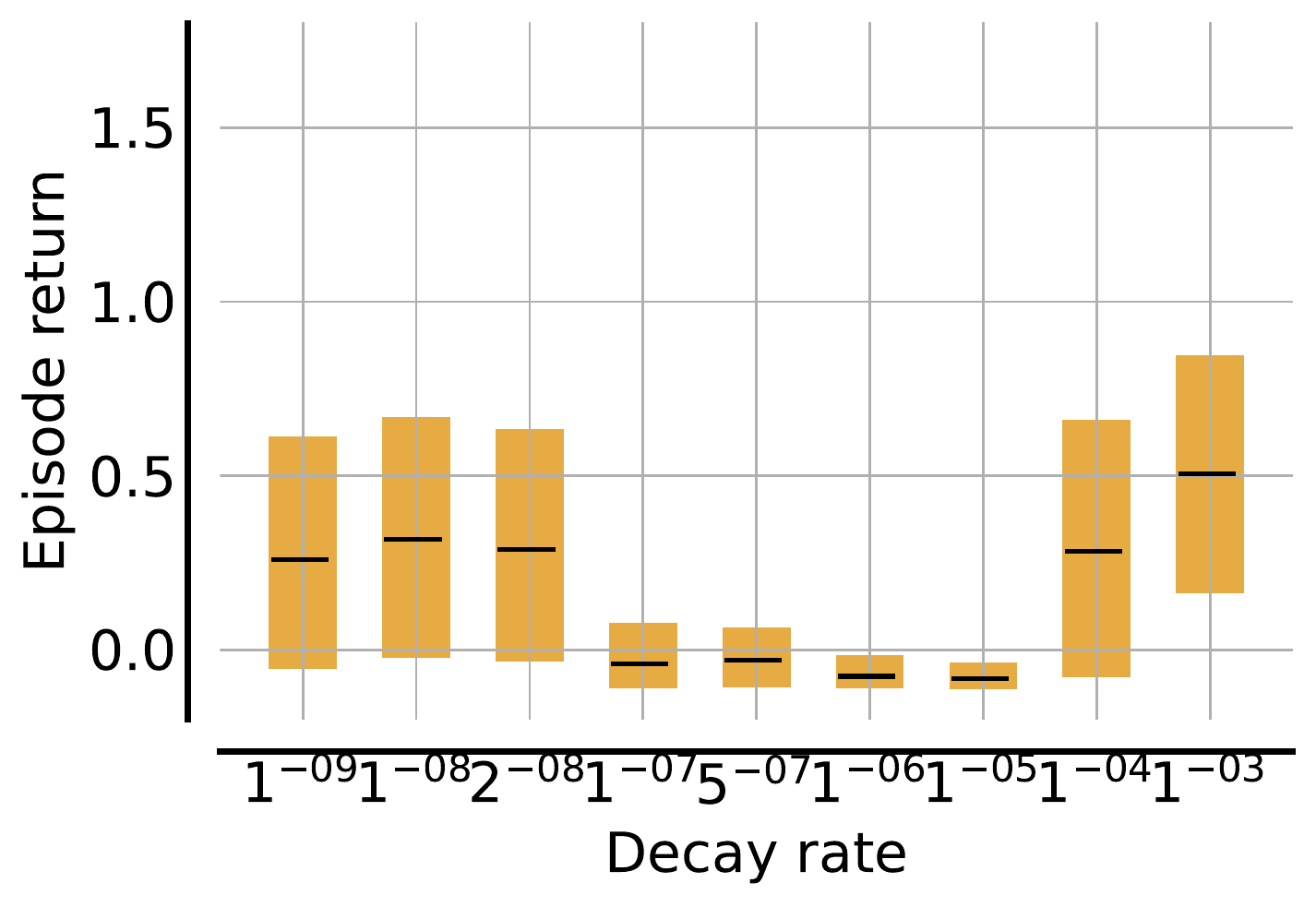}
        \caption{PPO RIDE}
    \end{subfigure}
    \caption{Average evaluation returns for A2C and PPO with Hash-Count, RND and RIDE intrinsic rewards in Hallway $N_l=N_r=10$ with varying count increments and learning rates. Shading indicates 95\% confidence intervals.}
\end{figure*}

\begin{table*}[hbt!]
	\centering
	\caption{Maximum evaluation returns with a single standard deviation in Hallway $N_l = N_r = 10$ with count-based intrinsic rewards and varying count increments.}
	\resizebox{\linewidth}{!}{
	\robustify\bf
	\begin{tabular}{l S S S S S S S}
		\toprule
		{Algorithm \textbackslash \ Count increment} & {0.01} & {0.1} & {0.2} & {1.0} & {5.0} & {10.0} & {100.0} \\
		\midrule
		 A2C Count &  0.80(0) & \bf 0.80(0) &  0.17(26) & \bf 0.85(0) &  0.68(26) &  0.68(26) & \bf 0.68(26) \\
		 A2C Hash-Count &  0.34(34) & \bf 0.85(0) & \bf 0.85(0) & \bf 0.85(0) & \bf 0.85(0) & \bf 0.85(0) & \bf 0.51(34) \\
		 \midrule
		 PPO Count &  0.00(0) &  0.00(0) & 0.16(25) &  0.00(0) &  0.00(0) &  0.00(0) &  0.00(0) \\
		 PPO Hash-Count &  0.16(24) &  0.26(38) & 0.80(0) &  0.17(26) &  0.00(0) &  0.00(0) &  0.00(0) \\
		 \midrule
		 DeA2C Count & \bf 0.85(0) & \bf 1.00(30) &  0.83(2) & \bf 0.85(0) & \bf 0.85(0) &  0.34(34) &  0.17(26) \\
		 DePPO Count & \bf 1.23(38) & \bf 0.80(0) &  0.25(37) & \bf 0.85(0) &  0.68(26) &  0.51(34) &  0.00(0) \\
		 DeDQN Count & \bf 1.02(60) & \bf 0.85(0) & \bf 1.23(38) &  0.43(42) &  0.24(40) &  0.05(34) &  0.26(42) \\
		\bottomrule
	\end{tabular}
	}
	\label{tab:decay_sensitivity_count_hallway10_10_max}
\end{table*}

\begin{table*}[hbt!]
	\centering
    \caption{Average evaluation returns with a single standard deviation in Hallway $N_l = N_r = 10$ with count-based intrinsic rewards and varying count increments.}
	\resizebox{\linewidth}{!}{
	\robustify\bf
	\begin{tabular}{l S S S S S S S}
		\toprule
		{Algorithm \textbackslash \ Count increment} & {0.01} & {0.1} & {0.2} & {1.0} & {5.0} & {10.0} & {100.0} \\
		\midrule
		 A2C Count &  -0.06(26) &  -0.11(10) &  0.10(9) & \bf 0.85(2) &  0.67(7) &  0.67(7) & \bf 0.67(7) \\
		 A2C Hash-Count &  0.27(14) & \bf 0.82(16) & \bf 0.84(10) & \bf 0.85(2) & \bf 0.85(1) & \bf 0.84(7) &  0.50(5) \\
		 \midrule
		 PPO Count &  -0.09(4) &  -0.02(4) &  -0.01(3) & 0.00(0) & 0.00(0) & 0.00(0) & 0.00(0) \\
		 PPO Hash-Count &  -0.00(2) &  0.00(3) & 0.50(6) &  0.04(7) &  0.00(0) &  0.00(0) &  0.00(0) \\
		 \midrule
		 DeA2C Count & \bf 0.84(7) & \bf 0.83(19) & \bf 0.81(12) & \bf 0.84(9) &  0.67(6) &  0.33(4) &  0.17(2) \\
		 DePPO Count &  -0.07(28) &  -0.11(10) &  -0.05(5) & 0.75(12) &  0.67(7) &  0.42(12) &  0.00(0) \\
		 DeDQN Count &  0.25(28) & 0.34(28) &  0.03(30) &  -0.16(7) &  -0.14(5) &  -0.07(2) &  -0.11(5) \\
		\bottomrule
	\end{tabular}
	}
	\label{tab:decay_sensitivity_count_hallway10_10_mean}
\end{table*}

\begin{table*}[hbt!]
	\centering
	\caption{Maximum evaluation returns with a single standard deviation in Hallway $N_l = N_r = 10$ with prediction-based intrinsic rewards and varying learning rates.}
	\resizebox{\linewidth}{!}{
	\robustify\bf
	\begin{tabular}{l S S S S S S S S S}
		\toprule
		{Algorithm \textbackslash \ Learning rate} & {1e-09} & {1e-08} & {2e-08} & {1e-07} & {5e-07} & {1e-06} & {1e-05} & {0.0001} & {0.001} \\
		\midrule
		 A2C ICM &  -0.04(6) &  -0.04(6) &  -0.04(6) &  0.15(28) &  0.68(26) &  0.51(34) & \bf 0.85(0) & \bf 0.85(0) &  0.34(34) \\
		 A2C RND &  0.00(0) &  0.00(0) &  0.00(0) &  0.00(0) &  0.00(0) &  0.00(0) &  -0.04(6) &  -0.08(8) & 0.17(25) \\
		 A2C RIDE &  -0.11(1) &  -0.11(1) &  -0.11(1) & \bf 0.84(2) &  0.66(28) & \bf 0.85(0) & \bf 0.85(0) &  0.66(28) &  0.17(26) \\
		 \midrule
		 PPO ICM &  0.40(40) &  0.40(40) &  0.22(40) &  0.24(40) &  0.40(40) &  0.26(38) &  0.33(33) &  0.50(34) & 0.67(26) \\
		 PPO RND & \bf 0.80(0) & \bf 0.80(0) & \bf 0.64(24) & \bf 0.44(38) & \bf 0.80(0) &  0.23(39) &  0.12(30) &  0.24(40) &  0.44(38) \\
		 PPO RIDE &  0.44(36) &  0.62(28) & \bf 0.65(24) & \bf 0.43(37) &  0.44(37) &  0.10(30) &  0.26(37) &  0.46(38) &  0.51(34) \\
		 \midrule
		 DeA2C ICM &  0.00(0) &  0.00(0) &  0.00(0) & \bf 0.49(33) &  0.67(26) & \bf 0.85(0) & \bf 0.85(0) & \bf 0.85(0) & \bf 0.85(0) \\
		 DePPO ICM &  -0.04(6) &  -0.04(6) &  -0.04(6) & \bf 0.42(40) &  0.48(38) &  0.68(26) & \bf 0.85(0) &  0.49(33) &  0.00(0) \\
		 DeDQN ICM &  -0.04(6) &  0.00(0) &  -0.04(6) & \bf 1.04(78) &  0.34(57) &  0.31(57) &  0.30(38) &  0.13(32) &  0.17(26) \\
		\bottomrule
	\end{tabular}
	}
	\label{tab:decay_sensitivity_deep_hallway10_10_max}
\end{table*}

\begin{table*}[hbt!]
	\centering
	\caption{Average evaluation returns with a single standard deviation in Hallway $N_l = N_r = 10$ with prediction-based intrinsic rewards and varying learning rates.}
	\resizebox{\linewidth}{!}{
	\robustify\bf
	\begin{tabular}{l S S S S S S S S S}
		\toprule
		{Algorithm \textbackslash \ Learning rate} & {1e-09} & {1e-08} & {2e-08} & {1e-07} & {5e-07} & {1e-06} & {1e-05} & {0.0001} & {0.001} \\
		\midrule
		 A2C ICM &  -0.08(2) &  -0.09(2) &  -0.10(4) &  -0.07(12) & \bf 0.56(26) &  0.44(14) & \bf 0.83(11) & \bf 0.84(9) &  0.34(0) \\
		 A2C RND &  0.00(0) &  -0.05(4) &  -0.06(3) &  -0.10(3) &  -0.12(3) &  -0.14(2) &  -0.04(0) &  -0.08(0) & 0.16(3) \\
		 A2C RIDE &  -0.11(0) &  -0.11(0) &  -0.11(0) & \bf 0.36(40) & \bf 0.58(19) & \bf 0.82(16) & \bf 0.85(2) &  0.66(3) &  0.17(0) \\
		 \midrule
		 PPO ICM &  -0.06(10) &  -0.03(10) &  -0.09(7) &  -0.10(8) &  -0.03(11) &  -0.06(10) &  0.05(13) &  0.19(14) & 0.32(13) \\
		 PPO RND &  -0.04(10) &  -0.02(12) & 0.07(12) &  -0.05(14) &  -0.04(17) &  -0.04(9) &  -0.06(5) &  -0.07(9) &  -0.02(6) \\
		 PPO RIDE & \bf 0.26(14) & \bf 0.32(14) & \bf 0.29(16) & \bf -0.04(10) &  -0.03(15) &  -0.08(6) &  -0.08(6) &  0.28(6) & 0.51(2) \\
		 \midrule
		 DeA2C ICM &  0.00(0) &  0.00(0) &  0.00(0) & \bf 0.23(22) & \bf 0.61(17) & \bf 0.78(20) &  0.70(18) & \bf 0.84(7) & \bf 0.68(3) \\
		 DePPO ICM &  -0.04(0) &  -0.10(4) &  -0.14(6) & \bf 0.06(18) &  0.25(13) &  0.63(16) & 0.65(19) &  0.40(15) &  0.00(0) \\
		 DeDQN ICM &  -0.11(2) &  -0.05(2) &  -0.11(2) & \bf 0.19(28) &  -0.03(16) &  -0.05(13) &  -0.12(6) &  -0.11(5) &  -0.15(5) \\
		\bottomrule
	\end{tabular}
	}
	\label{tab:decay_sensitivity_deep_hallway10_10_mean}
\end{table*}

\clearpage
\section{KL-Divergence Constraint Regularisation}
\label{app:kl_divergence_constraint}
In this section, we first provide figures showing evaluation returns of the evaluation policy (top left), training returns of the exploration policy (bottom left), IS weights (top right) and KL-divergence of exploration and exploitation policy (bottom right) for DeA2C trained with Count intrinsic rewards and KL-divergence constraints in the DeepSea $10$ and Hallway $N_l=N_r=20$ tasks for 20,000 episodes. Each figure corresponds to DeA2C being trained with KL-divergence constraint coefficients $\alpha_\beta$ and $\alpha_e$ for the regulariser terms for the exploration policy \pibeh and exploitation policy \piexp, respectively. We present training with the exploration policy being trained using only intrinsic rewards (blue) or using the sum of intrinsic and extrinsic rewards (orange). Average returns and 95\% confidence intervals are computed over 1,000 samples across three seeds and we consider regularisation coefficients $\alpha_\beta, \alpha_e \in \{0.0, 0.0001, 0.001, 0.01, 0.1\}$.

Following, we analyse the sensitivity of DeA2C trained with Count intrinsic rewards and KL-divergence constraints in DeepSea $10$ and Hallway $N_l=N_r=10$ tasks following the evaluation procedure outlined in \Cref{sec:results_sensitivity}.

\subsection{DeepSea 10}
\label{app:kl_divergence_constraint_ds10}
\begin{figure}[!ht]
    \centering
    \begin{subfigure}{.33\textwidth}
        \includegraphics[width=\textwidth]{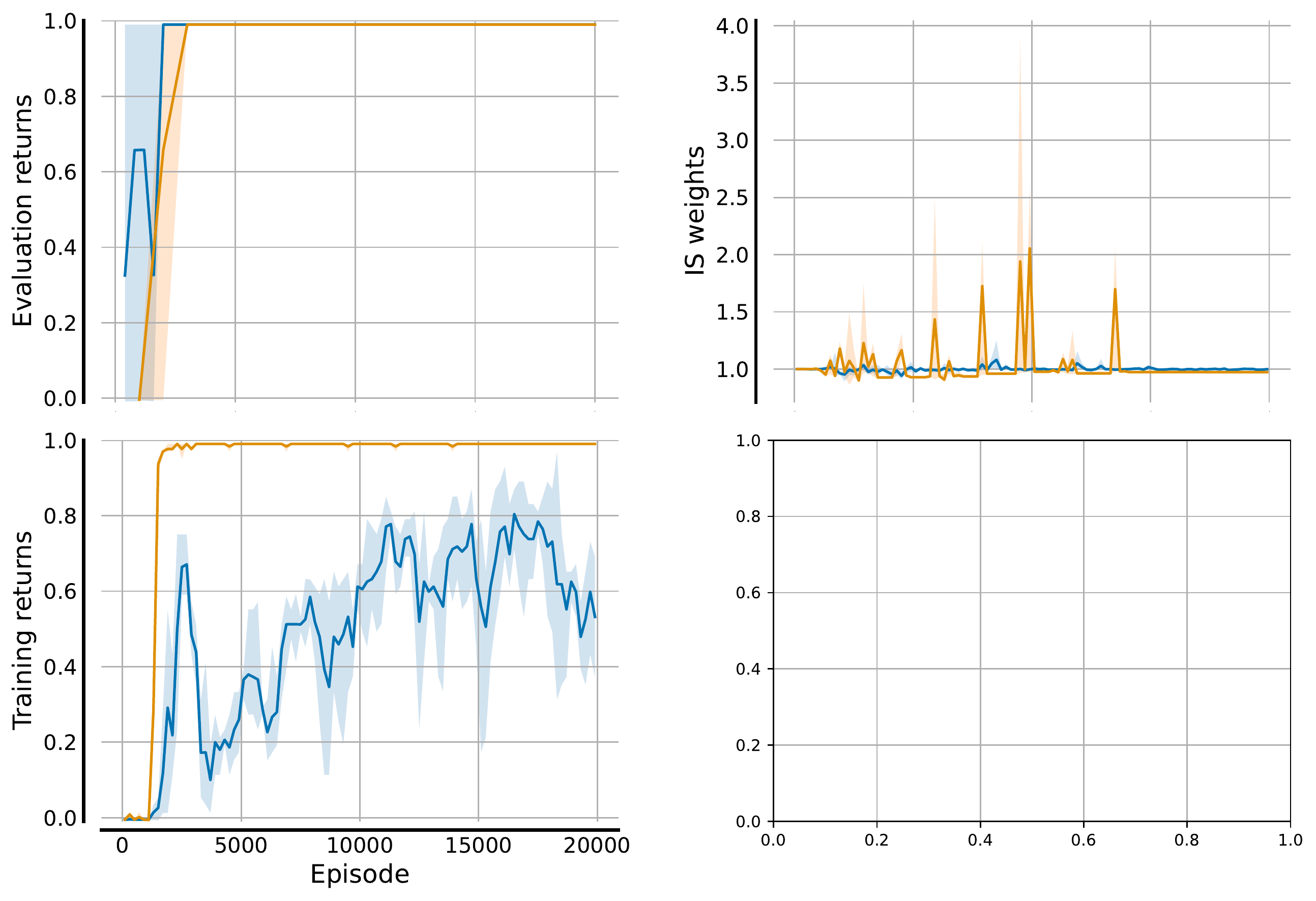}
        \caption {$\alpha_\beta=0.0, \alpha_e=0.0$}
    \end{subfigure}
    \hfill
    \begin{subfigure}{.33\textwidth}
        \includegraphics[width=\textwidth]{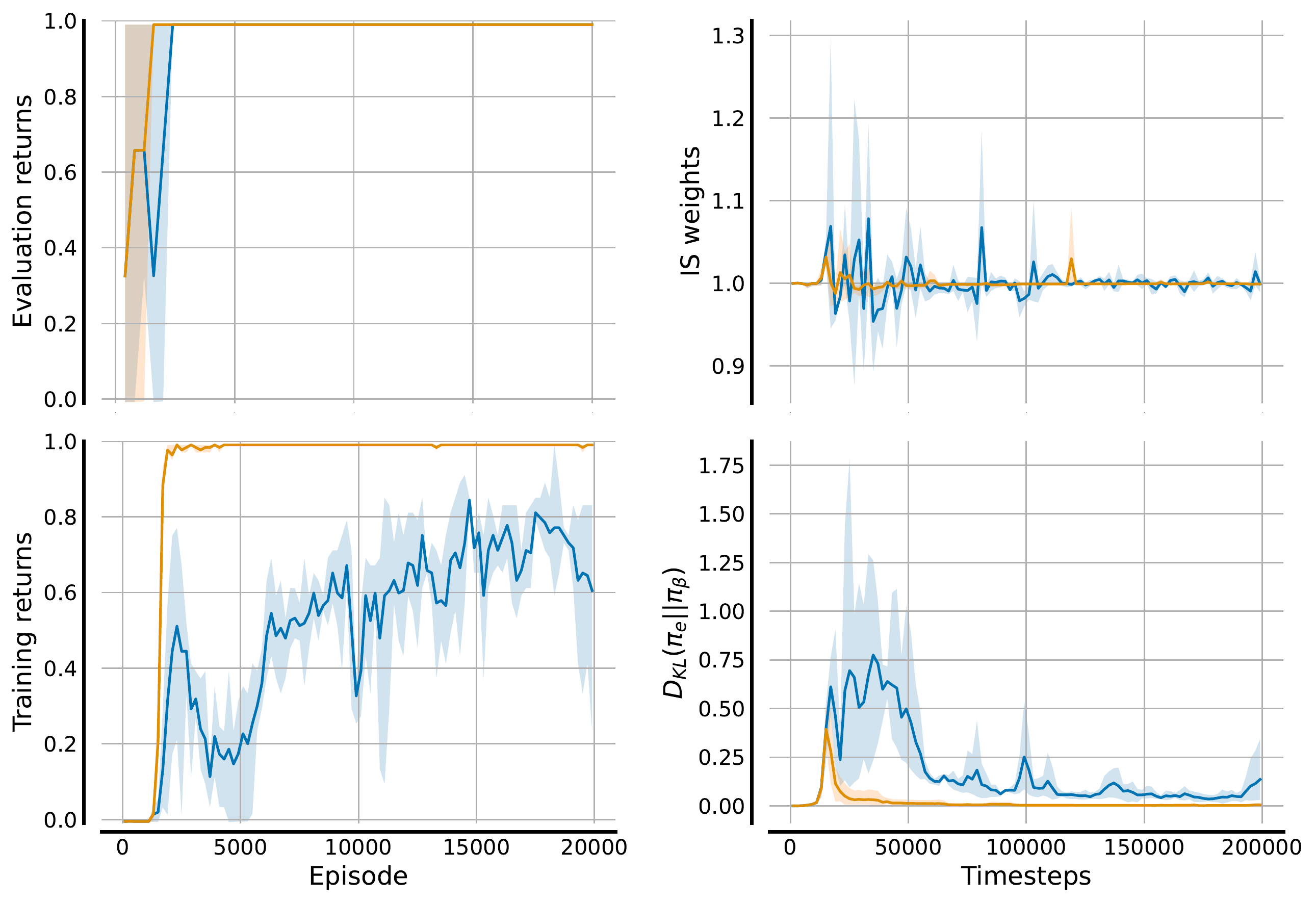}
        \caption {$\alpha_\beta=0.0, \alpha_e=0.0001$}
    \end{subfigure}
    \hfill
    \begin{subfigure}{.33\textwidth}
        \includegraphics[width=\textwidth]{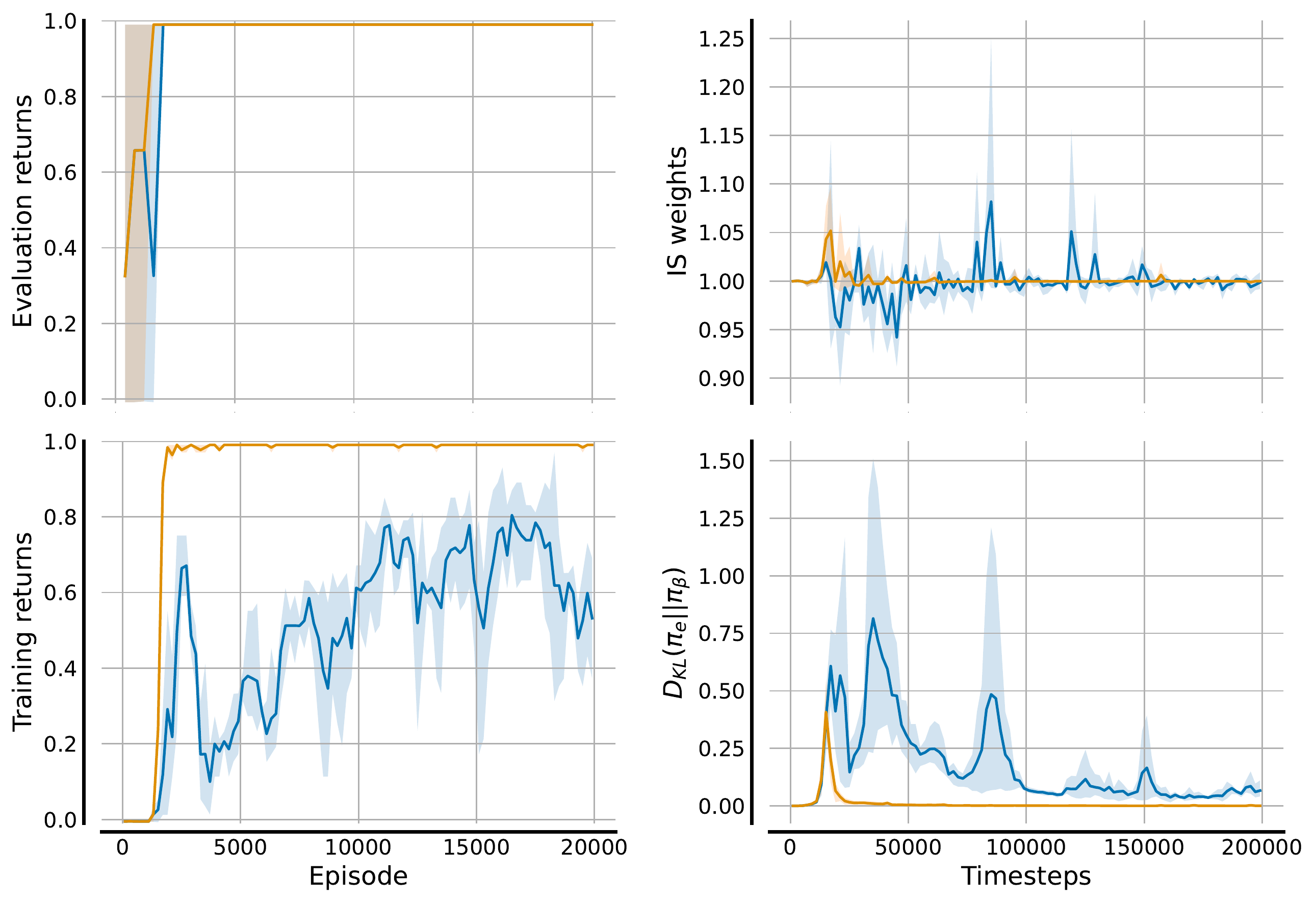}
        \caption {$\alpha_\beta=0.0, \alpha_e=0.001$}
    \end{subfigure}
    
    \ \vspace{1em}
    
    \begin{subfigure}{.33\textwidth}
        \includegraphics[width=\textwidth]{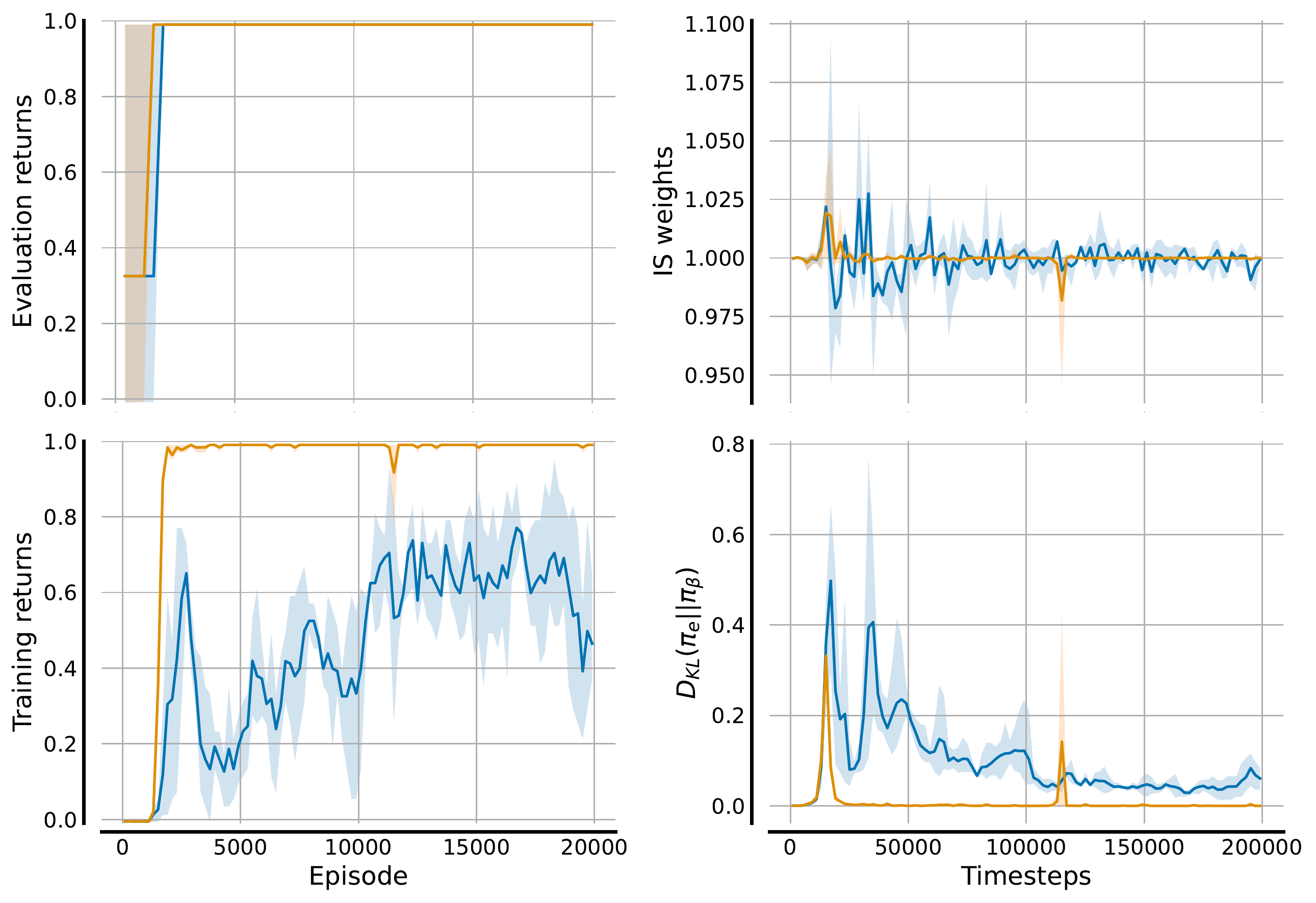}
        \caption {$\alpha_\beta=0.0, \alpha_e=0.01$}
    \end{subfigure}
    \hfill
    \begin{subfigure}{.33\textwidth}
        \includegraphics[width=\textwidth]{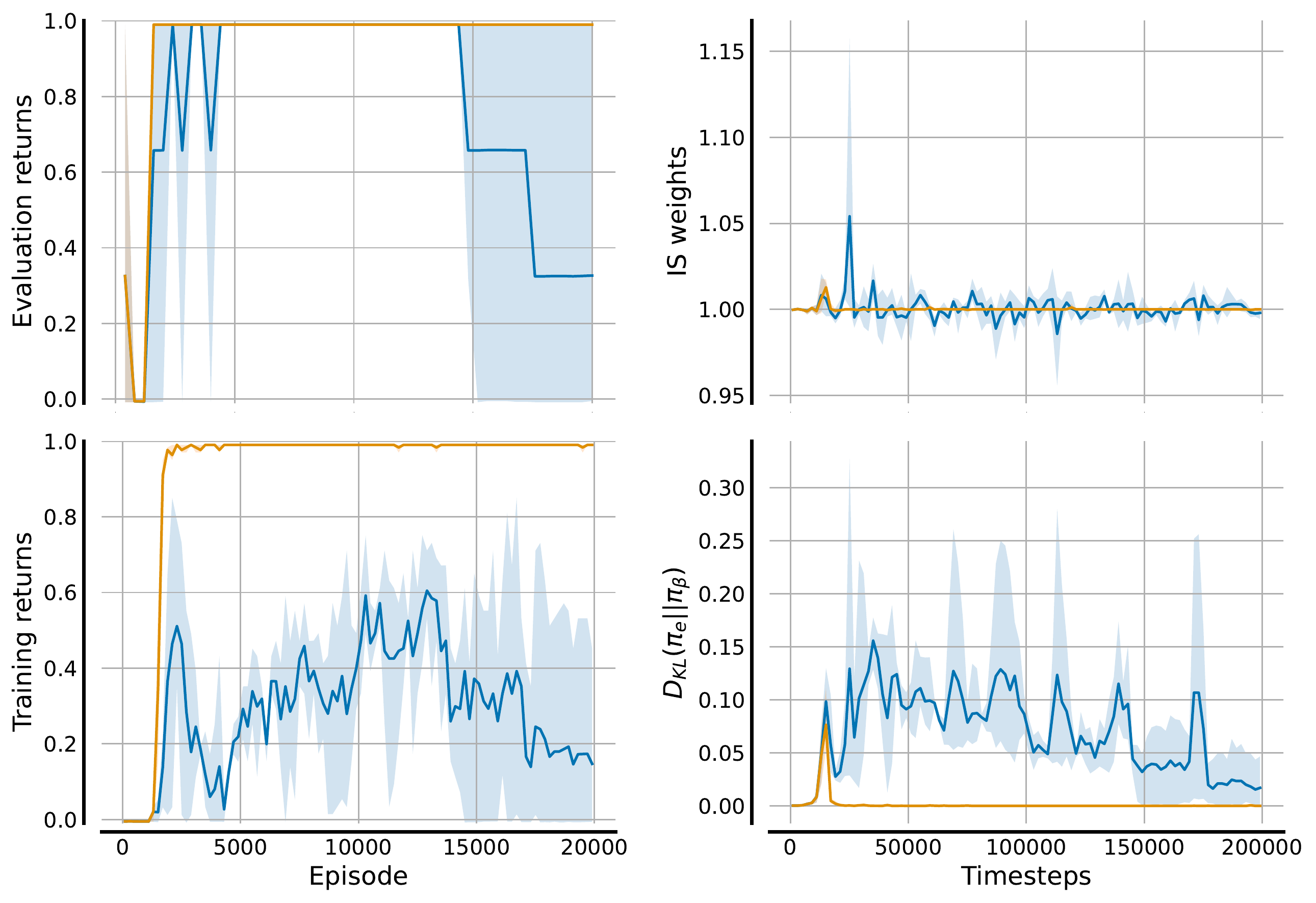}
        \caption {$\alpha_\beta=0.0, \alpha_e=0.1$}
    \end{subfigure}
    \hfill
    \begin{subfigure}{.33\textwidth}
        \includegraphics[width=\textwidth]{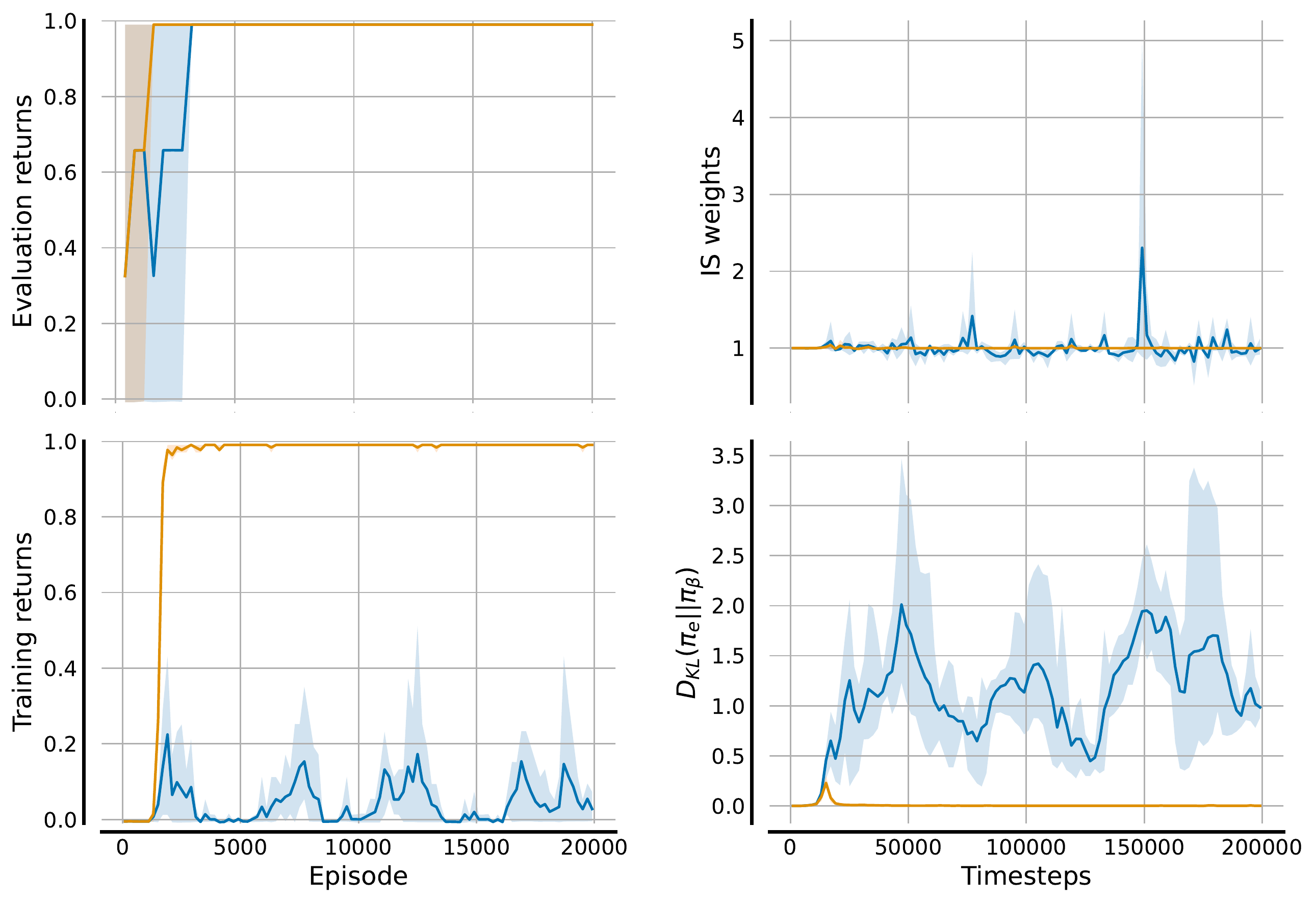}
        \caption {$\alpha_\beta=0.0001, \alpha_e=0.0$}
    \end{subfigure}
    
    \ \vspace{1em}
    
    \begin{subfigure}{.33\textwidth}
        \includegraphics[width=\textwidth]{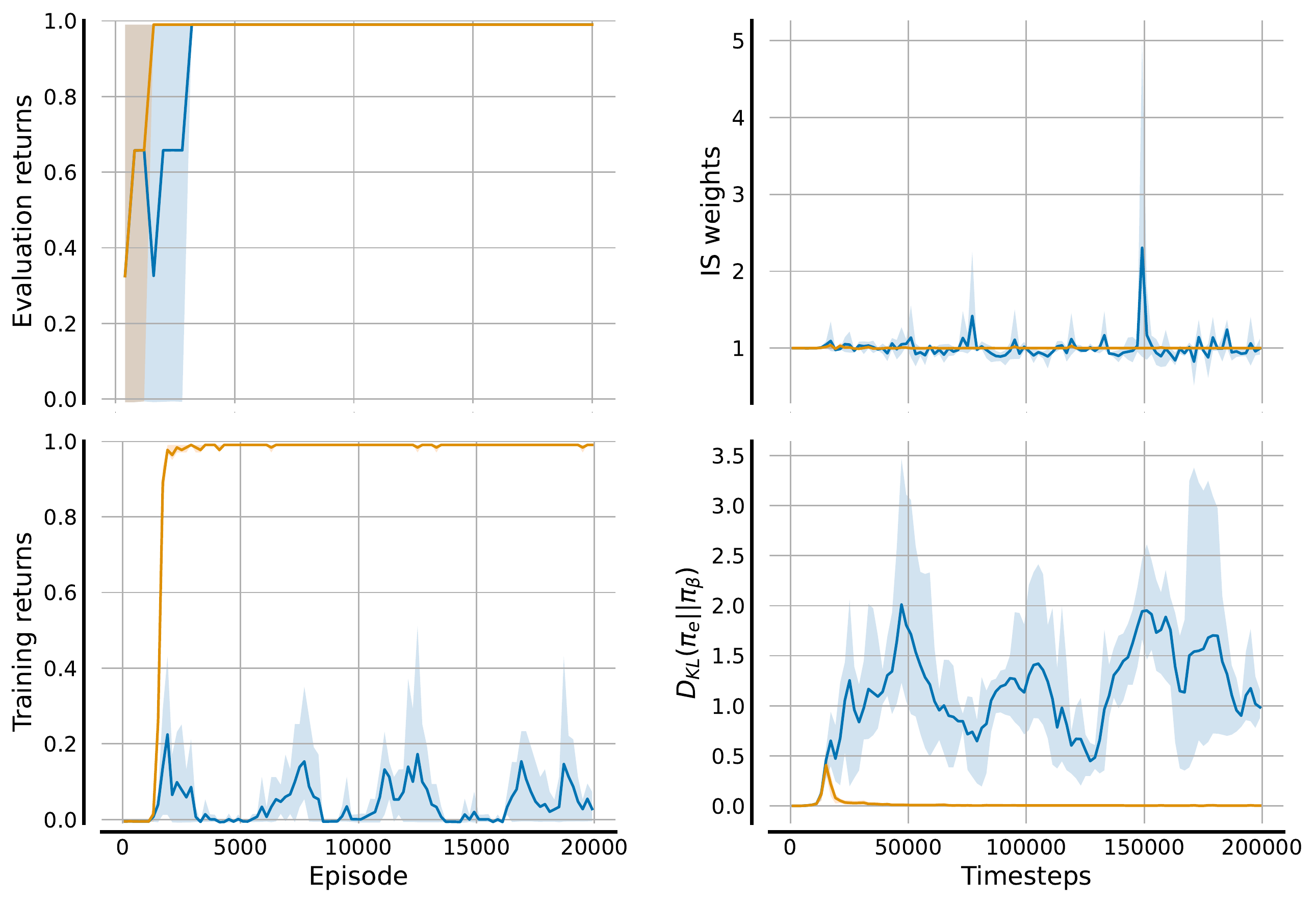}
        \caption {$\alpha_\beta=0.0001, \alpha_e=0.0001$}
    \end{subfigure}
    \hfill
    \begin{subfigure}{.33\textwidth}
        \includegraphics[width=\textwidth]{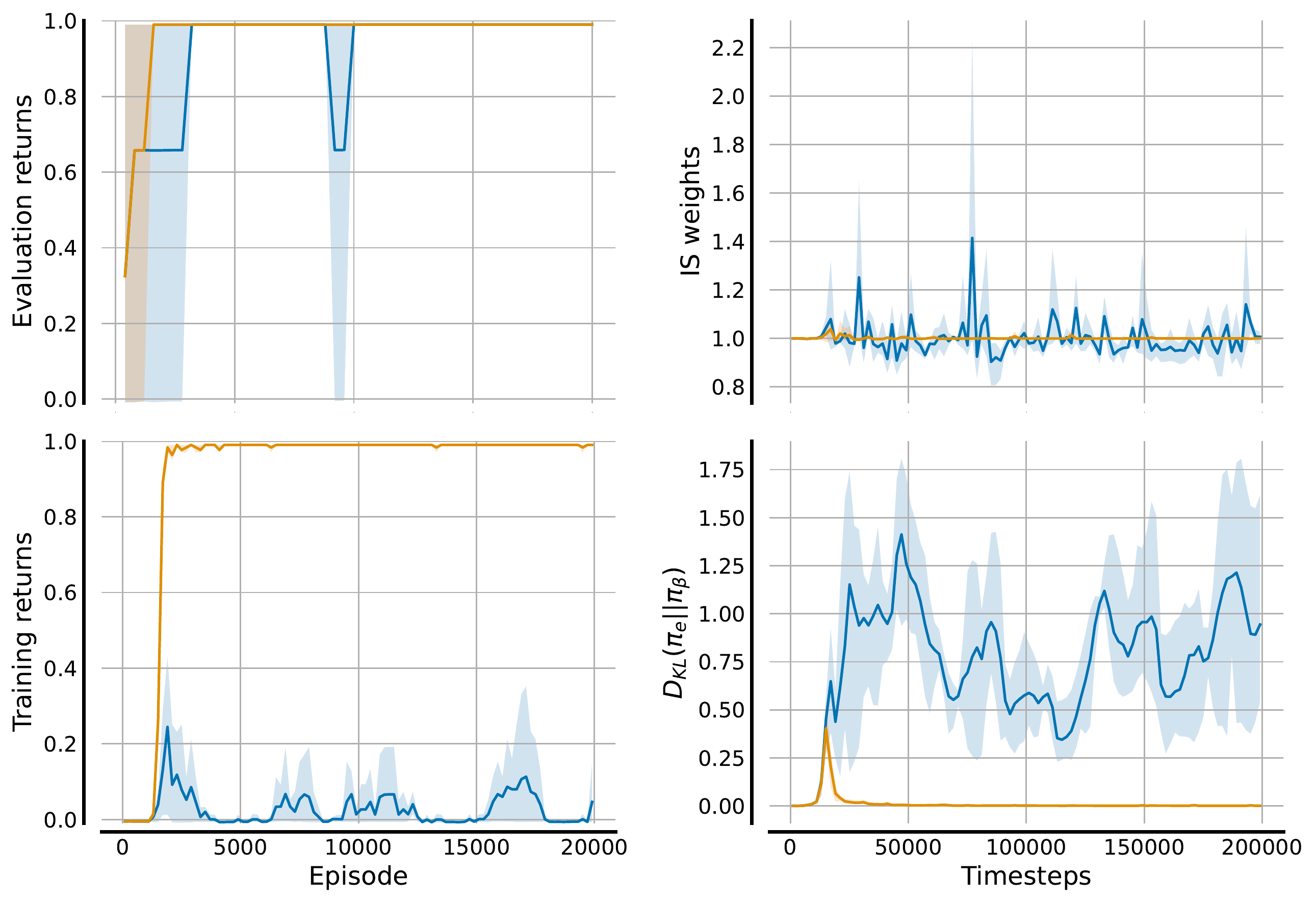}
        \caption {$\alpha_\beta=0.0001, \alpha_e=0.001$}
    \end{subfigure}
    \hfill
    \begin{subfigure}{.33\textwidth}
        \includegraphics[width=\textwidth]{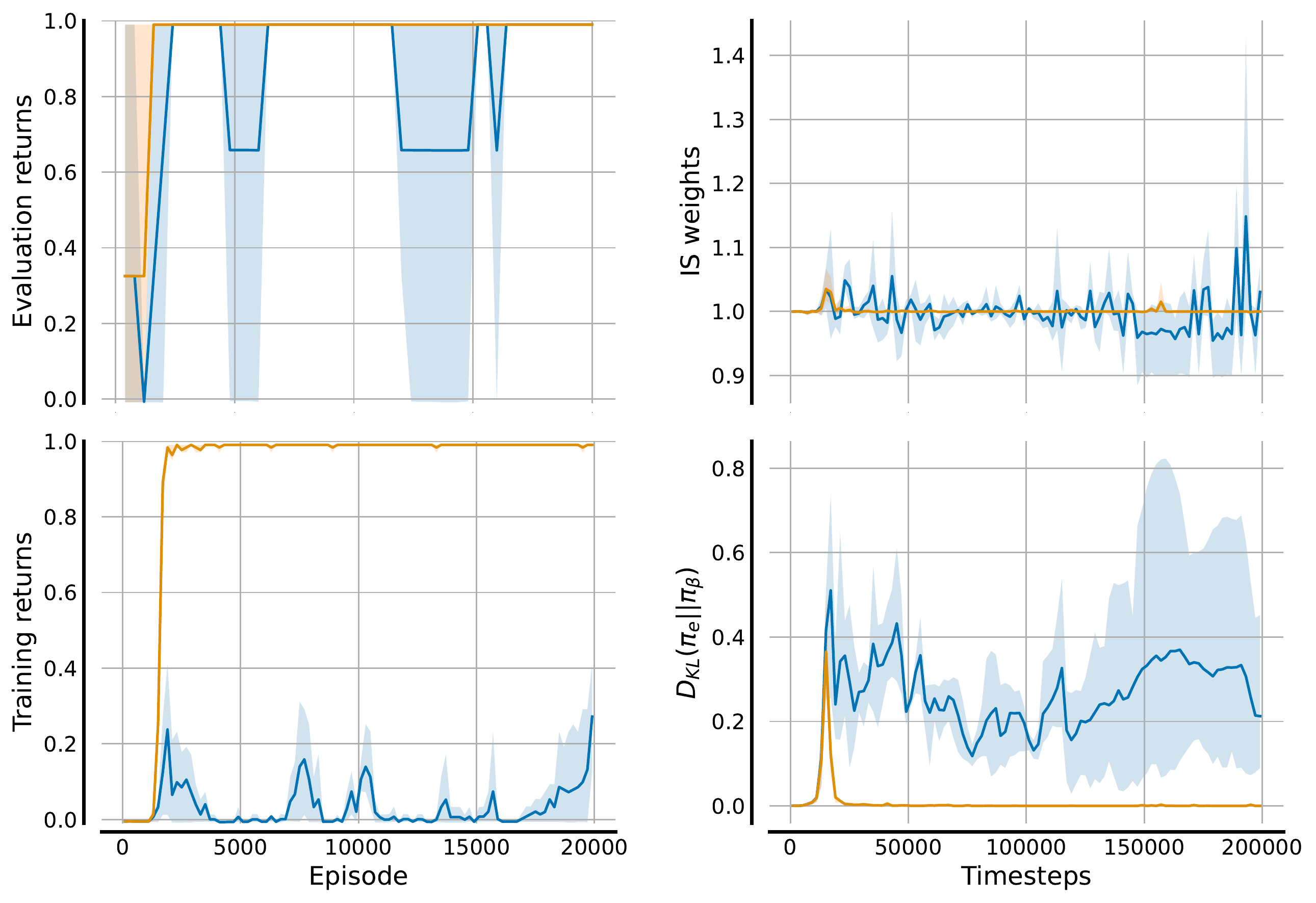}
        \caption {$\alpha_\beta=0.0001, \alpha_e=0.01$}
    \end{subfigure}
    \begin{subfigure}{.5\textwidth}
        \centering
        \includegraphics[width=.5\textwidth]{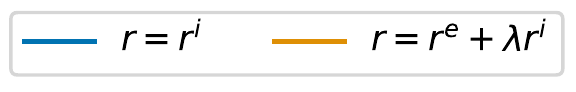}
    \end{subfigure}
    \caption{DeepSea 10 evaluation with divergence constraint regularisation coefficients $\alpha_\beta$ and $\alpha_e$.  Shading indicates 95\% confidence intervals; Part 1}
    \label{fig:kl_divergence_constraint_ds10_1}
\end{figure}

\begin{figure}[!ht]
    \centering
    \begin{subfigure}{.33\textwidth}
        \includegraphics[width=\textwidth]{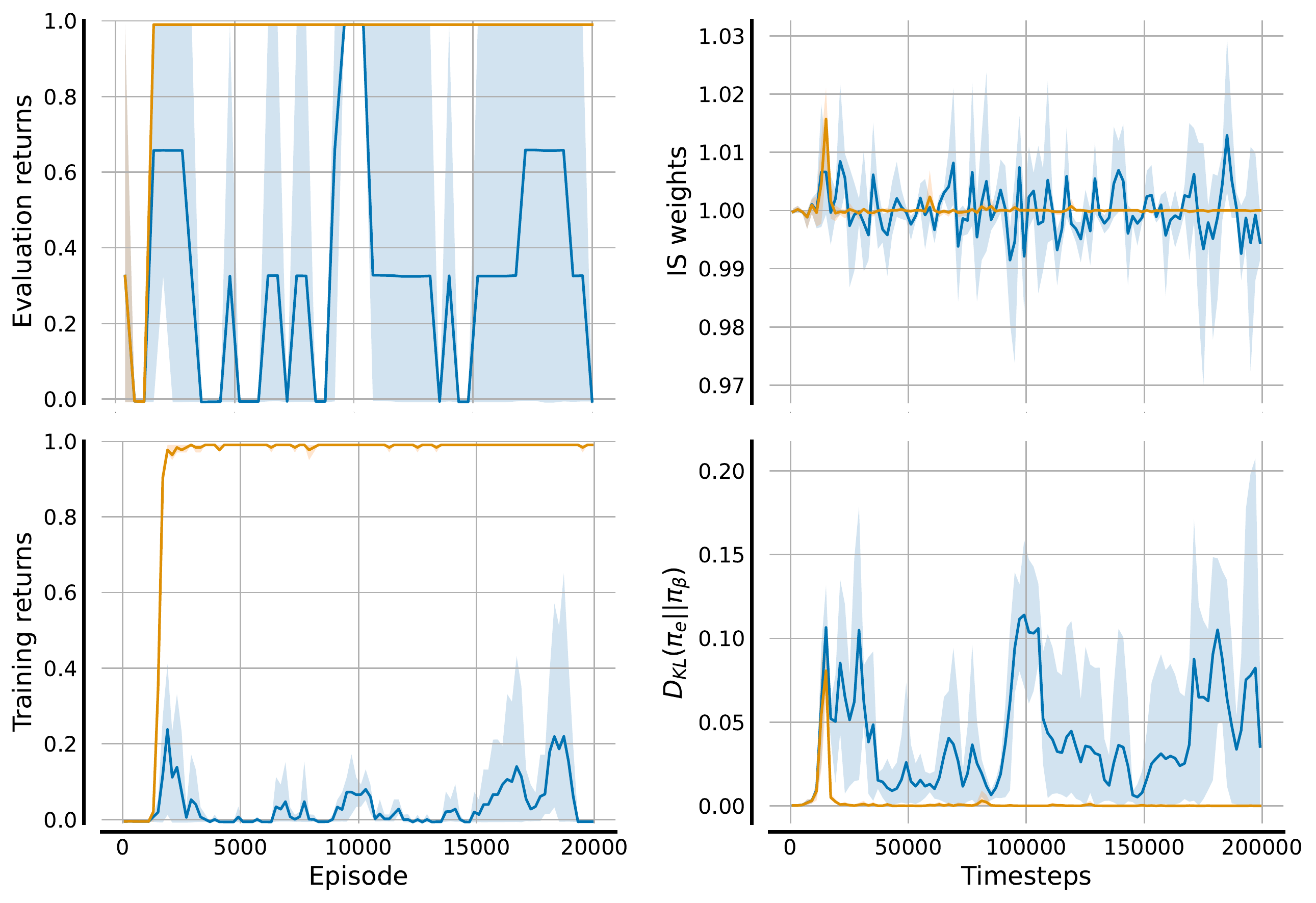}
        \caption {$\alpha_\beta=0.0001, \alpha_e=0.1$}
    \end{subfigure}
    \hfill
    \begin{subfigure}{.33\textwidth}
        \includegraphics[width=\textwidth]{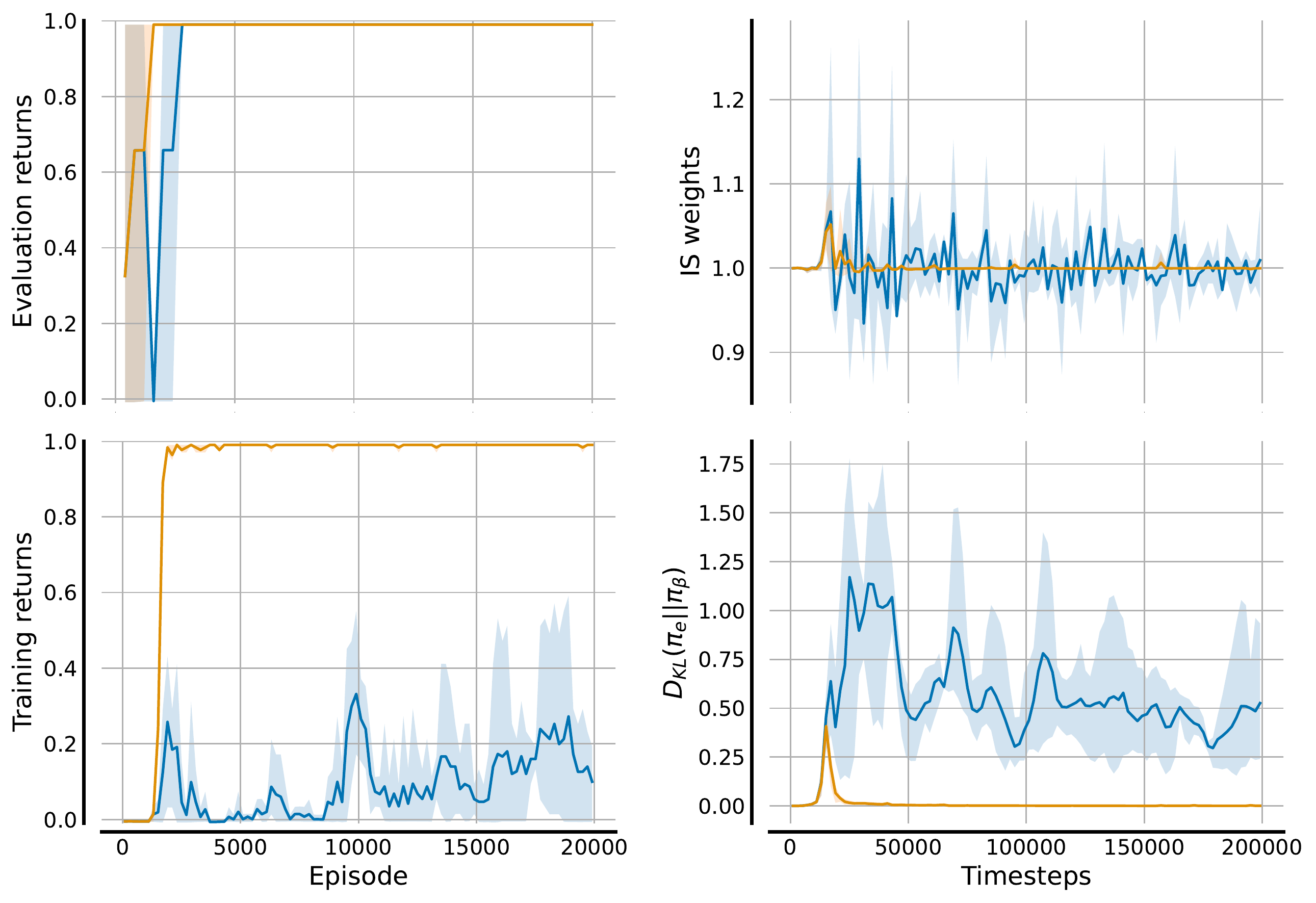}
        \caption {$\alpha_\beta=0.001, \alpha_e=0.0$}
    \end{subfigure}
    \hfill
    \begin{subfigure}{.33\textwidth}
        \includegraphics[width=\textwidth]{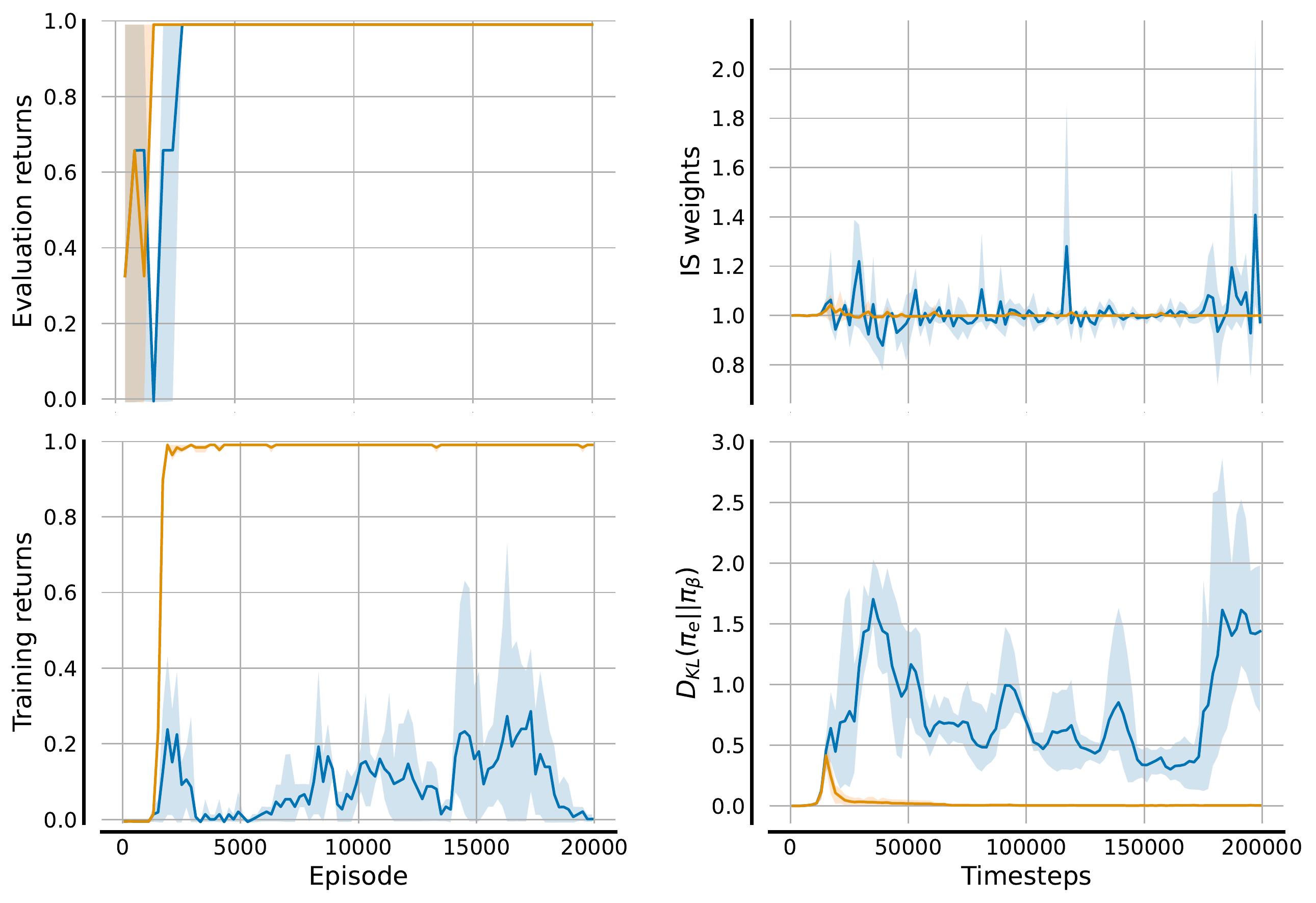}
        \caption {$\alpha_\beta=0.001, \alpha_e=0.0001$}
    \end{subfigure}
    
    \ \vspace{1em}
    
    \begin{subfigure}{.33\textwidth}
        \includegraphics[width=\textwidth]{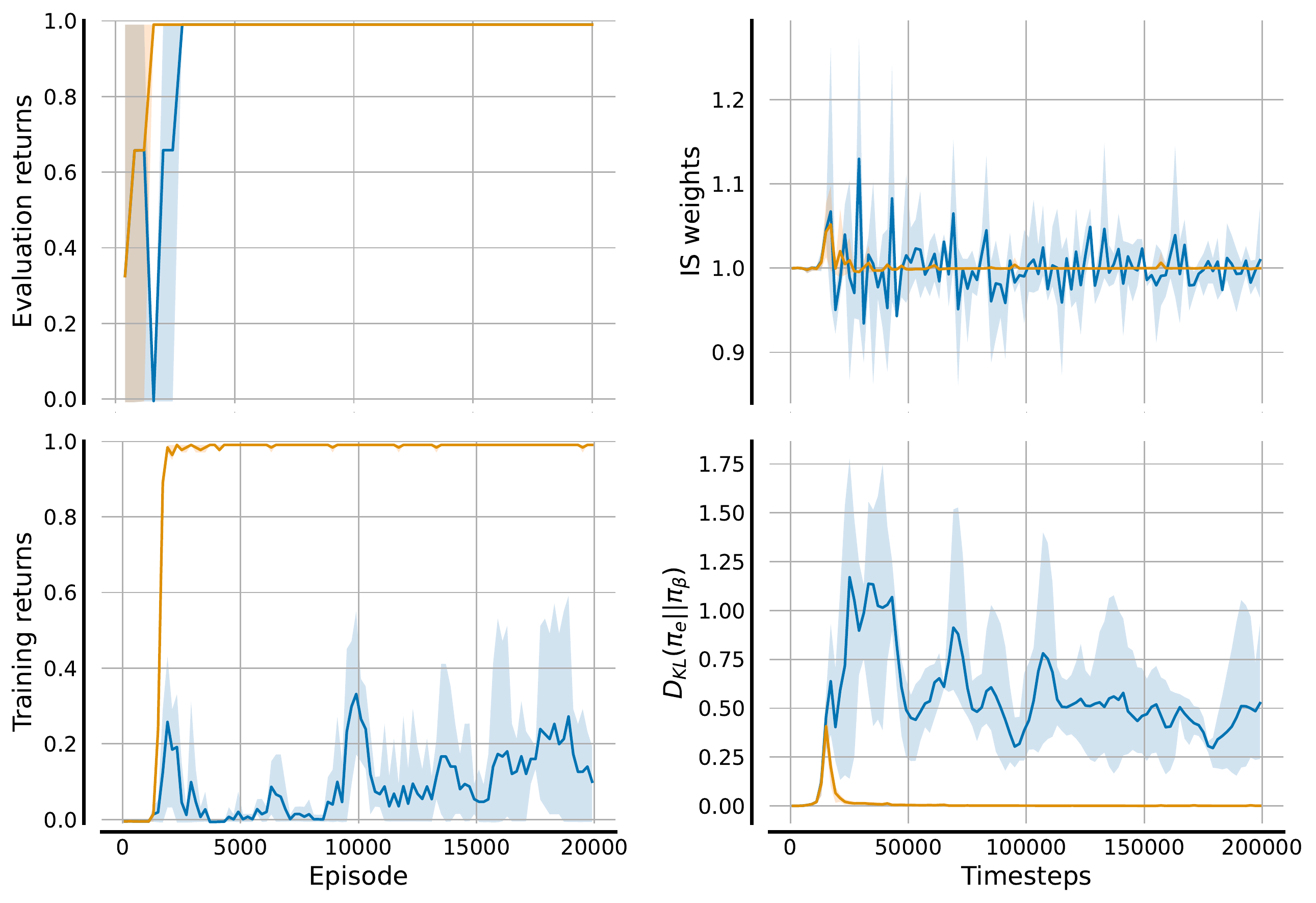}
        \caption {$\alpha_\beta=0.001, \alpha_e=0.001$}
    \end{subfigure}
    \hfill
    \begin{subfigure}{.33\textwidth}
        \includegraphics[width=\textwidth]{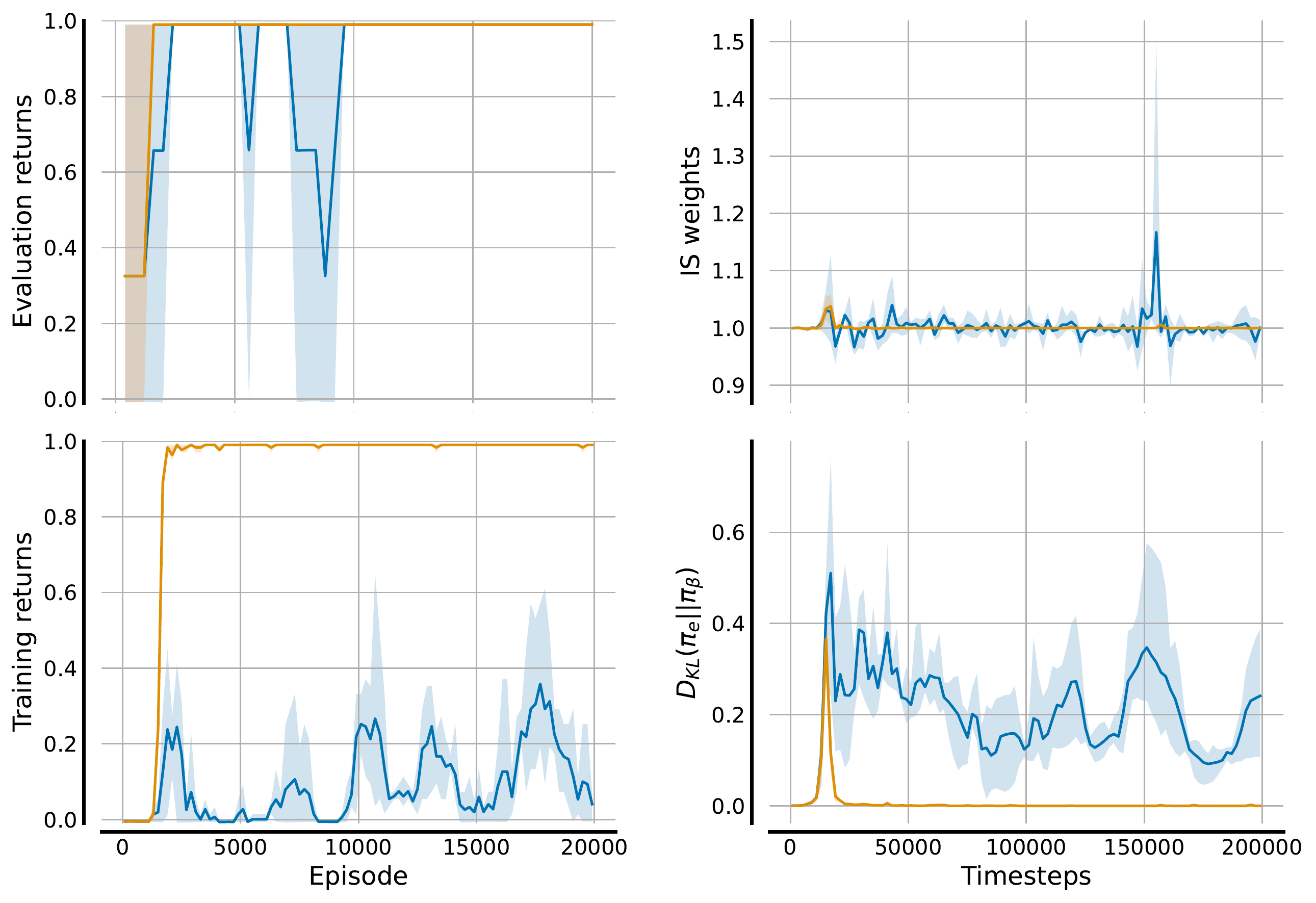}
        \caption {$\alpha_\beta=0.001, \alpha_e=0.01$}
    \end{subfigure}
    \hfill
    \begin{subfigure}{.33\textwidth}
        \includegraphics[width=\textwidth]{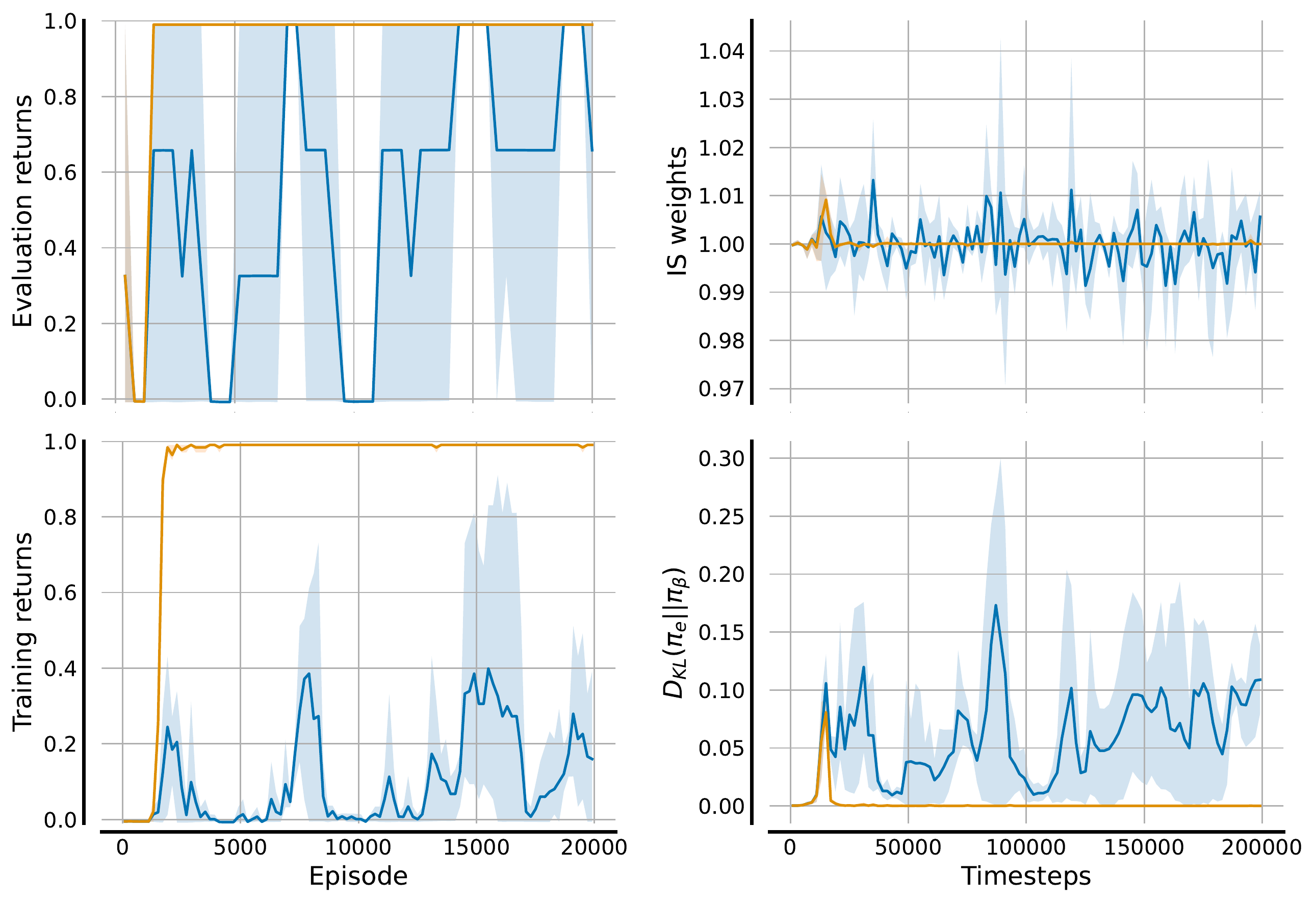}
        \caption {$\alpha_\beta=0.001, \alpha_e=0.1$}
    \end{subfigure}
    
    \ \vspace{1em}
    
    \begin{subfigure}{.33\textwidth}
        \includegraphics[width=\textwidth]{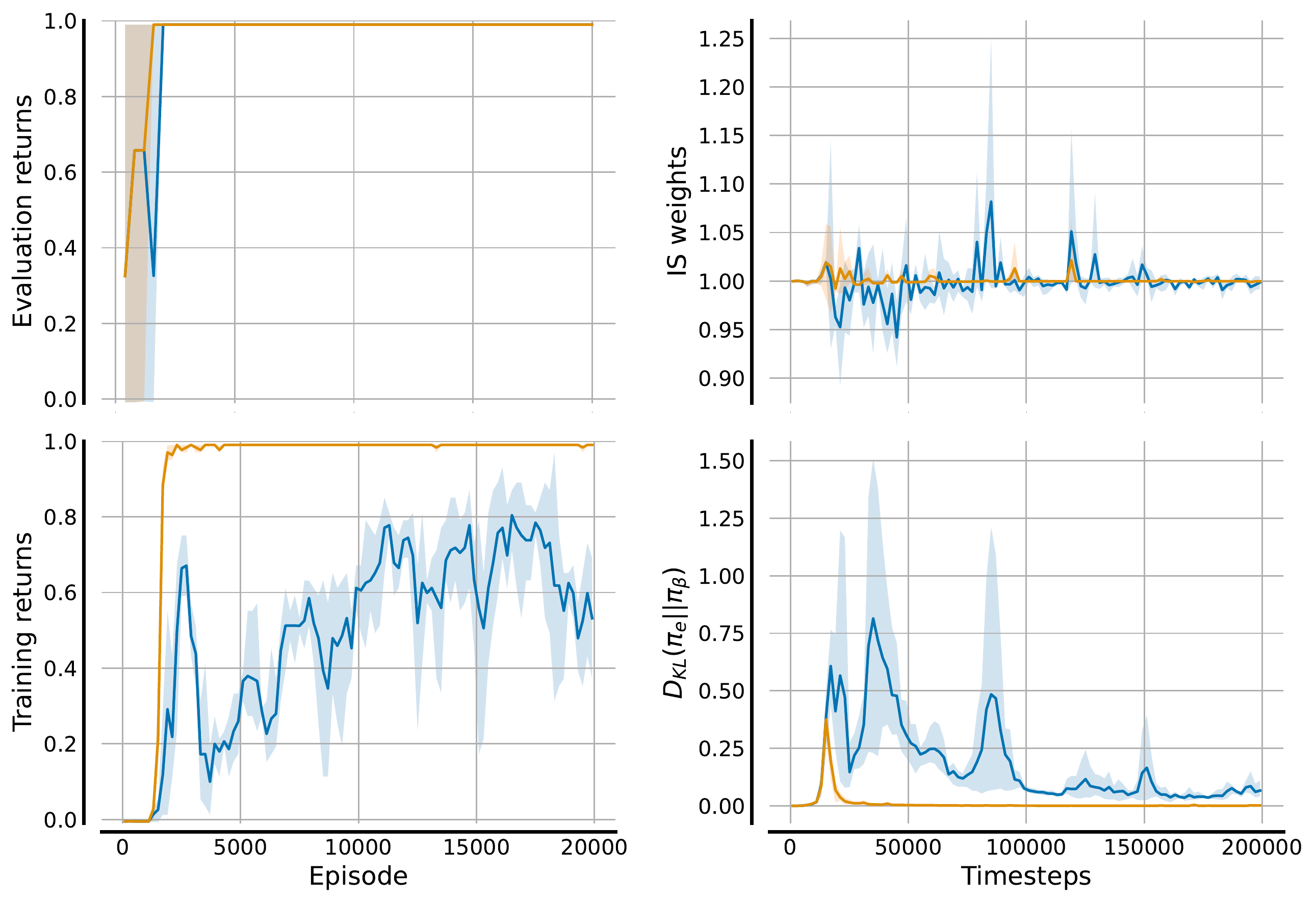}
        \caption {$\alpha_\beta=0.01, \alpha_e=0.0$}
    \end{subfigure}
    \hfill
    \begin{subfigure}{.33\textwidth}
        \includegraphics[width=\textwidth]{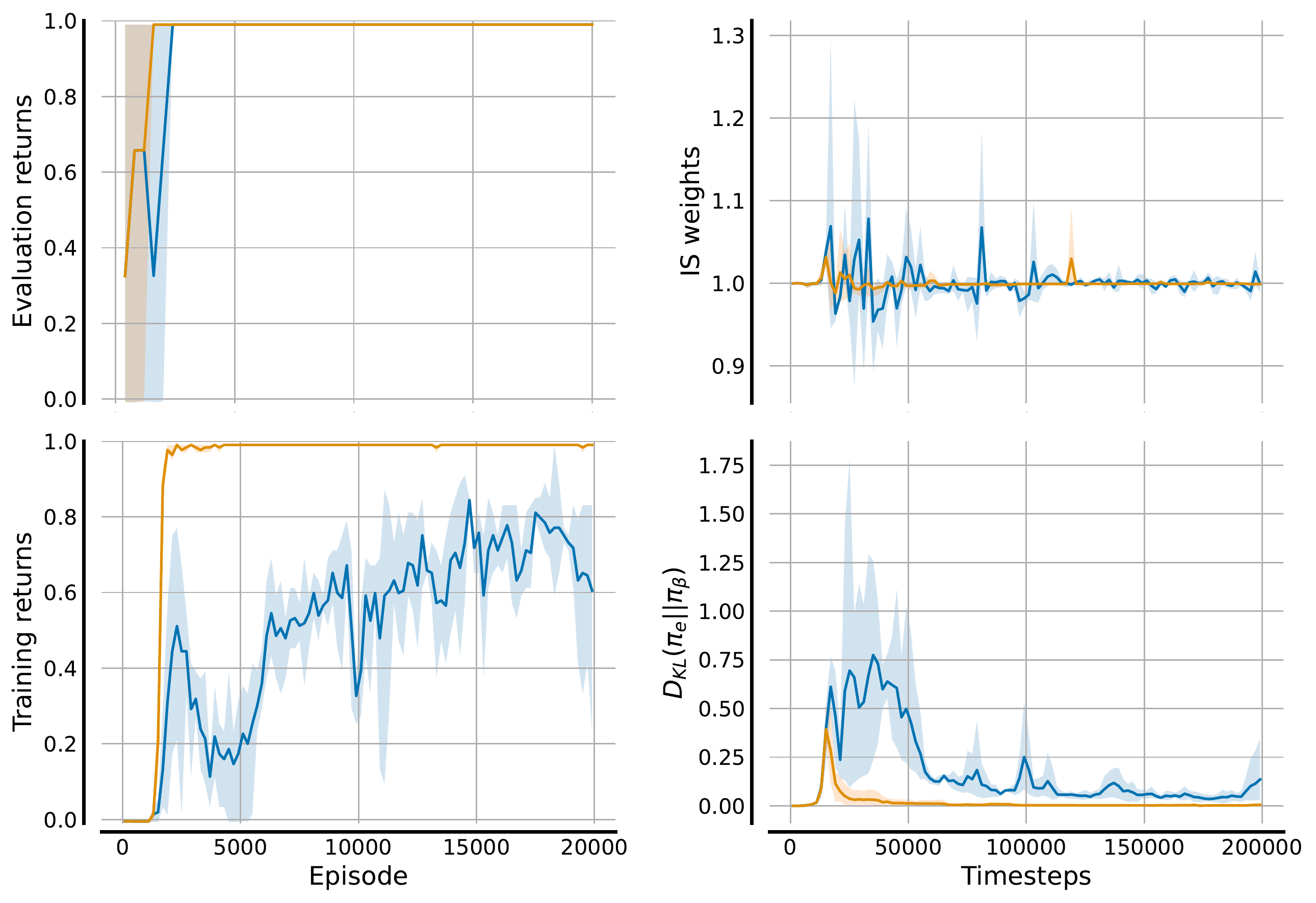}
        \caption {$\alpha_\beta=0.01, \alpha_e=0.0001$}
    \end{subfigure}
    \hfill
    \begin{subfigure}{.33\textwidth}
        \includegraphics[width=\textwidth]{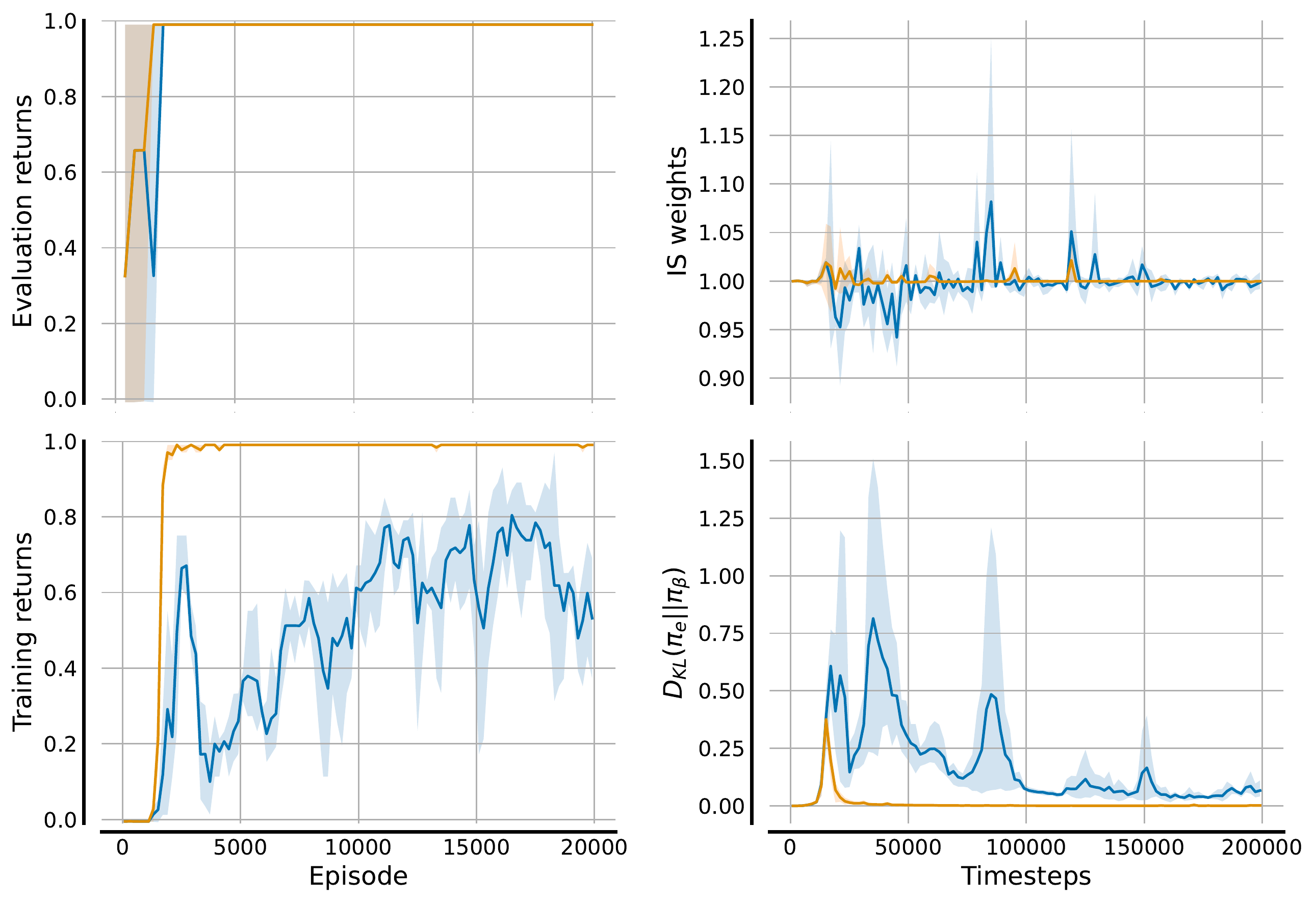}
        \caption {$\alpha_\beta=0.01, \alpha_e=0.001$}
    \end{subfigure}
    
    \ \vspace{1em}
    
    \begin{subfigure}{.33\textwidth}
        \includegraphics[width=\textwidth]{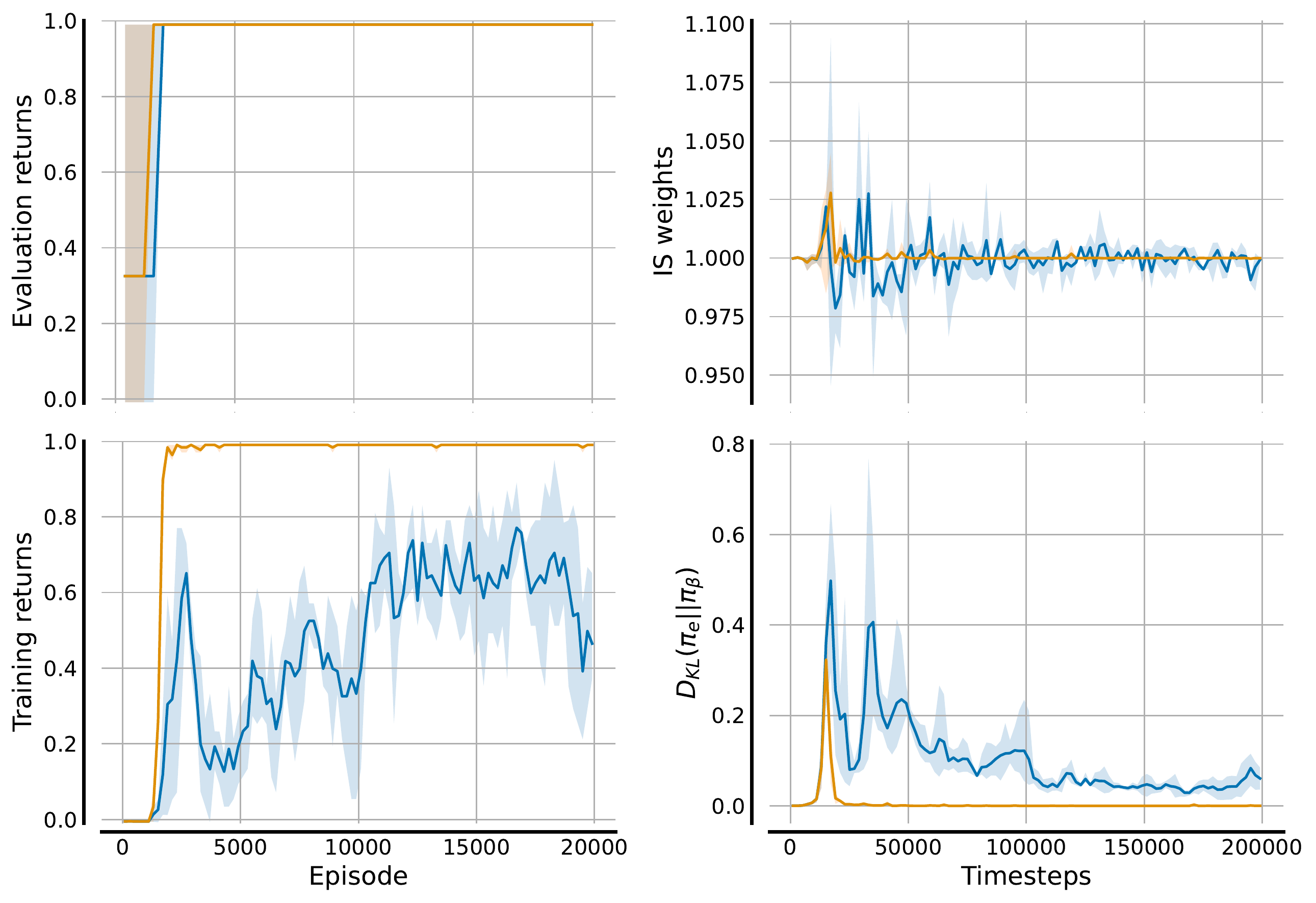}
        \caption {$\alpha_\beta=0.01, \alpha_e=0.01$}
    \end{subfigure}
    \hfill
    \begin{subfigure}{.33\textwidth}
        \includegraphics[width=\textwidth]{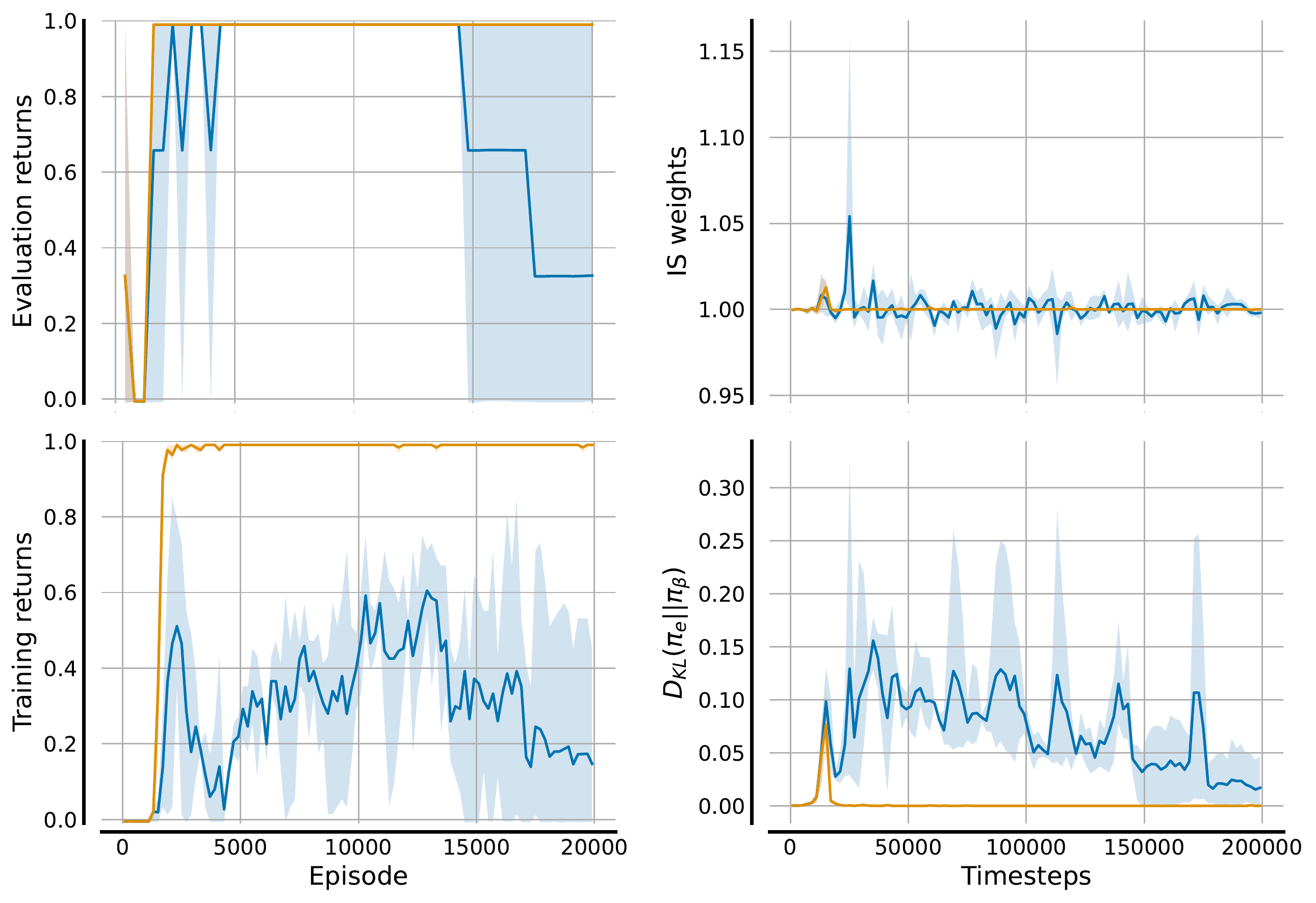}
        \caption {$\alpha_\beta=0.01, \alpha_e=0.1$}
    \end{subfigure}
    \hfill
    \begin{subfigure}{.33\textwidth}
        \includegraphics[width=\textwidth]{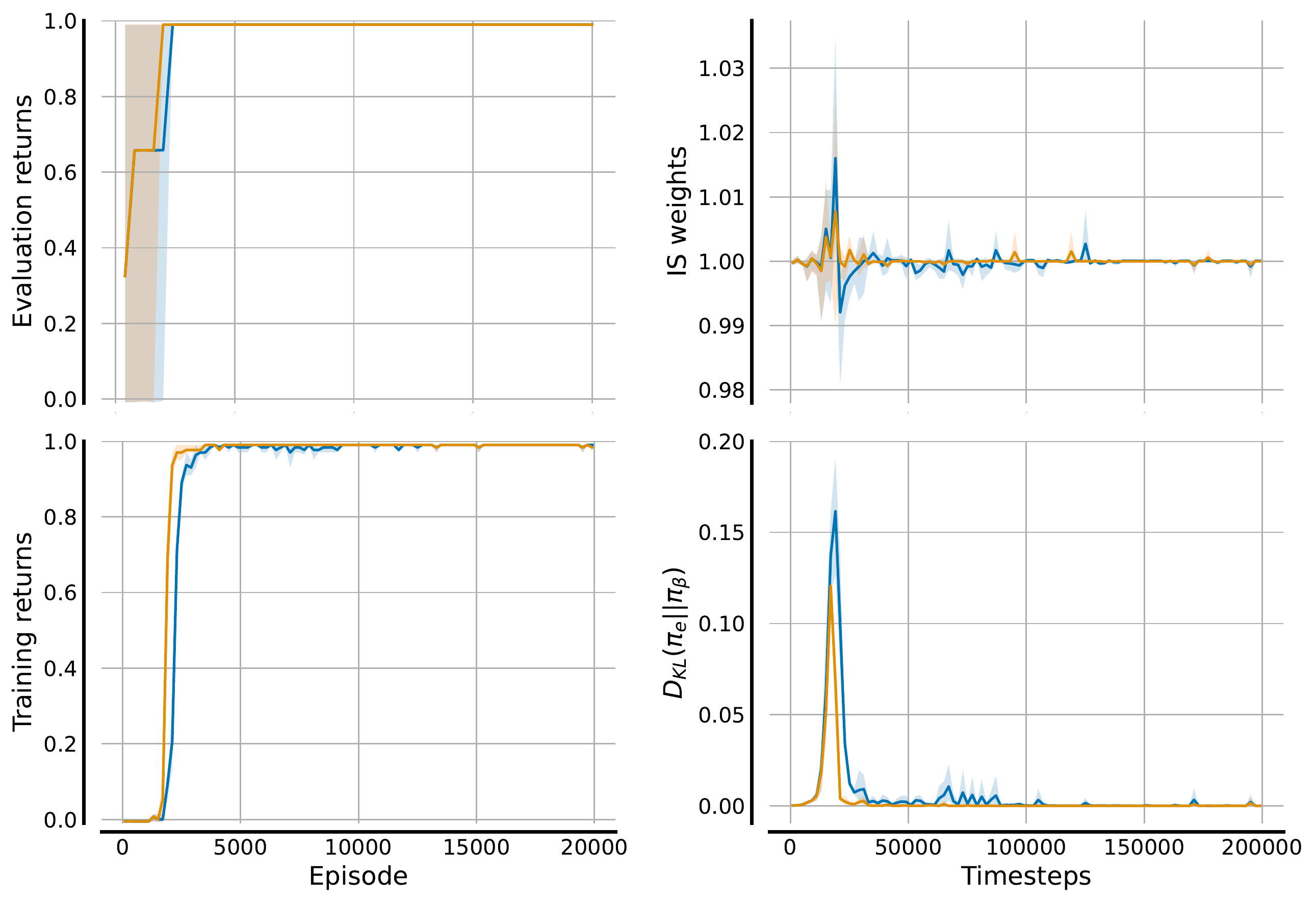}
        \caption {$\alpha_\beta=0.1, \alpha_e=0.0$}
    \end{subfigure}
    \begin{subfigure}{.5\textwidth}
        \centering
        \includegraphics[width=.5\textwidth]{media/kl_legend.pdf}
    \end{subfigure}
    \caption{DeepSea 10 evaluation with divergence constraint regularisation coefficients $\alpha_\beta$ and $\alpha_e$.  Shading indicates 95\% confidence intervals; Part 2}
    \label{fig:kl_divergence_constraint_ds10_2}
\end{figure}

\begin{figure}[!ht]
    \centering
    \begin{subfigure}{.33\textwidth}
        \includegraphics[width=\textwidth]{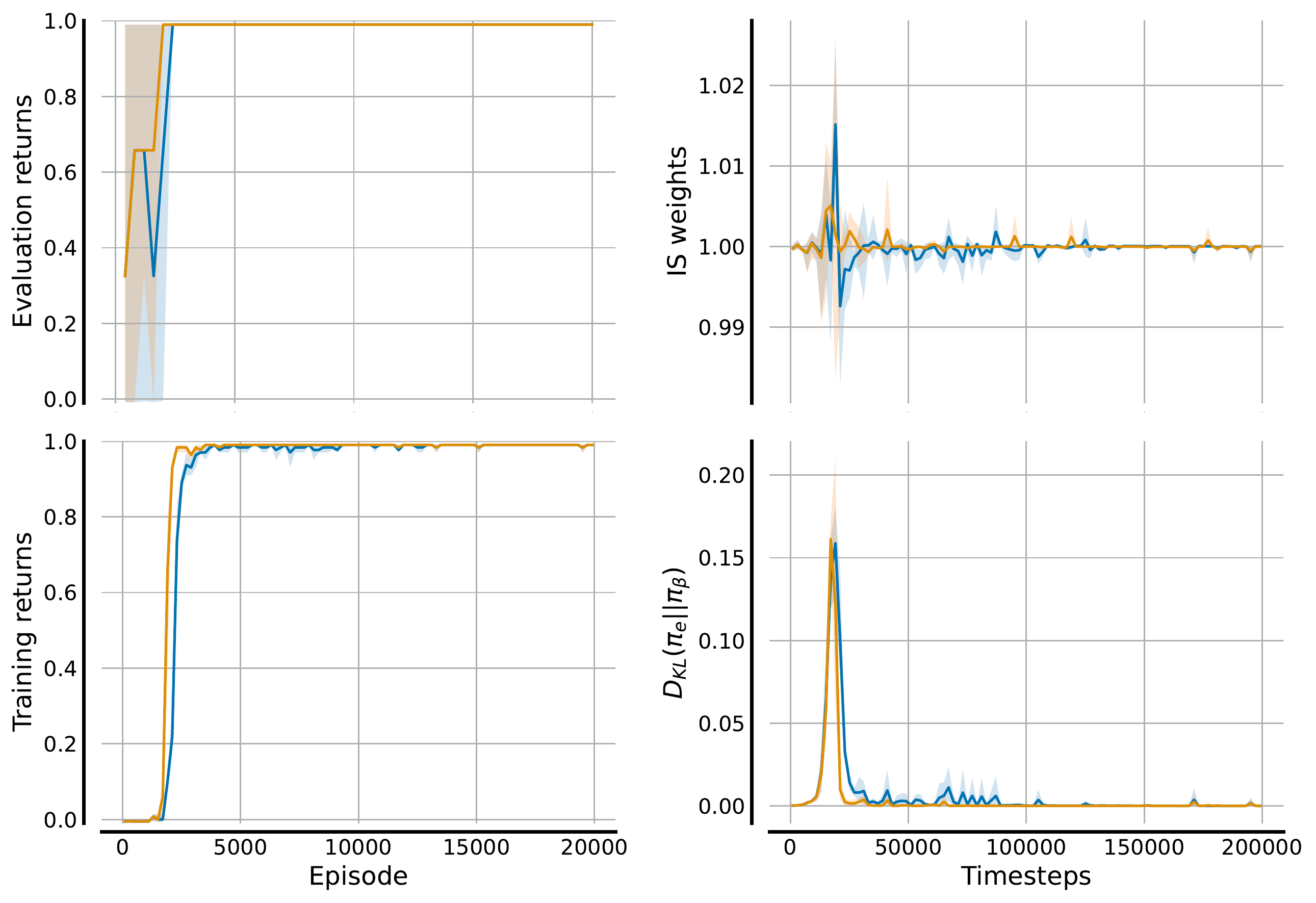}
        \caption {$\alpha_\beta=0.1, \alpha_e=0.0001$}
    \end{subfigure}
    \hfill
    \begin{subfigure}{.33\textwidth}
        \includegraphics[width=\textwidth]{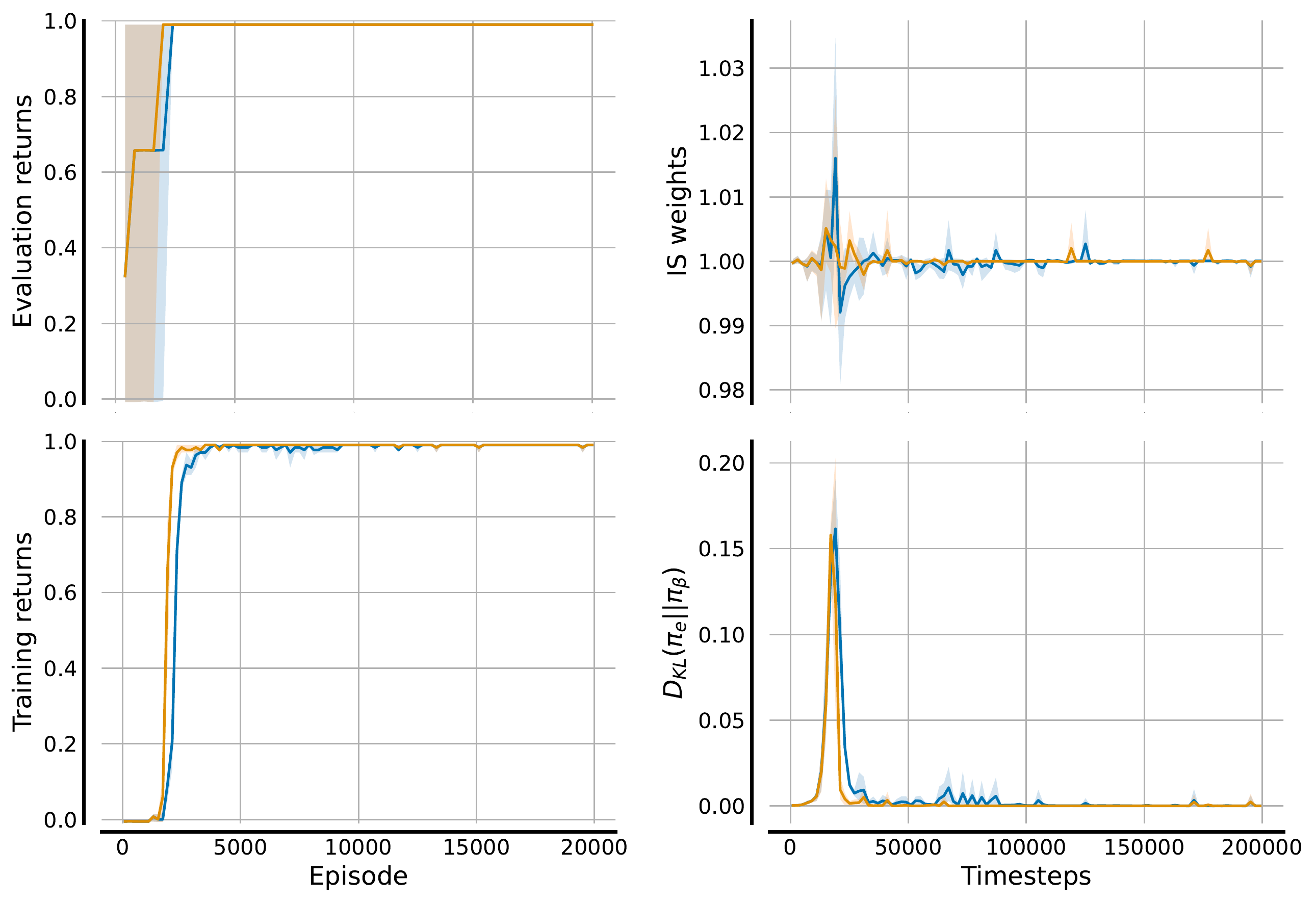}
        \caption {$\alpha_\beta=0.1, \alpha_e=0.001$}
    \end{subfigure}
    \hfill
    \begin{subfigure}{.33\textwidth}
        \includegraphics[width=\textwidth]{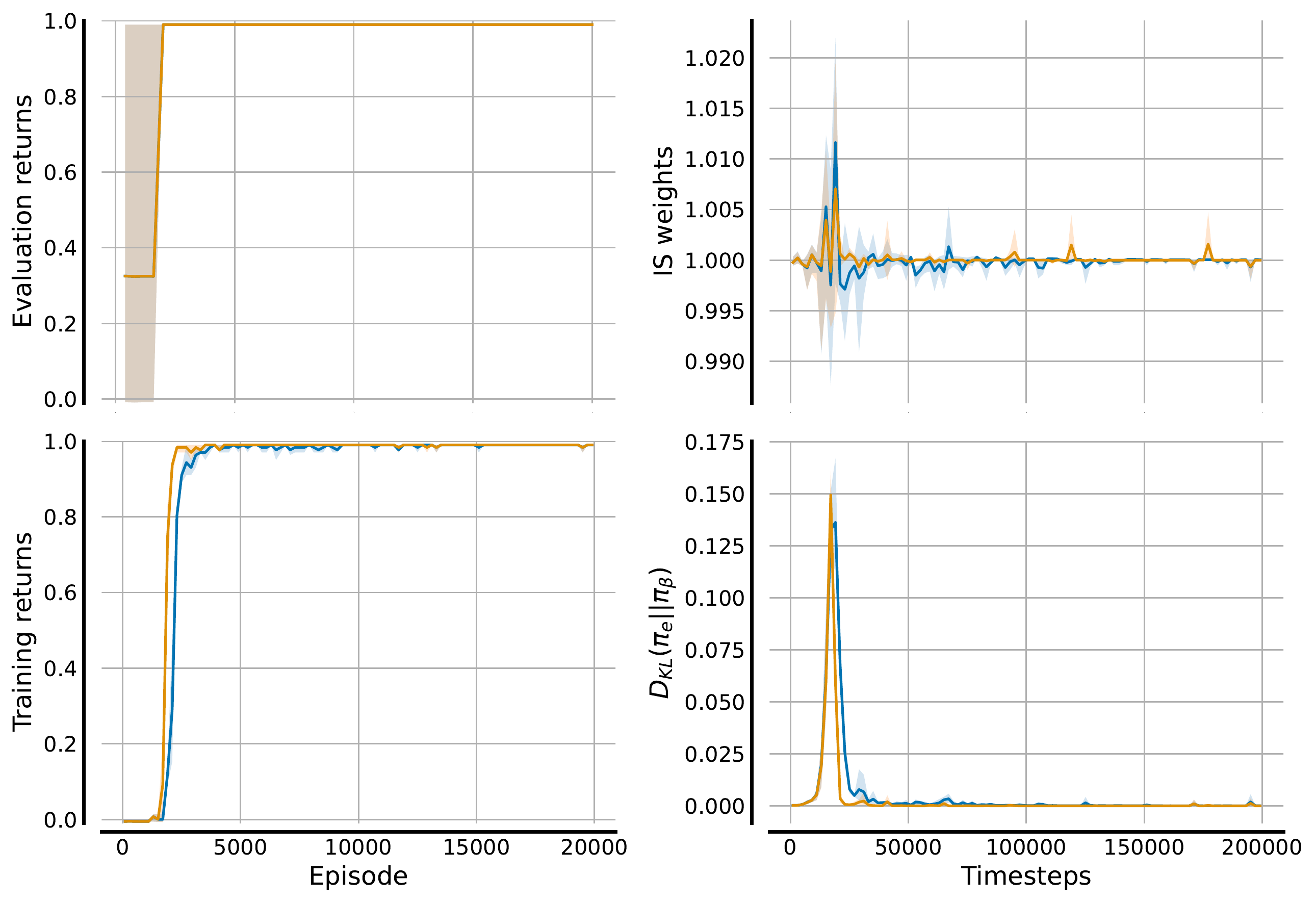}
        \caption {$\alpha_\beta=0.1, \alpha_e=0.01$}
    \end{subfigure}
    
    \ \vspace{1em}
    
    \begin{subfigure}{.33\textwidth}
        \includegraphics[width=\textwidth]{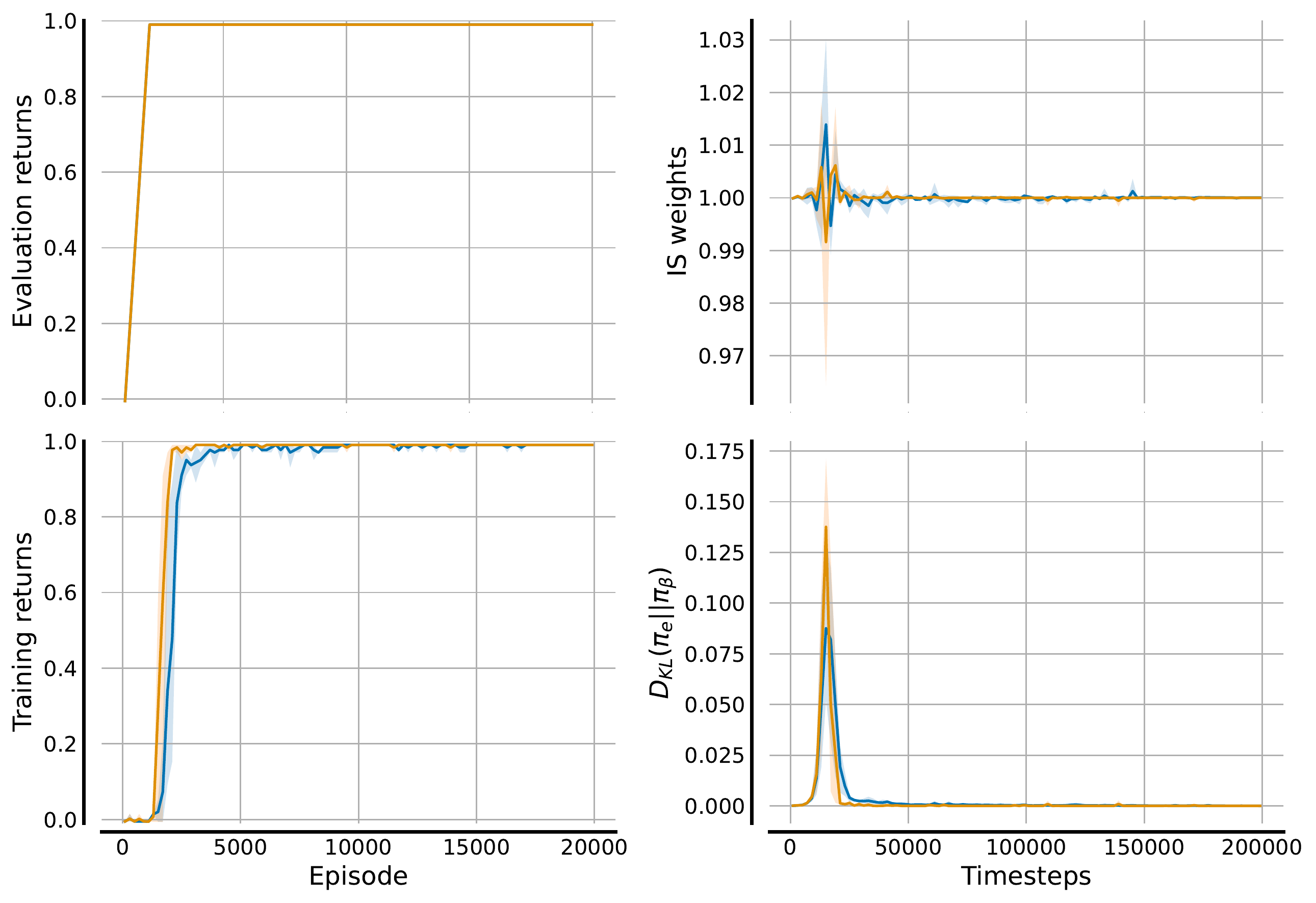}
        \caption {$\alpha_\beta=0.1, \alpha_e=0.1$}
    \end{subfigure}
    \hfill
    
    \begin{subfigure}{.5\textwidth}
        \centering
        \includegraphics[width=.5\textwidth]{media/kl_legend.pdf}
    \end{subfigure}
    \caption{DeepSea 10 evaluation with divergence constraint regularisation coefficients $\alpha_\beta$ and $\alpha_e$.  Shading indicates 95\% confidence intervals; Part 3}
    \label{fig:kl_divergence_constraint_ds10_3}
\end{figure}

\clearpage
\subsection{Hallway $N_l=N_r=20$}
\label{app:kl_divergence_constraint_hw20}

\begin{figure}[!ht]
    \centering
    \begin{subfigure}{.33\textwidth}
        \includegraphics[width=\textwidth]{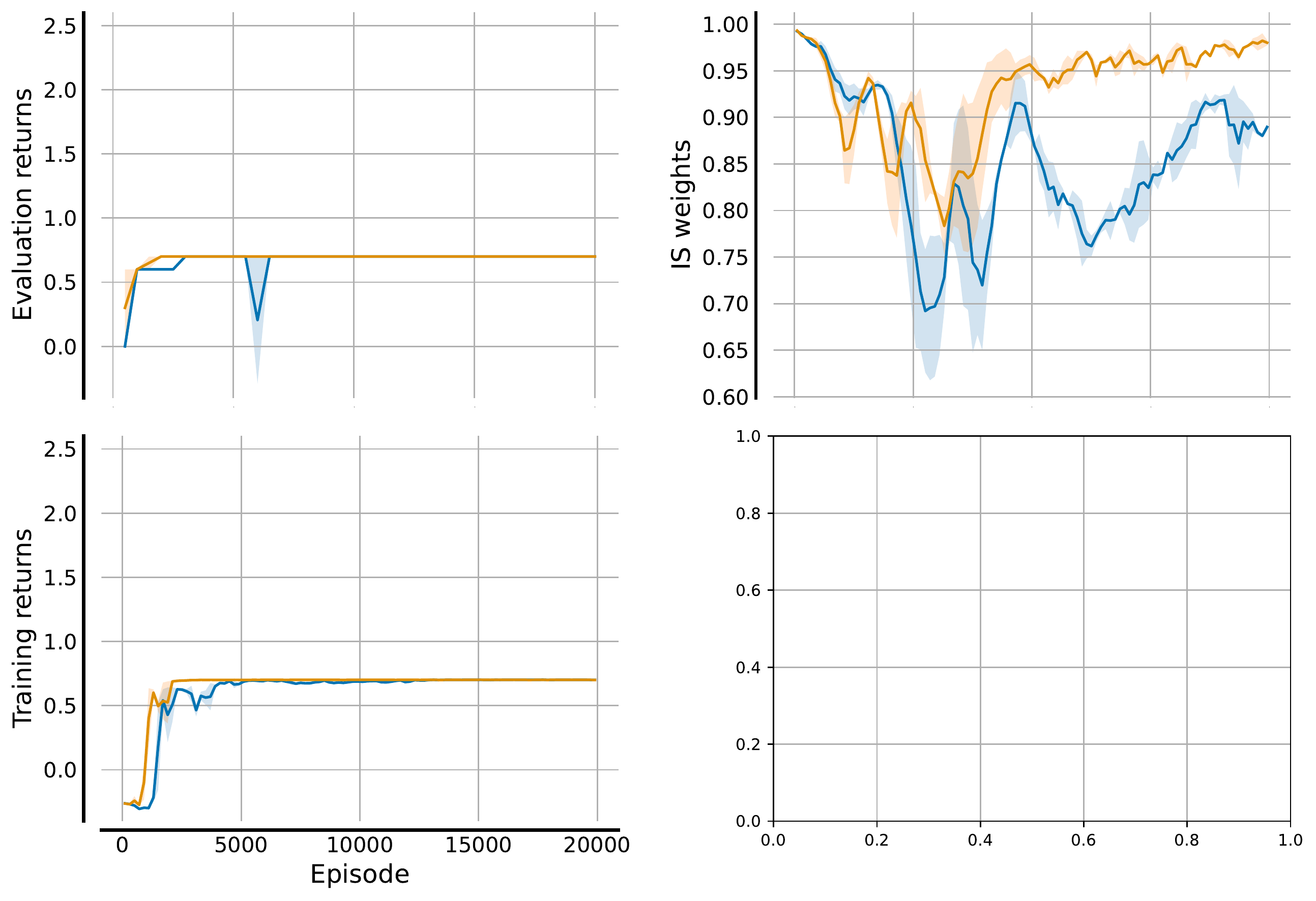}
        \caption {$\alpha_\beta=0.0, \alpha_e=0.0$}
    \end{subfigure}
    \hfill
    \begin{subfigure}{.33\textwidth}
        \includegraphics[width=\textwidth]{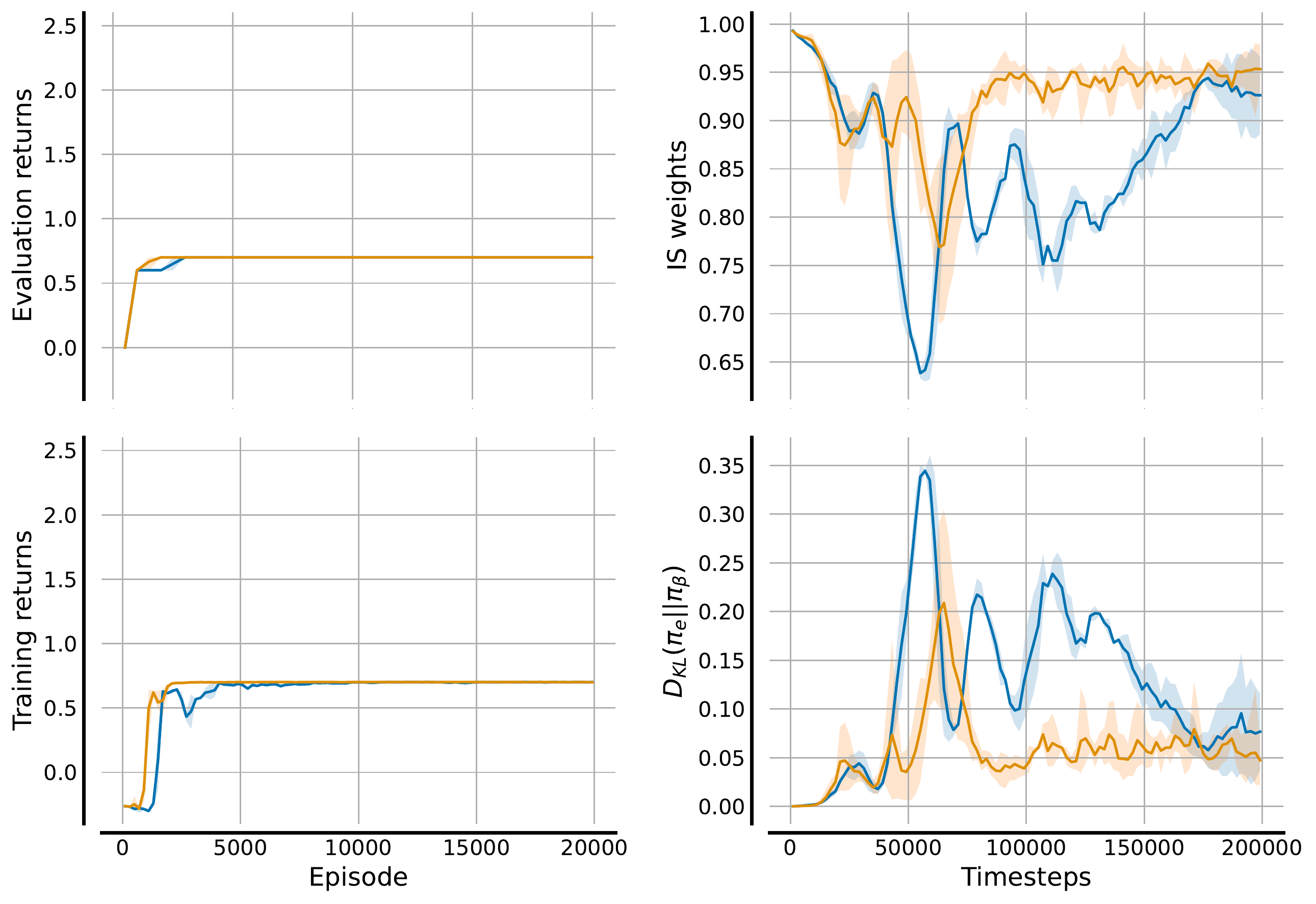}
        \caption {$\alpha_\beta=0.0, \alpha_e=0.0001$}
    \end{subfigure}
    \hfill
    \begin{subfigure}{.33\textwidth}
        \includegraphics[width=\textwidth]{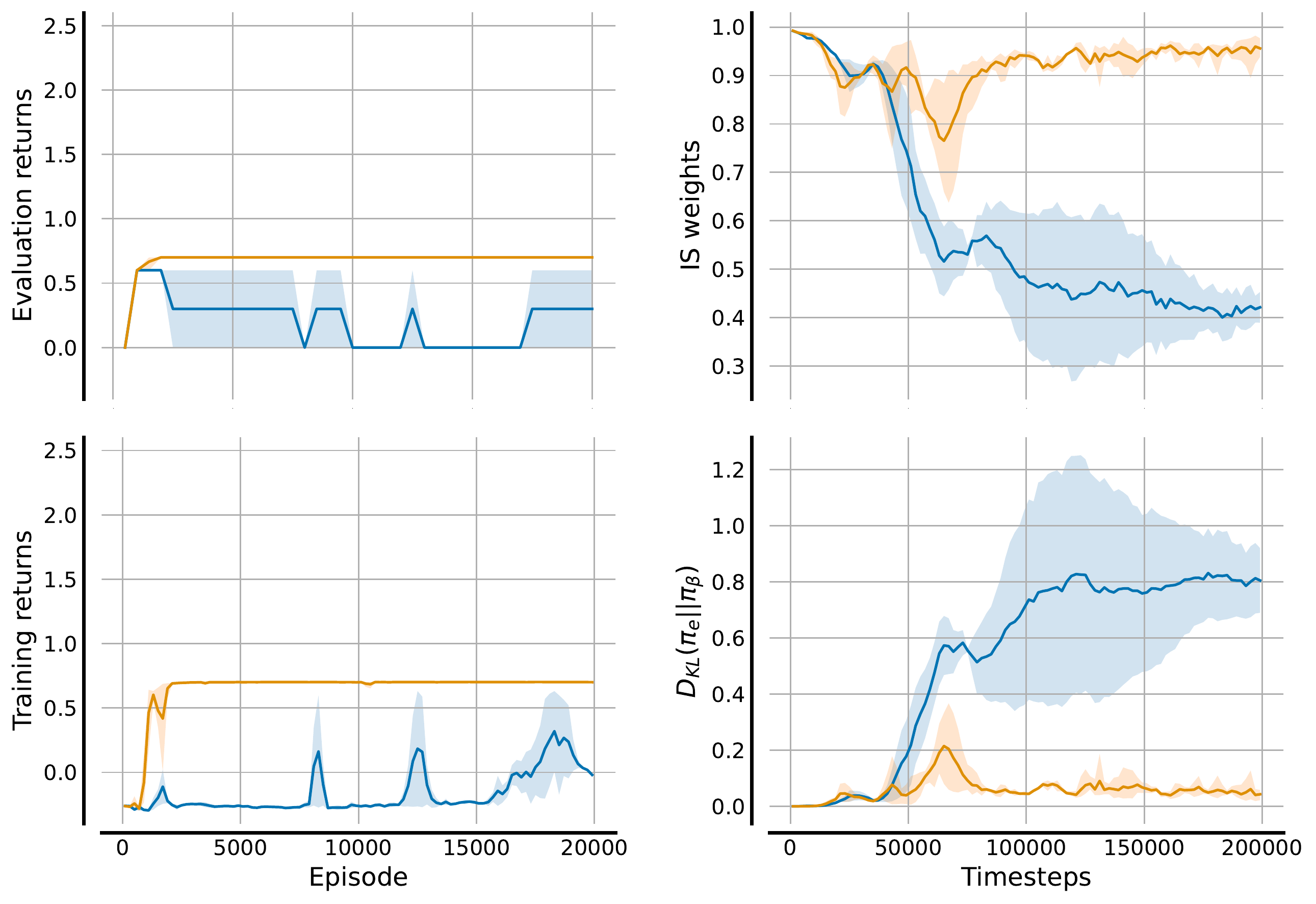}
        \caption {$\alpha_\beta=0.0, \alpha_e=0.001$}
    \end{subfigure}
    
    \ \vspace{1em}
    
    \begin{subfigure}{.33\textwidth}
        \includegraphics[width=\textwidth]{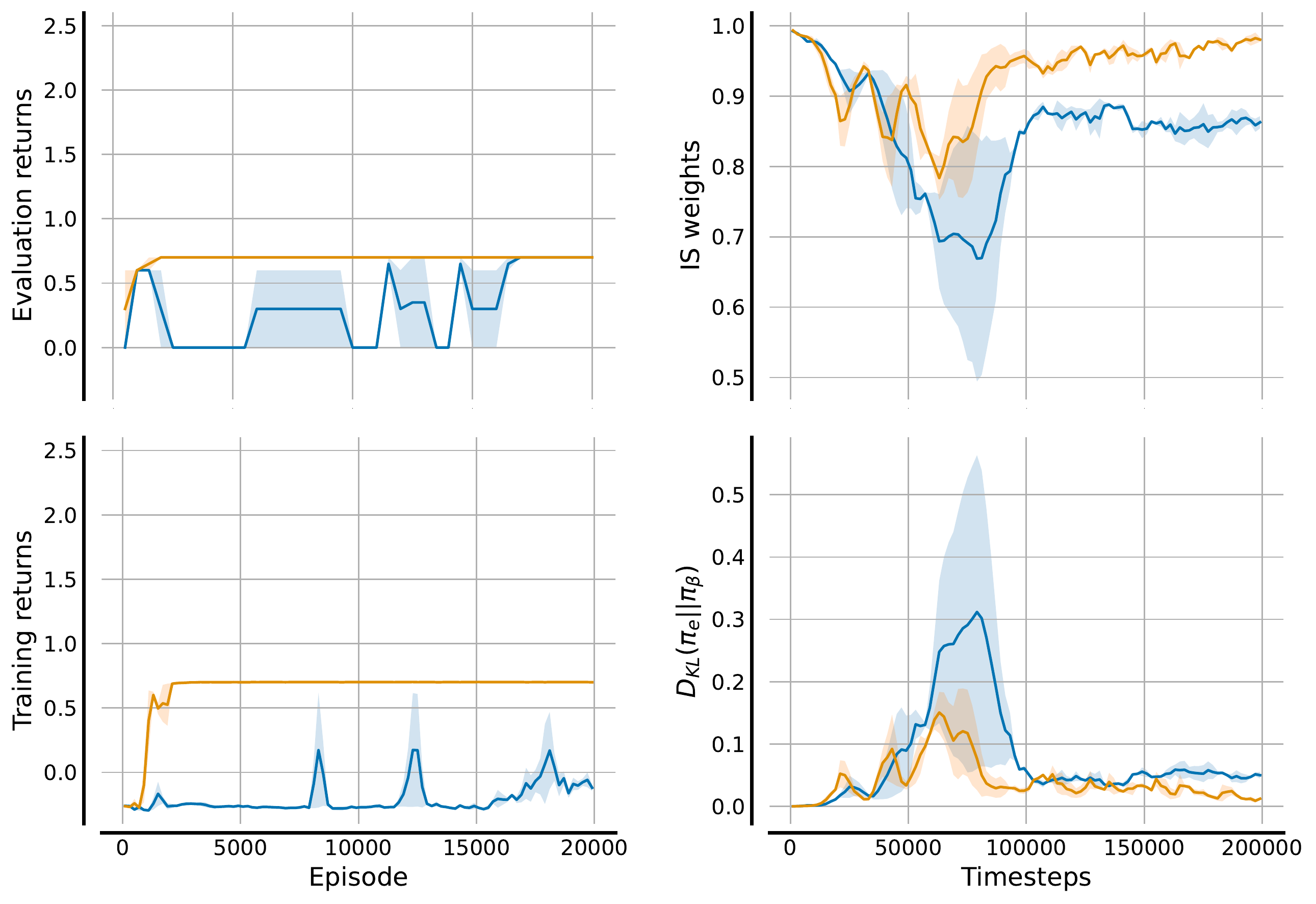}
        \caption {$\alpha_\beta=0.0, \alpha_e=0.01$}
    \end{subfigure}
    \hfill
    \begin{subfigure}{.33\textwidth}
        \includegraphics[width=\textwidth]{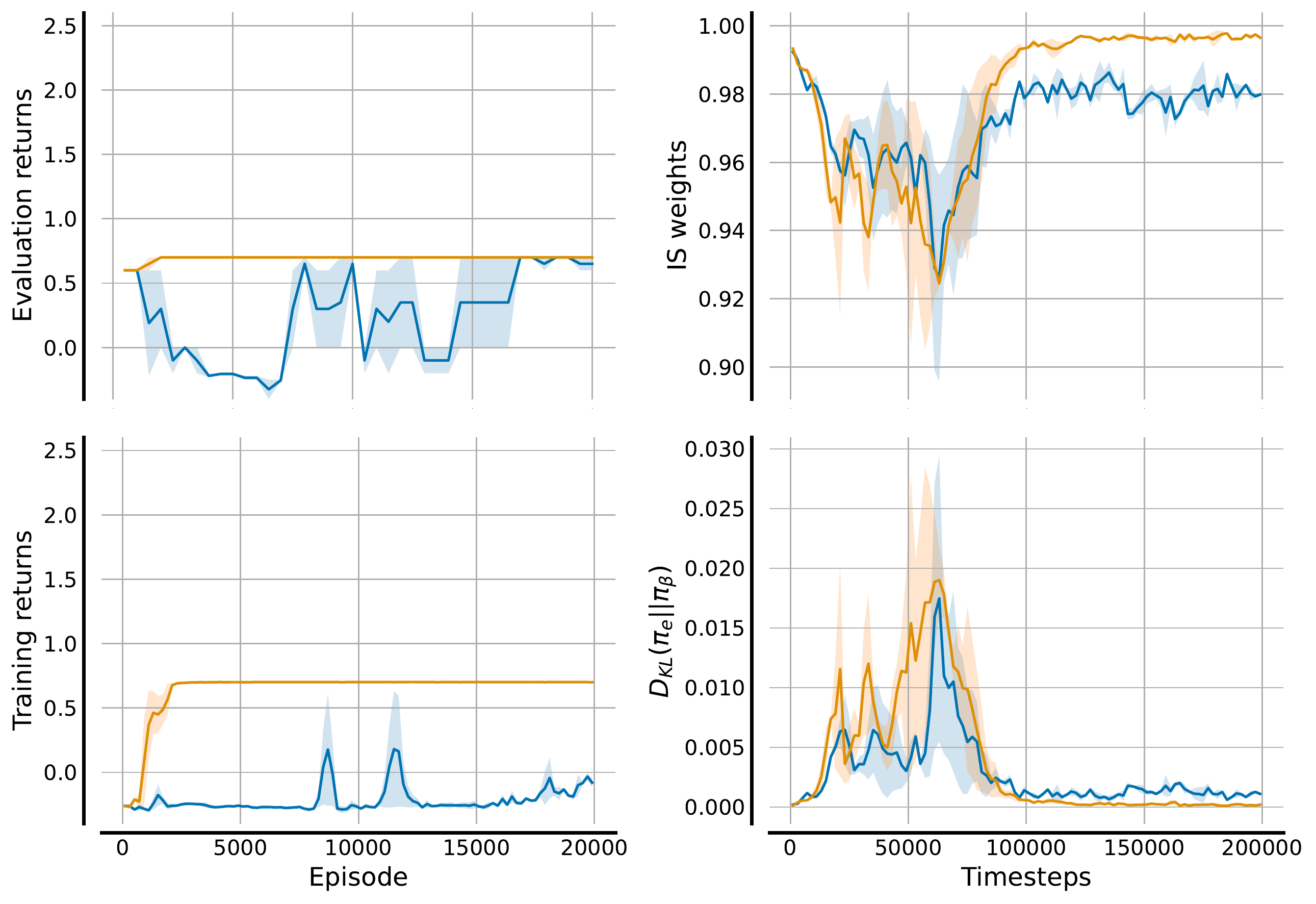}
        \caption {$\alpha_\beta=0.0, \alpha_e=0.1$}
    \end{subfigure}
    \hfill
    \begin{subfigure}{.33\textwidth}
        \includegraphics[width=\textwidth]{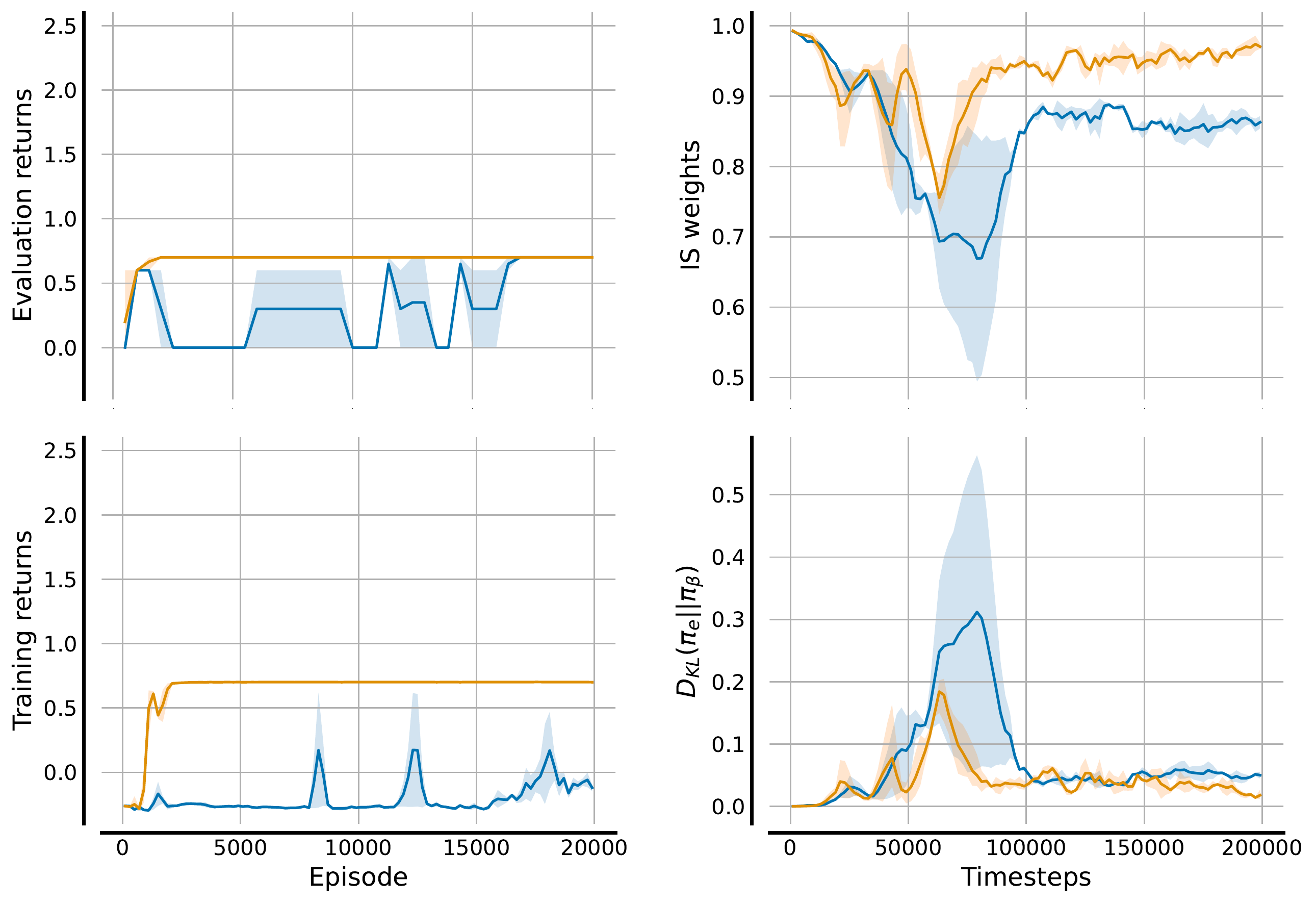}
        \caption {$\alpha_\beta=0.0001, \alpha_e=0.0$}
    \end{subfigure}
    
    \ \vspace{1em}
    
    \begin{subfigure}{.33\textwidth}
        \includegraphics[width=\textwidth]{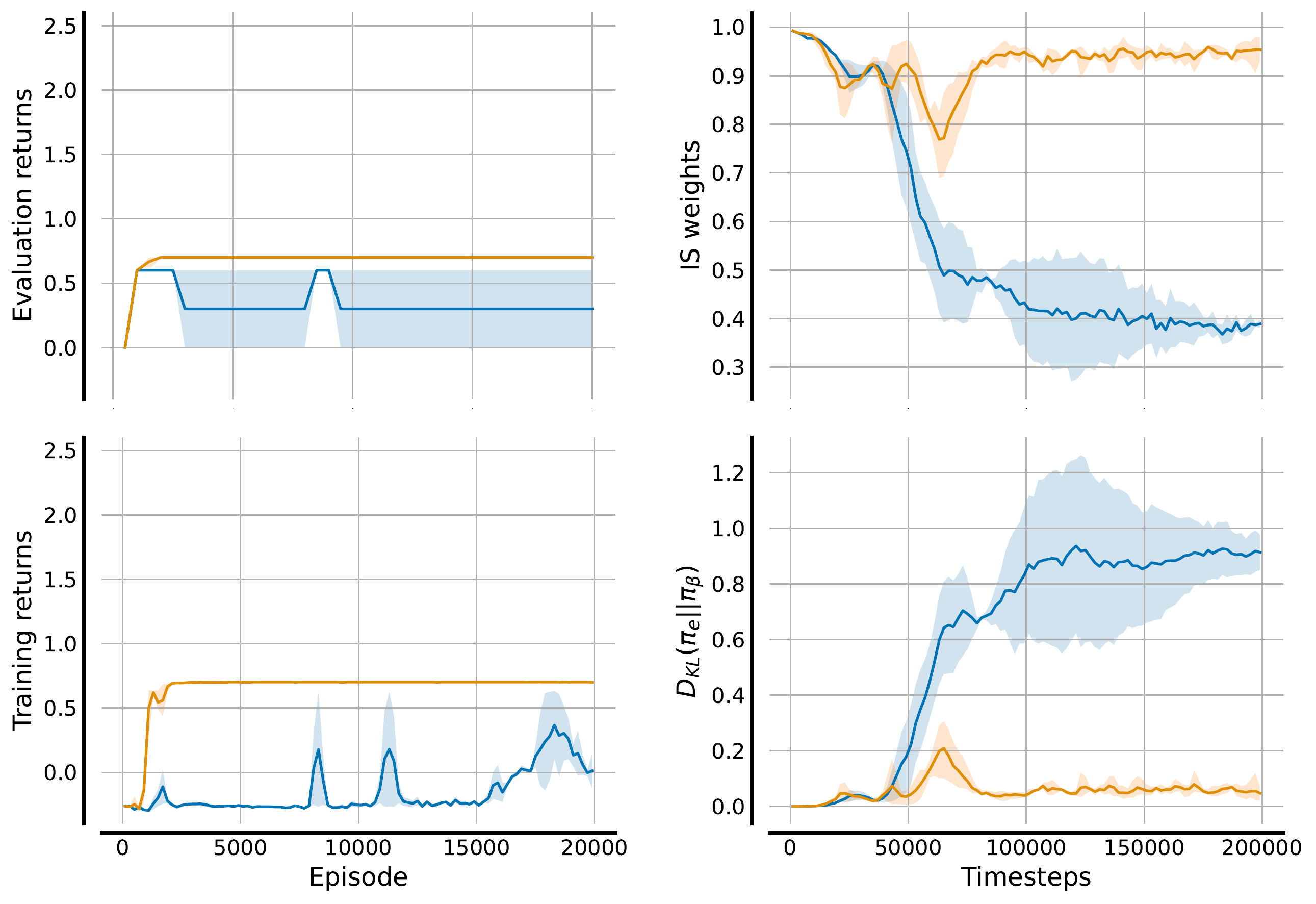}
        \caption {$\alpha_\beta=0.0001, \alpha_e=0.0001$}
    \end{subfigure}
    \hfill
    \begin{subfigure}{.33\textwidth}
        \includegraphics[width=\textwidth]{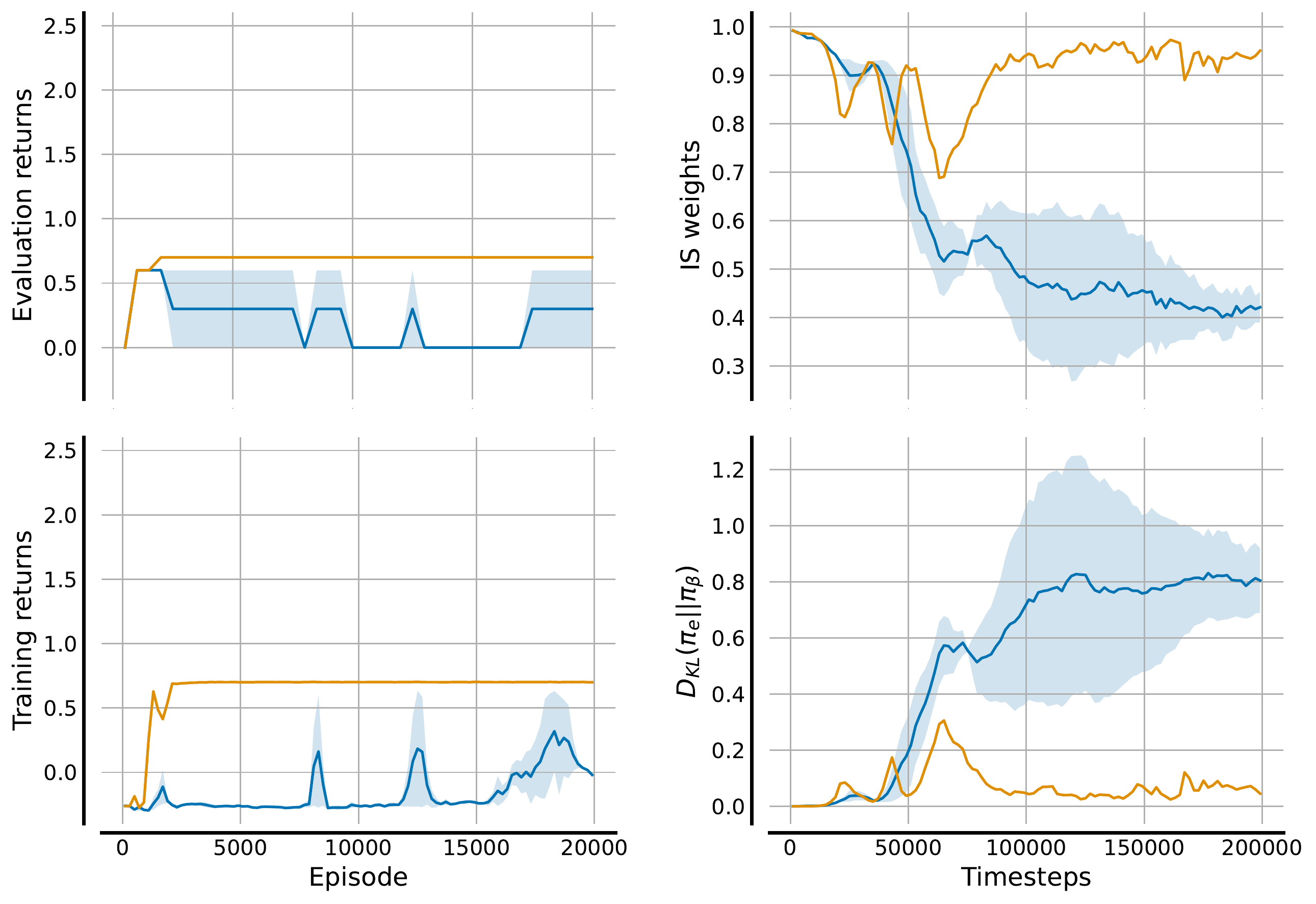}
        \caption {$\alpha_\beta=0.0001, \alpha_e=0.001$}
    \end{subfigure}
    \hfill
    \begin{subfigure}{.33\textwidth}
        \includegraphics[width=\textwidth]{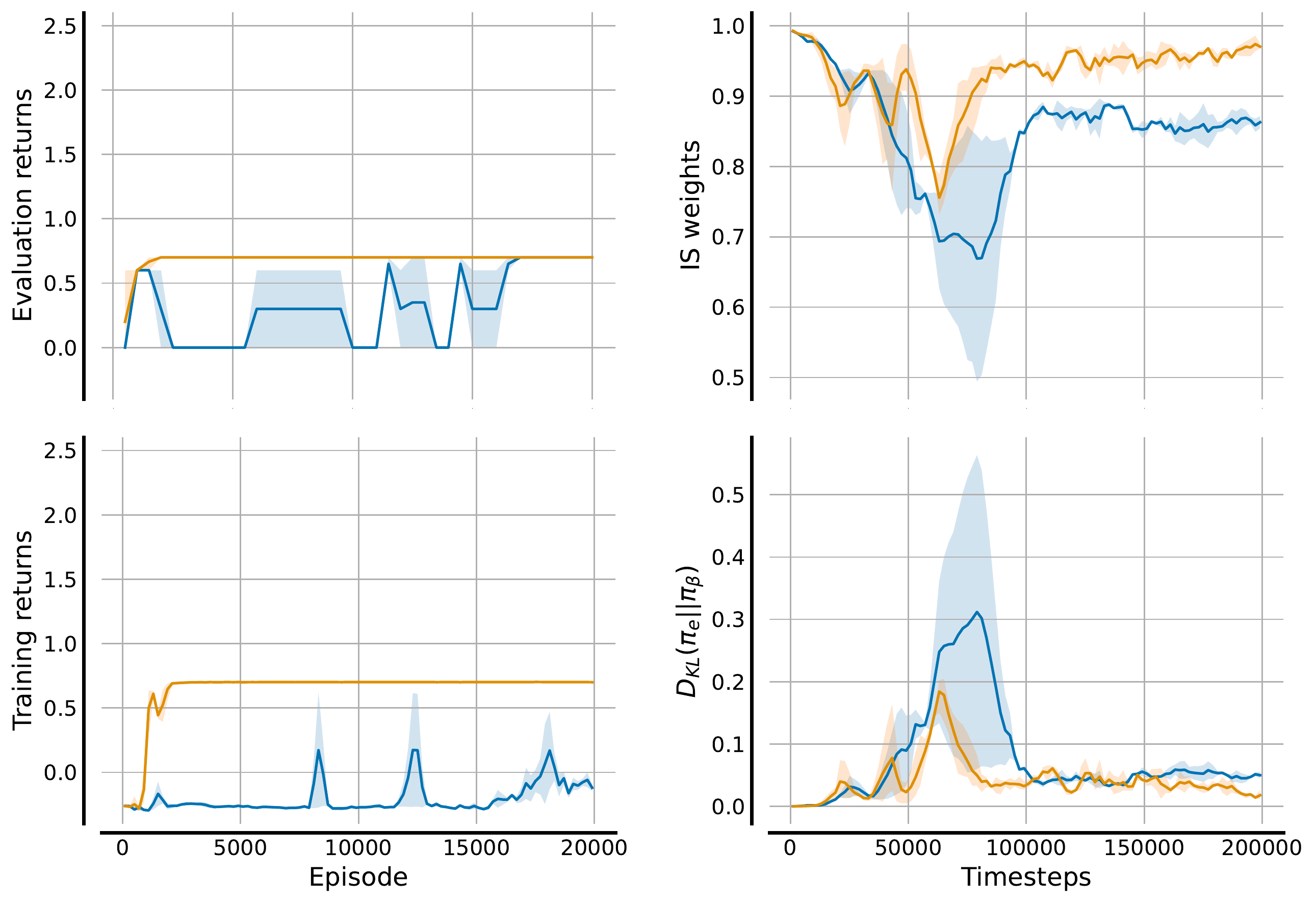}
        \caption {$\alpha_\beta=0.0001, \alpha_e=0.01$}
    \end{subfigure}
    
    \begin{subfigure}{.5\textwidth}
        \centering
        \includegraphics[width=.5\textwidth]{media/kl_legend.pdf}
    \end{subfigure}
    \caption{Hallway $N_l=N_r=20$ evaluation with divergence constraint regularisation coefficients $\alpha_\beta$ and $\alpha_e$.  Shading indicates 95\% confidence intervals; Part 1}
    \label{fig:kl_divergence_constraint_hw20_1}
\end{figure}

\begin{figure}[!ht]
    \centering
    \begin{subfigure}{.33\textwidth}
        \includegraphics[width=\textwidth]{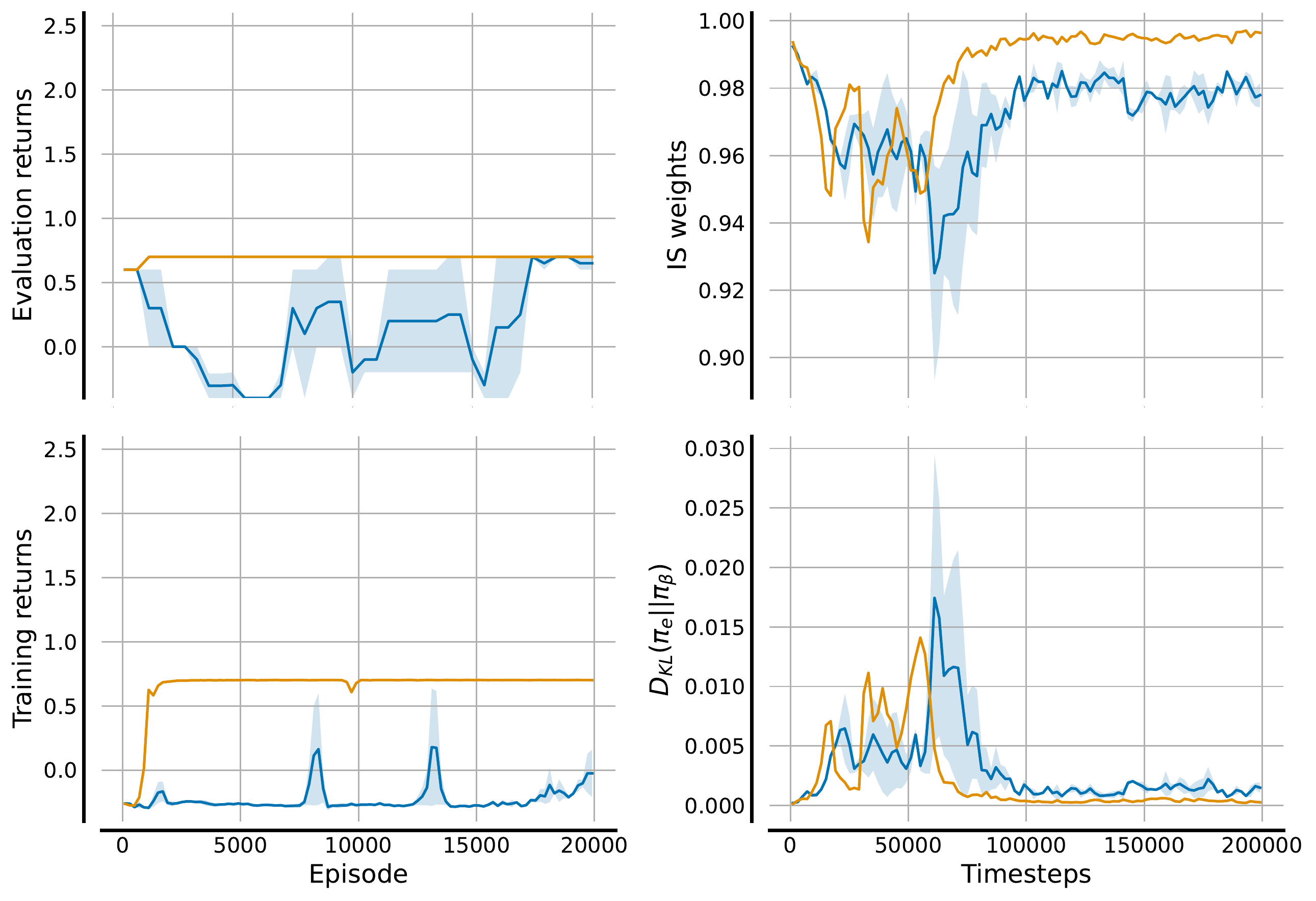}
        \caption {$\alpha_\beta=0.0001, \alpha_e=0.1$}
    \end{subfigure}
    \hfill
    \begin{subfigure}{.33\textwidth}
        \includegraphics[width=\textwidth]{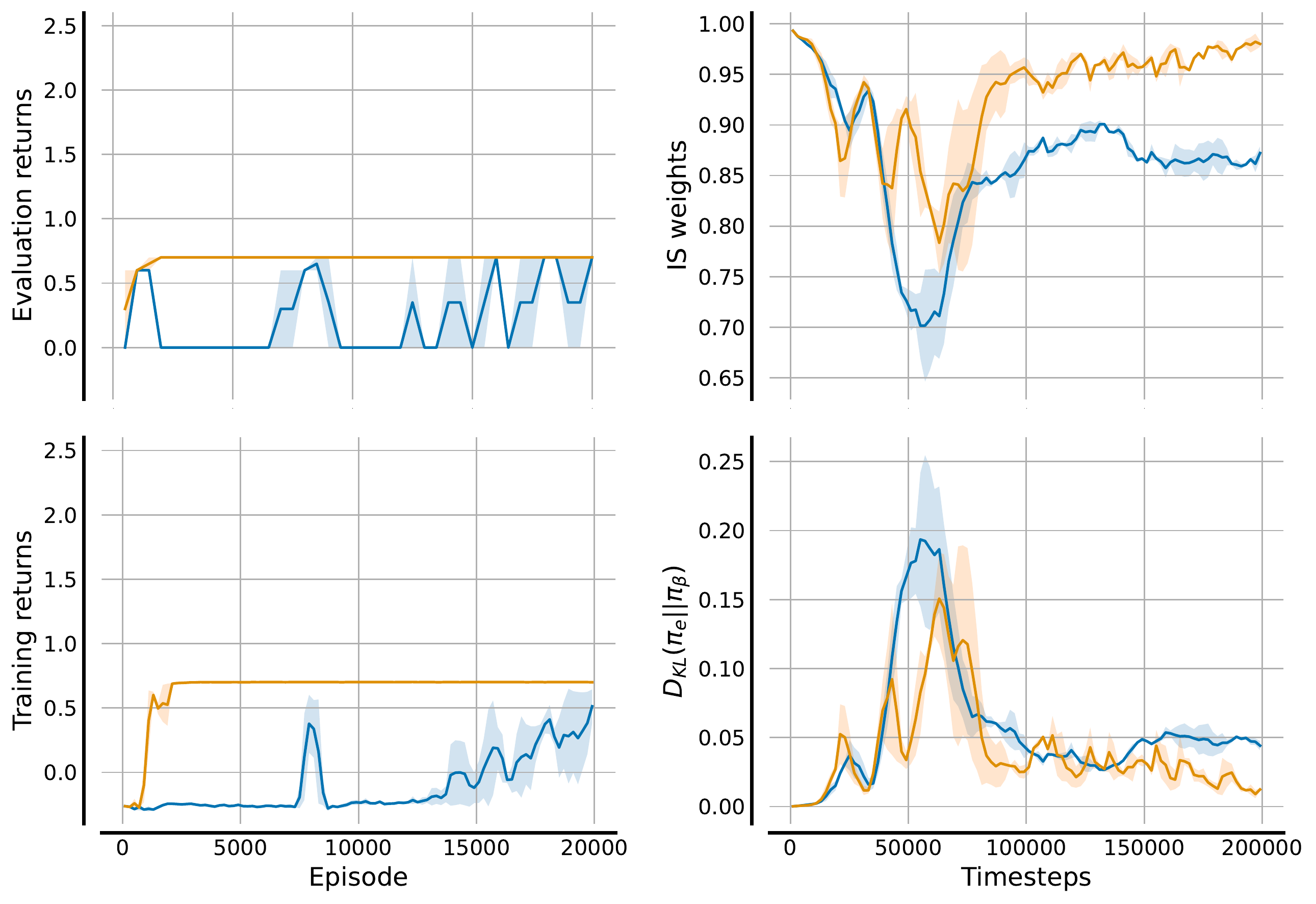}
        \caption {$\alpha_\beta=0.001, \alpha_e=0.0$}
    \end{subfigure}
    \hfill
    \begin{subfigure}{.33\textwidth}
        \includegraphics[width=\textwidth]{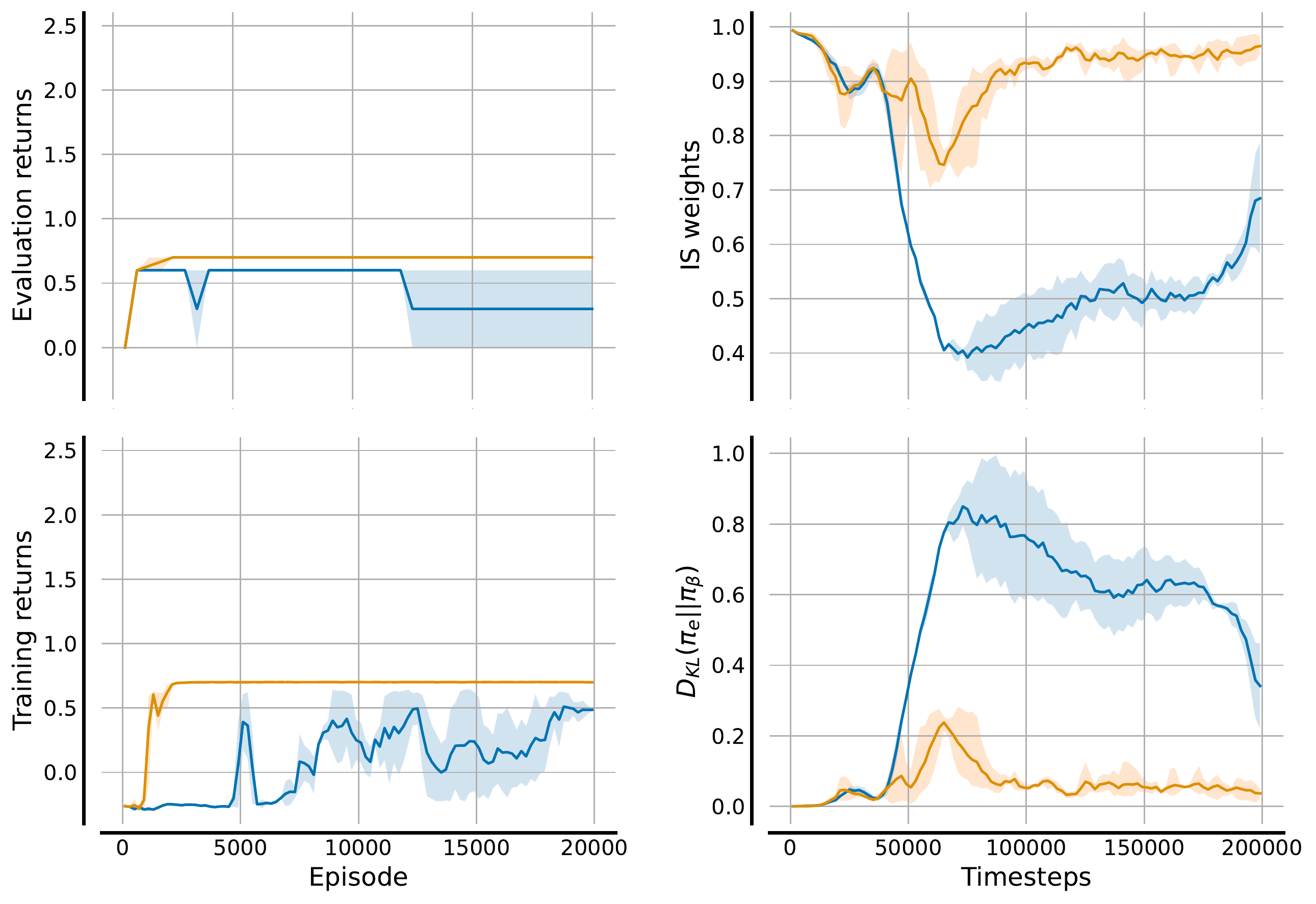}
        \caption {$\alpha_\beta=0.001, \alpha_e=0.0001$}
    \end{subfigure}
    
    \ \vspace{1em}
            
    \begin{subfigure}{.33\textwidth}
        \includegraphics[width=\textwidth]{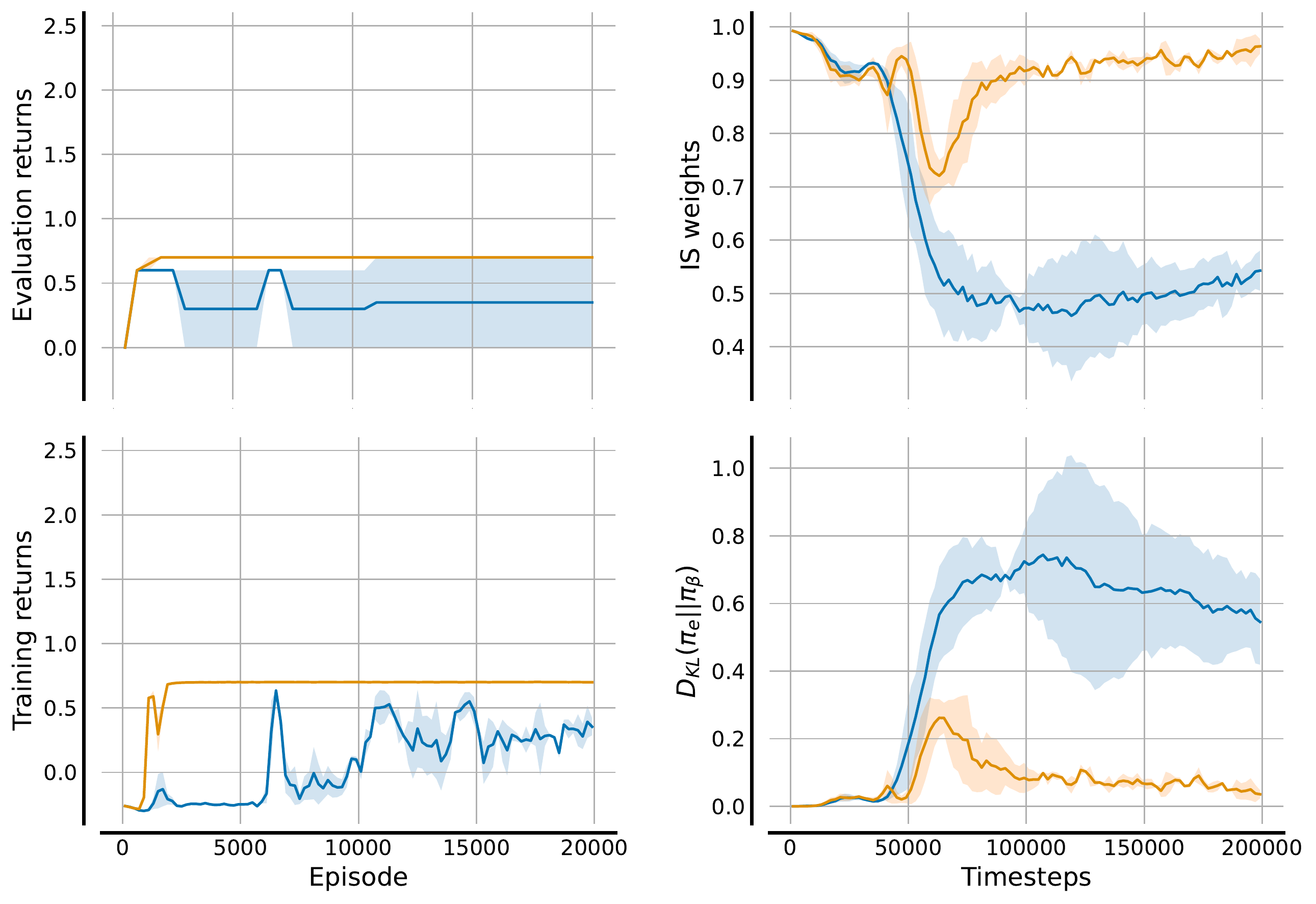}
        \caption {$\alpha_\beta=0.001, \alpha_e=0.001$}
    \end{subfigure}
    \hfill
    \begin{subfigure}{.33\textwidth}
        \includegraphics[width=\textwidth]{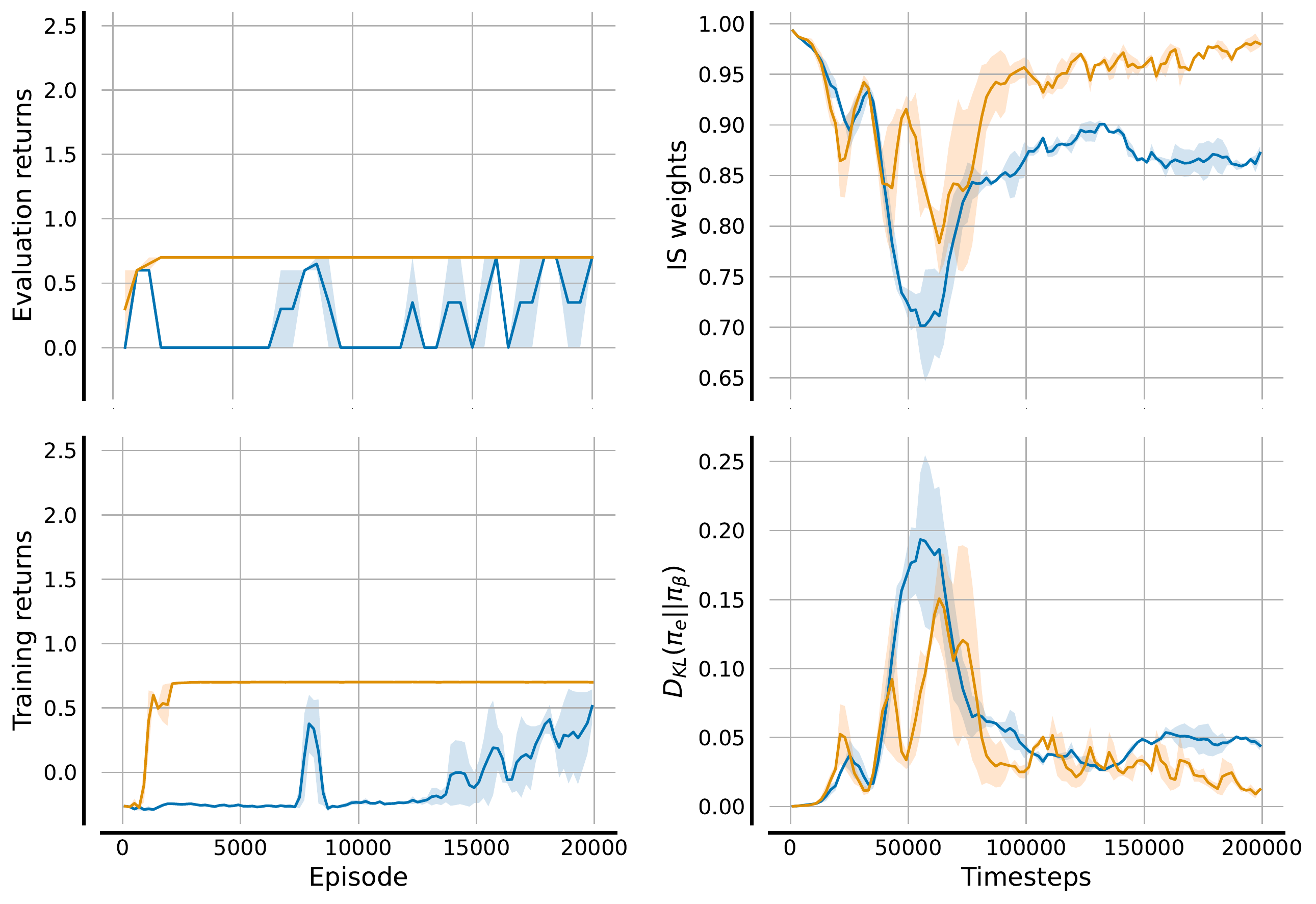}
        \caption {$\alpha_\beta=0.001, \alpha_e=0.01$}
    \end{subfigure}
    \hfill
    \begin{subfigure}{.33\textwidth}
        \includegraphics[width=\textwidth]{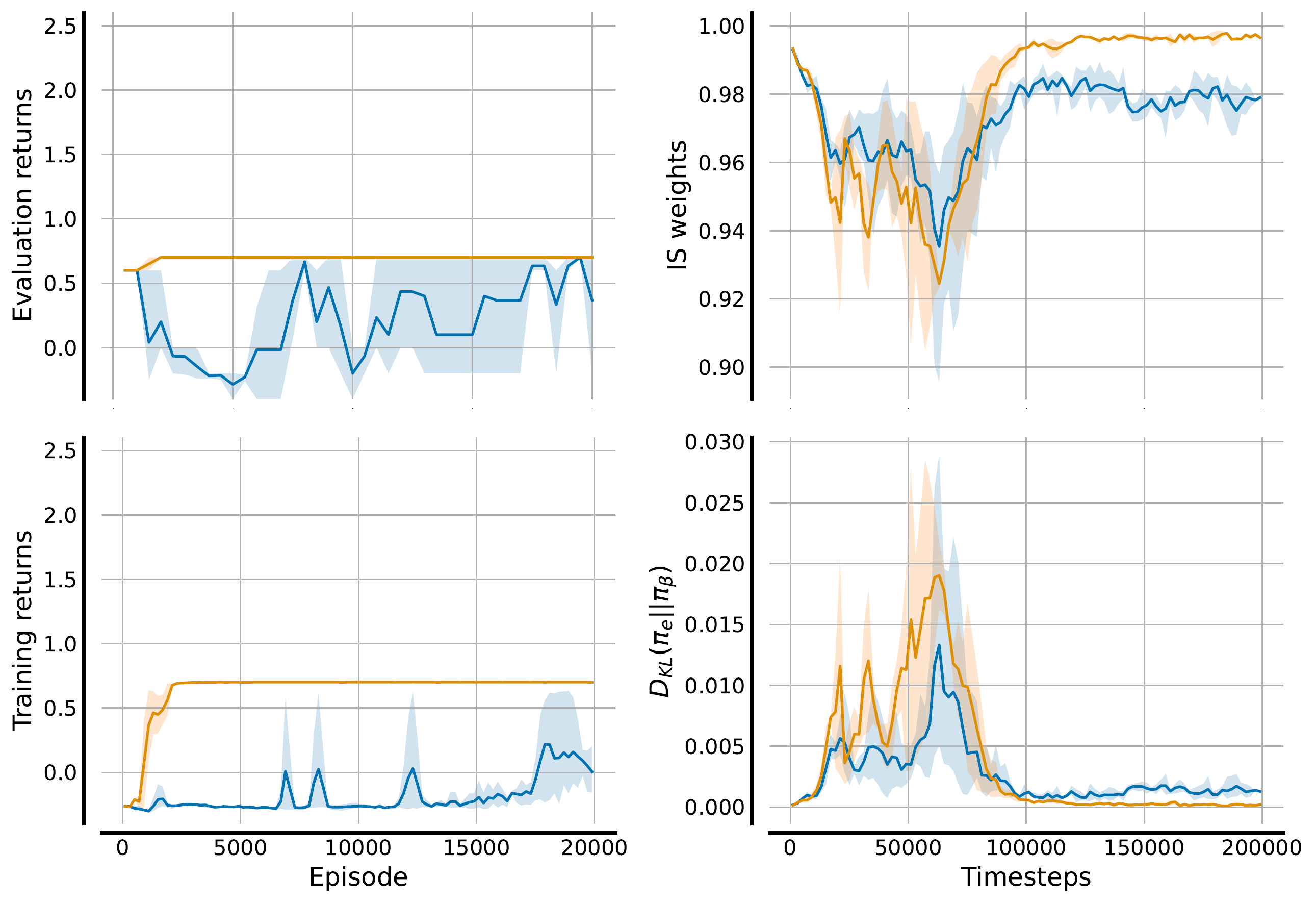}
        \caption {$\alpha_\beta=0.001, \alpha_e=0.1$}
    \end{subfigure}
    
    \ \vspace{1em}
    
    \begin{subfigure}{.33\textwidth}
        \includegraphics[width=\textwidth]{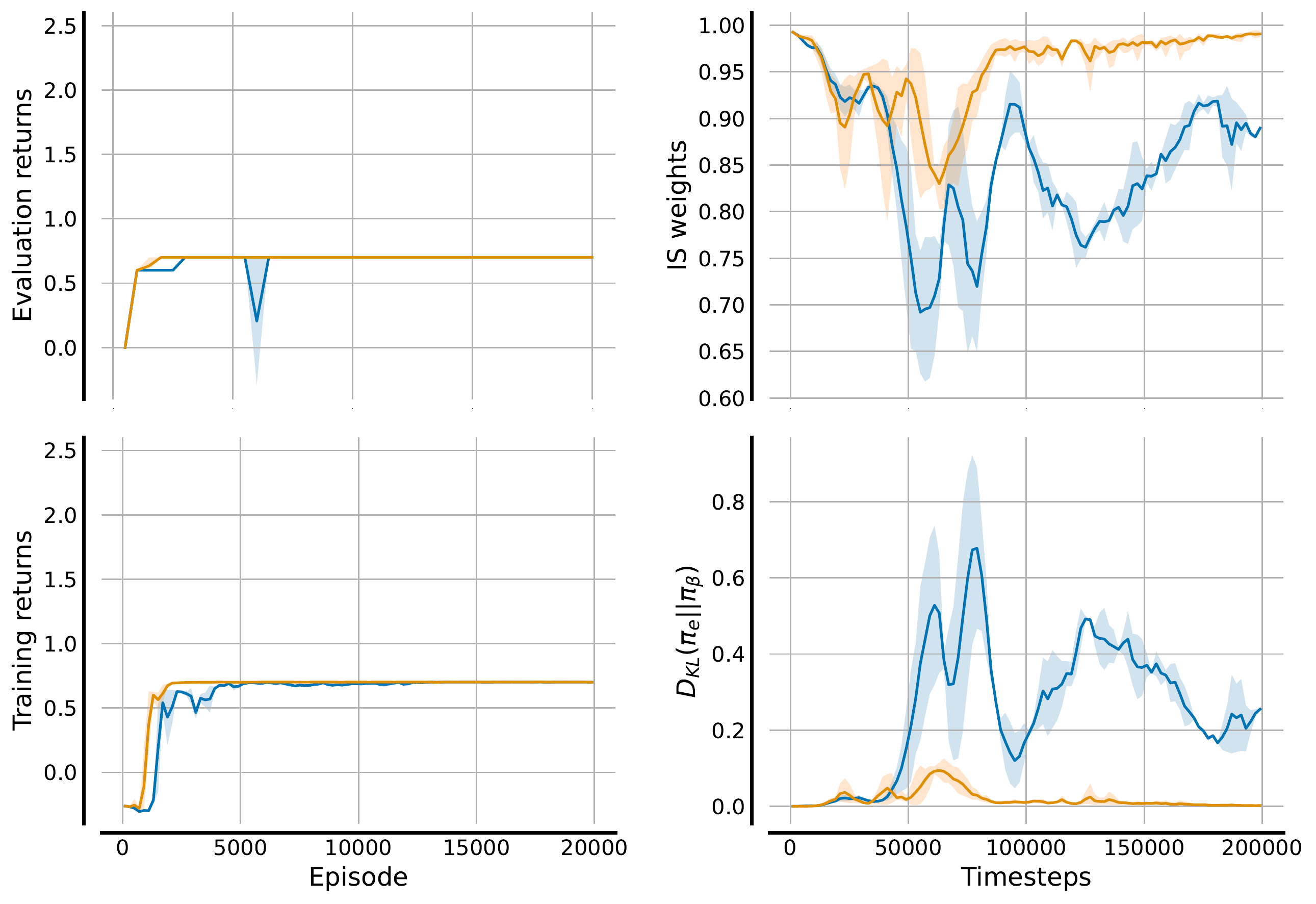}
        \caption {$\alpha_\beta=0.01, \alpha_e=0.0$}
    \end{subfigure}
    \hfill
    \begin{subfigure}{.33\textwidth}
        \includegraphics[width=\textwidth]{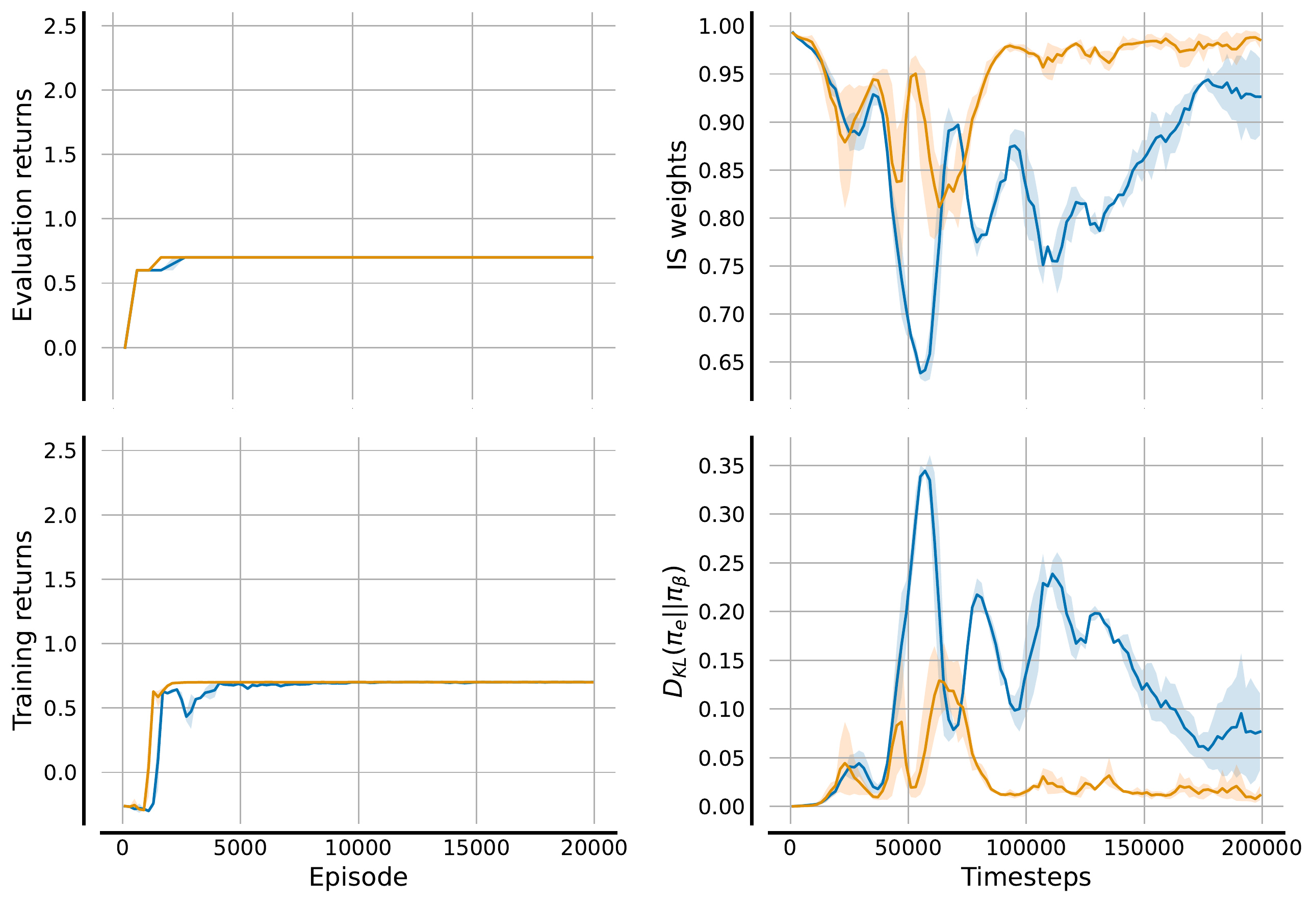}
        \caption {$\alpha_\beta=0.01, \alpha_e=0.0001$}
    \end{subfigure}
    \hfill
    \begin{subfigure}{.33\textwidth}
        \includegraphics[width=\textwidth]{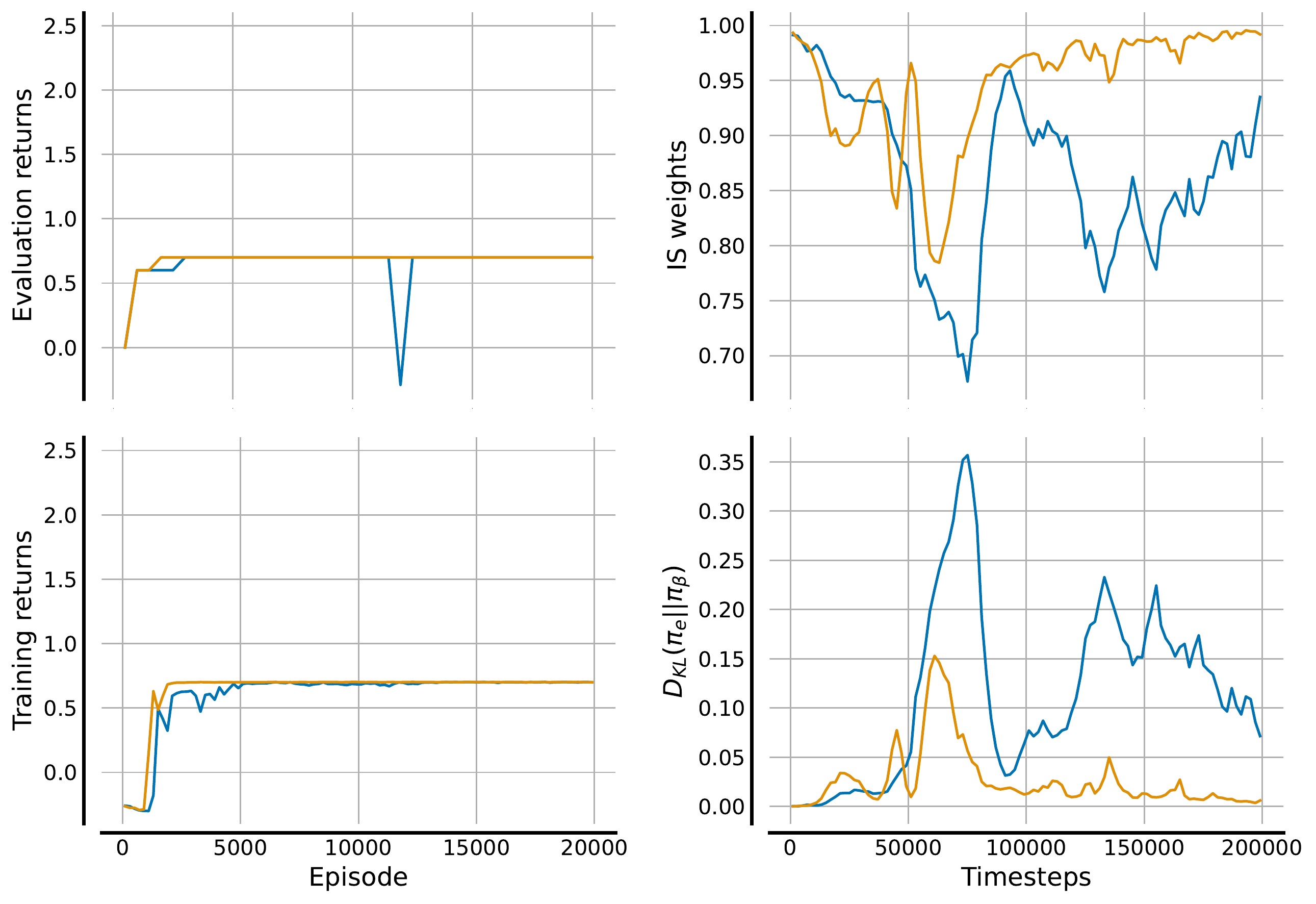}
        \caption {$\alpha_\beta=0.01, \alpha_e=0.001$}
    \end{subfigure}
    
    \ \vspace{1em}
    
    \begin{subfigure}{.33\textwidth}
        \includegraphics[width=\textwidth]{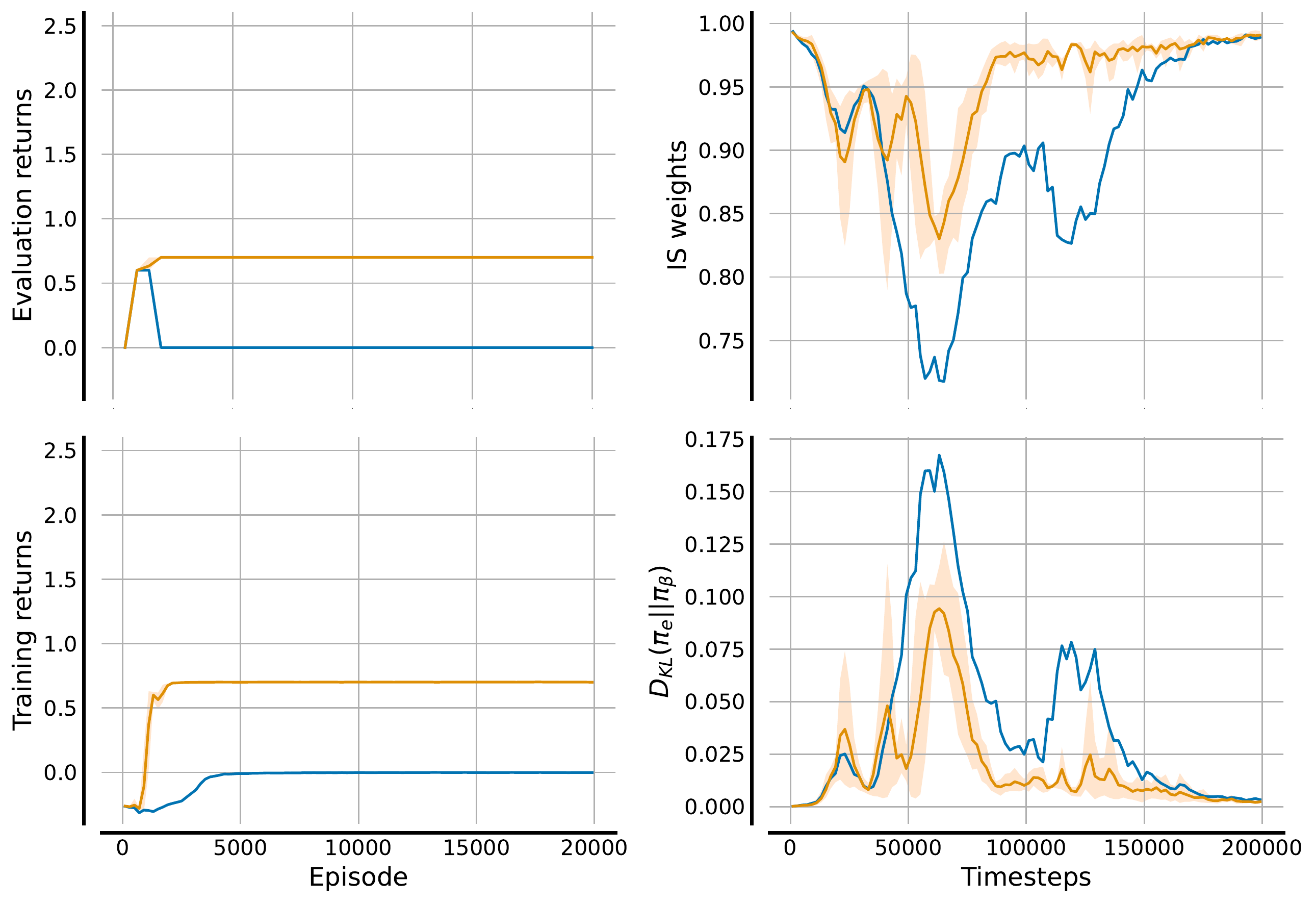}
        \caption {$\alpha_\beta=0.01, \alpha_e=0.01$}
    \end{subfigure}
    \hfill
    \begin{subfigure}{.33\textwidth}
        \includegraphics[width=\textwidth]{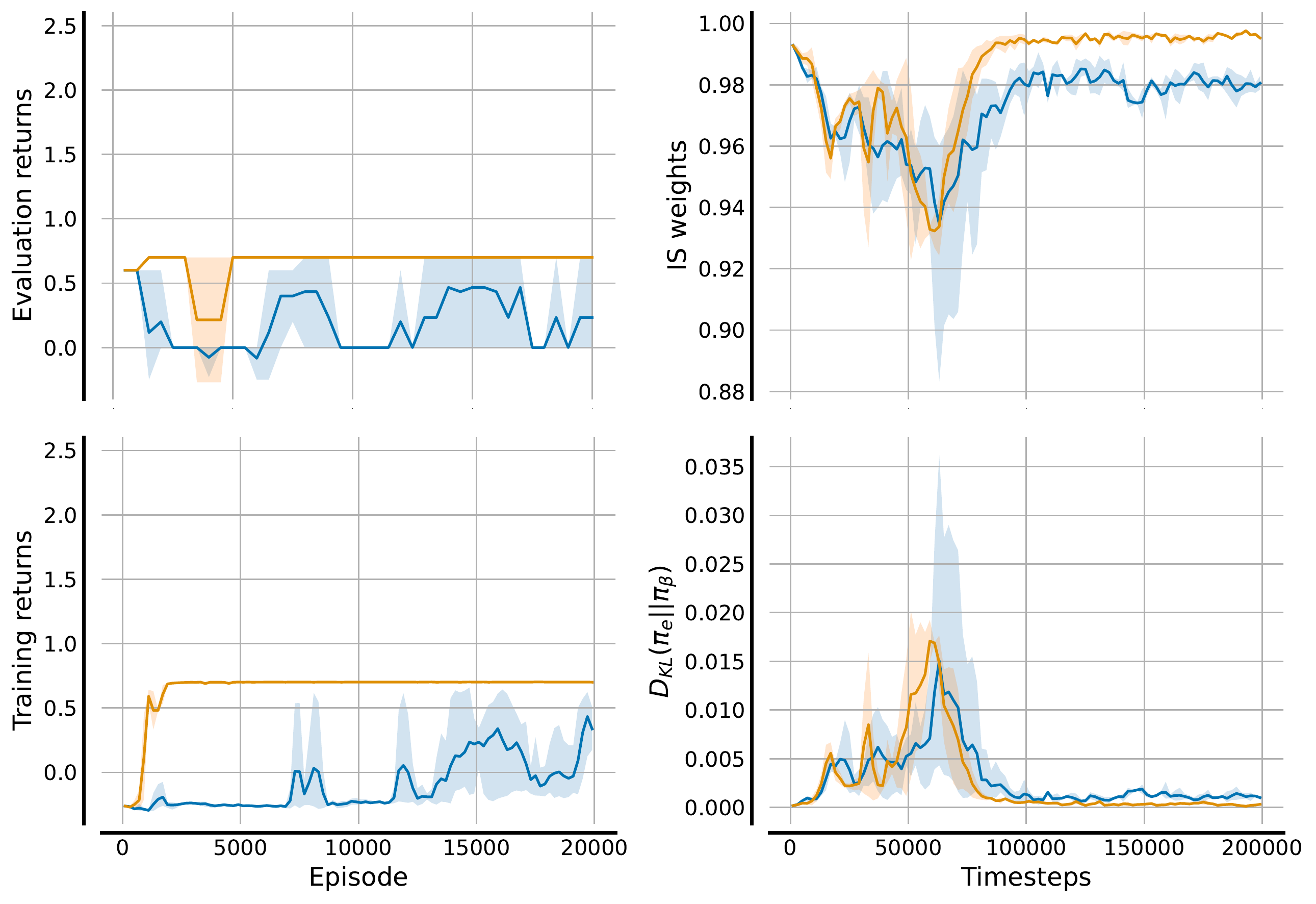}
        \caption {$\alpha_\beta=0.01, \alpha_e=0.1$}
    \end{subfigure}
    \hfill
    \begin{subfigure}{.33\textwidth}
        \includegraphics[width=\textwidth]{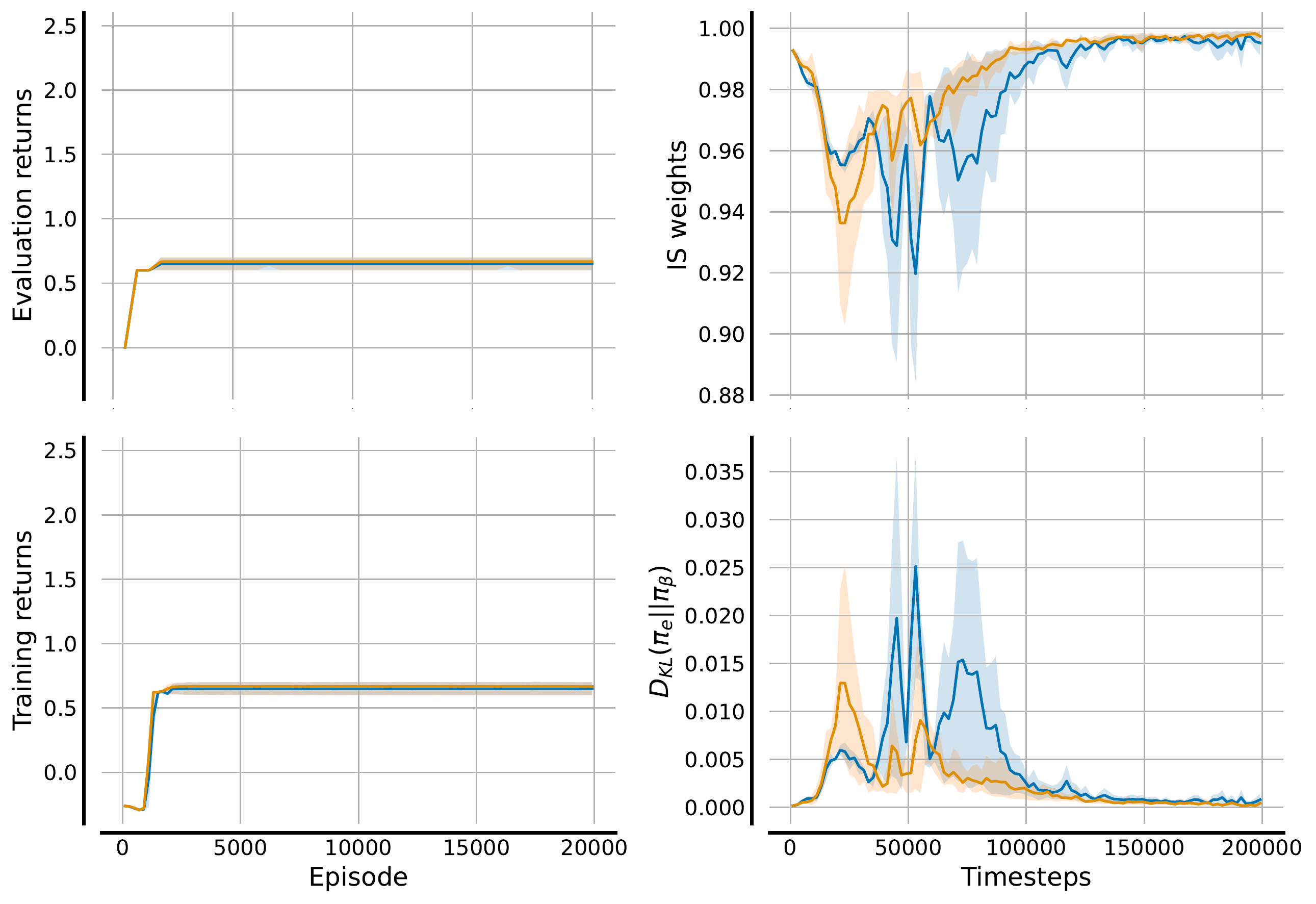}
        \caption {$\alpha_\beta=0.1, \alpha_e=0.0$}
    \end{subfigure}
    \begin{subfigure}{.5\textwidth}
        \centering
        \includegraphics[width=.5\textwidth]{media/kl_legend.pdf}
    \end{subfigure}
    \caption{Hallway $N_l=N_r=20$ evaluation with divergence constraint regularisation coefficients $\alpha_\beta$ and $\alpha_e$.  Shading indicates 95\% confidence intervals; Part 2}
    \label{fig:kl_divergence_constraint_hw20_2}
\end{figure}

\begin{figure}[!ht]
    \centering
    
    \begin{subfigure}{.33\textwidth}
        \includegraphics[width=\textwidth]{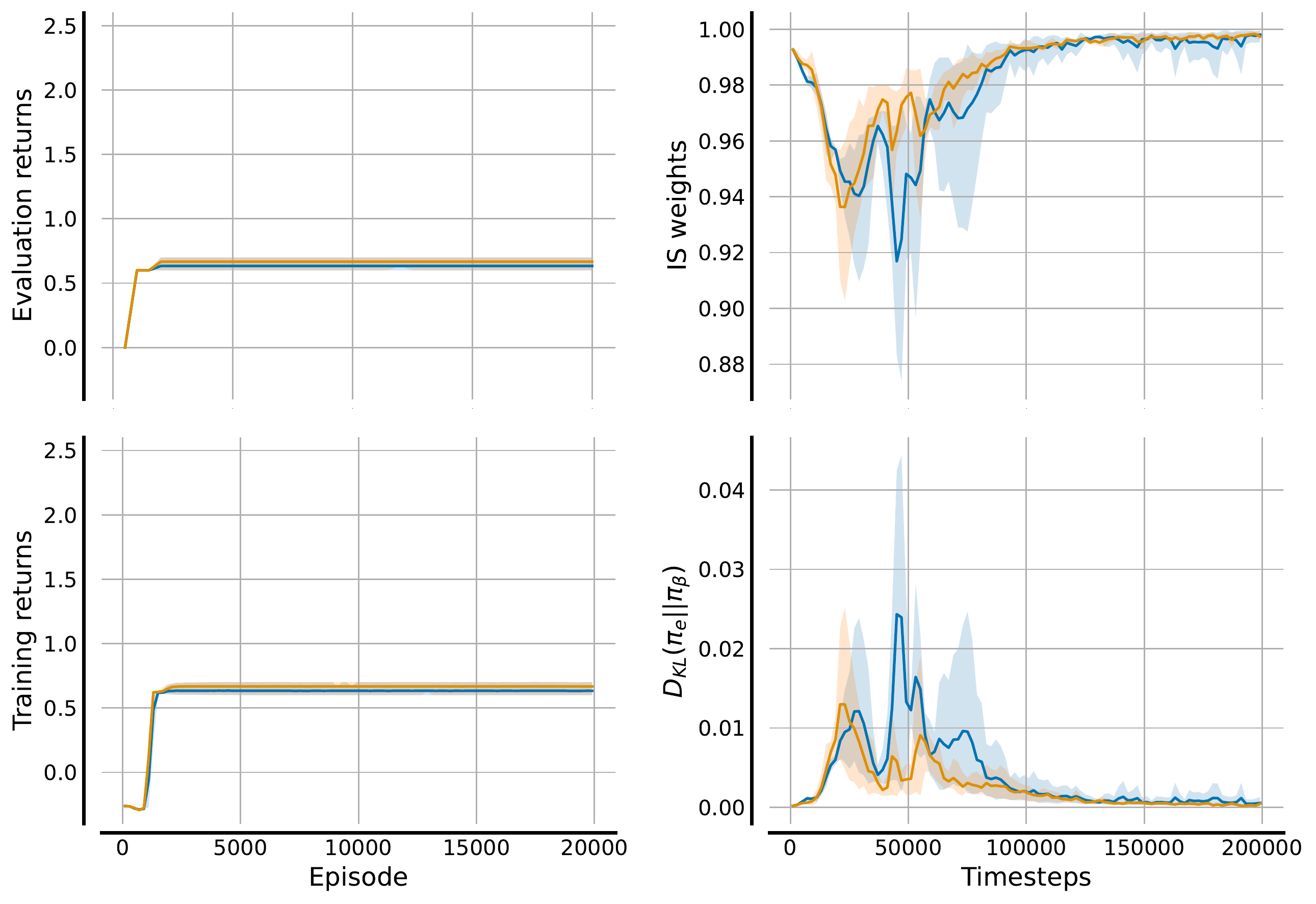}
        \caption {$\alpha_\beta=0.1, \alpha_e=0.0001$}
    \end{subfigure}
    \hfill
    \begin{subfigure}{.33\textwidth}
        \includegraphics[width=\textwidth]{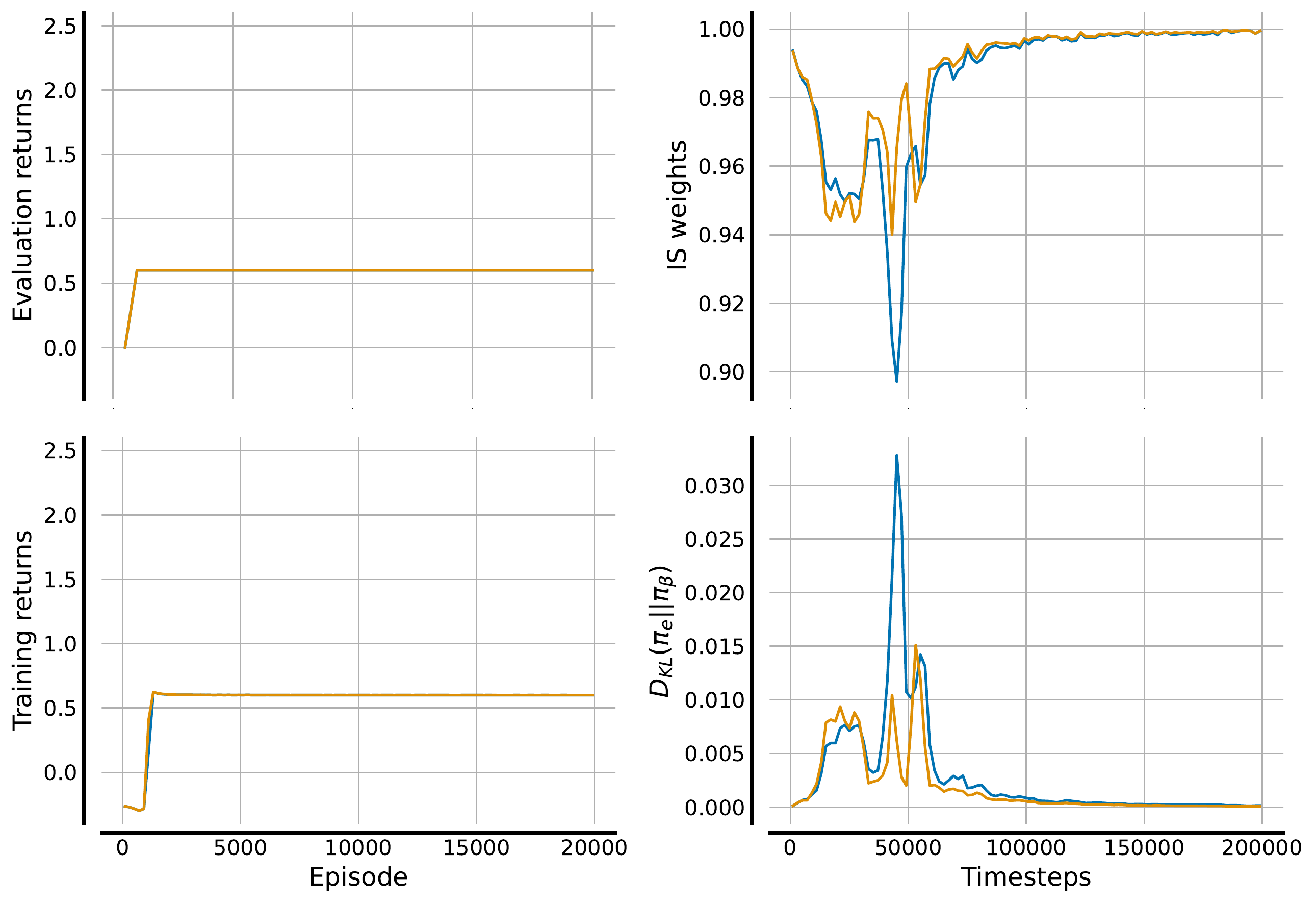}
        \caption {$\alpha_\beta=0.1, \alpha_e=0.001$}
    \end{subfigure}
    \hfill
    \begin{subfigure}{.33\textwidth}
        \includegraphics[width=\textwidth]{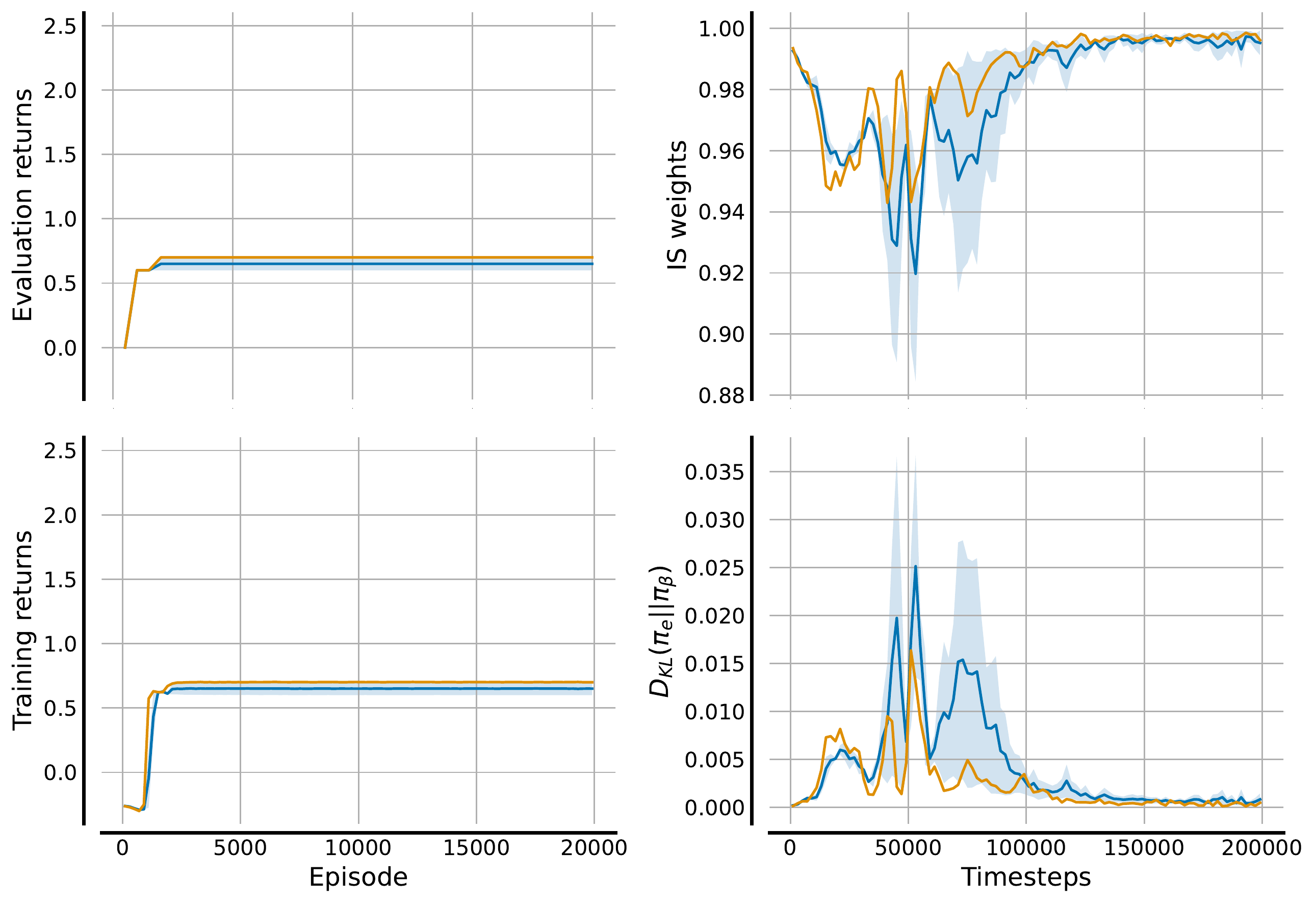}
        \caption {$\alpha_\beta=0.1, \alpha_e=0.01$}
    \end{subfigure}
            
    \ \vspace{1em}
    
    \begin{subfigure}{.33\textwidth}
        \includegraphics[width=\textwidth]{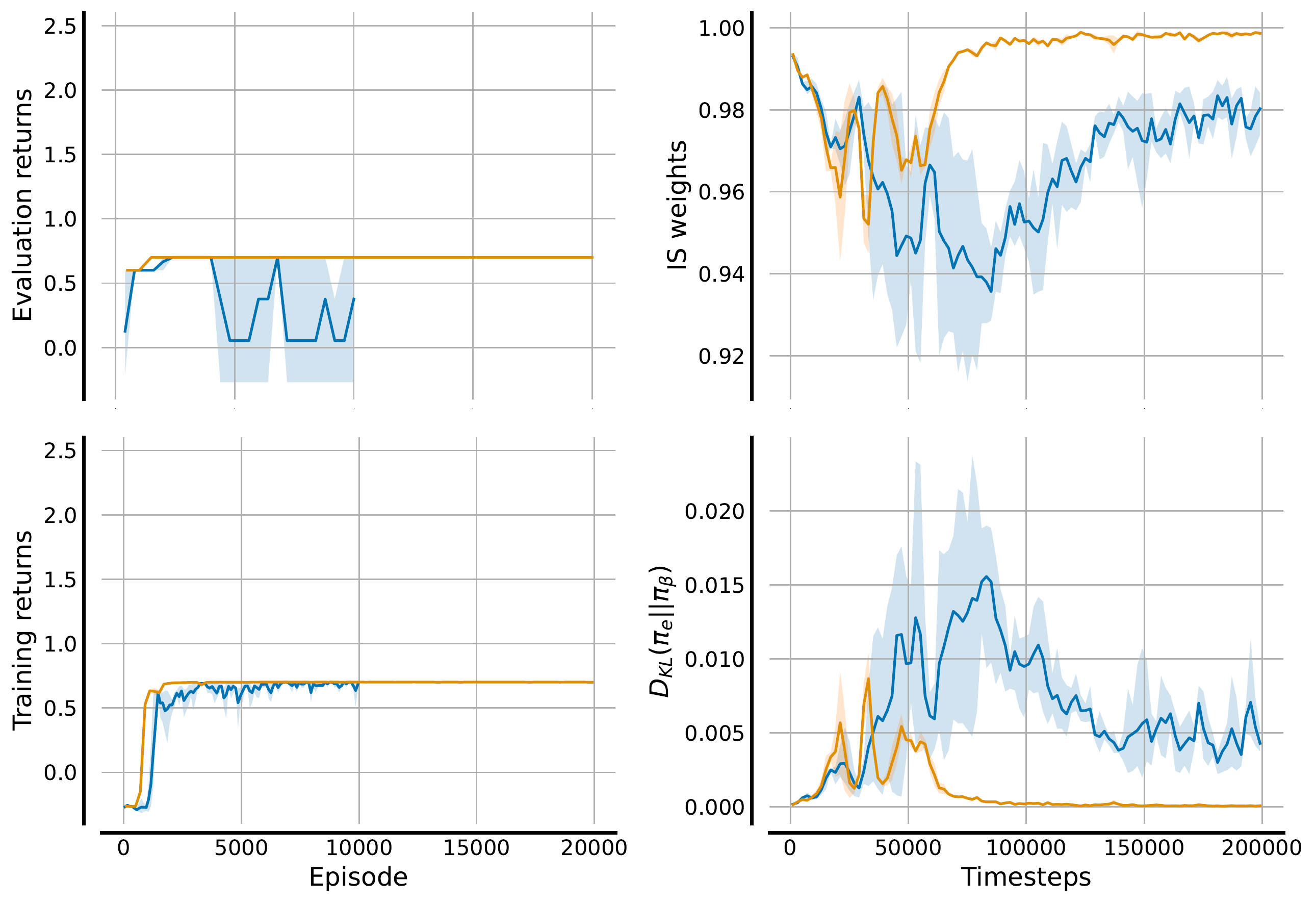}
        \caption {$\alpha_\beta=0.1, \alpha_e=0.1$}
    \end{subfigure}
    
    \begin{subfigure}{.5\textwidth}
        \centering
        \includegraphics[width=.5\textwidth]{media/kl_legend.pdf}
    \end{subfigure}
    \caption{Hallway $N_l=N_r=20$ evaluation with divergence constraint regularisation coefficients $\alpha_\beta$ and $\alpha_e$.  Shading indicates 95\% confidence intervals; Part 3}
    \label{fig:kl_divergence_constraint_hw20_3}
\end{figure}

\clearpage

\subsection{Sensitivity Analysis}
Below, we analyse the sensitivity of DeA2C trained with Count intrinsic rewards and KL-divergence constraints in DeepSea $10$ and Hallway $N_l=N_r=10$ tasks following the evaluation procedure outlined in \Cref{sec:results_sensitivity}. We consider both cases in which the exploration policy \pibeh is trained using extrinsic and intrinsic rewards or is trained using only intrinsic rewards. All results are computed over five random seeds.

\begin{figure*}[!ht]
    \begin{subfigure}{.33\textwidth}
        \centering
        \includegraphics[width=\linewidth]{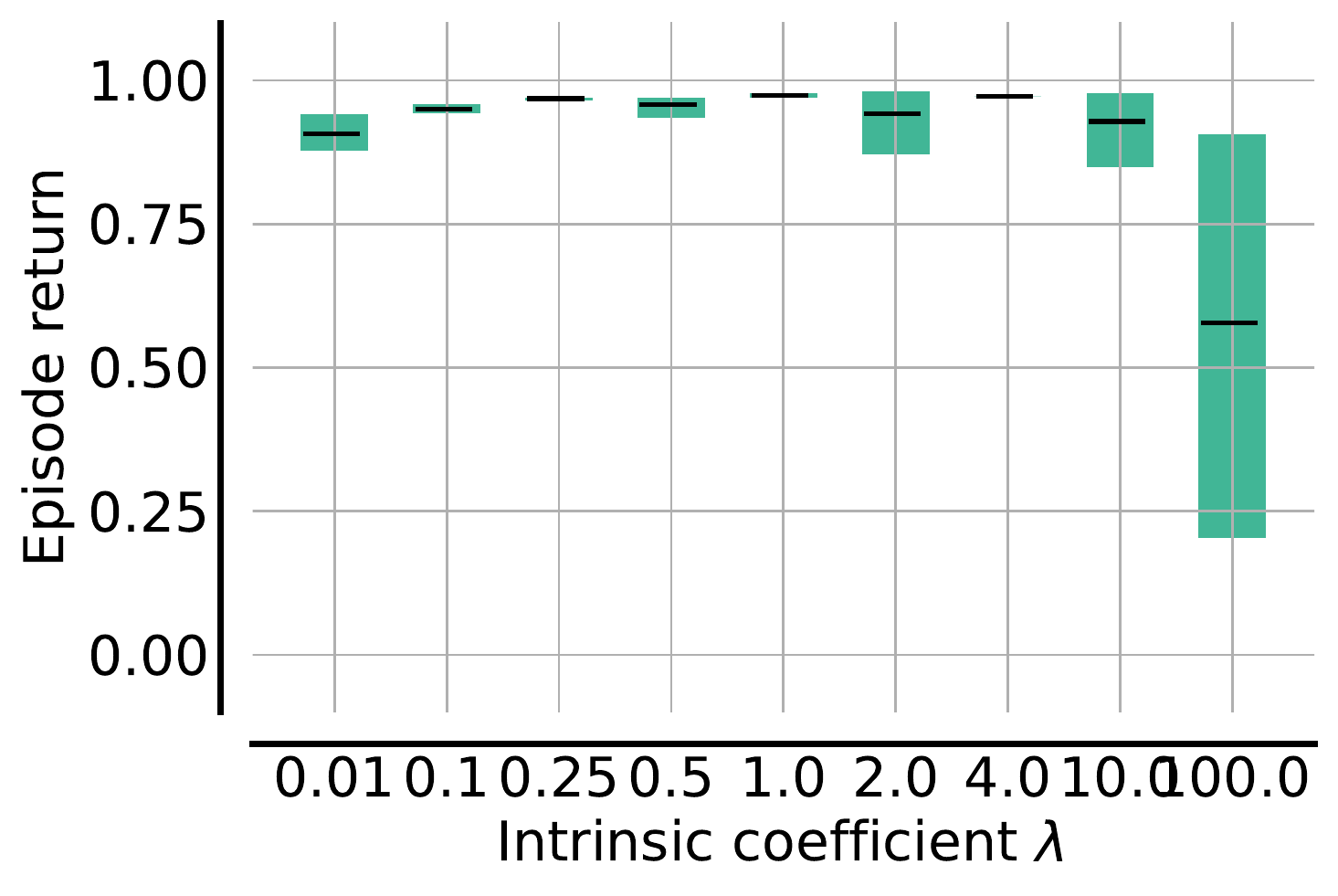}
        \caption{$\alpha_\beta=\alpha_e=0$, $r=r^e + \lambda r^i$}
    \end{subfigure}
    \hfill
    \begin{subfigure}{.33\textwidth}
        \centering
        \includegraphics[width=\linewidth]{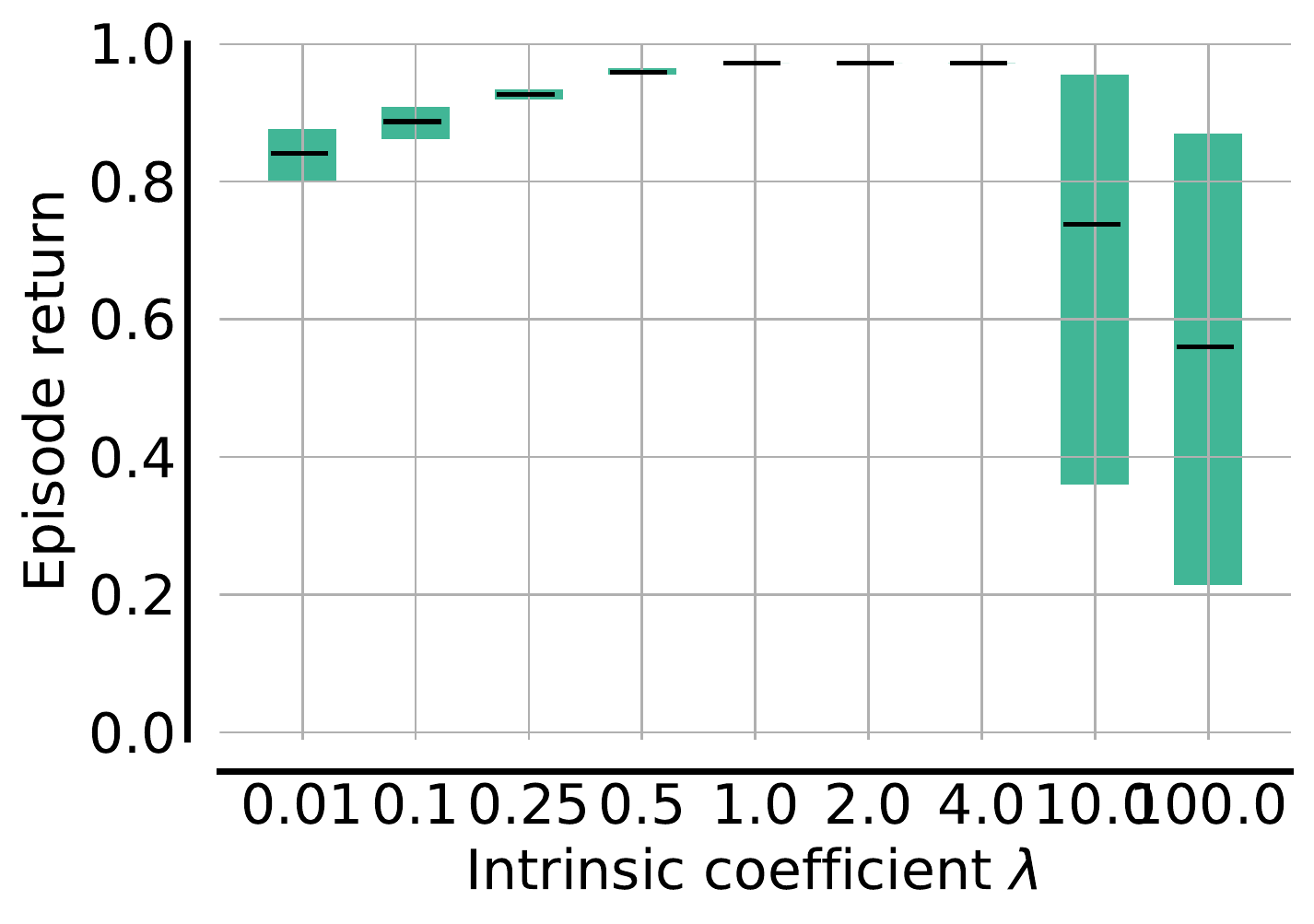}
        \caption{$\alpha_\beta=\alpha_e=0.1$, $r=r^e + \lambda r^i$}
    \end{subfigure}
    \hfill
    \begin{subfigure}{.33\textwidth}
        \centering
        \includegraphics[width=\linewidth]{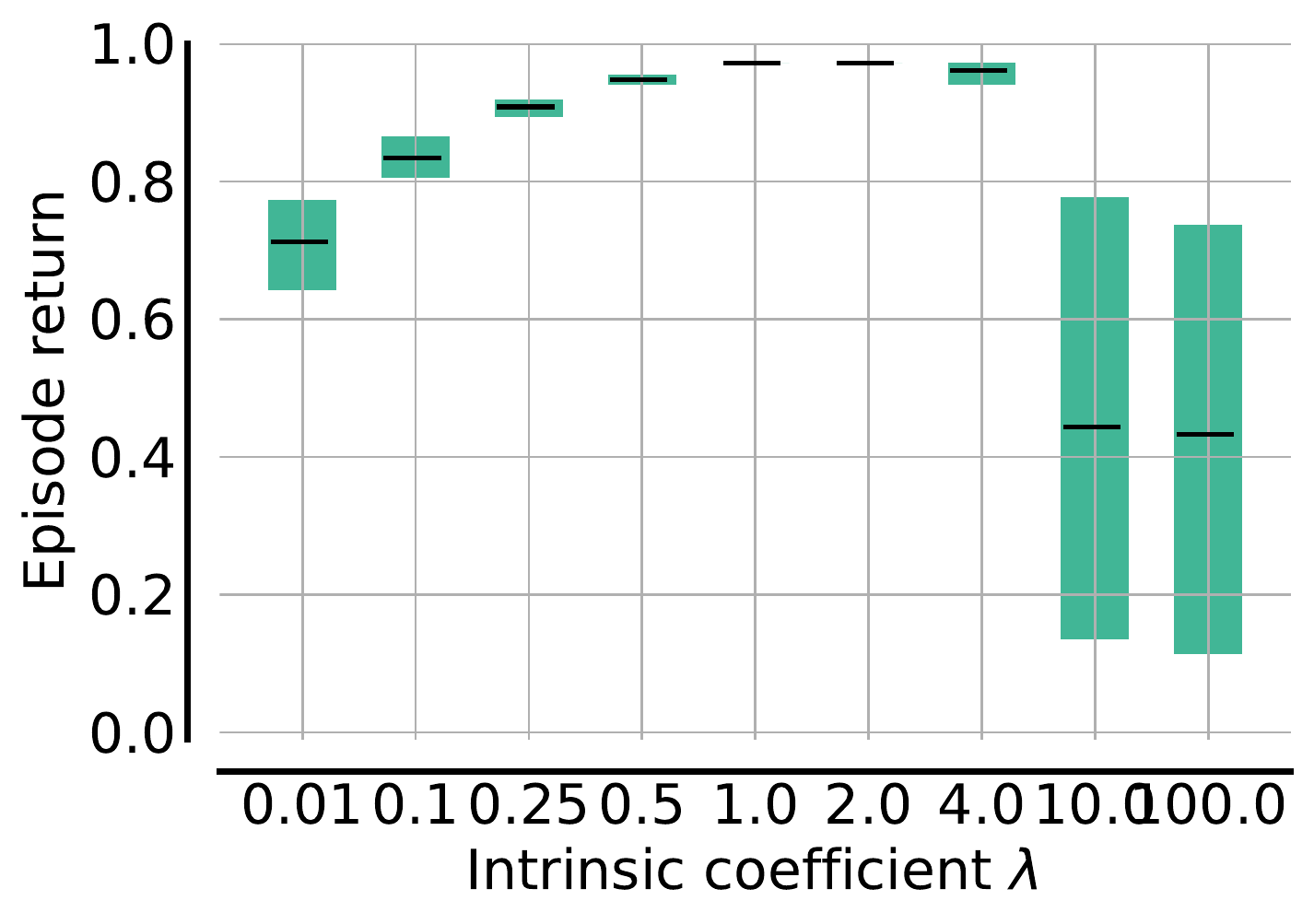}
        \caption{$\alpha_\beta=\alpha_e=0.1$, $r=\lambda r^i$}
    \end{subfigure}
    \caption{Average evaluation returns for DeA2C with Count in DeepSea 10 with $\lambda \in \{0.01, 0.1, 0.25, 0.5, 1.0, 2.0, 4.0, 10.0, 100.0\}$. Shading indicates 95\% confidence intervals.}
    \label{fig:deepsea_sensitivity_intcoefs_kl}
\end{figure*}

\begin{table}[!ht]
	\centering
	\caption{Maximum evaluation returns with a single standard deviation of DeA2C with Count in DeepSea 10 for $\lambda \in \{0.01, 0.1, 0.25, 0.5, 1.0, 2.0, 4.0, 10.0, 100.0\}$ and with or without KL-divergence constraints.}
	\resizebox{\textwidth}{!}{
	\robustify\bf
	\begin{tabular}{l S S S S S S S S S}
		\toprule
		{Algorithm \textbackslash \ $\lambda$} & {0.01} & {0.1} & {0.25} & {0.5} & {1.0} & {2.0} & {4.0} & {10.0} & {100.0} \\
		\midrule
		DeA2C Count (standard) & \bf 0.99(0) & \bf 0.99(0) & \bf 0.99(0) & \bf 0.99(0) & \bf 0.99(0) & \bf 0.99(0) & \bf 0.99(0) & \bf 0.99(0) & \bf 0.99(0) \\
		\midrule
		DeA2C Count ($\alpha_\beta=\alpha_e=0.1$, $r=r^e + \lambda r^i$) & \bf 0.99(0) & \bf 0.99(0) & \bf 0.99(0) & \bf 0.99(0) & \bf 0.99(0) & \bf 0.99(0) & \bf 0.99(0) & \bf 0.99(0) & \bf 0.99(0) \\
		DeA2C Count ($\alpha_\beta=\alpha_e=0.1$, $r=\lambda r^i$) & \bf 0.99(0) & \bf 0.99(0) & \bf 0.99(0) & \bf 0.99(0) & \bf 0.99(0) & \bf 0.99(0) & \bf 0.99(0) & \bf 0.99(0) & \bf 0.99(0) \\
		\bottomrule
	\end{tabular}
	}
	\label{tab:deepsea_10_intrinsic_coef_max_kl}
\end{table}

\begin{table}[!ht]
	\centering
	\caption{Average evaluation returns with a single standard deviation of DeA2C with Count in DeepSea 10 for $\lambda \in \{0.01, 0.1, 0.25, 0.5, 1.0, 2.0, 4.0, 10.0, 100.0\}$ and with or without KL-divergence constraints.}
	\resizebox{\textwidth}{!}{
	\robustify\bf
	\begin{tabular}{l S S S S S S S S S}
		\toprule
		{Algorithm \textbackslash \ $\lambda$} & {0.01} & {0.1} & {0.25} & {0.5} & {1.0} & {2.0} & {4.0} & {10.0} & {100.0} \\
		\midrule
		DeA2C Count (standard) & \bf 0.91(24) & \bf 0.95(18) & \bf 0.97(14) & \bf 0.96(15) & \bf 0.97(11) & \bf 0.94(13) & \bf 0.97(13) & \bf 0.93(15) & \bf 0.58(19) \\
		\midrule
		DeA2C Count ($\alpha_\beta=\alpha_e=0.1$, $r=r^e + \lambda r^i$) & \bf 0.84(32) & \bf 0.89(28) & \bf 0.93(24) & \bf 0.96(17) & \bf 0.97(13) & \bf 0.97(13) & \bf 0.97(13) &  0.74(16) & \bf 0.56(23) \\
		DeA2C Count ($\alpha_\beta=\alpha_e=0.1$, $r=\lambda r^i$) & \bf 0.71(40) & \bf 0.83(34) & \bf 0.91(26) & \bf 0.95(19) & \bf 0.97(13) & \bf 0.97(13) & \bf 0.96(14) &  0.44(24) & \bf 0.43(26) \\
		\bottomrule
	\end{tabular}
	}
	\label{tab:deepsea_10_intrinsic_coef_mean_kl}
\end{table}

\clearpage

\begin{figure*}[!ht]
    \begin{subfigure}{.33\textwidth}
        \centering
        \includegraphics[width=\linewidth]{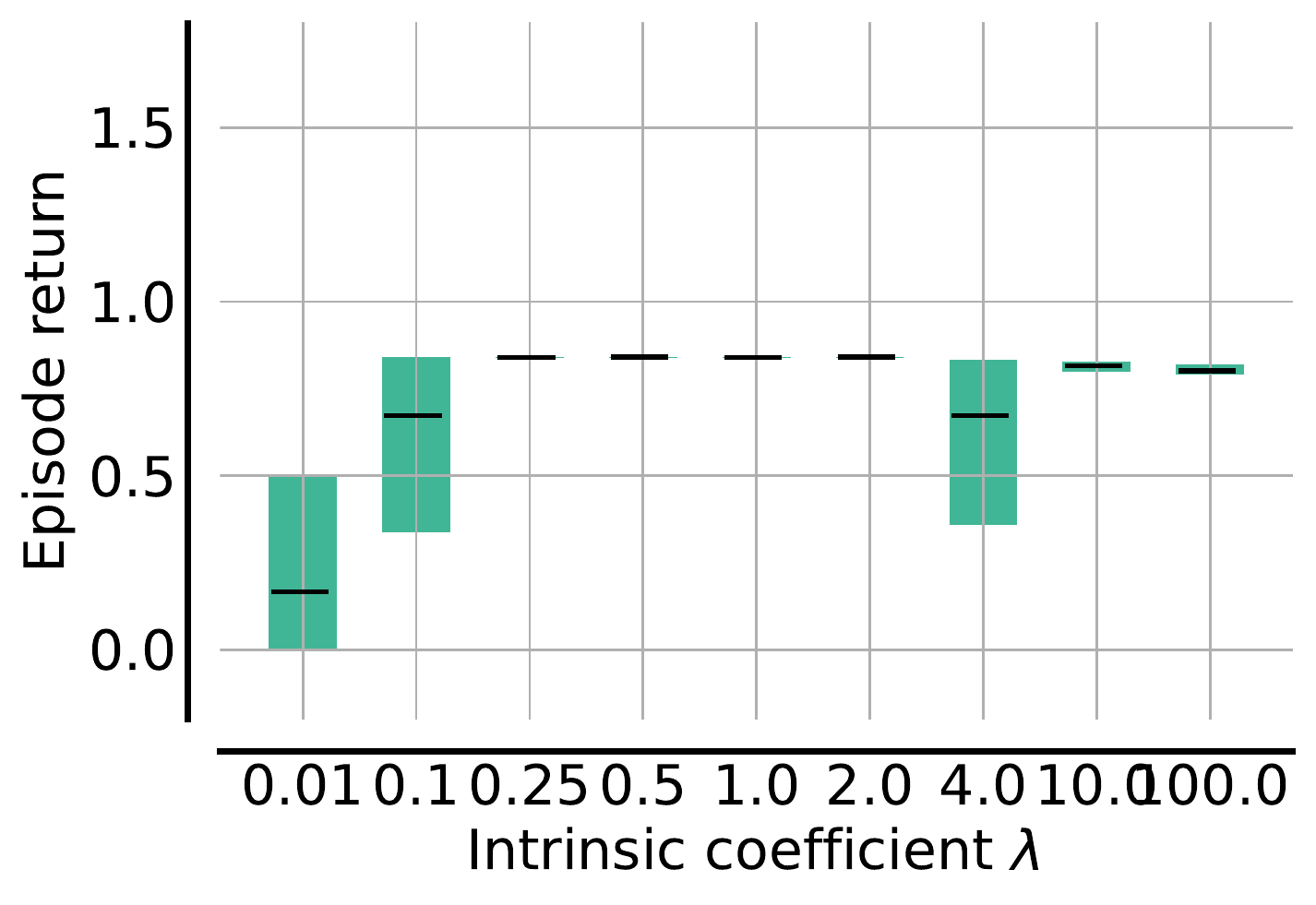}
        \caption{$\alpha_\beta=\alpha_e=0$, $r=r^e + \lambda r^i$}
    \end{subfigure}
    \hfill
    \begin{subfigure}{.33\textwidth}
        \centering
        \includegraphics[width=\linewidth]{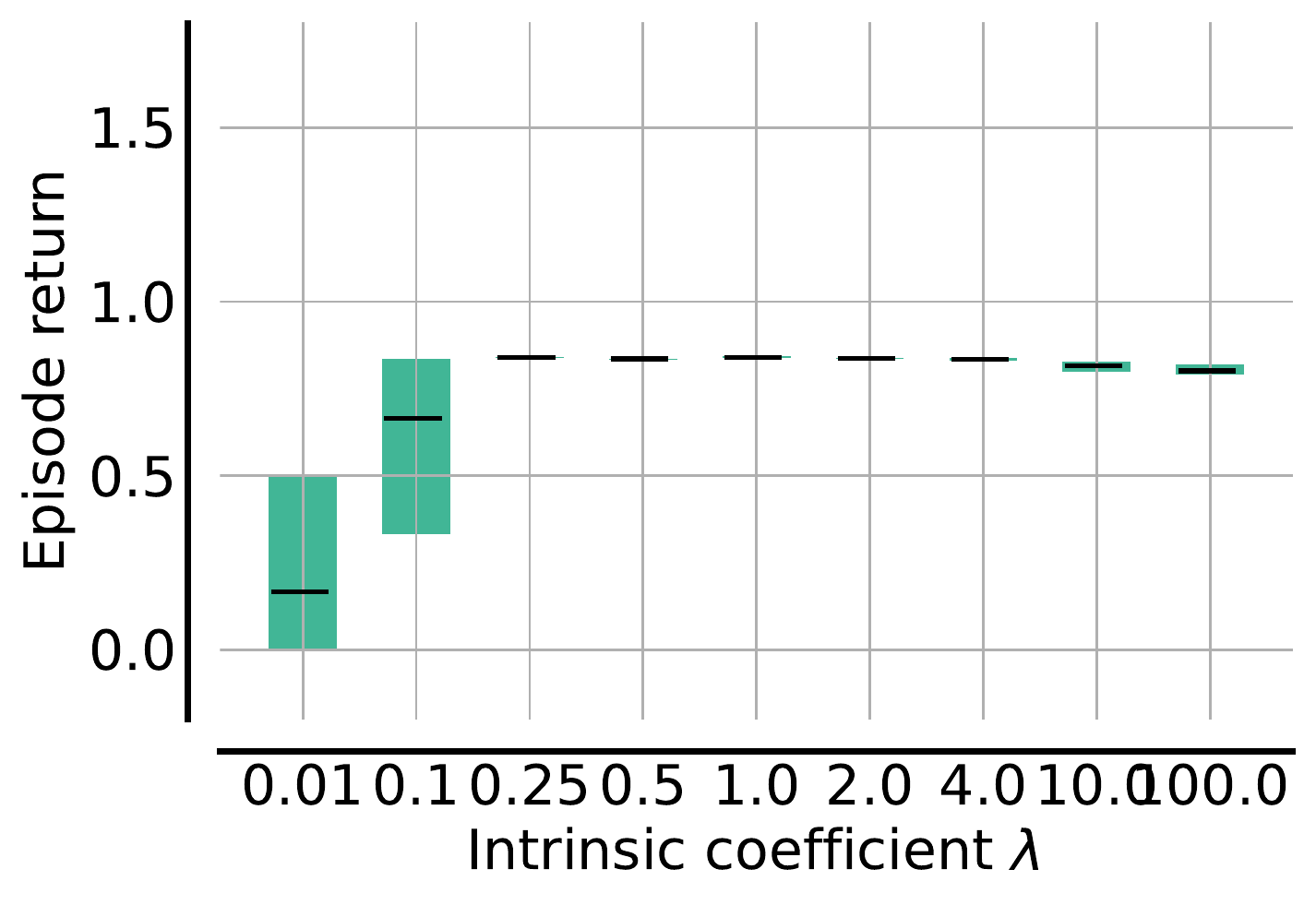}
        \caption{$\alpha_\beta=0.01, \alpha_e=0.0001$, $r=r^e + \lambda r^i$}
    \end{subfigure}
    \hfill
    \begin{subfigure}{.33\textwidth}
        \centering
        \includegraphics[width=\linewidth]{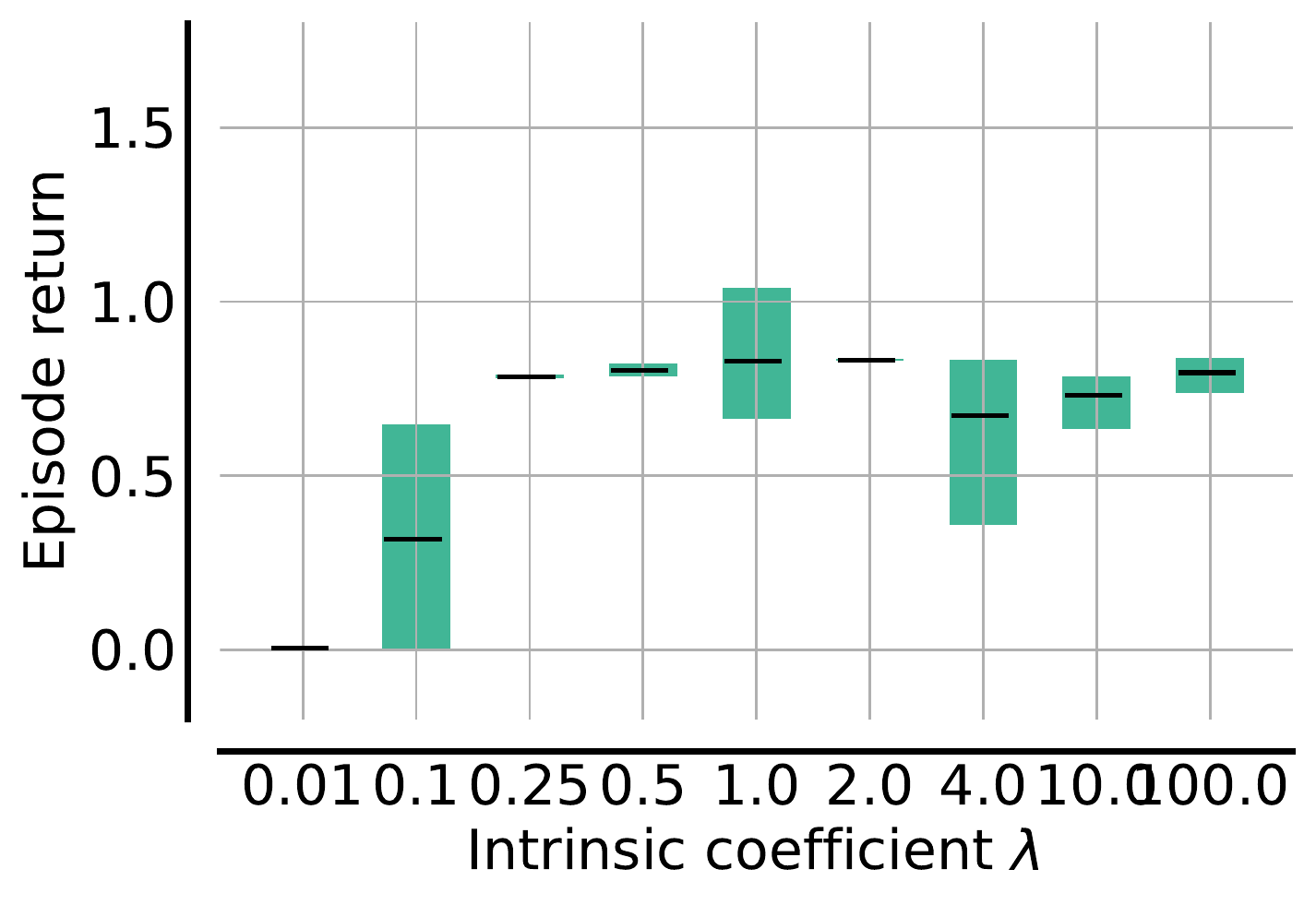}
        \caption{$\alpha_\beta=0.01, \alpha_e=0.0001$, $r=\lambda r^i$}
    \end{subfigure}
    \caption{Average evaluation returns for DeA2C with Count in Hallway $N_l=N_r=10$ with $\lambda \in \{0.01, 0.1, 0.25, 0.5, 1.0, 2.0, 4.0, 10.0, 100.0\}$. Shading indicates 95\% confidence intervals.}
    \label{fig:hallway_sensitivity_intcoefs_kl}
\end{figure*}

\begin{table}[!ht]
	\centering
	\caption{Maximum evaluation returns with a single standard deviation of DeA2C with Count in Hallway $N_l=N_r=10$ for $\lambda \in \{0.01, 0.1, 0.25, 0.5, 1.0, 2.0, 4.0, 10.0, 100.0\}$ and with or without KL-divergence constraints.}
	\resizebox{\textwidth}{!}{
	\robustify\bf
	\begin{tabular}{l S S S S S S S S S}
		\toprule
		{Algorithm \textbackslash \ $\lambda$} & {0.01} & {0.1} & {0.25} & {0.5} & {1.0} & {2.0} & {4.0} & {10.0} & {100.0} \\
		\midrule
	    DeA2C Count (standard) & \bf 0.17(26) & \bf 0.68(26) & \bf 0.85(0) & \bf 0.85(0) & \bf 0.85(0) & \bf 0.85(0) &  0.82(2) & \bf 0.84(2) & \bf 0.82(2) \\
		\midrule
		DeA2C Count ($\alpha_\beta=0.01, \alpha_e=0.0001$, $r=r^e + \lambda r^i$) & \bf 0.17(34) & \bf 0.68(34) & \bf 0.85(0) & \bf 0.85(0) & \bf 0.85(0) & \bf 0.85(0) & \bf 0.85(0) & \bf 0.84(2) & \bf 0.82(2) \\
		DeA2C Count ($\alpha_\beta=0.01, \alpha_e=0.0001$, $r=\lambda r^i$) & \bf 0.00(0) &  0.33(40) &  0.81(2) &  0.83(2) &  0.84(2) & \bf 0.85(0) &  0.82(2) &  0.81(2) & \bf 0.83(2) \\
		\bottomrule
	\end{tabular}
	}
	\label{tab:hallway10_10_intrinsic_coef_max_kl}
\end{table}

\begin{table}[!ht]
	\centering
	\caption{Average evaluation returns with a single standard deviation of DeA2C with Count in Hallway $N_l=N_r=10$ for $\lambda \in \{0.01, 0.1, 0.25, 0.5, 1.0, 2.0, 4.0, 10.0, 100.0\}$ and with or without KL-divergence constraints.}
	\resizebox{\textwidth}{!}{
	\robustify\bf
	\begin{tabular}{l S S S S S S S S S}
		\toprule
		{Algorithm \textbackslash \ $\lambda$} & {0.01} & {0.1} & {0.25} & {0.5} & {1.0} & {2.0} & {4.0} & {10.0} & {100.0} \\
		\midrule
		DeA2C Count (standard) & \bf 0.17(2) & \bf 0.67(7) & \bf 0.84(9) & \bf 0.84(8) & \bf 0.84(9) & \bf 0.84(8) &  0.67(10) & \bf 0.82(12) & \bf 0.80(6) \\
		\midrule
		DeA2C Count ($\alpha_\beta=0.01, \alpha_e=0.0001$, $r=r^e + \lambda r^i$) & \bf 0.17(2) & \bf 0.67(9) & \bf 0.84(9) & \bf 0.84(11) & \bf 0.84(7) & \bf 0.84(10) & \bf 0.83(10) & \bf 0.82(12) & \bf 0.80(6) \\
		DeA2C Count ($\alpha_\beta=0.01, \alpha_e=0.0001$, $r=\lambda r^i$) & 0.00(0) &  0.32(6) & \bf 0.78(12) & \bf 0.80(11) & \bf 0.83(19) & \bf 0.83(11) &  0.67(10) & \bf 0.73(15) & \bf 0.80(9) \\
		\bottomrule
	\end{tabular}
	}
	\label{tab:hallway10_10_intrinsic_coef_mean_kl}
\end{table}

\clearpage

\begin{figure*}[!ht]
    \begin{subfigure}{.33\textwidth}
        \centering
        \includegraphics[width=\linewidth]{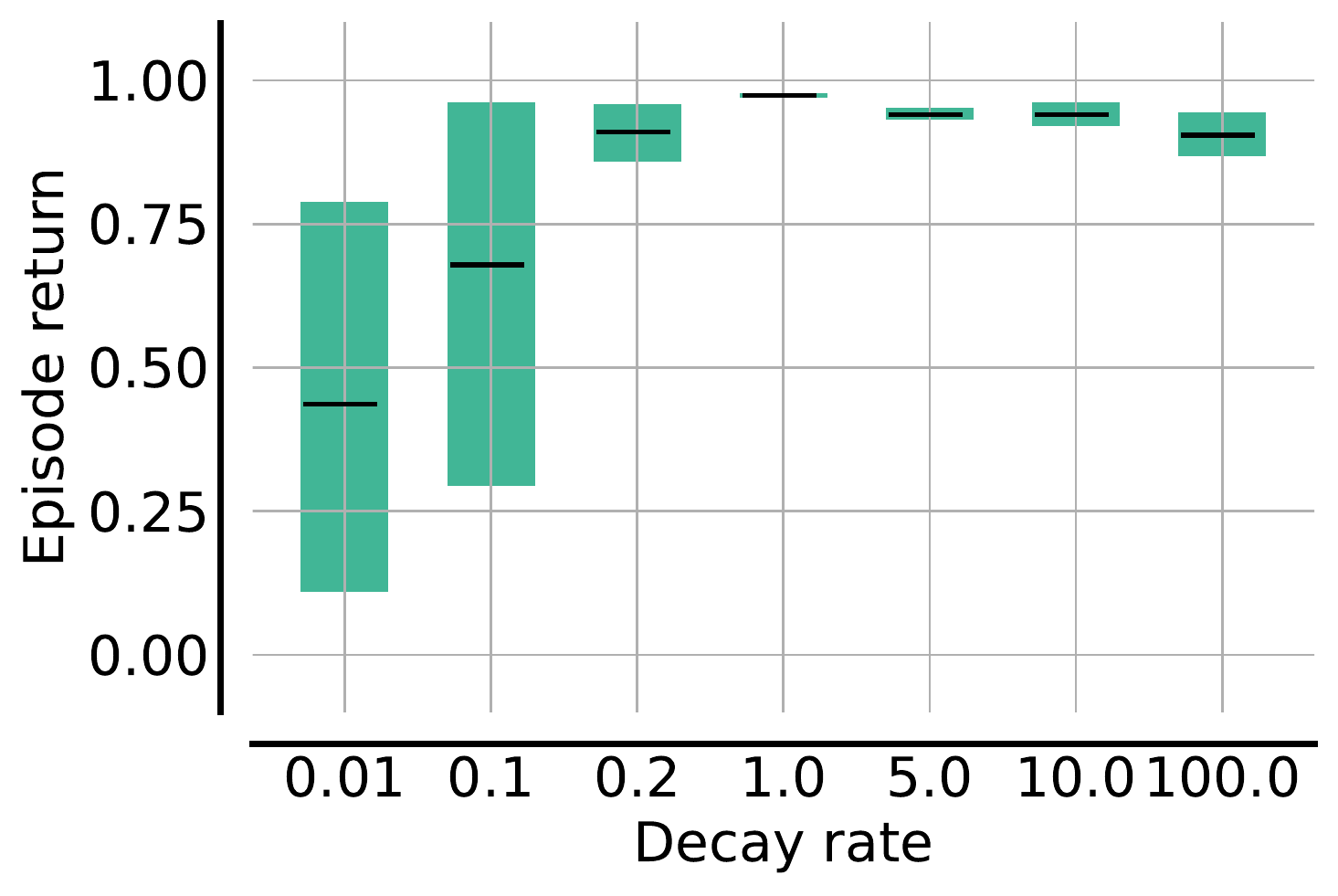}
        \caption{$\alpha_\beta=\alpha_e=0$, $r=r^e + \lambda r^i$}
    \end{subfigure}
    \hfill
    \begin{subfigure}{.33\textwidth}
        \centering
        \includegraphics[width=\linewidth]{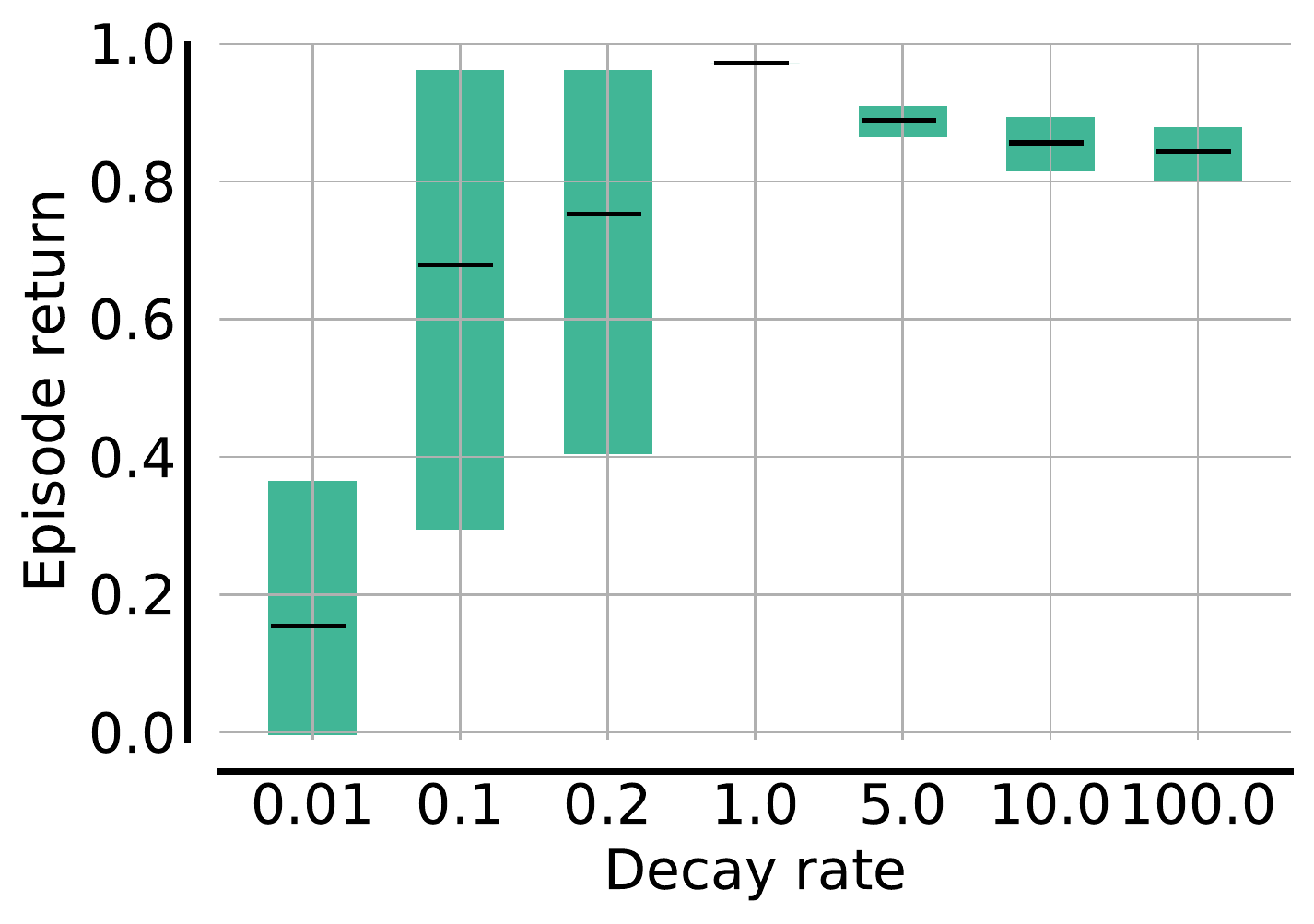}
        \caption{$\alpha_\beta=\alpha_e=0.1$, $r=r^e + \lambda r^i$}
    \end{subfigure}
    \hfill
    \begin{subfigure}{.33\textwidth}
        \centering
        \includegraphics[width=\linewidth]{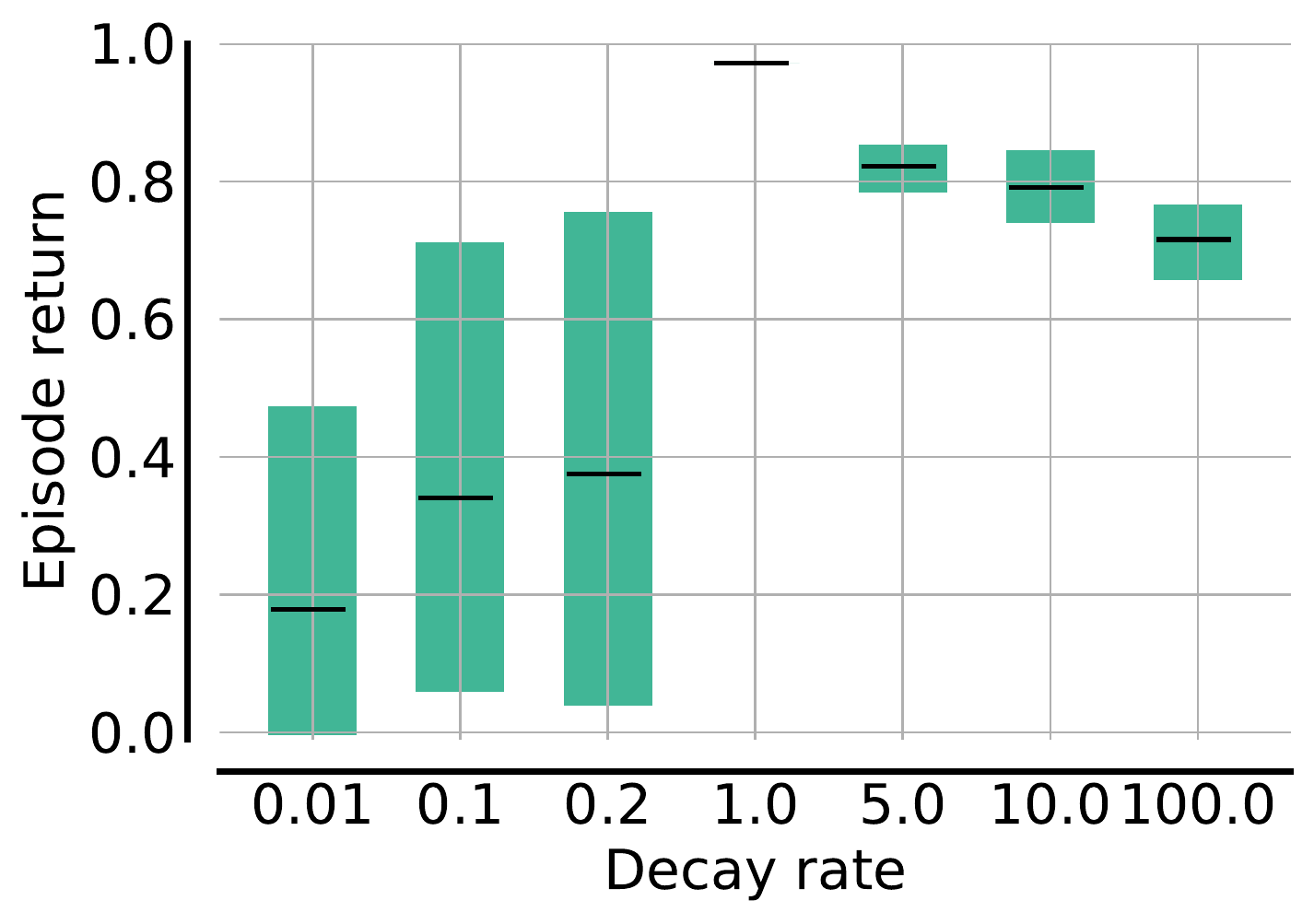}
        \caption{$\alpha_\beta=\alpha_e=0.1$, $r=\lambda r^i$}
    \end{subfigure}
    \caption{Average evaluation returns for DeA2C with Count in DeepSea 10 with varying count increments in $\{0.01, 0.1, 0.2, 1.0, 5.0, 10.0, 100.0\}$. Shading indicates 95\% confidence intervals.}
    \label{fig:deepsea_sensitivity_decay_kl}
\end{figure*}

\begin{table}[!ht]
	\centering
	\caption{Maximum evaluation returns with a single standard deviation of DeA2C with Count in DeepSea 10 with varying count increments and with or without KL-divergence constraints.}
	\resizebox{\linewidth}{!}{
	\robustify\bf
	\begin{tabular}{l S S S S S S S}
		\toprule
		{Algorithm \textbackslash \ Count increment} & {0.01} & {0.1} & {0.2} & {1.0} & {5.0} & {10.0} & {100.0} \\
		\midrule
		 DeA2C Count (standard) & \bf 0.99(0) & \bf 0.99(0) & \bf 0.99(0) & \bf 0.99(0) & \bf 0.99(0) & \bf 0.99(0) & \bf 0.99(0) \\
		 \midrule
		 DeA2C Count ($\alpha_\beta=\alpha_e=0.1$, $r=r^e + \lambda r^i$) & 0.59(40) & \bf 0.99(0) & \bf 0.99(0) & \bf 0.99(0) & \bf 0.99(0) & \bf 0.99(0) & \bf 0.99(0) \\
		 DeA2C Count ($\alpha_\beta=\alpha_e=0.1$, $r=\lambda r^i$) &  0.59(40) &  0.79(30) &  0.79(30) & \bf 0.99(0) & \bf 0.99(0) & \bf 0.99(0) & \bf 0.99(0) \\
		\bottomrule
	\end{tabular}
	}
	\label{tab:deepsea_decay_sensitivity_max_kl}
\end{table}

\begin{table}[!ht]
	\centering
	\caption{Average evaluation returns with a single standard deviation of DeA2C with Count in DeepSea 10 with varying count increments and with or without KL-divergence constraints.}
	\resizebox{\linewidth}{!}{
	\robustify\bf
	\begin{tabular}{l S S S S S S S}
		\toprule
		{Algorithm \textbackslash \ Count increment} & {0.01} & {0.1} & {0.2} & {1.0} & {5.0} & {10.0} & {100.0} \\
		\midrule
		 DeA2C Count (standard) & \bf 0.55(20) & \bf 0.68(15) & \bf 0.82(21) & \bf 0.98(10) & \bf 0.95(18) & \bf 0.94(18) & \bf 0.92(21) \\
		 \midrule
		 DeA2C Count ($\alpha_\beta=\alpha_e=0.1$, $r=r^e + \lambda r^i$) & 0.15(17) & \bf 0.68(15) & \bf 0.75(16) & \bf 0.97(13) & \bf 0.89(28) & \bf 0.86(30) & \bf 0.84(32) \\
		 DeA2C Count ($\alpha_\beta=\alpha_e=0.1$, $r=\lambda r^i$) & 0.18(14) &  0.34(20) &  0.38(16) & \bf 0.97(13) & \bf 0.82(35) & \bf 0.79(36) & \bf 0.72(41) \\
		\bottomrule
	\end{tabular}
	}
	\label{tab:deepsea_decay_sensitivity_mean_kl}
\end{table}

\clearpage

\begin{figure*}[!ht]
    \begin{subfigure}{.33\textwidth}
        \centering
        \includegraphics[width=\linewidth]{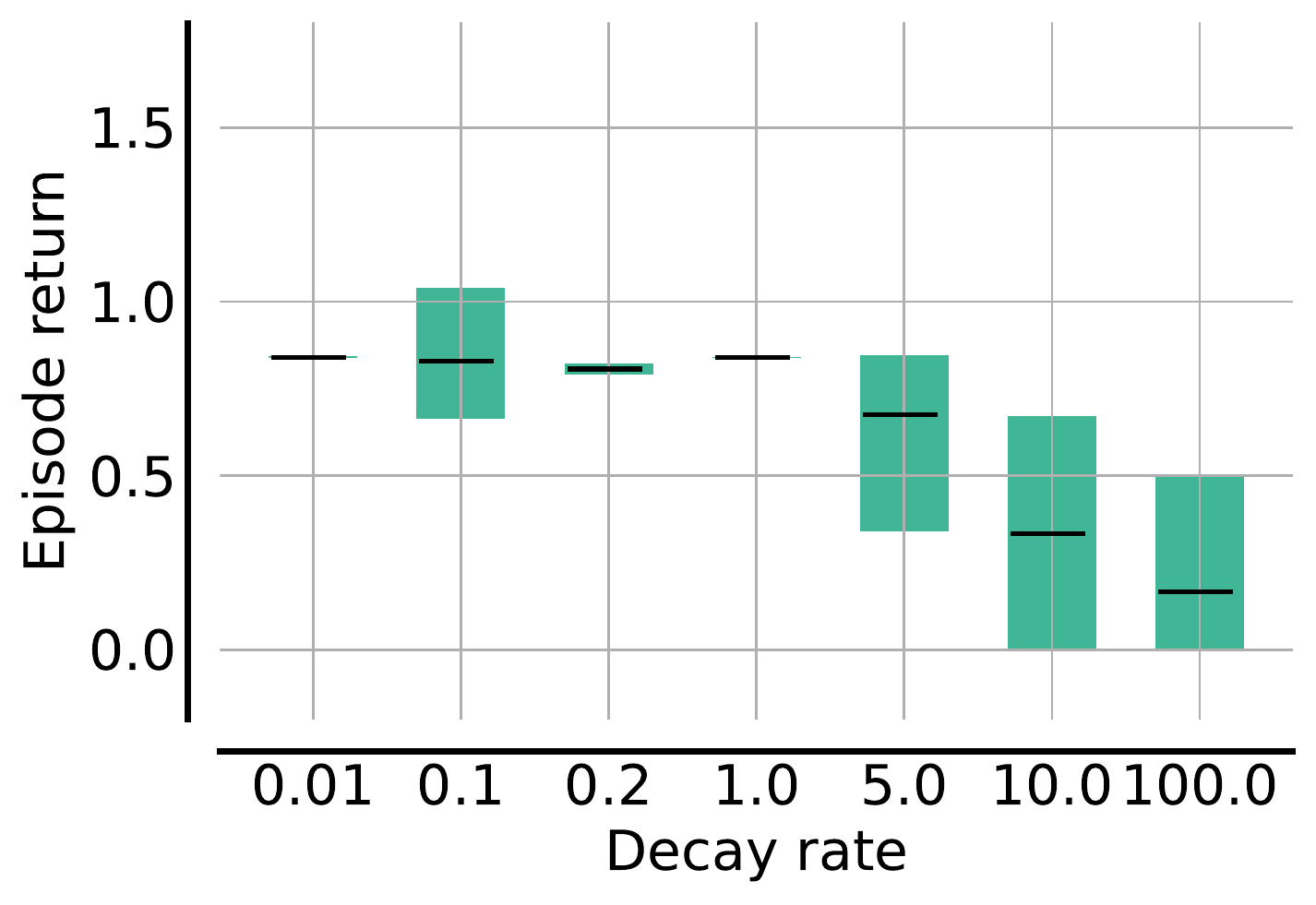}
        \caption{$\alpha_\beta=\alpha_e=0$, $r=r^e + \lambda r^i$}
    \end{subfigure}
    \hfill
    \begin{subfigure}{.33\textwidth}
        \centering
        \includegraphics[width=\linewidth]{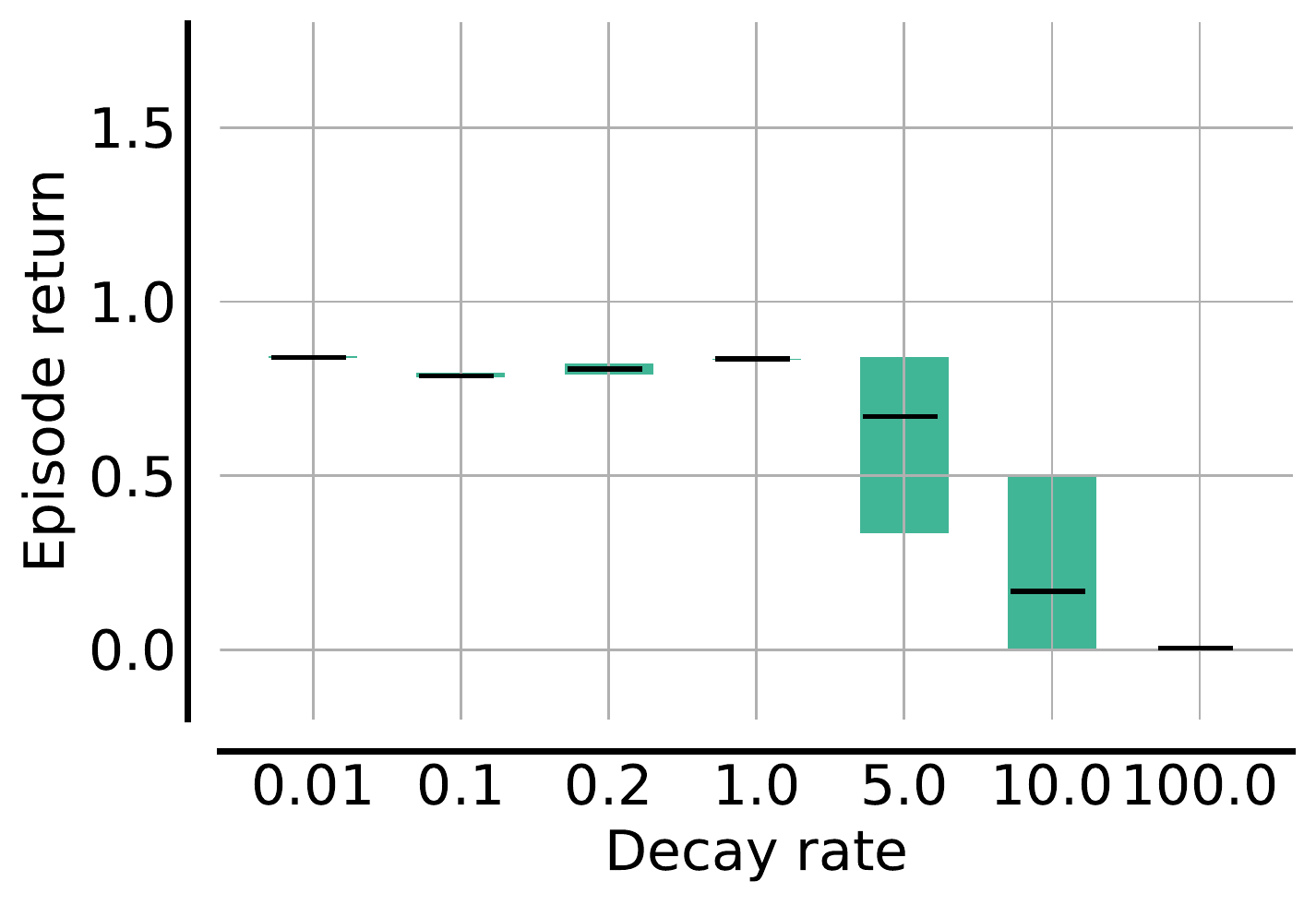}
        \caption{$\alpha_\beta=0.01, \alpha_e=0.0001$, $r=r^e + \lambda r^i$}
    \end{subfigure}
    \hfill
    \begin{subfigure}{.33\textwidth}
        \centering
        \includegraphics[width=\linewidth]{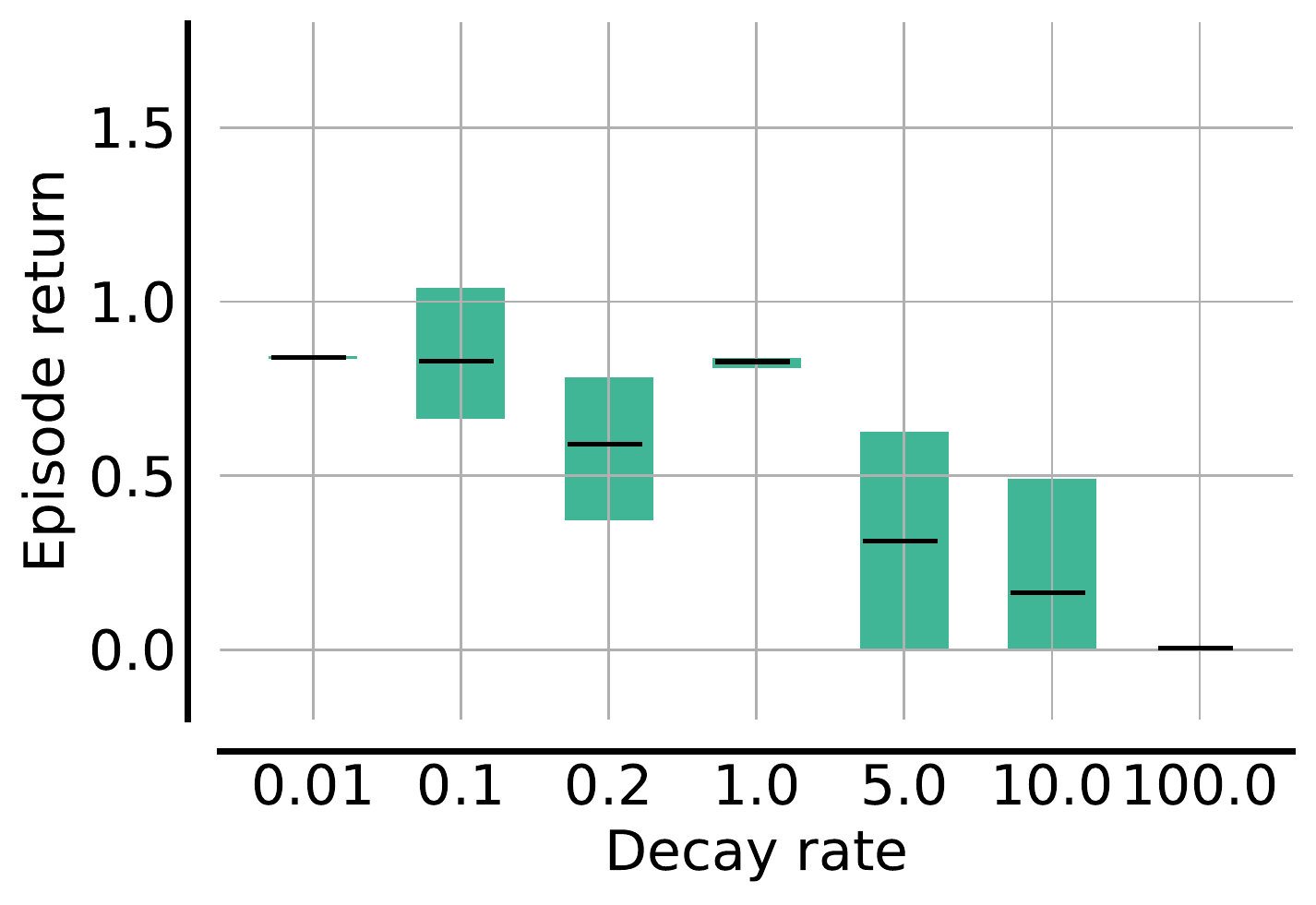}
        \caption{$\alpha_\beta=0.01, \alpha_e=0.0001$, $r=\lambda r^i$}
    \end{subfigure}
    \caption{Average evaluation returns for DeA2C with Count in Hallway $N_l=N_r=10$ with varying count increments in $\{0.01, 0.1, 0.2, 1.0, 5.0, 10.0, 100.0\}$. Shading indicates 95\% confidence intervals.}
    \label{fig:hallway_sensitivity_decay_kl}
\end{figure*}

\begin{table}[!ht]
	\centering
	\caption{Maximum evaluation returns with a single standard deviation of DeA2C with Count in Hallway $N_l=N_r=10$ with varying count increments and with or without KL-divergence constraints.}
	\resizebox{\linewidth}{!}{
	\robustify\bf
	\begin{tabular}{l S S S S S S S}
		\toprule
		{Algorithm \textbackslash \ Count increment} & {0.01} & {0.1} & {0.2} & {1.0} & {5.0} & {10.0} & {100.0} \\
		\midrule
		 DeA2C Count (standard) & \bf 0.85(0) & \bf 0.82(2) & \bf 0.82(2) & \bf 0.85(0) & \bf 0.57(40) & \bf 0.00(0) & \bf 0.00(0) \\
		 \midrule
		 DeA2C Count ($\alpha_\beta=0.01, \alpha_e=0.0001$, $r=r^e + \lambda r^i$) & \bf 0.85(0) & \bf 0.81(1) & \bf 0.83(2) & \bf 0.85(0) & \bf 0.68(26) & \bf 0.17(26) & \bf 0.00(0) \\
		 DeA2C Count ($\alpha_\beta=0.01, \alpha_e=0.0001$, $r=\lambda r^i$) & \bf 0.85(0) & \bf 1.00(30) &  0.80(0) &  0.84(2) &  0.32(32) & \bf 0.17(26) & \bf 0.00(0) \\
		\bottomrule
	\end{tabular}
	}
	\label{tab:hallway_decay_sensitivity_max_kl}
\end{table}

\begin{table}[!ht]
	\centering
	\caption{Average evaluation returns with a single standard deviation of DeA2C with Count in Hallway $N_l=N_r=10$ with varying count increments and with or without KL-divergence constraints.}
	\resizebox{\linewidth}{!}{
	\robustify\bf
	\begin{tabular}{l S S S S S S S}
		\toprule
		{Algorithm \textbackslash \ Count increment} & {0.01} & {0.1} & {0.2} & {1.0} & {5.0} & {10.0} & {100.0} \\
		\midrule
		 DeA2C Count (standard) & \bf 0.84(7) & \bf 0.83(19) & \bf 0.81(12) & \bf 0.84(9) & \bf 0.67(5) & \bf 0.33(4) & \bf 0.17(2) \\
		 \midrule
		 DeA2C Count ($\alpha_\beta=0.01, \alpha_e=0.0001$, $r=r^e + \lambda r^i$) & \bf 0.84(7) & \bf 0.79(12) & \bf 0.81(12) & \bf 0.84(11) & \bf 0.67(7) & 0.17(2) & 0.00(0) \\
		 DeA2C Count ($\alpha_\beta=0.01, \alpha_e=0.0001$, $r=\lambda r^i$) & \bf 0.84(7) & \bf 0.83(19) &  0.59(17) & \bf 0.83(10) &  0.31(5) & 0.16(3) & 0.00(0) \\
		\bottomrule
	\end{tabular}
	}
	\label{tab:hallway_decay_sensitivity_average_kl}
\end{table}

\end{document}